\documentclass[lettersize,journal]{IEEEtran}

\usepackage{cite}
\usepackage{amsmath,amsfonts}
\usepackage{algorithmic}
\usepackage{array}
\usepackage{textcomp}
\usepackage{stfloats}
\usepackage{url}
\usepackage{verbatim}
\usepackage{graphicx}
\usepackage{balance}
\usepackage{epsfig}
\usepackage{amssymb}

\usepackage{soul}
\usepackage[utf8]{inputenc}
\usepackage[list=true]{subcaption}
\usepackage{booktabs}
\usepackage{tabu}
\usepackage{wrapfig}
\usepackage{arydshln}
\usepackage[normalem]{ulem}
\usepackage{enumitem}
\usepackage{ragged2e}
\usepackage{hyperref}

\begin{document}

\title{Benchmarking Visual-Inertial Deep Multimodal Fusion for Relative Pose Regression and Odometry-aided Absolute Pose Regression}

\author{Felix Ott$^{*}$, Nisha Lakshmana Raichur$^{*}$, David Rügamer, Tobias Feigl, Heiko Neumann, Bernd Bischl, \\ Christopher Mutschler

\thanks{*F. Ott and N. L. Raichur contributed equally.}
\thanks{F. Ott, N. L. Raichur, T. Feigl, and C. Mutschler are with the Fraunhofer IIS, Fraunhofer Institute for Integrated Circuits IIS, Nuremberg, Germany. E-mail: \{felix.ott, nisha.lakshmana.raichur, tobias.feigl, christopher.mutschler\}@iis.fraunhofer.de}
\thanks{F. Ott, D. Rügamer, and B. Bischl are with the Statistical Learning and Data Science Chair, LMU Munich, Munich, Germany, and with the Munich Center for Machine Learning (MCML), Munich, Germany. E-mail: \{david.ruegamer, bernd.bischl\}@stat.uni-muenchen.de}
\thanks{D. Rügamer is with the Technical University of Dortmund, Germany}
\thanks{H. Neumann is with the Institute of Neural Information Processing, Ulm University, Ulm, Germany. E-mail: heiko.neumann@uni-ulm.de}
}

\markboth{August 2023}
{Felix Ott, Nisha L. Raichur \MakeLowercase{\textit{(et al.)}:
Benchmarking Visual-Inertial Deep Multimodal Fusion for Relative Pose Regression and Odometry-aided Absolute Pose Regression}}

\maketitle

\begin{abstract}
Visual-inertial localization is a key problem in computer vision and robotics applications such as virtual reality, self-driving cars, and aerial vehicles. The goal is to estimate an accurate pose of an object when either the environment or the dynamics are known. Absolute pose regression (APR) techniques directly regress the absolute pose from an image input in a known scene using convolutional and spatio-temporal networks. Odometry methods perform relative pose regression (RPR) that predicts the relative pose from a known object dynamic (visual or inertial inputs). The localization task can be improved by retrieving information from both data sources for a cross-modal setup, which is a challenging problem due to contradictory tasks. In this work, we conduct a benchmark to evaluate deep multimodal fusion based on pose graph optimization and attention networks. Auxiliary and Bayesian learning are utilized for the APR task. We show accuracy improvements for the APR-RPR task and for the RPR-RPR task for aerial vehicles and hand-held devices. We conduct experiments on the EuRoC MAV and PennCOSYVIO datasets and record and evaluate a novel industry dataset.\footnote{Datasets: \url{https://gitlab.cc-asp.fraunhofer.de/ottf/industry_datasets}}
\end{abstract}


\section{Introduction}
\label{chap_introduction}

\IEEEPARstart{L}{ocalization} is important for intelligent systems such as virtual and mixed reality, increasingly deployed in areas of tourism, education, and entertainment \cite{han, bower, stapleton}. Accurately localizing objects is key to many path planning applications to determine future movements \cite{williams, mur2, butyrev2019} of mobile objects, e.g., robots or micro aerial vehicles (MAVs) \cite{yan, yol}. This allows for monitoring and optimizing workflows as well as tracking goods for automated inventory management in real-time. A prerequisite for success is a highly accurate pose recognition (i.e., position and orientation) of the object. Environments in which such objects are typically used include large warehouses, factory buildings, and shopping centers. Localization solutions often use a combination of \mbox{LiDAR-,} \mbox{radio-,} or radar-based systems \cite{marker, marker2}, which, however, either require additional infrastructure or are costly in their operation. An alternative approach is an optical pose estimation. The accuracy of the pose estimation depends to a large extent on suitable invariance properties of the available features such that these features can be reliably detected \cite{loeffler}. For image-based localization, additional contextual information such as 3D models of the scene or pre-recorded landmark databases can be used when the environment is known. This potentially improves the pose accuracy but also increases the system's complexity. State-of-the-art mobile setups often use cheap sensors such as an inertial measurement unit (IMU) \cite{mohamed}, from which different motion dynamics (such as the slow movement of a robot or fast walking and rotation of a human) can be learned in advance. The goal of approaches using both sensors simultaneously is to utilize the advantages of both image and inertial data to improve the self-localization.

\begin{figure*}[t!]
	\centering
	\begin{minipage}[t]{0.195\linewidth}
        \centering
    	\includegraphics[trim=1 2 1 2, clip, width=1.0\linewidth]{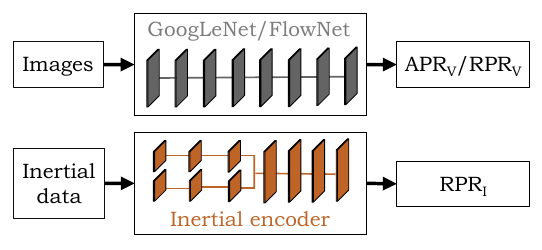}
    	\subcaption{Baseline models ($\text{APR}_\text{V}$, $\text{RPR}_\text{V}$, $\text{RPR}_\text{I}$).}
    	\label{image_overview_method1}
    \end{minipage}
    \hfill
	\begin{minipage}[t]{0.195\linewidth}
        \centering
    	\includegraphics[trim=18 12 18 8, clip, width=1.0\linewidth]{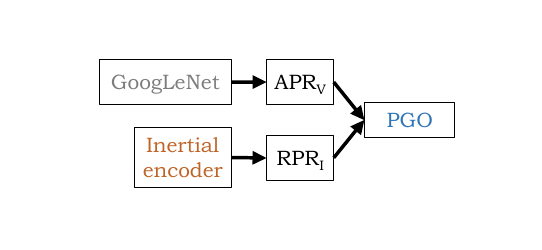}
    	\subcaption{$\text{APR}_\text{V}$-$\text{RPR}_\text{I}$, pose graph optimization \cite{brahmbhatt}.}
    	\label{image_overview_method2}
    \end{minipage}
    \hfill
	\begin{minipage}[t]{0.195\linewidth}
        \centering
    	\includegraphics[trim=18 12 18 8, clip, width=1.0\linewidth]{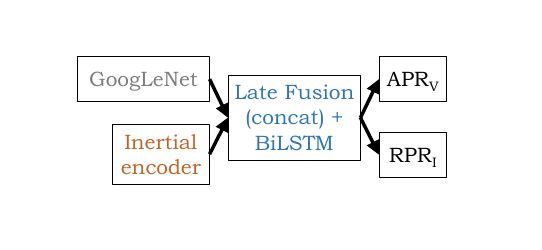}
    	\subcaption{$\text{APR}_\text{V}$-$\text{RPR}_\text{I}$, late fusion (concatenation).}
    	\label{image_overview_method3}
    \end{minipage}
    \hfill
	\begin{minipage}[t]{0.195\linewidth}
        \centering
    	\includegraphics[trim=17 12 18 8, clip, width=1.0\linewidth]{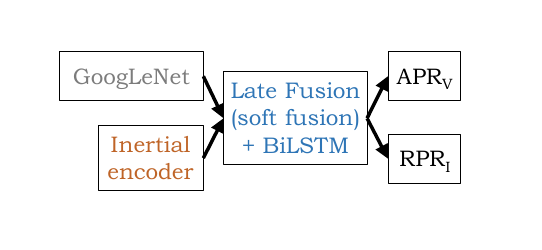}
    	\subcaption{$\text{APR}_\text{V}$-$\text{RPR}_\text{I}$, late fusion (soft fusion) \cite{chen}.}
    	\label{image_overview_method4}
    \end{minipage}
    \hfill
	\begin{minipage}[t]{0.195\linewidth}
        \centering
    	\includegraphics[trim=7 6 7 2, clip, width=1.0\linewidth]{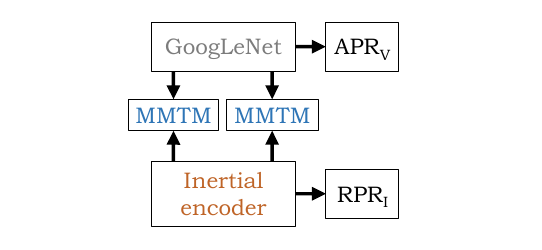}
    	\subcaption{$\text{APR}_\text{V}$-$\text{RPR}_\text{I}$, intermediate fusion (MMTM) \cite{joze}.}
    	\label{image_overview_method5}
    \end{minipage}
    
	\begin{minipage}[t]{0.195\linewidth}
        \centering
    	\includegraphics[trim=1 2 1 2, clip, width=1.0\linewidth]{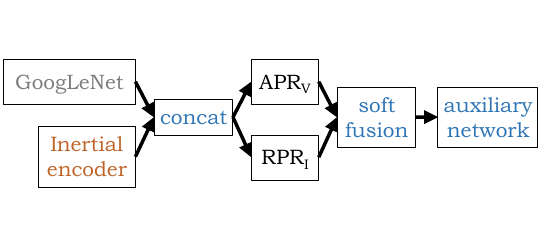}
    	\subcaption{$\text{APR}_\text{V}$-$\text{RPR}_\text{I}$, auxiliary network \cite{navon_aux}.}
    	\label{image_overview_method6}
    \end{minipage}
    \hfill
	\begin{minipage}[t]{0.195\linewidth}
        \centering
    	\includegraphics[trim=16 12 16 8, clip, width=1.0\linewidth]{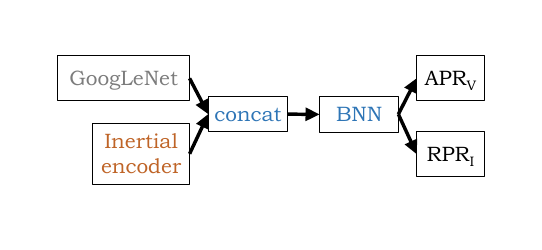}
    	\subcaption{$\text{APR}_\text{V}$-$\text{RPR}_\text{I}$, Bayesian neural network (BNN) \cite{kendall_uncertainty}.}
    	\label{image_overview_method7}
    \end{minipage}
    \hfill
	\begin{minipage}[t]{0.195\linewidth}
        \centering
    	\includegraphics[trim=18 12 18 8, clip, width=1.0\linewidth]{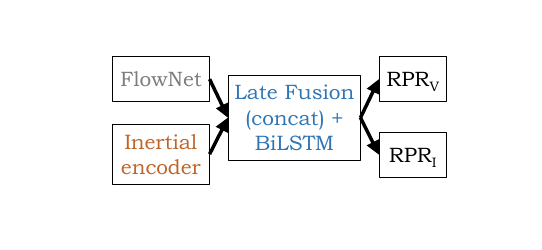}
    	\subcaption{$\text{RPR}_\text{V}$-$\text{RPR}_\text{I}$, late fusion (concatenation).}
    	\label{image_overview_method8}
    \end{minipage}
    \hfill
	\begin{minipage}[t]{0.195\linewidth}
        \centering
    	\includegraphics[trim=18 12 18 8, clip, width=1.0\linewidth]{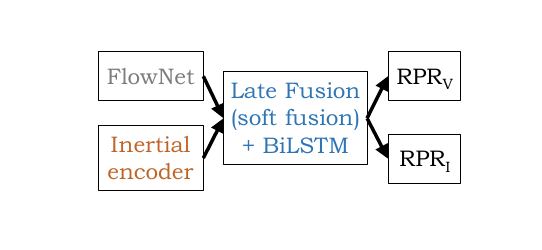}
    	\subcaption{$\text{RPR}_\text{V}$-$\text{RPR}_\text{I}$, late fusion (soft fusion) \cite{chen}.}
    	\label{image_overview_method9}
    \end{minipage}
    \hfill
	\begin{minipage}[t]{0.195\linewidth}
        \centering
    	\includegraphics[trim=7 6 7 2, clip, width=1.0\linewidth]{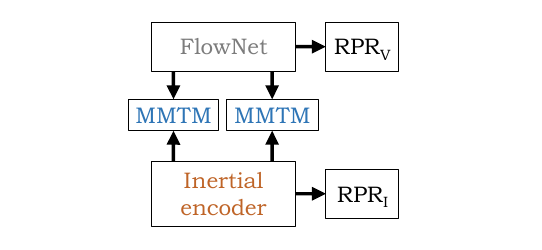}
    	\subcaption{$\text{RPR}_\text{V}$-$\text{RPR}_\text{I}$, intermediate fusion (MMTM) \cite{joze}.}
    	\label{image_overview_method10}
    \end{minipage}
    \caption{Structures of the benchmark methods for visual-inertial fusion for absolute (APR) and relative (RPR) pose regression. The baselines $\text{APR}_\text{V}$ (PoseNet~\cite{kendall}), $\text{RPR}_\text{V}$ (FlowNet~\cite{dosovitskiy}), and $\text{RPR}_\text{I}$ (IMUNet~\cite{silva}) models are fused for a common representation.}
    \label{image_overview_method}
\end{figure*}

Absolute pose regression (APR) techniques directly regress the absolute six degree-of-freedom (6DoF) pose from images ($\text{APR}_{\text{V}}$) and have become popular in recent years. These techniques are based on convolutional neural networks (CNNs) \cite{kendall, sattler_apr, camnet_ding, melekhov, kendall_cipolla, laskar, naseer, radwan, Liwei} in combination with recurrent neural networks (RNNs) \cite{walch, ott, contextualnet, seifi, clark_vidloc}. However, they do not achieve the same level of pose accuracy as 3D structure-based methods \cite{sattler_apr}. On the other hand, visual odometry (VO) techniques predict the 6DoF relative pose between image pairs of consecutive time steps. Recently, end-to-end approaches utilize CNNs in combination with RNNs for relative pose regression ($\text{RPR}_{\text{V}}$) \cite{clark, lin, constante, pcnn, ctcnet, distancenet, mansur, muller_savakis, deepvo_vo, zhao}. Another approach is inertial odometry (IO), which estimates the 6DoF relative pose from IMUs of consecutive time steps. Classical (non ML-) approaches are \cite{inertial_navigation, el, hu_imu}. In the context of inertial RPR ($\text{RPR}_{\text{I}}$), recent techniques predict the relative pose with CNNs or RNNs \cite{clark, Lu, silva}.

Odometry techniques typically suffer from accumulated errors and high drifting errors. For IO systems, the method continually integrates acceleration and angular velocities with respect to time to calculate the pose changes \cite{el}. Measurement errors -- even if small individually -- accumulate over time and lead to long-term drifts, i.e., due to temperature changes or loosely placed sensors \cite{li_ou_wei}. The drifting error of VO techniques arises from fast movement changes and image blur that is handled by loop closure \cite{campos_elvira}. Although VO has made remarkable progress over the last decade, it still suffers greatly from scaling errors \cite{F6, kasyanov, klein, F8, mur2, F15}. While $\text{APR}_{\text{V}}$ is highly accurate by relying on distinct observed features in the environment (e.g., texture-rich scenes with perfect illumination), its accuracy is largely degraded by appearance changes caused by intermittent occlusions such as moving objects, photometric calibration, low-light conditions, and illumination changes \cite{yang_wang}. Recent techniques tackle this problem with a fusion of camera and IMU sensors \cite{clark, lin, ott, chen, radwan}.

As both sensor types have different advantages in different situations, visual and inertial data can be used simultaneously to achieve a highly accurate pose in long-term use. The goal of fusion approaches is to reduce the drifting error of odometry techniques by utilizing the absolute pose and to reduce the error of APR in texture-less environments utilizing the relative pose. Classical fusion methods of visual-inertial (VI) systems are based on Bayesian filtering \cite{sibley, mourikis, bloesch} or on optimization-based methods \cite{sibley}. Traditionally, these methods rely on 3D maps and local features \cite{li, lim, sattler_activesearch}. However, naively using all the features before fusion will lead to unreliable state estimation, incorrect feature extraction, or a matching that cripples the entire system \cite{zeisl}. Hybrid methods combine geometric and ML approaches to predict the 3D position of each pixel in world coordinates \cite{shotton_scene, brachmann_dsac}. With more computing power, VI SLAM partly resolves the scale ambiguity to provide motion cues without visual features \cite{F6, F144, kasyanov} and to make tracking more robust \cite{F18}. Multiple works combine global localization in a scene with VO or IO \cite{lynen, mur2, castle, engel, mur_vislam, nuetzi}.

For multimodal learning, several streams are constructed to optimally perform individual tasks at different levels -- i.e., early, intermediate, and late fusion. Although, studies suggest the use of intermediate fusion \cite{karpathy, owens}, late fusion is still the predominant method due to practical reasons. In the field of VI-based learning, intermediate features of the encoders have unaligned spatial dimensions, which make intermediate fusion more challenging \cite{clark, lin, ott, chen, radwan}. Commonly, 1D feature vectors from each unimodal stream are concatenated and an attention mechanism chooses the best expert for each input signal \cite{chen, joze}, or dense networks are used for hierarchical joint feature learning \cite{hu}. With multi-task learning (MTL), a model learns multiple tasks simultaneously \cite{chen_vijay, liu, standley} with a shared representation that contains the mutual concepts between multiple related tasks. In contrast, auxiliary learning models can be trained on the main task of interest with multiple auxiliary tasks \cite{navon_aux, liebel, valada}. This adds an inductive bias that pushes models to capture meaningful representations and improves generalization. Training APR and RPR networks can be interpreted as MTL with equal tasks, while -- in the context of auxiliary learning -- APR is the main task and RPR is the auxiliary task. A different field is uncertainty quantification. Estimating the uncertainty in position estimation provides significant insight into the model reliability. One possibility to explain models better is to estimate their aleatoric and epistemic uncertainty \cite{kendall_uncertainty, gal}. Bayesian methods show robustness to noisy data and provide a practical framework for understanding uncertainty in models \cite{smith, kaess, gposenet, brachmann_michel}. For APR and RPR fusion, the model can learn to rely on the absolute or relative pose prediction dependent on the quantified aleatoric uncertainty.

\textbf{Contributions.} In this work, our main objective is to evaluate a wide range of different, fundamental fusion techniques (see Figure~\ref{image_overview_method}), that proved to be effective in different fields, for the VI pose regression problems ($\text{APR}_\text{V}$-$\text{RPR}_\text{I}$ and $\text{RPR}_\text{V}$-$\text{RPR}_\text{I}$). The issue at hand involves the global and relative pose in order to optimize the global pose, and mitigating the effects of environmental factors on the fusion techniques. (1) We provide an overview of $\text{APR}_\text{V}$, $\text{RPR}_\text{V}$, and $\text{RPR}_\text{I}$ methods, and use PoseNet~\cite{kendall}, FlowNet~\cite{dosovitskiy}, and IMUNet~\cite{silva} as baseline models. Indeed, there are more advanced models that yield a lower localization error in the context of APR and RPR. Instead, we benchmark different fusion techniques and highlight their influence on the pose regression tasks. (2) We apply pose graph optimization (PGO) \cite{matthew} for absolute pose refinement (see MapNet~\cite{brahmbhatt}), and absolute and relative pose fusion. (3) We evaluate attention-based methods for late fusion (concatenation and soft fusion \cite{chen}) and intermediate fusion, i.e., multimodal transfer module (MMTM) \cite{joze}, for a cross-modal feature representation. (4) We utilize non-linear and convolutional auxiliary learning \cite{navon_aux} and quantify the aleatoric uncertainty using Bayesian networds \cite{kendall_uncertainty} to improve the loss of the main $\text{APR}_\text{V}$ task. (5) We record a large indoor industrial dataset and benchmark the EuRoC MAV~\cite{burri}, the PennCOSYVIO~\cite{pfrommer}, and our IndustryVI datasets. To further enhance the results, advanced techniques may serve as black box models in place of the baseline models. Instead, our contribution is to provide insights into the role of environmental changes and motion dynamics on the localization task as well as the robustness of the fusion techniques by exploring results on these three different datasets.

The remainder of our paper is organized as follows. We first discuss related work in Section~\ref{chap_related_work}. Section~\ref{chap_method} introduces our framework and methods. We present a novel dataset in Section~\ref{chap_datasets}, before showing experimental results in Section~\ref{chap_evaluation}.

\section{Related Work}
\label{chap_related_work}

\noindent We address methods for VI self-localization in Section~\ref{chap_rw_methods}, focus on multimodal fusion techniques in Section~\ref{chap_rw_fusion}, and briefly summarize uncertainty estimation techniques in Section~\ref{chap_rw_uncert}. In Section~\ref{chap_rw_datasets}, we give an overview of datasets.

\subsection{Methods for VI Self-Localization}
\label{chap_rw_methods}

\noindent For self-localization, we separate between odometry methods, and learning-based APR and RPR methods. Odometry methods can be partitioned into VO, IO, and VI odometry, each separated into classical and regression-based techniques. As this paper provides a benchmark for APR and RPR fusion techniques, in this section, we briefly summarize methods that purely rely on SLAM, odometry, RPR, or APR (that build our baseline models). We focus on the combination of the classical VO and IO methods, the combination of $\text{RPR}_{\text{V}}$ and $\text{RPR}_{\text{I}}$, and the combination of $\text{APR}_{\text{V}}$ and RPR (see Section~\ref{chap_rw_fusion}).

\textbf{VO \& $\text{RPR}_{\text{V}}$.} Classical VO methods are based on a dead reckoning system using image features \cite{mansur}, or use point features between pairs of frames from stereo cameras \cite{bergen_visual_2004}. For an overview, see \cite{F14}. DeepVO~\cite{deepvo_vo} is one of the first networks that combined CNNs with RNNs (two stacked LSTMs) to model sequential dynamics and relations. While CTCNet~\cite{ctcnet} predicts relative poses from a CNN+LSTM model, DistanceNet~\cite{distancenet} uses a CNN+BiLSTM to estimate traveled distances divided into classes. Several methods exist that use the optical flow (OF) to regress the relative pose from consecutive image pairs -- such as Flowdometry~\cite{muller, muller_savakis}, ViPR~\cite{ott}, DeepVIO~\cite{han_lim}, and KFNet~\cite{zhou_luo}. These methods either use FlowNet~\cite{dosovitskiy} or FlowNet2~\cite{flownet2}. While LS-VO~\cite{constante} uses an autoencoder for OF prediction, P-CNN~\cite{pcnn} uses the Brox algorithm~\cite{brox} with a standard CNN. P-CNN+OF~\cite{zhao} combines FlowNet2 and P-CNN. DF-VO~\cite{zhan_weerasekera} proposed a method integrating epipolar geometry from CNNs (single-view depth and OF) and the Perspective-n-Point \cite{gao_hou} method. D3VO~\cite{yang_stumberg} learns the image depth, models the pixel uncertainties which improves the depth estimation, and predicts the pose with PoseNet. In this paper, we use FlowNetSimple~\cite{dosovitskiy} to predict the $\text{RPR}_{\text{V}}$ relative pose (see Figure~\ref{image_overview_method1}).

\textbf{IO \& $\text{RPR}_{\text{I}}$.} Classical IO includes pedestrian dead reckoning (often combined with GPS signals) \cite{beauregard}, is based on walk detection and step counting \cite{brajdic}, or is designed as a strapdown inertial navigation system \cite{inertial_navigation, shu}. While IONet~\cite{Lu} segments inertial data into independent windows and learns the relative pose with a CNN+RNN, VINet~\cite{clark} directly estimates features from IMU data with LSTMs. Silva et al.~\cite{silva} propose a CNN+BiLSTM model (IMUNet) and evaluate different loss functions (i.e., based on a vector in the spherical coordinate system, or based on a translation vector and a unit quaternion). As the CNN+BiLSTM model \cite{silva} yields state-of-the-art results, we use this model as $\text{RPR}_{\text{I}}$ baseline (see Figure~\ref{image_overview_method1}).

\textbf{APR.} APR methods are based on CNNs for (re-)localization such as GoogLeNet~\cite{szegedy} or ResNet~\cite{resnet}. Methods as (Dense) PoseNet~\cite{kendall}, BranchNet~\cite{naseer}, Hourglass~\cite{melekhov}, Geometric PoseNet~\cite{kendall_cipolla} and Bayesian PoseNet~\cite{kendall_modelling} are partially insensitive to occlusions, light changes, and motion blur. As PoseNet~\cite{kendall} evolved as a simple and effective APR technique, we use PoseNet as $\text{APR}_{\text{V}}$ baseline method (see Figure~\ref{image_overview_method1}). \cite{Liwei} dealt with the coupling of orientation and translation by splitting the network into two branches. LSTMs~\cite{hochreiter} and BiLSTM~\cite{graves_liwicki} were utilized to extract the temporal context, e.g., PoseNet+LSTM~\cite{walch}, ViPR~\cite{ott}, ContextualNet~\cite{contextualnet} and Seifi et al.~\cite{seifi} use LSTMs, while VidLoc~\cite{clark_vidloc} uses BiLSTMs. RelocNet~\cite{relocnet} learns metrics continuously from global image features through a camera frustum overlap loss. CamNet~\cite{camnet_ding} is a coarse-to-fine retrieval-based model that includes relative pose regression to get close to the best database entry that contains extracted features of images. NNet~\cite{laskar} queries a database for similar images to predict the relative pose between images and RANSAC solves the triangulation to provide a position. The CNN by \cite{brachmann} densely regresses so-called scene coordinates, defining the correspondence between the input image and the 3D scene space. Sattler et al.~\cite{sattler_apr} showed that pose regression is more closely related to pose approximation via image retrieval than to accurate pose estimation via 3D structure by predicting failure cases. \cite{brachmann_humenberger} showed that learning-based scene coordinate regression outperforms classical feature-based methods. MapNet~\cite{brahmbhatt} learns a map representation by geometric constraints that are formulated as loss terms. \cite{huang_xu} add a prior guided dropout module before PoseNet with spatial and channel attention modules to guide CNNs to ignore foreground objects. \cite{piasco} inferred a depth map from a CNN encoder and predicted the pose from the most similar image with nearest neighbor indexing. AtLoc~\cite{wang_chen} consists of a visual encoder that extracts features and an attention module that computed the attention and re-weights features.

\begin{table*}
\begin{center}
\setlength{\tabcolsep}{1.4pt}
    \caption{Overview of visual and inertial datasets and its sensors and ground truth properties.}
    \label{table_datasets}
    \footnotesize \begin{tabular}{ p{3.9cm} | p{0.5cm} | p{0.5cm} | p{0.5cm} | p{0.5cm} | p{0.5cm} | p{0.5cm} }
    \multicolumn{1}{c|}{\textbf{Dataset}} & \multicolumn{1}{c|}{\textbf{Ref.}} & \multicolumn{1}{c|}{\textbf{Year}} & \multicolumn{1}{c|}{\textbf{Environment}} & \multicolumn{1}{c|}{\textbf{Carrier}} & \multicolumn{1}{c|}{\textbf{Sensors}} & \multicolumn{1}{c}{\textbf{Ground truth}} \\ \hline
    \multicolumn{1}{l|}{TUM RGB-D} & \multicolumn{1}{r|}{\cite{sturm}} & \multicolumn{1}{l|}{2012} & \multicolumn{1}{l|}{Indoors} & \multicolumn{1}{l|}{Pioneer Robot} & \multicolumn{1}{l|}{Cam: 2 stereo RGB-D (30\,Hz), IMU: 1 acc. (500\,Hz)} & \multicolumn{1}{l}{Motion capture (300\,Hz)} \\
    \multicolumn{1}{l|}{KITTI} & \multicolumn{1}{r|}{\cite{geiger_kitti}} & \multicolumn{1}{l|}{2012} & \multicolumn{1}{l|}{Outdoors} & \multicolumn{1}{l|}{Car} & \multicolumn{1}{l|}{Cam: 1 stereo RGB/gray, IMU: 1 acc./gyr., laser} & \multicolumn{1}{l}{INS/GNSS (10\,Hz)} \\
    \multicolumn{1}{l|}{Microsoft 7-Scenes} & \multicolumn{1}{r|}{\cite{shotton_scene}} & \multicolumn{1}{l|}{2013} & \multicolumn{1}{l|}{Indoors} & \multicolumn{1}{l|}{Handheld} & \multicolumn{1}{l|}{Cam: RGB-D} & \multicolumn{1}{l}{KinectFusion} \\
    \multicolumn{1}{l|}{M\'{a}laga Urban} & \multicolumn{1}{r|}{\cite{blanco}} & \multicolumn{1}{l|}{2014} & \multicolumn{1}{l|}{Outdoors} & \multicolumn{1}{l|}{Car} & \multicolumn{1}{l|}{Cam: 1 stereo RGB, IMU: 1 acc., 2 gyr.} & \multicolumn{1}{l}{GPS (1\,Hz)} \\
    \multicolumn{1}{l|}{Cambridge Landmarks} & \multicolumn{1}{r|}{\cite{kendall}} & \multicolumn{1}{l|}{2015} & \multicolumn{1}{l|}{Outdoors} & \multicolumn{1}{l|}{Handheld, urban} & \multicolumn{1}{l|}{Cam: Google LG Nexus 5 smartphone (2\,Hz)} & \multicolumn{1}{l}{From SfM} \\
    \multicolumn{1}{l|}{UMich NCLT} & \multicolumn{1}{r|}{\cite{carlevaris}} & \multicolumn{1}{l|}{2016} & \multicolumn{1}{l|}{In-/outdoors} & \multicolumn{1}{l|}{Segway} & \multicolumn{1}{l|}{Cam: 6 omnid. RGB (5\,Hz), IMU: 1 acc., 2 gyr., laser} & \multicolumn{1}{l}{GPS/IMU/laser} \\
    \multicolumn{1}{l|}{EuRoC MAV} & \multicolumn{1}{r|}{\cite{burri}} & \multicolumn{1}{l|}{2016} & \multicolumn{1}{l|}{Indoors} & \multicolumn{1}{l|}{MAV hexacopter} & \multicolumn{1}{l|}{Cam: 1 stereo gray (20\,Hz), IMU: 1 acc./gyr. (200\,Hz)} & \multicolumn{1}{l}{MoCap Laser (20\,Hz)} \\
    \multicolumn{1}{l|}{12-Scenes} & \multicolumn{1}{r|}{\cite{12scenes}} & \multicolumn{1}{l|}{2016} & \multicolumn{1}{l|}{Indoors} & \multicolumn{1}{l|}{Handheld} & \multicolumn{1}{l|}{Cam: 1 RGB, 1 RGB-D} & \multicolumn{1}{l}{VoxelHashing} \\
    \multicolumn{1}{l|}{Oxford RobotCar} & \multicolumn{1}{r|}{\cite{robotcar}} & \multicolumn{1}{l|}{2016} & \multicolumn{1}{l|}{Outdoors} & \multicolumn{1}{l|}{RobotCar} & \multicolumn{1}{l|}{Cam: Bumblebee XB3} & \multicolumn{1}{l}{GPS, IMU (5\,Hz)} \\
    \multicolumn{1}{l|}{PennCOSYVIO} & \multicolumn{1}{r|}{\cite{pfrommer}} & \multicolumn{1}{l|}{2016} & \multicolumn{1}{l|}{In-/outdoors} & \multicolumn{1}{l|}{Handheld} & \multicolumn{1}{l|}{Cam: 4 RGB, 1 stereo, 1 fisheye, IMU: 1 acc./gyr.} & \multicolumn{1}{l}{Visual tags (30\,Hz)} \\
    \multicolumn{1}{l|}{Zurich Urban MAV} & \multicolumn{1}{r|}{\cite{majdik}} & \multicolumn{1}{l|}{2017} & \multicolumn{1}{l|}{Outdoors} & \multicolumn{1}{l|}{MAV} & \multicolumn{1}{l|}{Cam: 1 RGB (30\,Hz), IMU: 1 acc./gyr. (10\,Hz)} & \multicolumn{1}{l}{Pix4D visual pose} \\
    \multicolumn{1}{l|}{Aalto University} & \multicolumn{1}{r|}{\cite{laskar}} & \multicolumn{1}{l|}{2017} & \multicolumn{1}{l|}{Indoors} & \multicolumn{1}{l|}{Handheld} & \multicolumn{1}{l|}{Cam: iPhone 6S smartphone} & \multicolumn{1}{l}{Google Project Tango’s} \\
    \multicolumn{1}{l|}{TUM-LSI} & \multicolumn{1}{r|}{\cite{walch}} & \multicolumn{1}{l|}{2017} & \multicolumn{1}{l|}{Indoors} & \multicolumn{1}{l|}{NavVis sytem} & \multicolumn{1}{l|}{Cam: 6 Panasonic wide-angle from NavVis M3 system} & \multicolumn{1}{l}{Hokuyo laser (SLAM)} \\
    \multicolumn{1}{l|}{Warehouse} & \multicolumn{1}{r|}{\cite{loeffler}} & \multicolumn{1}{l|}{2018} & \multicolumn{1}{l|}{Indoors} & \multicolumn{1}{l|}{Pos. system} & \multicolumn{1}{l|}{Cam: 8 $60^{\circ}$ RGB} & \multicolumn{1}{l}{Nikon iGPS} \\
    \multicolumn{1}{l|}{DeepLoc} & \multicolumn{1}{r|}{\cite{radwan}} & \multicolumn{1}{l|}{2018} & \multicolumn{1}{l|}{Outdoors} & \multicolumn{1}{l|}{Robot platform} & \multicolumn{1}{l|}{Cam: RGB-D ZED stereo (20\,Hz), IMU: XSens, LiDARs} & \multicolumn{1}{l}{GPS} \\
    \multicolumn{1}{l|}{TUM VI} & \multicolumn{1}{r|}{\cite{schubert}} & \multicolumn{1}{l|}{2018} & \multicolumn{1}{l|}{In-/outdoors} & \multicolumn{1}{l|}{Handheld} & \multicolumn{1}{l|}{Cam: 1 stereo gray (20\,Hz), IMU: 1 acc./gyr. (200\,Hz)} & \multicolumn{1}{l}{Partial motion} \\
    \multicolumn{1}{l|}{CMU Seasons} & \multicolumn{1}{r|}{\cite{sattler_cmu}} & \multicolumn{1}{l|}{2018} & \multicolumn{1}{l|}{Outdoors} & \multicolumn{1}{l|}{Car, suburban} & \multicolumn{1}{l|}{Cam: $45^{\circ}$ forward/left and forward/right} & \multicolumn{1}{l}{SIFT, BA} \\
    \multicolumn{1}{l|}{Industry} & \multicolumn{1}{r|}{\cite{ott}} & \multicolumn{1}{l|}{2020} & \multicolumn{1}{l|}{Indoors} & \multicolumn{1}{l|}{Forklift, pos. sys.} & \multicolumn{1}{l|}{Cam: 4 $170^{\circ}$ RGB on fork., 3 $170^{\circ}$ RGB on pos. sys.} & \multicolumn{1}{l}{Qualisys (140\,Hz)} \\
    \multicolumn{1}{l|}{UMA-VI} & \multicolumn{1}{r|}{\cite{zuniga_noel}} & \multicolumn{1}{l|}{2020} & \multicolumn{1}{l|}{In-/outdoors} & \multicolumn{1}{l|}{Handheld} & \multicolumn{1}{l|}{Cam: RGB (12.5\,Hz), gray (25\,Hz), IMU: acc./gyr. (250\,Hz)} & \multicolumn{1}{l}{Visual pose (partial)} \\
    \multicolumn{1}{l|}{\textbf{IndustryVI}} & \multicolumn{1}{c|}{ours} & \multicolumn{1}{l|}{2022} & \multicolumn{1}{l|}{Indoors} & \multicolumn{1}{l|}{Handheld} & \multicolumn{1}{l|}{Cam: Orbbec3D (23\,Hz), IMU: 1 acc./gyr./mag. (140\,Hz)} & \multicolumn{1}{l}{Qualisys (140\,Hz)} \\
    \end{tabular}
    \vspace{-0.25cm}
\end{center}
\end{table*}

\subsection{Multimodal Fusion for Self-Localization}
\label{chap_rw_fusion}

\noindent \textbf{Classical VI Odometry.} Classical methods for sensor fusion are the (extended) Kalman filter (KF) \cite{magnusson, goel} or pose graphs \cite{das}. \cite{hu_imu} use a trifocal tensor geometry between three images without estimating the 3D position of feature points (without reconstructing the environment). The poses are refined with a multi-state KF in combination with a RANSAC algorithm. They transform the camera frame w.r.t.~the IMU frame. VINS-Mono~\cite{qin} is a nonlinear optimization-based method for VI odometry by fusing pre-integrated IMU measurements and feature observations that merge maps by PGO~\cite{matthew}.

\textbf{RPR-based VI Odometry.} DeepVIO~\cite{han_lim} learns the OF from consecutive images and the relative pose from an inertial-based network, fused with fully connected layers for VI ego-motion. This is supported by a network with stereoscopic image inputs. SelfVIO~\cite{almalioglu} combined VO and IO networks with an adaptive fusion model that concatenates features of both networks, selects features and predicts the absolute pose with an LSTM. Similarly, the selective sensor fusion (SSF) approach by \cite{chen} extracts features from image and IMU data, uses a soft or hard (based on Gumbel softmax) fusion approach to select features and regresses the pose with an LSTM. We use the soft fusion approach of SSF to combine $\text{APR}_\text{V}$-$\text{RPR}_\text{I}$ (see Figure~\ref{image_overview_method4}) and $\text{RPR}_\text{V}$-$\text{RPR}_\text{I}$ (see Figure~\ref{image_overview_method9}).

\textbf{APR \& RPR Fusion.} Learning-based methods are based on APR or RPR networks. VI-DSO~\cite{stumberg} jointly estimates camera poses and sparse scene geometry by minimizing the photometric and the IMU measurement error in a combined energy functional. The loss formulation of LM-Reloc~\cite{stumberg_wenzel} is inspired by the Levenberg-Marquardt algorithm, such that the learned features significantly improve the robustness of direct image alignment. Additionally, their network performs RPR to bootstrap the direct image alignment. VINet~\cite{clark} incorporates relative features from an LSTMs inertial encoder with absolute features from a visual encoder by concatenation. ViPR~\cite{ott} concatenates relative poses (from OF) and absolute poses to refine the absolute poses with an LSTM network. MapNet+PGO~\cite{brahmbhatt} uses PGO~\cite{matthew} to refine predicted poses from absolute and relative pose predictions (we use PGO in Figure~\ref{image_overview_method2}). VLocNet~\cite{valada} estimates a global pose and combines it with VO. To further improve the (re-)localization accuracy, VLocNet++~\cite{radwan} uses features from a semantic segmentation.\footnote{Public code is not available for VLocNet. Re-implementations lead to subpar performance \cite{github_decayale}, but close to MapNet~\cite{brahmbhatt}.} RCNN~\cite{lin} fuses relative and global networks -- while the relative sub-networks smooth the VO trajectory, the global sub-networks avoid the drift problem. Their cross transformation constraints represent the temporal geometric consistency of consecutive frames. We use concatenation as a baseline fusion technique (see Figure~\ref{image_overview_method3} and \ref{image_overview_method8}).

\subsection{Uncertainty Estimation for Multimodal Fusion}
\label{chap_rw_uncert}

\noindent KFNet~\cite{zhou_luo} extends the scene coordinate regression problem to the time domain based on KF, OF, and Bayesian learning. \cite{russel, xu_davison} show that the training can allow uncertainty predictions through a Gaussian density loss in combination with a KF. ToDayGAN~\cite{anoosheh} is a generative network that alters nighttime driving images to a more useful daytime representation captured from two trajectories of the same area in both day and night. The dropout module by \cite{huang_xu} enables the pose regressor to output multiple hypotheses from which the uncertainty of pose estimates can be quantified. CoordiNet~\cite{moreau} predicts the pose and uncertainty from a single loss function for visual relocalization that are fused with a KF to embed the scene geometry. \cite{brieger_ion_gnss} utilized MMTM modules to fuse features derived from images (ResNet) and features extracted from time-series data (TS Transformer) while assessing Monte-Carlo dropout. This approach bears similarity to our own setup, demonstrating that MMTM can be employed in diverse areas of study.

\subsection{Datasets}
\label{chap_rw_datasets}

\noindent Each application covers different characteristics that have to be represented by the dataset, i.e., properties of the environment (small or large scale, features), properties of the object (movement patterns such as direction, velocity, and acceleration), and size of the dataset. We provide a benchmark on different datasets to evaluate the performance of models in certain scenarios. A dataset can be classified with the following properties: Indoor or outdoor environment, small- or large-scale environment, a high accuracy ($<1cm$) of the ground truth trajectory, availability of datasets, same spatial range of training and test trajectories, and whether the training dataset allows a generalized training.

Table~\ref{table_datasets} summarizes all VI self-localization datasets and their characteristics. As the TUM RGB-D~\cite{sturm}, Microsoft 7-Scenes~\cite{shotton_scene}, Cambridge Landmarks~\cite{kendall}, 12-Scenes~\cite{12scenes}, Aalto University~\cite{laskar}, DeepLoc~\cite{radwan}, TUM-LSI~\cite{walch}, Warehouse~\cite{loeffler} and Industry~\cite{ott} datasets contain only images, we cannot make use of these datasets for our VI multimodal setup. The KITTI~\cite{geiger, geiger_kitti}, M\'{a}laga Urban \cite{blanco}, Oxford RobotCar~\cite{robotcar}, and UMich NCLT~\cite{carlevaris} datasets contain IMU and LiDAR data, but cannot be used for APR as odometry is the main task. The Aachen day-night, CMU seasons, and RobotCar seasons datasets \cite{sattler_cmu} address viewing conditions such as weather and seasonal variations and day-night changes for visual localization. Similarly, the Zurich Urban MAV~\cite{majdik} dataset addresses only evaluations for VO and SLAM. UMA-VI~\cite{zuniga_noel} is a handheld indoor and outdoor dataset with VI data but contains only partial ground truth poses. Also, the ground truth systems of TUM VI~\cite{schubert} cover only a single room, and hence, longer trajectories do not cover ground truth poses.

The EuRoC MAV~\cite{burri} dataset contains VI data recorded with an MAV in indoor small-scale environments and is suitable for our APR-RPR benchmark. In contrast, PennCOSYVIO~\cite{pfrommer} was recorded handheld in a large-scale outdoor and indoor environment with VI sensors. As this dataset does not allow evaluations across various movement patterns, we record the IndustryVI dataset in a challenging large-scale indoor industrial environment with ground truth accuracies below $1mm$. We let two persons walk with a handheld device with various movement patterns. Having two different persons, in particular, allows an evaluation of the generalizability between different walking styles. For more information, see Section~\ref{chap_datasets}. Hence, we use the EuRoC MAV, the PennCOSYVIO, and our IndustryVI datasets to benchmark different fusion models on indoor and outdoor applications.
\section{Methodology}
\label{chap_method}

\noindent In this section, we present different deep multimodal fusion techniques for odometry-aided APR and VI odometry. Section~\ref{section_baseline_models} presents the baseline models for $\text{APR}_{\text{V}}$, $\text{RPR}_{\text{I}}$, and $\text{RPR}_{\text{V}}$. We describe PGO for absolute pose refinement in Section~\ref{section_pgo}. Section~\ref{section_attention_fusion} proposes attention-based fusion methods. We use auxiliary learning for $\text{APR}_{\text{V}}$-$\text{RPR}_{\text{I}}$ fusion in Section~\ref{section_auxiliary_learning}, and use Bayesian neural networks (BNNs) for aleatoric uncertainty estimation in Section~\ref{setcion_bayesian_learning}. We describe fusion techniques for $\text{RPR}_{\text{V}}$ in Section~\ref{setcion_flownet_odometry}. Table~\ref{table_notations} summarizes the notations.

\begin{table}
\begin{center}
\setlength{\tabcolsep}{2.7pt}
    \caption{Key parameters and their descriptions.}
    \label{table_notations}
    \footnotesize \begin{tabular}{ p{3.1cm} | p{0.5cm}  }
    \multicolumn{1}{c|}{\textbf{Parameter}} & \multicolumn{1}{c}{\textbf{Description}} \\ \hline
    \multicolumn{1}{l|}{$\text{APR}_{\text{V}}$} & \multicolumn{1}{l}{Absolute pose regression based on visual input} \\
    \multicolumn{1}{l|}{$\text{RPR}_{\text{V}}$} & \multicolumn{1}{l}{Relative pose regression based on visual input} \\
    \multicolumn{1}{l|}{$\text{RPR}_{\text{I}}$} & \multicolumn{1}{l}{Relative pose regression based on inertial input} \\
    \multicolumn{1}{l|}{$\mathbf{x} = [\mathbf{p}, \mathbf{q}]$} & \multicolumn{1}{l}{Absolute pose} \\
    \multicolumn{1}{l|}{$\Delta \mathbf{x} = [\Delta \mathbf{p}, \Delta \mathbf{q}]$} & \multicolumn{1}{l}{Relative pose} \\
    \multicolumn{1}{l|}{$\mathbf{p} \in \mathbb{R}^3$} & \multicolumn{1}{l}{Absolute 3D position in Euclidean space} \\
    \multicolumn{1}{l|}{$\mathbf{q} \in \mathbb{R}^4$} & \multicolumn{1}{l}{Absolute orientation as quaternion} \\
    \multicolumn{1}{l|}{$\Delta \mathbf{p} \in \mathbb{R}^3$} & \multicolumn{1}{l}{Relative position (translation)} \\
    \multicolumn{1}{l|}{$\Delta \mathbf{q} \in \mathbb{R}^4$} & \multicolumn{1}{l}{Relative orientation (rotation)} \\
    \multicolumn{1}{l|}{$\mathbf{v}_{ij}$} & \multicolumn{1}{l}{Relative pose between predicted poses $\mathbf{x}_{i}$ and $\mathbf{x}_{j}$} \\
    \multicolumn{1}{l|}{$\mathcal{L}$} & \multicolumn{1}{l}{Loss function} \\
    \multicolumn{1}{l|}{$E_{c}$} & \multicolumn{1}{l}{Constraint energy} \\
    \multicolumn{1}{l|}{$f(c)$} & \multicolumn{1}{l}{Prediction function} \\
    \multicolumn{1}{l|}{$\mathbf{S}_{c}$} & \multicolumn{1}{l}{Distance matrix} \\
    \multicolumn{1}{l|}{$\mathbf{J}$} & \multicolumn{1}{l}{Jacobian} \\
    \multicolumn{1}{l|}{$\mathbf{r}$} & \multicolumn{1}{l}{Residual} \\
    \multicolumn{1}{l|}{$\boxplus$} & \multicolumn{1}{l}{Manifold update operations for quaternions} \\
    \multicolumn{1}{l|}{$\odot$} & \multicolumn{1}{l}{Element-wise multiplication} \\
    \multicolumn{1}{l|}{$D$} & \multicolumn{1}{l}{Dataset of size $|D|$} \\
    \multicolumn{1}{l|}{$\mathbf{a}_{I}, \mathbf{a}_{V}$} & \multicolumn{1}{l}{Inertial and visual features} \\
    \multicolumn{1}{l|}{$\alpha, \beta, \gamma$} & \multicolumn{1}{l}{Hyperparamters for loss weighting} \\
    \multicolumn{1}{l|}{$\mathcal{S}$} & \multicolumn{1}{l}{Soft fusion operator} \\
    \multicolumn{1}{l|}{$g$} & \multicolumn{1}{l}{Sigmoid function} \\
    \multicolumn{1}{l|}{$\mathbf{A}, \mathbf{B}$} & \multicolumn{1}{l}{Feature at any given level of APR or RPR} \\
    \multicolumn{1}{l|}{$Z$} & \multicolumn{1}{l}{Latent representation} \\
    \multicolumn{1}{l|}{$E_{\mathbf{A}}, E_{\mathbf{B}}$} & \multicolumn{1}{l}{Excitation signals} \\
    \multicolumn{1}{l|}{$\mathbf{W}$} & \multicolumn{1}{l}{Weights of the network} \\
    \multicolumn{1}{l|}{$s$} & \multicolumn{1}{l}{Log variance} \\
    \end{tabular}
\end{center}
\end{table}

\subsection{Baseline Models}
\label{section_baseline_models}

\noindent The baseline of our fusion models is established through the utilization of APR and RPR models. A CNN-based APR model is capable of learning to directly regress the camera pose from a single image or a set of training images in conjunction with their corresponding ground truth poses. In contrast, an RPR model performs 6DoF odometry through the utilization of either inertial data obtained from an IMU ($\text{RPR}_{\text{I}}$) or visual data obtained from a camera ($\text{RPR}_{\text{V}}$).

\begin{figure}[t!]
	\centering
    \includegraphics[trim=2 13 2 13, clip, width=1.0\linewidth]{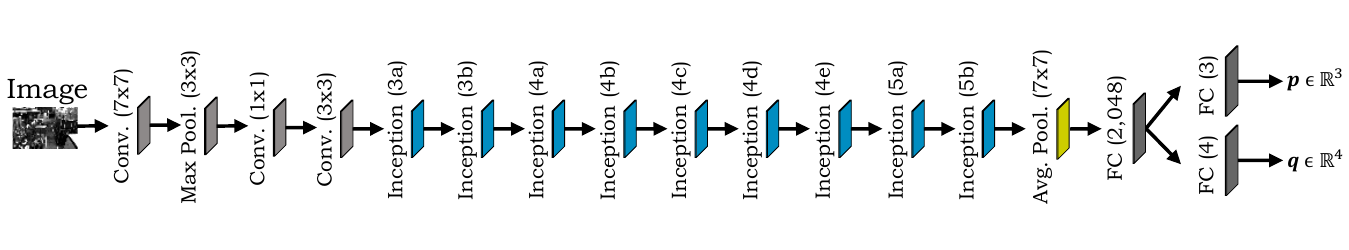}
    \caption{Structure of PoseNet~\cite{kendall} with a modified GoogLeNet~\cite{szegedy} network as the $\text{APR}_{\text{V}}$ module with input images $I$ and output position $\mathbf{p}$ and orientation $\mathbf{q}$.}
    \label{image_apr}
\end{figure}

\textbf{$\text{APR}_{\text{V}}$.} We use PoseNet~\cite{kendall} with time-distribution \cite{ott} to predict the absolute positions $\mathbf{p} \in \mathbb{R}^3$  and the absolute orientations $\mathbf{q} \in \mathbb{R}^4$  as illustrated in Figure~\ref{image_apr}. PoseNet is a CNN architecture that utilizes GoogLeNet~\cite{szegedy}, which is pre-trained on a huge classification dataset such as ImageNet~\cite{dengt}. PoseNet adds a fully connected (FC) layer of $2,048$ units on top of the last inception module to form a localization feature vector that may be trained for generalization. In addition, we replace the FC layer and softmax layers of GoogLeNet with two parallel FC layers, each with three and four units. These units are utilized to regress the pose, represented by the $\mathbf{p}$ = [$x, y, z$] coordinates of the position in Euclidean space and $\mathbf{q}$ =  [$w, p, q, r$] as a quaternion for orientation \cite{kendall_cipolla, zhou_barnes}.

\begin{figure}[t!]
	\centering
    \includegraphics[trim=0 1 0 1, clip, width=1.0\linewidth]{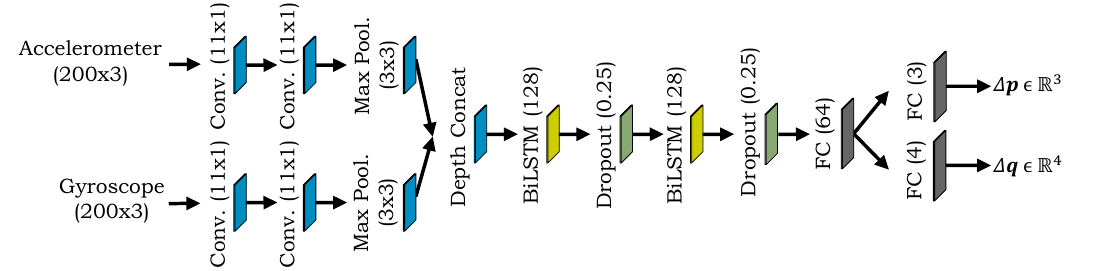}
    \caption{Structure of the $\text{RPR}_{\text{I}}$ module \cite{silva} with inertial input data (accelerometer and gyroscope data from 200 timesteps) and output translation $\Delta \mathbf{p}$ and change in orientation $\Delta \mathbf{q}$.}
    \label{image_rpr}
\end{figure}

\textbf{$\text{RPR}_{\text{I}}$.} Inspired by \cite{silva}, our $\text{RPR}_{\text{I}}$ model is designed based on a CNN combined with two BiLSTM units, as depicted in Figure~\ref{image_rpr}. This design is highly suitable for problems that require the processing of sequential data. The model comprises of 1D-convolutional layers, each with 128 features and a kernel size of 11, which separately process the gyroscope and accelerometer data. After two 1D convolutions, the feature vectors are down-sampled in size through a max-pooling layer with a size of 3. The outputs of the two separate processing streams are concatenated and fed into a BiLSTM layer, consisting of 128 hidden units. A BiLSTM is utilized to enable the past and future IMU readings to influence the regressed relative pose. To prevent overfitting, a dropout layer with a rate of 25\% is added after each BiLSTM layer. Finally, the relative pose -- $\Delta \mathbf{p} \in \mathbb{R}^3$ and $\Delta \mathbf{q} \in \mathbb{R}^4$ -- is regressed through the use of FC layers that take the output from the BiLSTM layers as input.

\textbf{$\text{RPR}_{\text{V}}$.} As the feature encoder in our model, we utilize FlowNetSimple~\cite{dosovitskiy} pre-trained on Flying Chairs \cite{dosovitskiy} dataset. This encoder predicts the relative position $\Delta \mathbf{p} \in \mathbb{R}^3$ and orientation $\Delta \mathbf{q} \in \mathbb{R}^4$ from a pair of consecutive monocular images as input. The network is comprised of 9 convolutional layers and is designed to have sufficient capacity to learn the prediction of the OF. The size of the receptive fields gradually decreases from $7 \times 7$  to $5 \times 5$, and finally $3 \times 3$. We use the features from the last convolution layer ($\text{conv6}$) as an input to a FC layer consisting of $2,048$ units, which forms a localization feature vector. Finally, we add two parallel FC layers, each containing three and four units. These units perform regression of the relative pose represented by $\Delta \mathbf{p}$ and $\Delta \mathbf{q}$.

\subsection{Pose Graph Optimization (PGO)}
\label{section_pgo}

\noindent PGO estimates smooth and globally consistent pose predictions from absolute and relative pose measurements during inference. The method is formulated as a non-convex minimization problem, which can be represented as a graph with vertices corresponding to the estimated global poses edges representing the relative measurements. The objective of PGO is to refine the predicted poses during inference such that the refined poses are close to the actual poses and ensuring agreement between the refined relative poses and the input relative poses \cite{matthew}. The algorithm utilizes the predicted absolute poses and the relative poses between them as inputs. In the following, we represent the poses
\begin{equation}
\label{eq_1}
    \mathbf{p}_{i} = (t_{x}, t_{y}, t_{z}, q_{w}, q_{p}, q_{q}, q_{r})
\end{equation}
as a vector for translation $\mathbf{t} \in \mathbb{R}^3$ and a vector for orientation represented as quaternion $\mathbf{q} \in \mathbb{R}^4$. To perform PGO for the predicted absolute poses (from the $\text{APR}_{\text{V}}$ model), we initially collect all the absolute poses in a single vector $\mathbf{z}$. We define the objective function, which is the sum of the costs of all the constraints $E_{c}$. The constraints can be either for the absolute poses or for the relatives poses (between a pair of absolute poses) for both translation and rotation. The constraint energy
\begin{equation}
\label{eq_2}
    E_{c}(\mathbf{z}) = (f_{c}(\mathbf{z}) - k_{c})^{T} \mathbf{S}_{c}(f_{c}(\mathbf{z}) - k_{c})
\end{equation}
is represented as a quadratic penalty on the difference between $f(c)$ and its desired value $k_{c}$ where $f(c)$ is a prediction function that maps the state vector $\mathbf{z}$ to the quantity relevant for constraint $c$, weighted by distance matrix $\mathbf{S}_{c}$ (the inverse of the covariance matrix). Equation~\eqref{eq_1} represents the relative position constraint $k_{c}$ by having $f_{c}(\mathbf{z})$ to be the relative position between two poses. We initially linearize $f_{c}$ around $\overline{\mathbf{z}}$, the current value of the state vector $\mathbf{z}$, using the substitution $\mathbf{z} = \overline{\mathbf{z}} + \mathbf{x}$ , where $\mathbf{x}$ represents the parameter update we solve for \cite{matthew}. We take the Cholesky decomposition of the stiffness matrices $\mathbf{S}_{c} = \mathbf{L}_{c}\mathbf{L}_{c}^{T}$. With
\begin{equation}
\label{eq_3}
    f_{c}(\mathbf{z}) \approx f_{c}(\overline{\mathbf{z}}) + \frac{\partial f}{\partial \mathbf{z}}\Bigr|_{\overline{\mathbf{z}}}\mathbf{x},
\end{equation}
we get
\begin{equation}
    \label{eq_4}
    E = \sum_{c} \bigg\| \mathbf{L}_{c}^{T} \Big( f_{c}(\overline{\mathbf{z}}) + \frac{\partial f}{\partial \mathbf{z}}\Bigr|_{\overline{z}} \mathbf{x} - \mathbf{k}_{c} \Big) \bigg\|^{2}.
\end{equation}
Let $\mathbf{J}_{c} = \mathbf{L}^{T}_{c} \frac{\partial f}{\partial \mathbf{z}}\Bigr|_{\overline{\mathbf{z}}}$ and $\mathbf{r}_{c} = \mathbf{L}_{T}^{c}(k_{c}-f_{c}(\overline{\mathbf{z}}))$. We get
\begin{equation}
\label{eq_5}
    E = \sum_{c} \Big\| \mathbf{J}_{c}\mathbf{x} - \mathbf{r}_{c} \Big\|^2 = \| \mathbf{J}\mathbf{x} - \mathbf{r} \|^2.
\end{equation}
Stacking the individual Jacobians $\mathbf{J}$ and the residuals $\mathbf{r}$, we arrive at the least squares problem
\begin{equation}
\label{eq_6}
    \Delta \mathbf{z} = \min_{\Delta \mathbf{z}} \| \mathbf{J} \Delta \mathbf{z} - \mathbf{r} \|^2
\end{equation}
to solve for $\Delta \mathbf{z}$. This can be solved by $\Delta \mathbf{z} = (\mathbf{J}^{T}\mathbf{J})^{-1}\mathbf{J}^{T}\mathbf{r}$. Finally, the predicted absolute pose state vector $\mathbf{z}$ is updated using $\mathbf{z} = \mathbf{z} \boxplus \Delta \mathbf{z}$, where $\boxplus$ represents the manifold update operations for quaternions as described in \cite{brahmbhatt}.

\textbf{PGO during Inference.} During inference, the absolute pose predictions and the relative poses  between them are used to obtain the optimal absolute poses using PGO. The algorithm runs iteratively utilizing a moving temporal window size of $T$ frames. Suppose the absolute predictions for $T$ frames are $\{\mathbf{p}_{i}\}_{i=1}^{T}$ and the relative poses between them are $\{\Delta \mathbf{p}_{ij}\}_{i=1}^{T-1}$, the optimal poses $\{\mathbf{p}^{o}\}_{i=1}^{T}$ are solved by minimizing the following cost
\begin{equation}
\label{eq_7}
    \mathcal{L}_{\text{PGO}}(\{\mathbf{p}^{o}\}_{i=1}^{T}) = \sum_{i=1}^{T} E_{c}(\mathbf{p}_{i}) + \sum_{i=1}^{T-1} E_{c} (\Delta \mathbf{p}_{ij}),
\end{equation}
where $E_{c}$ is the constraint energy from Equation~\eqref{eq_1}. We use PGO for absolute pose refinement from consecutive relative poses (see Section~\ref{section_pgo_for_apr}) and absolute pose refinement from relative poses from the RPR model (see Section~\ref{section_pgo_for_fusion}).

\subsubsection{PGO for APR}
\label{section_pgo_for_apr}

\begin{figure}[t!]
	\centering
    \includegraphics[width=0.8\linewidth]{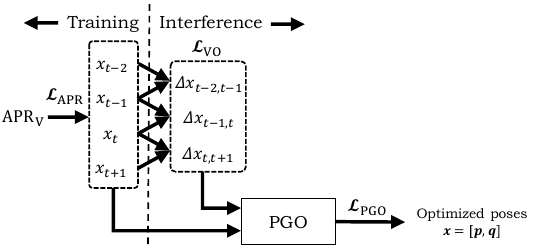}
    \caption{Overview of the $\text{APR}_{\text{V}}$ model with PGO used for absolute pose refinement using the consecutive relative poses between the predicted absolute poses.}
    \label{image_pgo_for_apr}
\end{figure}

\noindent PGO refines the predicted absolute poses such that the relative transforms between them agree with the relative camera pose between the predicted poses. To achieve this, we use a time-distributed image encoder (GoogLeNet) similar to MapNet~\cite{brahmbhatt} while using PGO during inference (see Figure~\ref{image_pgo_for_apr}). We learn to estimate the 6DoF camera pose from a tuple of images and additionally enforce constraints between pose predictions for each  image pair. While the APR encoder minimizes the absolute pose loss per image, MapNet proposes to minimize both the absolute pose loss per image and the relative pose loss between the consecutive image pair as
\begin{equation}
\label{eq_8}
    \mathcal{L}_{\text{total}} = \sum_{i=1}^{|D|} h(\mathbf{p}_{i}, \hat{\mathbf{p}}_{i}) + \sum_{i=1}^{|D|} h(\mathbf{v}_{ij}, \hat{\mathbf{v}}_{ij}),
\end{equation}
where the relative pose between predicted absolute poses $\mathbf{p}_{i}$ and $\mathbf{p}_{j}$ for image pairs $(i, j)$ is represented by $\mathbf{v}_{ij} = (t_{i}-t_{j}, q_{i}-q_{j})$. $h(\cdot)$ is a metric that measures the distance between the actual pose $\hat{\mathbf{p}}$ and the predicted camera pose $\mathbf{p}$ as defined in \cite{kendall_cipolla}. During inference, we use PGO to  fuse the predicted absolute poses $\mathbf{p}_{i}$ and the relative poses ($\mathbf{v}_{ij}$) between the consecutive image pairs to get smooth and globally consistent absolute poses $\mathbf{p}_{i}^{o}$.

\subsubsection{$\text{APR}_{\text{V}}$-$\text{RPR}_{\text{I}}$+PGO}
\label{section_pgo_for_fusion}

\begin{figure}[t!]
	\centering
    \includegraphics[width=0.8\linewidth]{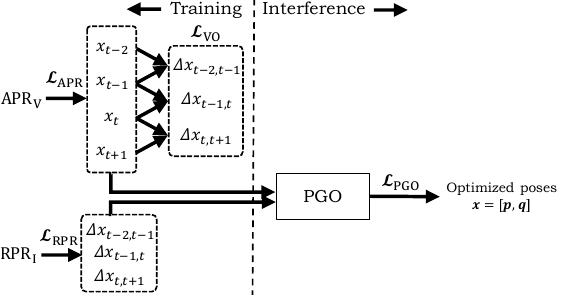}
    \caption{Overview of the $\text{APR}_{\text{V}}$-$\text{RPR}_{\text{I}}$ fusion model with PGO used for absolute pose refinement using the predicted relative poses from the $\text{RPR}_{\text{I}}$ model.}
    \label{image_pgo_for_fusion}
\end{figure}

\noindent MapNet+~\cite{brahmbhatt} relies on visual data to regress the absolute poses (that fail to provide accurate information in challenging conditions) and show how the geometric constraints between pairs of observations can be included as an additional loss term (i.e., VO from pairs of images, from GPS readings, or from IMU readings). Suppose we have IMU sequences of the same scene as additional data. The relative poses can be regressed using these sequences in order to efficiently update the graph, particularly in challenging lighting conditions. Therefore, our fusion model consists of $\text{APR}_{\text{V}}$ and $\text{RPR}_{\text{I}}$ models to simultaneously regress absolute and relative poses for $S$ pairs of images and IMU sequences sampled with a gap of $k$ timesteps from the dataset $D$ (see Figure~\ref{image_pgo_for_fusion}). The $\text{APR}_{\text{V}}$ and $\text{RPR}_{\text{I}}$ networks process the images and IMU sequences separately in a time-distributed manner. The visual features ($\mathbf{a}_V$) from $\text{APR}_{\text{V}}$ and inertial features ($\mathbf{a}_I$) from $\text{RPR}_{\text{I}}$ of the last layers are fused based on the soft fusion mechanism discussed in Section \ref{section_soft_fusion}. The features from the soft fusion model are forwarded to BiLSTM layers and a pose regressor, one for each absolute and relative pose regression. We use a similar optimization method as proposed in MapNet~\cite{brahmbhatt}. The minor difference is that, while MapNet optimizes the prediction of absolute pose $\mathbf{p}_{i}$ and $\mathbf{p}_{j}$ for image pairs ($i$, $j$) and the relative pose $\mathbf{v}_{ij}$ between ($i$, $j$), our fusion model additionally optimizes the relative pose ($\Delta t_{ij}$, $\Delta q_{ij}$) regressed by the $\text{RPR}_{\text{I}}$ model. So the final loss function for the fusion model is
\begin{equation}
\label{eq_9}
\begin{aligned}
    \mathcal{L}_{\text{total}} = \,\, &\gamma \sum_{i=1}^{|D|} h(\mathbf{p}_{i}, \hat{\mathbf{p}}_{i}) + \alpha \sum_{i,j=1,i\neq j}^{|D|} h(\mathbf{v}_{ij}, \hat{\mathbf{v}}_{ij}) + \\
    &+ \beta \sum_{i,j=1,i\neq j}^{|D|} h(\Delta \mathbf{p}_{i}, \Delta \hat{\mathbf{p}}_{i}),
\end{aligned}
\end{equation}
the sum of losses for the predicted absolute poses $\mathbf{p}_{i}$, the relatives poses between the predicted absolute poses $\mathbf{v}_{ij}$, and the predicted relative poses $\Delta \mathbf{p}_{i}$ weighted by the hyperparameters $\alpha$, $\beta$ and $\gamma$. We utilize Optuna to search for optimal hyperparameters. During inference, we use PGO to fuse the predicted absolute poses $\mathbf{p}_{i}$ and the predicted relative pose $\Delta \mathbf{p}_{ij}$ from the fusion network to get smooth and globally consistent absolute poses $\mathbf{p}_{i}^{o}$.
 
\subsection{Attention-based Fusion Methods}
\label{section_attention_fusion}

\noindent Visual and interial features offer different strengths to pose regression. Hence, our objective is to extract meaningful information from the camera and IMU sensors and to obtain a precise estimate of the absolute poses through the use of a combined feature representation. Inspired by widely applied attention mechanisms \cite{vaswani, xu_show, li_tell}, we re-weight each feature by conditioning on both visual and inertial features. The selection process of attention-based fusion is conditioned on the measurement reliability and the dynamics of both sensors, which learns to keep the most relevant feature representations while discarding useless or noisy information. For decision level fusion of $\text{APR}_{\text{V}}$ and $\text{RPR}_{\text{I}}$ features, we use layer concatenation (see Section~\ref{section_concatenation}), and soft fusion \cite{chen} (see Section \ref{section_soft_fusion}). We use the architecture introduced by \cite{joze} to fuse the $\text{APR}_{\text{V}}$ and $\text{RPR}_{\text{I}}$ features at the intermediate levels (see Section \ref{section_intermediate_fusion}).

\subsubsection{Late Fusion (Layer Concatenation)}
\label{section_concatenation}

\noindent We visualize late fusion in Figure~\ref{image_concat_fusion} that combines the high-level features produced by the individual sources to extract meaningful information for future pose regression tasks. Late fusion is possible when the features have the same number of units and dimensionality. The structure of the late fusion based on concatenation is shown in Figure~\ref{image_late_fusion}. The fusion network consists of the baseline models $\text{APR}_{\text{V}}$ and $\text{RPR}_{\text{I}}$ (see Section \ref{section_baseline_models}), that process the image and IMU data separately. During fusion, we concatenate the 1D visual features $\mathbf{a}_V$ from $\text{APR}_{\text{V}}$ and 1D inertial features $\mathbf{a}_I$ from $\text{RPR}_{\text{I}}$. $\mathbf{a}_V$ and $\mathbf{a}_I$ are of size $128$. In order to model the temporal dependencies between the combined features, we add a two-layer BiLSTM. After the recurrent network, an FC layer is utilized to regress the absolute and relative poses in an MTL setup.

\subsubsection{Late Fusion (Soft Fusion)}
\label{section_soft_fusion}

\begin{figure*}[t!]
	\centering
	\begin{minipage}[b]{1.0\linewidth}
        \centering
    	\includegraphics[width=0.8\linewidth]{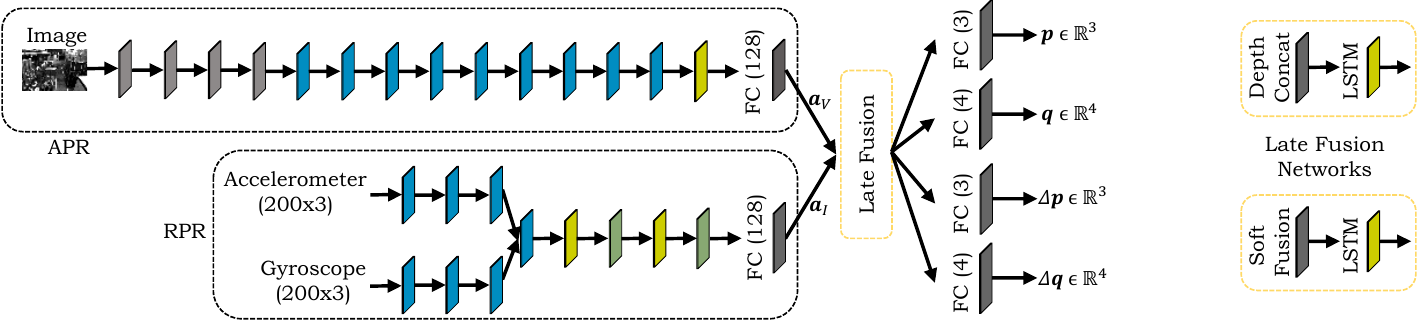}
        \caption{Overview of the fusion of the visual feature $\mathbf{a}_V$ of the $\text{APR}_{\text{V}}$ model and the inertial feature $\mathbf{a}_I$ of the $\text{RPR}_{\text{I}}$ model with layer concatenation. Four FC layers regress the absolute pose [$\mathbf{p}$, $\mathbf{q}$] and the relative pose [$\Delta \mathbf{p}$, $\Delta \mathbf{q}$].}
        \label{image_concat_fusion}
    \end{minipage}
    
	\begin{minipage}[b]{0.325\linewidth}
        \centering
    	\includegraphics[width=1.0\linewidth]{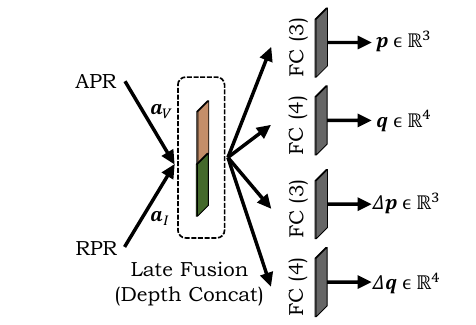}
        \caption{Overview of the fusion of the visual feature $\mathbf{a}_V$ of the $\text{APR}_{\text{V}}$ model and the inertial features $\mathbf{a}_I$ of the $\text{RPR}_{\text{I}}$ model using concatenation. Four FC layers regress the absolute pose [$\mathbf{p}$, $\mathbf{q}$] and the relative pose [$\Delta \mathbf{p}$, $\Delta \mathbf{q}$].}
        \label{image_late_fusion}
    \end{minipage}
    \hfill
	\begin{minipage}[b]{0.325\linewidth}
        \centering
    	\includegraphics[width=1.0\linewidth]{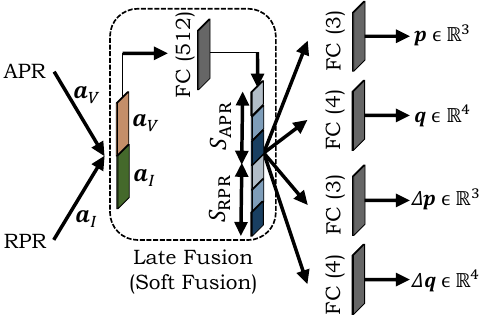}
        \caption{Overview of the fusion of the visual feature $\mathbf{a}_V$ of the $\text{APR}_{\text{V}}$ model and the inertial features $\mathbf{a}_I$ of the $\text{RPR}_{\text{I}}$ model using soft fusion \cite{chen}. Four FC layers regress the absolute pose [$\mathbf{p}$, $\mathbf{q}$] and the relative pose [$\Delta \mathbf{p}$, $\Delta \mathbf{q}$].}
        \label{image_soft_fusion}
    \end{minipage}
    \hfill
	\begin{minipage}[b]{0.325\linewidth}
        \centering
    	\includegraphics[width=0.70\linewidth]{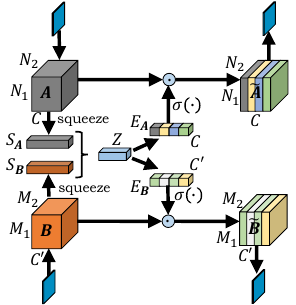}
        \caption{Overview of MMTM~\cite{joze} for two modalities. While $\mathbf{A}$ represents the visual features of $\text{APR}_{\text{V}}$, $\mathbf{B}$ represents the inertial features of $\text{RPR}_{\text{I}}$ at a given layer (blue).}
        \label{image_mmtm}
    \end{minipage}
\end{figure*}

\noindent The structure of the soft fusion module is shown in Figure~\ref{image_soft_fusion}. We combine the high-level features $\mathbf{a}_V$ and $\mathbf{a}_I$ produced by the $\text{APR}_{\text{V}}$ and $\text{RPR}_{\text{I}}$ models. The features are fused based on the selective sensor fusion (SSF) approach introduced by Chen et al.~\cite{chen} that contains an attention mechanism. A pair of continuous masks, $\mathcal{S}_{\text{APR}_{\text{V}}}$ and $\mathcal{S}_{\text{RPR}_{\text{I}}}$, are introduced by
\begin{equation}
\label{eq: soft_fusion_masks}
\begin{split}
    \mathcal{S}_{\text{APR}_{\text{V}}} = \text{Sigmoid}_{V}([\textbf{a}_{V};\textbf{a}_{I}]) \\
    \mathcal{S}_{\text{RPR}_{\text{I}}} = \,\text{Sigmoid}_{I}([\textbf{a}_{V};\textbf{a}_{I}])
\end{split}
\end{equation}
to perform soft fusion. $\mathcal{S}_{\text{APR}_{\text{V}}}$ and $\mathcal{S}_{\text{RPR}_{\text{I}}}$ are the masks applied (element-wise product $\odot$) to the features $\mathbf{a}_V$ and $\mathbf{a}_I$ learned by the CNNs by conditioning on both features by
\begin{equation}
\centering
g_{\text{soft}}(\mathbf{a}_V, \mathbf{a}_I) = [\mathbf{a}_V \odot \mathcal{S}_{\text{APR}} ; \mathbf{a}_I \odot \mathcal{S}_{\text{RPR}_{\text{I}}}].
\label{eq: soft_fusion}
\end{equation}
The sigmoid function finally re-weights each feature vector and preserves the order of coefficient values in the range $[0, 1]$ to produce the new re-weighted vectors. After the soft-fusion, the features are forwarded to the recurrent network and finally to FC layers to perform the pose regression.

\subsubsection{Intermediate Fusion (MMTM)}
\label{section_intermediate_fusion}

\begin{figure*}[t!]
	\centering
	\begin{minipage}[b]{1.0\linewidth}
        \centering
    	\includegraphics[width=0.78\linewidth]{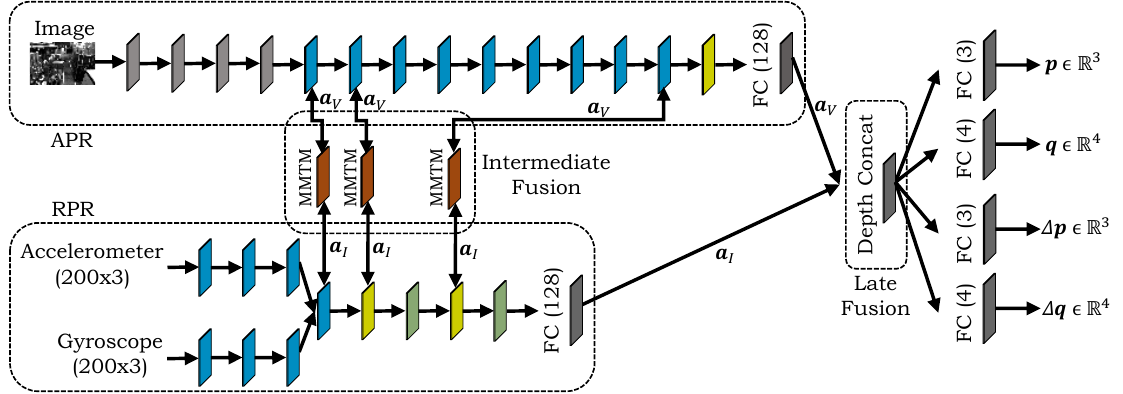}
        \caption{Overview of the fusion of intermediate visual features $\mathbf{a}_V$ of the $\text{APR}_{\text{V}}$ model and intermediate inertial features $\mathbf{a}_I$ of the $\text{RPR}_{\text{I}}$ model with three MMTMs~\cite{joze} (see Figure~\ref{image_mmtm}) for latent layer representations. After late fusion with layer concatenation, four FC layers regress the absolute pose [$\mathbf{p}$, $\mathbf{q}$] and the relative pose [$\Delta \mathbf{p}$, $\Delta \mathbf{q}$].}
        \label{image_mmtm_fusion}
    \end{minipage}
    
	\begin{minipage}[b]{0.7\linewidth}
        \centering
    	\includegraphics[trim=0 2 0 3, clip, width=0.96\linewidth]{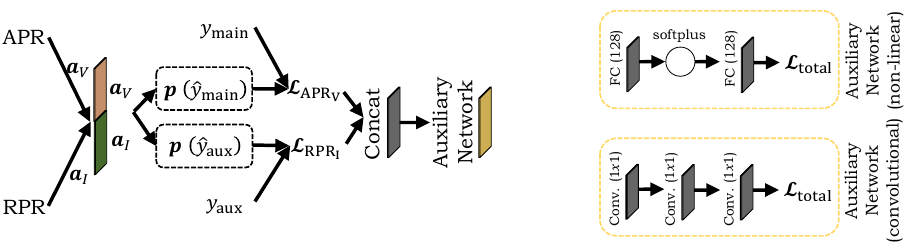}
        \caption{Left: Overview of the structure of $\text{APR}_{\text{V}}$ and $\text{RPR}_{\text{I}}$ fusion with the auxiliary learning model \cite{navon_aux}. Right: After layer concatenation a \textit{convolutional} or \textit{non-linear} auxiliary network is employed that learns the combinations of losses.}
        \label{image_auxiliary_learning}
    \end{minipage}
    \hfill
	\begin{minipage}[b]{0.27\linewidth}
        \centering
    	\includegraphics[trim=0 1 0 2, clip, width=1.0\linewidth]{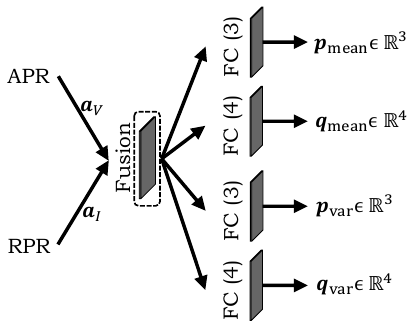}
        \caption{Overview of the structure of the BNN~\cite{kendall_uncertainty}.}
        \label{image_bayesian_model}
    \end{minipage}
\end{figure*}

\noindent As the $\text{APR}_{\text{V}}$ and $\text{RPR}_{\text{I}}$ models have unaligned spatial dimensions, they cannot directly be fused using the commonly used techniques like element-wise summation \cite{hu}, weighted average \cite{natarajan}, or more sophisticated methods like attention mechanisms \cite{chen} that assume identical spatial dimensions of different streams. Inspired by the squeeze-and-excitation (SE) \cite{hu2018squeeze} module for unimodal CNNs, Joze et al.~\cite{joze} proposed the multimodal transfer module (MMTM) that allows the fusion of modalities with different spatial dimensions. Importantly, the squeeze operation squeezes the spatial information into channel descriptors via a global average pooling operation over spatial dimensions of input features that enables fusion of modalities of arbitrarily feature dimension. The excitation operation generates the excitation signals using a simple gating mechanism as a sigmoid function, which allows the suppression or excitation of different filters in each stream. While MMTM was applied to RGB+depth, RGB+wave, and RGB+pose (both modalities correspond to each other without prerequisites), the fusion of APR and RPR with MMTM proved to be more challenging, as RPR requires knowledge of its global pose. In our approach, this issue was addressed by utilizing a time-distributed PoseNet, which provided absolute poses from three consecutive images, see \cite{ott}. The structure of MMTM~\cite{joze} is shown in Figure~\ref{image_mmtm}. The matrices $\mathbf{A} \in \mathbb{R}^{N_{1} \times \cdot \cdot \cdot \times N_{K} \times C}$ and $\mathbf{B} \in \mathbb{R}^{M_{1} \times \cdot \cdot \cdot \times M_{L} \times C^{'}}$ represent the features at any given level of the $\text{APR}_{\text{V}}$ and $\text{RPR}_{\text{I}}$ models that are the inputs to the MMTM module. $N_{i}$ and $M_{i}$ represent the spatial dimensions, and $C$ and $C^{'}$ represent the number of channels of $\text{APR}_{\text{V}}$ and $\text{RPR}_{\text{I}}$, respectively. MMTM learns the global multimodal embedding to re-calibrate the inputs $\mathbf{A}$ and $\mathbf{B}$ using the SE operation on the input tensors $\mathbf{A}$ and $\mathbf{B}$. The squeeze operation enables fusion between the modalities $S_{\mathbf{A}}$ and $S_{\mathbf{B}}$ that have arbitrary spatial dimensions.
\begin{equation}
    \begin{split}
        S_{\mathbf{A}}(c) & = \frac{1}{\prod_{i=1}^K N_{i}} \sum_{n_{1}, \ldots, n_{k}}\textbf{A}(n_{1}, \ldots, n_{K},c) \\
        S_{\mathbf{B}}(c) & = \frac{1}{\prod_{i=1}^L M_{i}} \sum_{m_{1}, \ldots, m_{L}}\textbf{B}(m_{1}, \ldots, m_{L},c) \\
    \end{split}
\end{equation}
that are further mapped into a joint representation $Z$ using concatenation and FC layers. Excitation signals, $E_{\mathbf{A}} \in \mathbb{R}_{C}$ and $E_{\mathbf{B}} \in \mathbb{R}_{C'}$ are generated using $Z$, which are used to re-calibrate the input features, $\mathbf{A}$ and $\mathbf{B}$, by a simple gating mechanism
\begin{equation}
    \begin{split}
         \Tilde{\textbf{A}} & = 2 \times \sigma(E_{\mathbf{A}}) \odot \textbf{A} \\
         \Tilde{\textbf{B}} & = 2 \times \sigma(E_{\mathbf{B}}) \odot \textbf{B}, \\
    \end{split}
\end{equation}
where $\sigma(\cdot)$ is a sigmoid function and $\odot$ is a channel-wise product operation. We use MMTM to fuse the features of the $\text{APR}_{\text{V}}$ and $\text{RPR}_{\text{I}}$ models as proposed in Figure~\ref{image_mmtm_fusion}. Learning the joint representation using MMTM allows the $\text{APR}_{\text{V}}$ model to re-calibrate the features of $\text{RPR}_{\text{I}}$ when the IMU sensor data is noisy, or vice versa when the images are blurred, texture-less, or are low light. Based on the experiments conducted in \cite{joze}, the best performance is achieved when the output of half of the last modules of two uni-modal streams are fused by MMTM modules. We insert the first MMTM module at layer $12$ of the $\text{APR}_{\text{V}}$ model and at layer $5$ of the $\text{RPR}_{\text{I}}$ model. The add the second MMTM at layer $14$ of $\text{APR}_{\text{V}}$ and layer $7$ of $\text{RPR}_{\text{I}}$, and the third MMTM at layer $15$ of $\text{APR}_{\text{V}}$ and at layer $9$ of $\text{RPR}_{\text{I}}$ (see Figure~\ref{image_mmtm_fusion}). Finally, similar to the late fusion, the combined features at the end are concatenated and forwarded to the FC layers to regress the poses.

\subsection{Auxiliary Learning}
\label{section_auxiliary_learning}

\noindent In the low data regime, where the main task overfits and generalizes poorly to unseen data, learning auxiliary tasks have proven to benefit the learning process \cite{jaderberg}. In this work, we use the auxiliary learning framework AuxiLearn~\cite{navon_aux} to optimize the learning of the main task ($\text{APR}_{\text{V}}$) while using $\text{RPR}_{\text{I}}$ as auxiliary task. The structure of the AuxiLearn network consists of the main network and an auxiliary network, as shown in Figure~\ref{image_auxiliary_learning}. The main network is an $\text{APR}_{\text{V}}$-$\text{RPR}_{\text{I}}$ soft fusion network (see Section~\ref{section_soft_fusion}) that regresses the absolute and relative poses separately. The main network minimizes the losses on the main task $\mathcal{L}_{\text{APR}_{\text{V}}}$ and the auxiliary task $\mathcal{L}_{\text{RPR}_{\text{I}}}$. The auxiliary network finally operates on the concatenated vector of losses from the main task. We employ two kinds of auxiliary networks (see Figure~\ref{image_auxiliary_learning}): The \textit{convolutional network} variant of the auxiliary network consists of stacked 1D convolutional layers that models the spatial relation among losses, whereas the \textit{non-linear} variant consists of stacked FC layers along with a softplus activation function that captures complex interactions between tasks and learns the non-linear combination of losses. To train the auxiliary learning framework, we use the training set $(\textit{\textbf{x}}^{t}, \textit{\textbf{y}}^{t})$, and the distinct, independent set $(\textit{\textbf{x}}^{a}, \textit{\textbf{y}}^{a})$ that represents the auxiliary set. The weights of the main network $\mathbf{W}$ are optimized on the training set $(\textit{\textbf{x}}^{t}, \textit{\textbf{y}}^{t})$ to minimize the total loss
\begin{equation}
    \centering
    \mathcal{L}_{\text{total}} = \mathcal{L}_{\text{APR}_{\text{V}}}(\textit{\textbf{x}}^{t}, \textit{\textbf{y}}^{t};\textit{\textbf{W}}) + h(\textit{\textbf{x}}^{t}, \textit{\textbf{y}}^{t}; \textit{\textbf{W}};\phi),
\end{equation}
where $\mathcal{L}_{\text{APR}_{\text{V}}}$ is the loss of the main task, and $h$  is the overall auxiliary loss controlled by $\phi$. The loss on the auxiliary set is defined as $\mathcal{L}_{\text{A}} = \mathcal{L}_{\text{APR}_{\text{V}}}(\textit{\textbf{x}}^{a}\textit{\textbf{y}}^{a};\textit{\textbf{W}})$
as we are interested in the generalization performance of the main task. Since there is an indirect dependence of the $\mathcal{L}_{\text{A}}$ on the auxiliary parameters $\phi$, we compute the \textit{bi-level optimization} \cite{navon_aux} over the main network's parameters $\mathbf{W}$. In practice, we simultaneously train both, $\mathbf{W}$ and $\phi$, by altering between optimizing $\mathbf{W}$ on $\mathcal{L}_{\text{total}}$ and $\phi$ on $\mathcal{L}_{\text{A}}$.

\subsection{Bayesian Learning}
\label{setcion_bayesian_learning}

\noindent Understanding the uncertainty of a model is a crucial part of many ML systems. Neural networks learn the powerful representations that can map high-dimensional data to an array of features \cite{szegedy, kendall}. However, the mappings are often assumed to be accurate, which is not always the case. In order to understand the confidence of the models' predictions, we use the Bayesian neural network (BNN) \cite{kendall_uncertainty} technique, which offers to understand the ML model's uncertainty. Kendall et al.~\cite{kendall_uncertainty} introduced two main kinds of uncertainty: \textit{Aleatoric uncertainty} that captures noise inherent in the training data. \textit{Epistemic uncertainty}, also known as model uncertainty, accounts for uncertainty in the model parameters; this uncertainty can be explained away given enough data. In this work, we model the aleatoric uncertainty by modifying the $\text{APR}_{\text{V}}$-$\text{RPR}_{\text{I}}$ fusion architecture to predict both, the mean pose values $\textbf{\textit{p}}_{\text{mean}}$ and the corresponding variance $\boldsymbol{\sigma}^{2}_{\textit{p}}$ (see Figure~\ref{image_bayesian_model}). This modification induces a new kind of minimization objective based on the aleatoric uncertainty as
\begin{equation}
    \centering
    \mathcal{L}_{\text{BNN}} = \frac{|| \textbf{\textit{p}} - \hat{\textbf{\textit{p}}}||^{2}}{2\boldsymbol{\sigma}^{2}_{p}} + \frac{1}{2}\log\boldsymbol{\sigma}^{2}_{p},
    \label{eq_aleatoric}
\end{equation}
where $\textit{\textbf{p}}$ and $\hat{\textbf{\textit{p}}}$ are the predicted and ground truth absolute poses and $\boldsymbol{\sigma}^{2}_{\textit{p}}$ are the predicted variances. We do not need the uncertainty labels to learn the uncertainty, rather, we only need to supervise the learning of the pose regression by learning the variances implicitly from the loss function. If the model is uncertain in pose prediction (first term of Equation~\eqref{eq_aleatoric}), the model learns to attenuate the total loss $\mathcal{L}_{\text{BNN}}$ by increasing the uncertainty $\boldsymbol{\sigma}^{2}_{p}$. The second regularization term of Equation~\eqref{eq_aleatoric}, however, prevents the network from predicting infinite uncertainty, and thus can be thought of as "learned loss attenuation". In practice, we train the network to predict the log variance $s_{p} := \log \boldsymbol{\sigma}_{p}^2$ with the loss
\begin{equation}
\centering
    \mathcal{L}_{\text{BNN}} = \frac{1}{2}\text{exp}(-s_{p})|| \textbf{\textit{p}} - \hat{\textbf{\textit{p}}}||^{2} + \frac{1}{2}s_{p}.
\end{equation}
To regress $s_{p}$ is more stable than to regress $\boldsymbol{\sigma}^{2}_{p}$ that avoids a potential division by zero and dampens the effect of log-functions.

\subsection{Deep Multimodal Fusion for Relative Pose Regression}
\label{setcion_flownet_odometry}

\noindent In this section, we propose techniques for VI odometry by fusing the relative pose regression models $\text{RPR}_{\text{V}}$ and $\text{RPR}_{\text{I}}$ as discussed in Section \ref{section_baseline_models}. $\text{RPR}_{\text{V}}$ extracts the latent representation from two consecutive monocular images, and $\text{RPR}_{\text{I}}$ extracts temporal information from the inertial data. Both $\text{RPR}_{\text{V}}$ and $\text{RPR}_{\text{I}}$ models are supervised to regress the relative change in pose. In order to combine the high-level features produced by the two encoders from raw data sequences, we perform the late fusion (Section \ref{section_soft_fusion}) and intermediate fusion (Section \ref{section_intermediate_fusion}) of $\text{RPR}_{\text{V}}$ and $\text{RPR}_{\text{I}}$ to regress the relative poses. Furthermore, we use the BNN~\cite{kendall_uncertainty} (Section \ref{setcion_bayesian_learning}) to model the aleatoric uncertainty in the relative pose estimation. It is not possible to perform PGO for $\text{RPR}_{\text{V}}$ and $\text{RPR}_{\text{I}}$ fusion since PGO aims to optimize the consecutive absolute poses. Similarly, we cannot use the auxiliary learning framework \cite{navon_aux} as it involves learning two related tasks namely, the main task of interest, while using another auxiliary task to aid the learning of the main task.

\textbf{Late Fusion.} To perform late fusion, we use a similar architecture as in Section~\ref{section_soft_fusion}. The fusion model takes high-level visual features $\mathbf{a}_{V}$ and inertial features $\mathbf{a}_{I}$ from image and IMU encoders, respectively. Contrarily, the visual features input $\mathbf{a}_{V}$ for the fusion model is obtained from the output of layer $\text{Conv6}$, rather than being derived from the last layer of $\text{APR}_{\text{V}}$. The inertial features $\mathbf{a}_{I}$ from the $\text{RPR}_{\text{I}}$ model remain the same. We perform soft fusion \cite{chen} by generating a pair of continuous masks, $\mathcal{S}_{\text{RPR}_{\text{V}}}$ and $\mathcal{S}_{\text{RPR}_{\text{I}}}$ from the visual features $\mathbf{a}_V$ and the inertial features $\mathbf{a}_I$ (see Equation~\eqref{eq: soft_fusion_masks}). Finally, the output of the fusion model $g_{\text{soft}}$ is propagated further to the FC layers to perform pose regression. 

\textbf{Intermediate Fusion.} We utilize MMTM~\cite{joze} as discussed in Section \ref{section_intermediate_fusion} to perform the intermediate fusion of the $\text{RPR}_{\text{V}}$ and $\text{RPR}_{\text{I}}$ models. We insert the first MMTM at layer $7$ of $\text{RPR}_{\text{V}}$ and layer $5$ of $\text{RPR}_{\text{I}}$, the second MMTM at layer $8$ of $\text{RPR}_{\text{V}}$ and layer $7$ of $\text{RPR}_{\text{I}}$, and the third MMTM at layer $9$ of $\text{RPR}_{\text{V}}$ and layer $9$ of $\text{RPR}_{\text{I}}$. Finally, similar to the late fusion, the features from the last layers of $\text{RPR}_{\text{V}}$ and $\text{RPR}_{\text{I}}$ are concatenated and forwarded to the FC layers to regress the relative pose.

\textbf{Bayesian Learning.} To model the aleatoric uncertainty inherent in the relative pose regression of the $\text{RPR}_{\text{V}}$-$\text{RPR}_{\text{I}}$ fusion model, we follow a similar method as discussed in Section \ref{setcion_bayesian_learning}. First, we modify the $\text{RPR}_{\text{V}}$-$\text{RPR}_{\text{I}}$ fusion architecture based on late fusion to predict both the mean relative poses and their corresponding variances. Finally, we adapt Equation \eqref{eq_aleatoric} based on the aleatoric uncertainty to minimize the loss in the pose regression as
\begin{equation}
\centering
    \mathcal{L}_{\text{BNN}} = \frac{1}{2}\text{exp}(-s_{\Delta p})|| \Delta\textbf{\textit{p}} - \Delta\hat{\textbf{\textit{p}}}||^{2} + \frac{1}{2}s_{\Delta p}.
\end{equation}
where $s_{\Delta p} := \log \boldsymbol{\sigma}_{\Delta p}^2$. $\Delta\textit{\textbf{p}}$ and $\Delta\hat{\textbf{\textit{p}}}$ are the predicted and ground truth relative poses, and $\boldsymbol{\sigma}^{2}_{\Delta\textit{p}}$ is the predicted variance.

\section{Datasets}
\label{chap_datasets}

\noindent We give details about the EuRoC MAV and PennCOSYVIO datasets in Section~\ref{chap_euroc}, respectively in Section~\ref{chap_penncosyvio}. We propose our novel IndustryVI dataset in Section~\ref{chap_Industry}.

\subsection{The EuRoC MAV Dataset}
\label{chap_euroc}

\begin{figure}[t!]
	\centering
	\begin{minipage}[b]{0.24\linewidth}
        \centering
    	\includegraphics[width=1.0\linewidth]{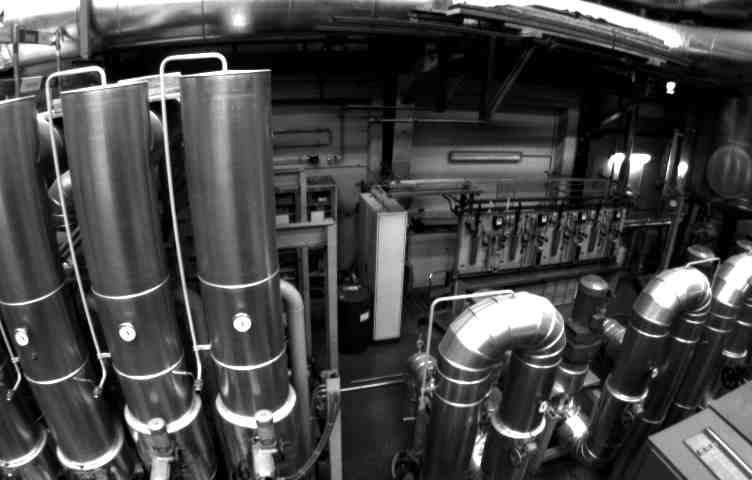}
    \end{minipage}
    \hfill
	\begin{minipage}[b]{0.24\linewidth}
        \centering
    	\includegraphics[width=1.0\linewidth]{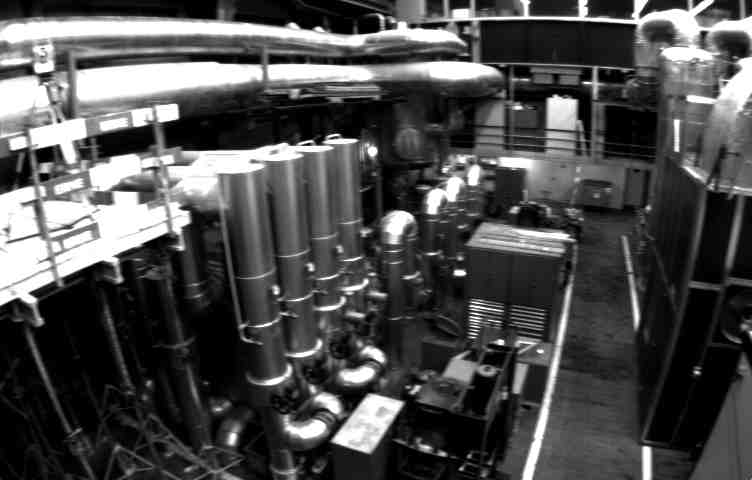}
    \end{minipage}
    \hfill
	\begin{minipage}[b]{0.24\linewidth}
        \centering
    	\includegraphics[width=1.0\linewidth]{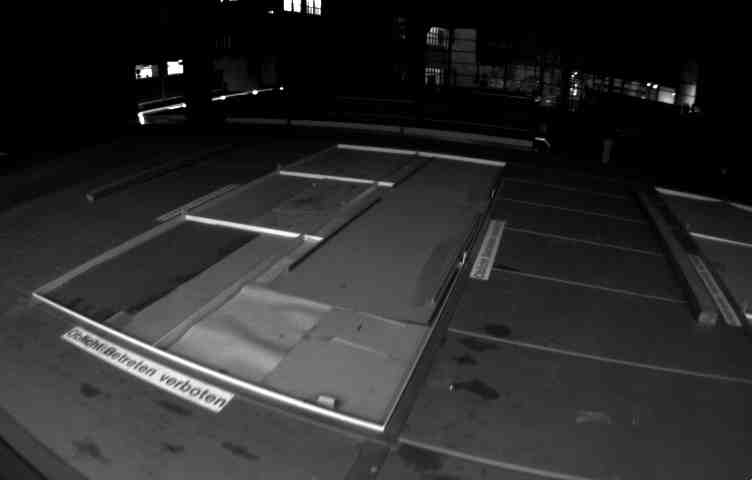}
    \end{minipage}
    \hfill
	\begin{minipage}[b]{0.24\linewidth}
        \centering
    	\includegraphics[width=1.0\linewidth]{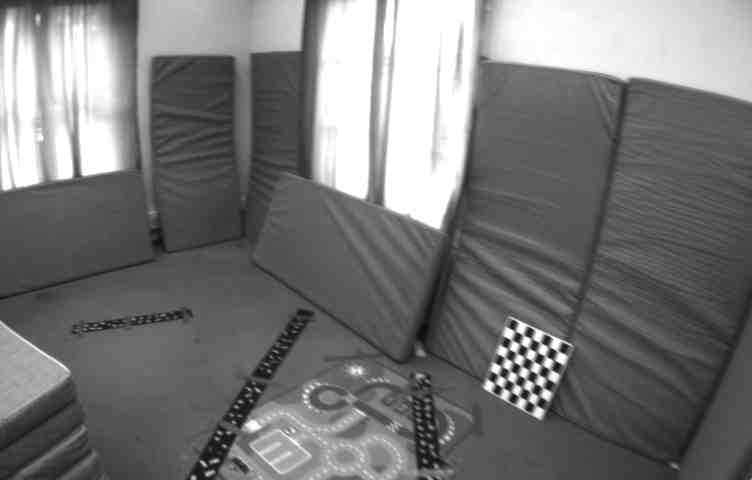}
    \end{minipage}
    \caption{Exemplary images of the EuRoC MAV dataset \cite{burri} of the machine hall (1-3) and the Vicon room (4).}
    \label{image_euroc_ex_images}
\end{figure}

\noindent The EuRoC MAV \cite{burri} dataset was collected on-board an MAV and was recorded in an industrial machine hall (MH) environment and in an indoor Vicon (V) room. The dataset contains synchronized images (from a front-down looking stereo camera), IMU measurements, and ground truth poses from a Leica Nova MS50 laser tracking system and a motion capture system. Exemplary images are given in Figure~\ref{image_euroc_ex_images}. 11 datasets range from slow flights under good visual conditions (MH-01, MH-02, V1-01, V2-01) to dynamic flights with poor illumination (MH-04, MH-05) and motion blur (MH-03, V1-02, V1-03, V2-02, V2-03). This presents a difficulty in terms of generalizing the dataset. The size of the Vicon room is small-scale ($8m \times 8.4m \times 4m$). Many (SLAM) methods are evaluated on this dataset as it contains different motion dynamics, but the dataset is not useful for many applications (i.e., robotics or handheld devices) as it is recorded on an MAV. The dataset contains 14,566 image and 145,660 IMU training samples and 12,481 image and 124,810 IMU test samples. We train on MH-01, MH-03, and MH-04 and test on MH-02 and MH-05 for APR techniques, and cross-validate all sequences for RPR techniques.

\begin{figure}[t!]
	\centering
	\begin{minipage}[b]{0.24\linewidth}
        \centering
    	\includegraphics[width=1.0\linewidth]{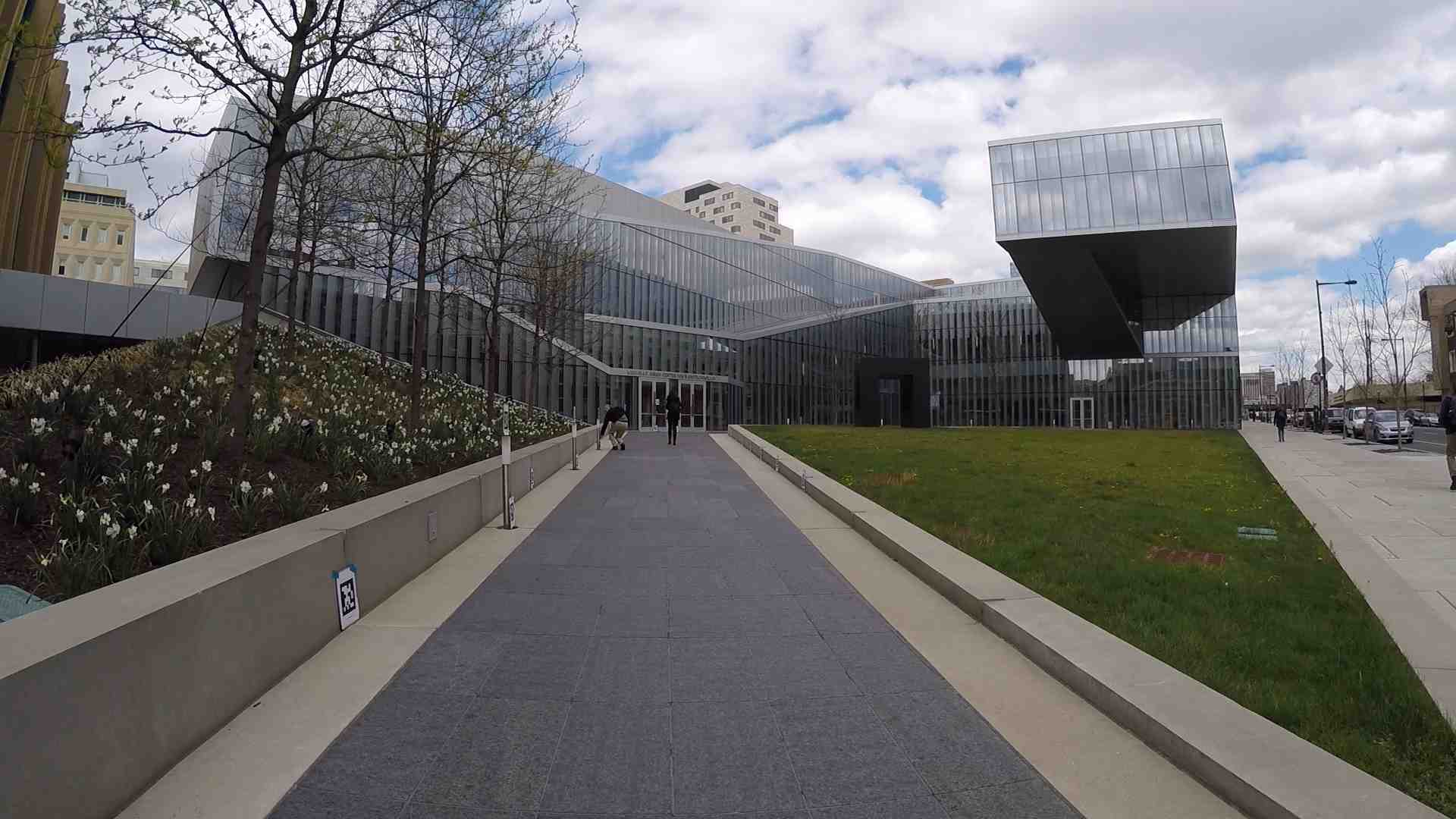}
    \end{minipage}
    \hfill
	\begin{minipage}[b]{0.24\linewidth}
        \centering
    	\includegraphics[width=1.0\linewidth]{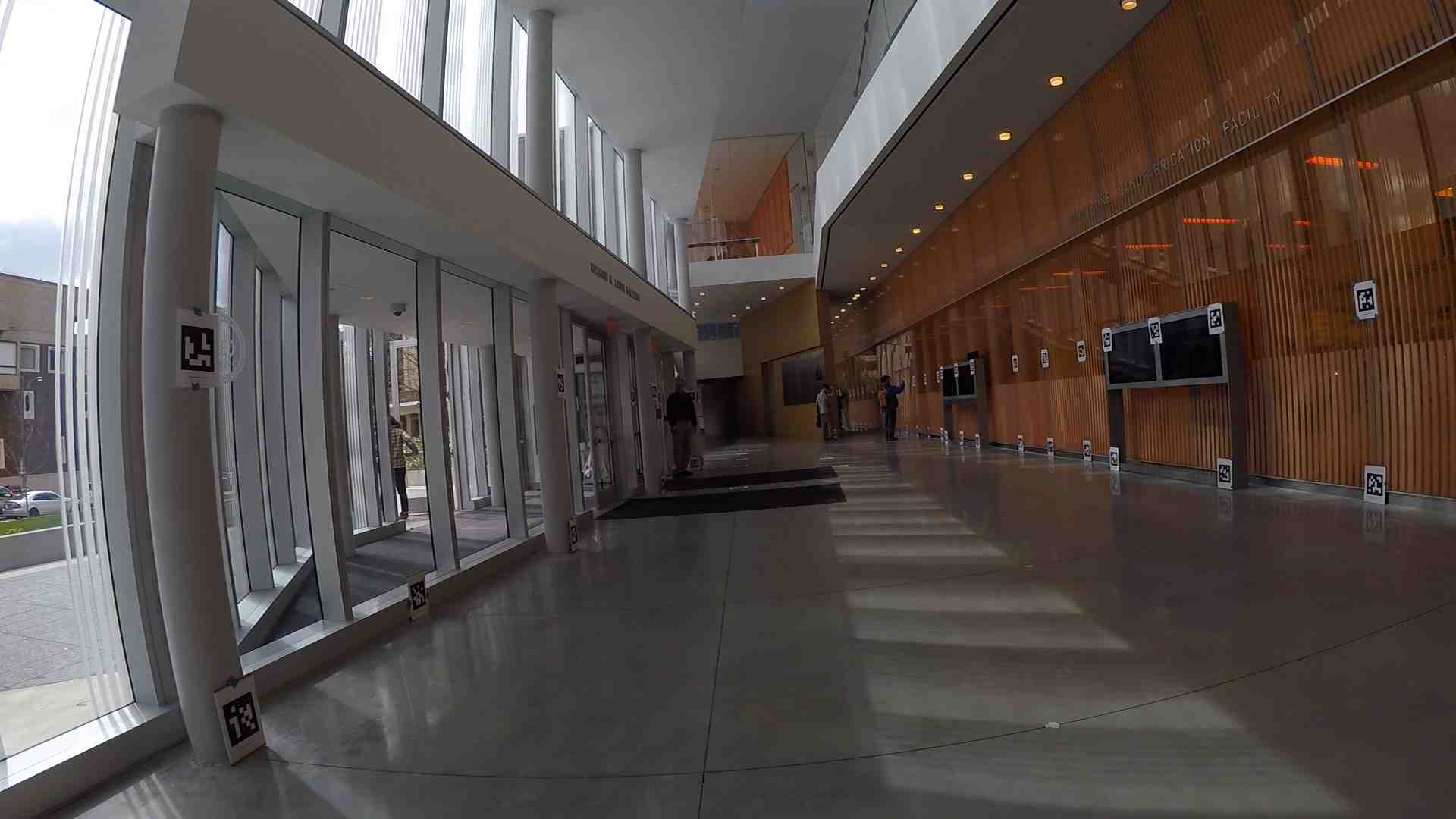}
    \end{minipage}
    \hfill
	\begin{minipage}[b]{0.24\linewidth}
        \centering
    	\includegraphics[width=1.0\linewidth]{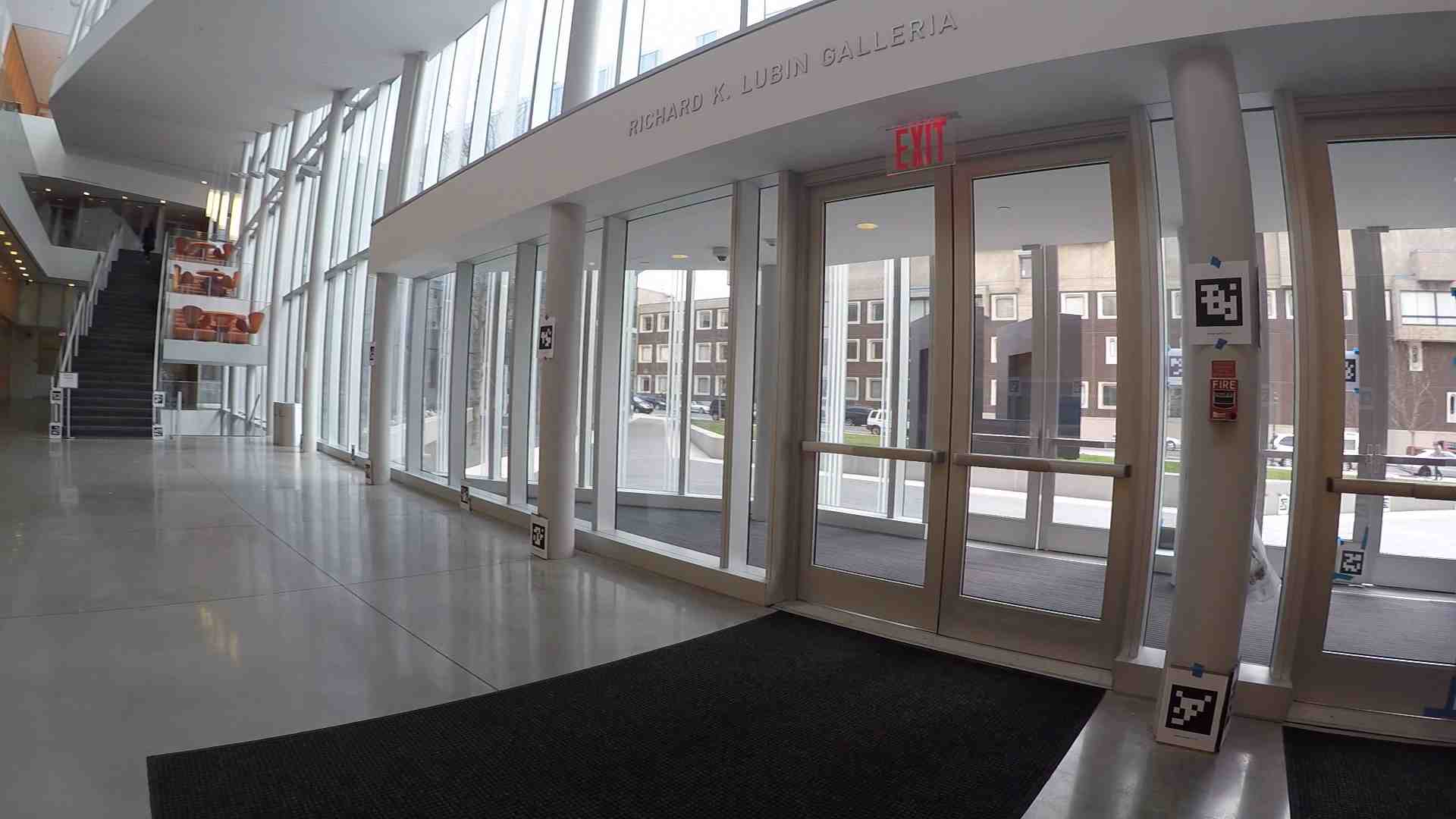}
    \end{minipage}
    \hfill
	\begin{minipage}[b]{0.24\linewidth}
        \centering
    	\includegraphics[width=1.0\linewidth]{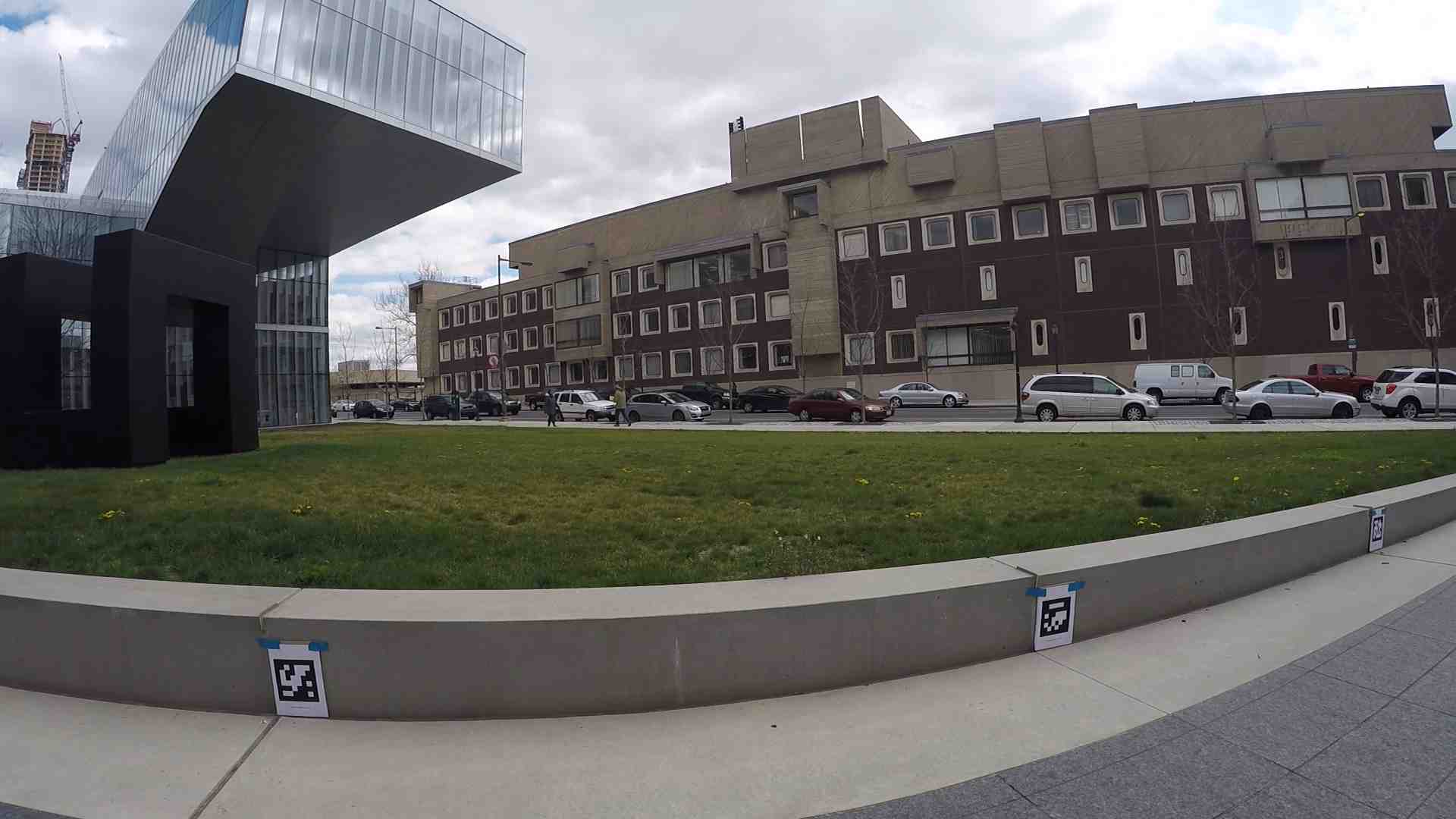}
    \end{minipage}
    \caption{Images of the PennCOSYVIO dataset \cite{pfrommer}.}
    \label{image_pennc_ex_images}
\end{figure}

\subsection{The PennCOSYVIO Dataset}
\label{chap_penncosyvio}

\noindent The PennCOSYVIO \cite{pfrommer} dataset was recorded with a handheld device at the University of Pennsylvania's Signh center. The device contains a stereo camera, an IMU, two Tango devices, and three GoPro cameras. An 150\textit{m} long path crosses from outdoors to indoors. Four sequences include rapid rotations, changes in lighting, repetitive structures such as large glass surfaces, and different textures (see Figure~\ref{image_pennc_ex_images}). The sequences AF and AS are for training and BF and BS are for testing. AS and BS are recorded at slow pace, and AF and BF at fast pace. We train and test on each pace separately. The dataset contains 5,035 image and 50,318 IMU training samples and 5,369 image and 53,670 IMU test samples.

\subsection{The IndustryVI Dataset}
\label{chap_Industry}

\begin{figure}[t!]
	\centering
	\begin{minipage}[b]{0.48\linewidth}
        \centering
    	\includegraphics[width=1.0\linewidth]{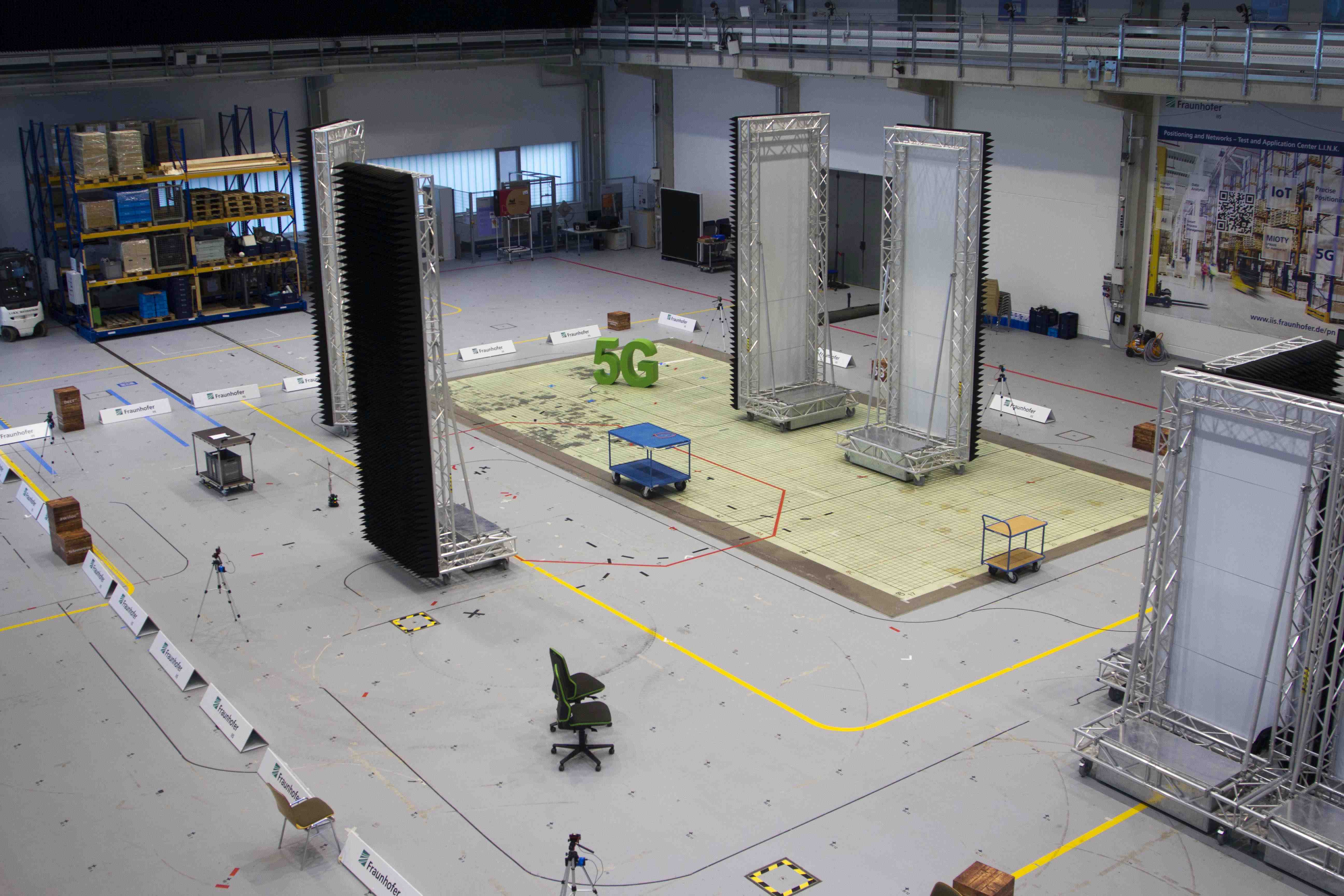}
    \end{minipage}
    \hfill
	\begin{minipage}[b]{0.48\linewidth}
        \centering
    	\includegraphics[width=1.0\linewidth]{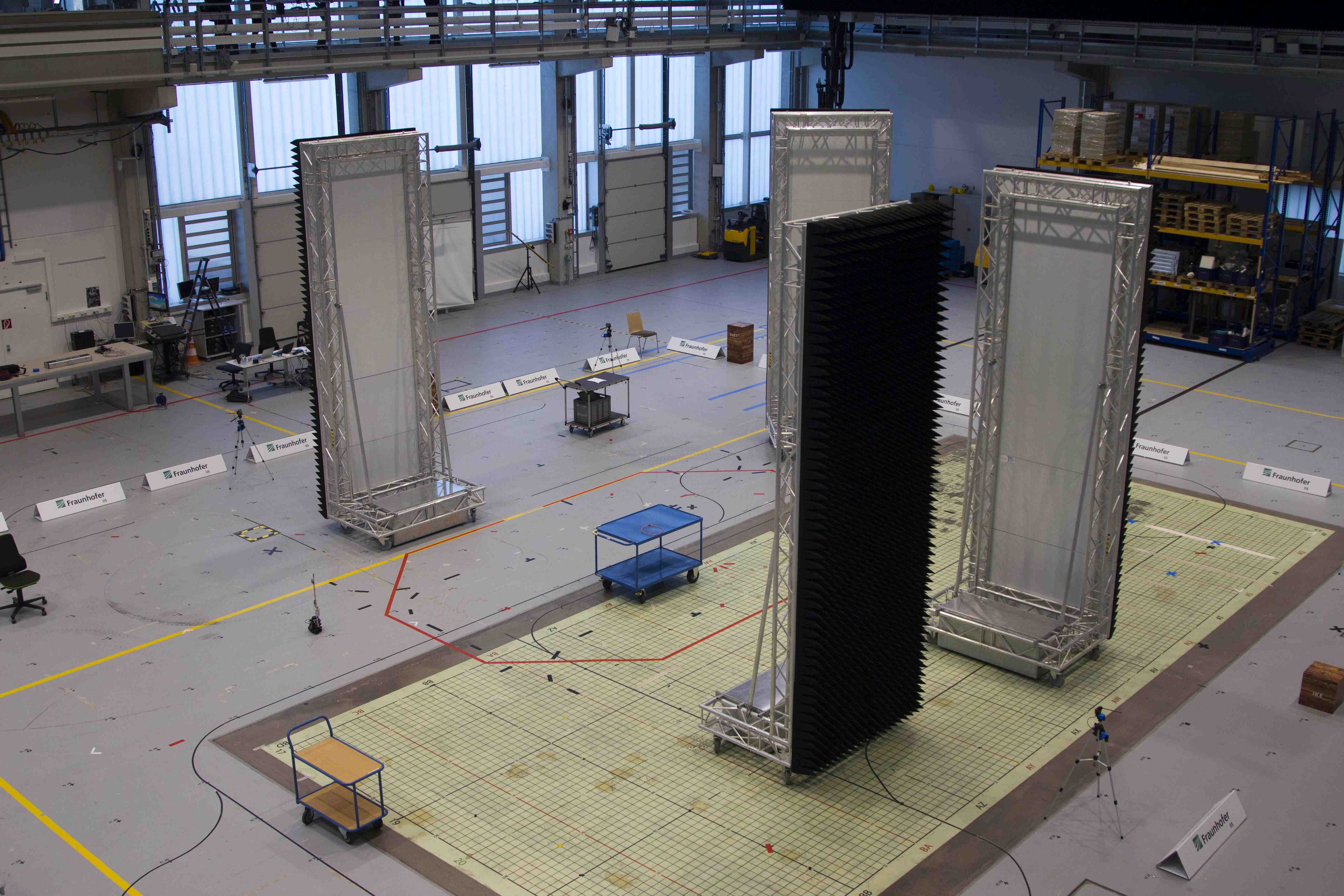}
    \end{minipage}
    \caption{Environment setup of the IndustryVI dataset with high absorber walls, randomly placed objects and warehouse racks.}
    \label{image_industry_env}
\end{figure}

\begin{figure}[t!]
	\centering
	\begin{minipage}[b]{0.32\linewidth}
        \centering
    	\includegraphics[width=1.0\linewidth]{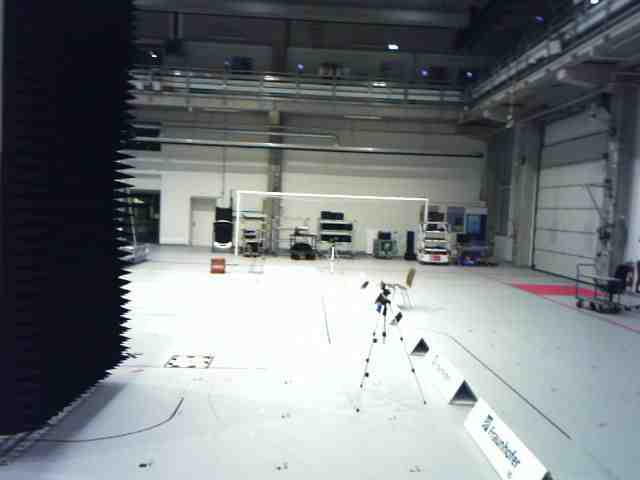}
    \end{minipage}
    \hfill
	\begin{minipage}[b]{0.32\linewidth}
        \centering
    	\includegraphics[width=1.0\linewidth]{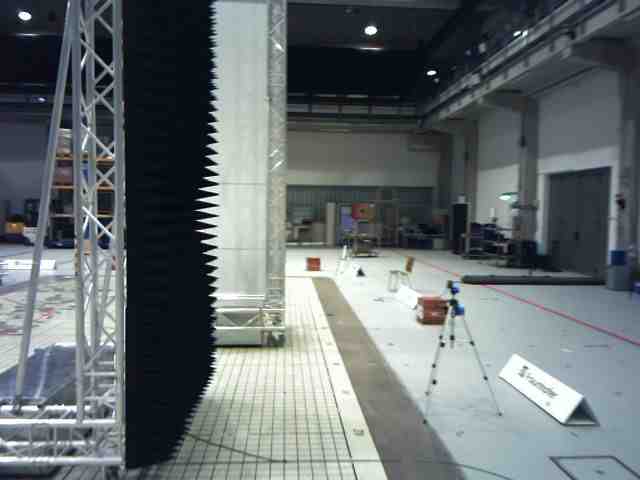}
    \end{minipage}
    \hfill
	\begin{minipage}[b]{0.32\linewidth}
        \centering
    	\includegraphics[width=1.0\linewidth]{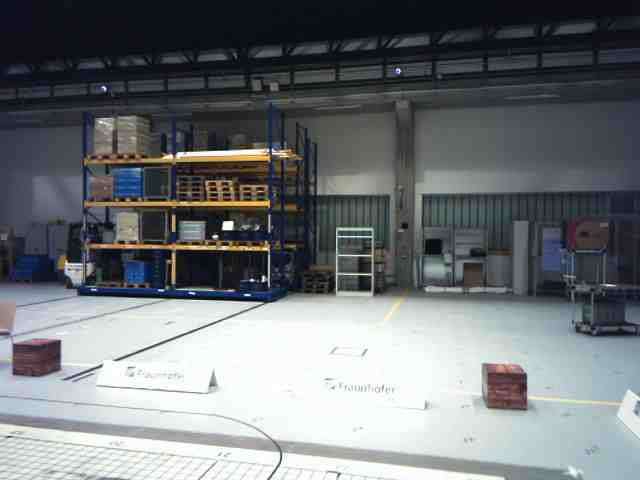}
    \end{minipage}
    \caption{Exemplary images with texture-less surfaces and variations in illuminations of the IndustryVI dataset.}
    \label{image_industry_ex_images}
\end{figure}

\begin{figure}[t!]
	\centering
	\begin{minipage}[b]{0.493\linewidth}
        \centering
    	\includegraphics[trim=10 10 10 10, clip, width=1.0\linewidth]{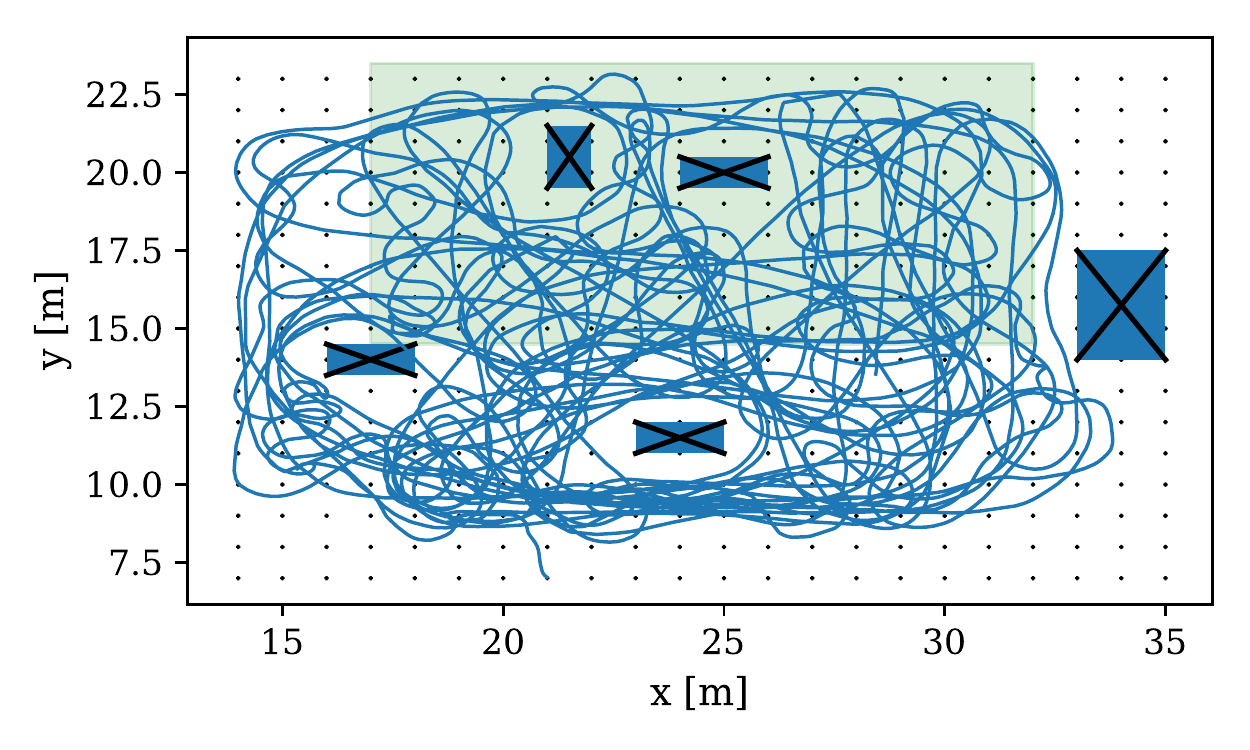}
    	\subcaption{Training trajectory 1.}
    	\label{image_traj_train1}
    \end{minipage}
    \hfill
	\begin{minipage}[b]{0.493\linewidth}
        \centering
    	\includegraphics[trim=10 10 10 10, clip, width=1.0\linewidth]{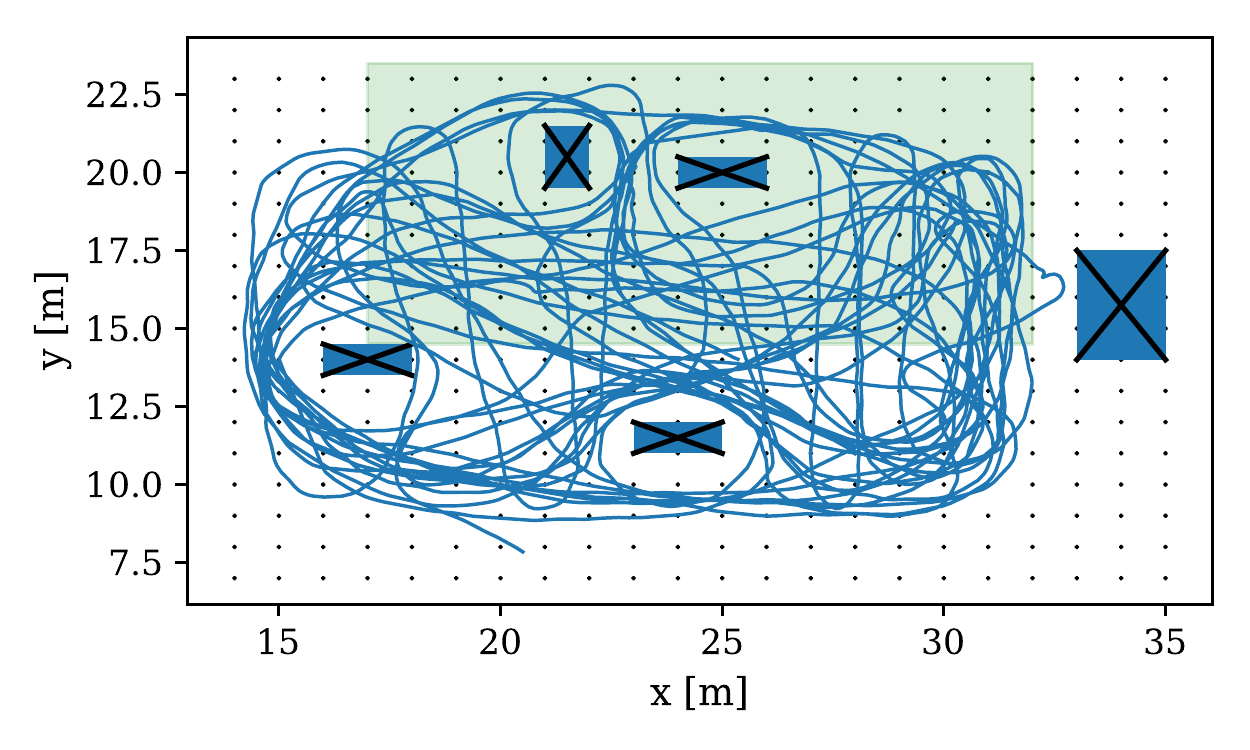}
    	\subcaption{Training trajectory 2.}
    	\label{image_traj_train2}
    \end{minipage}
	\begin{minipage}[b]{0.493\linewidth}
        \centering
    	\includegraphics[trim=10 10 10 0, clip, width=1.0\linewidth]{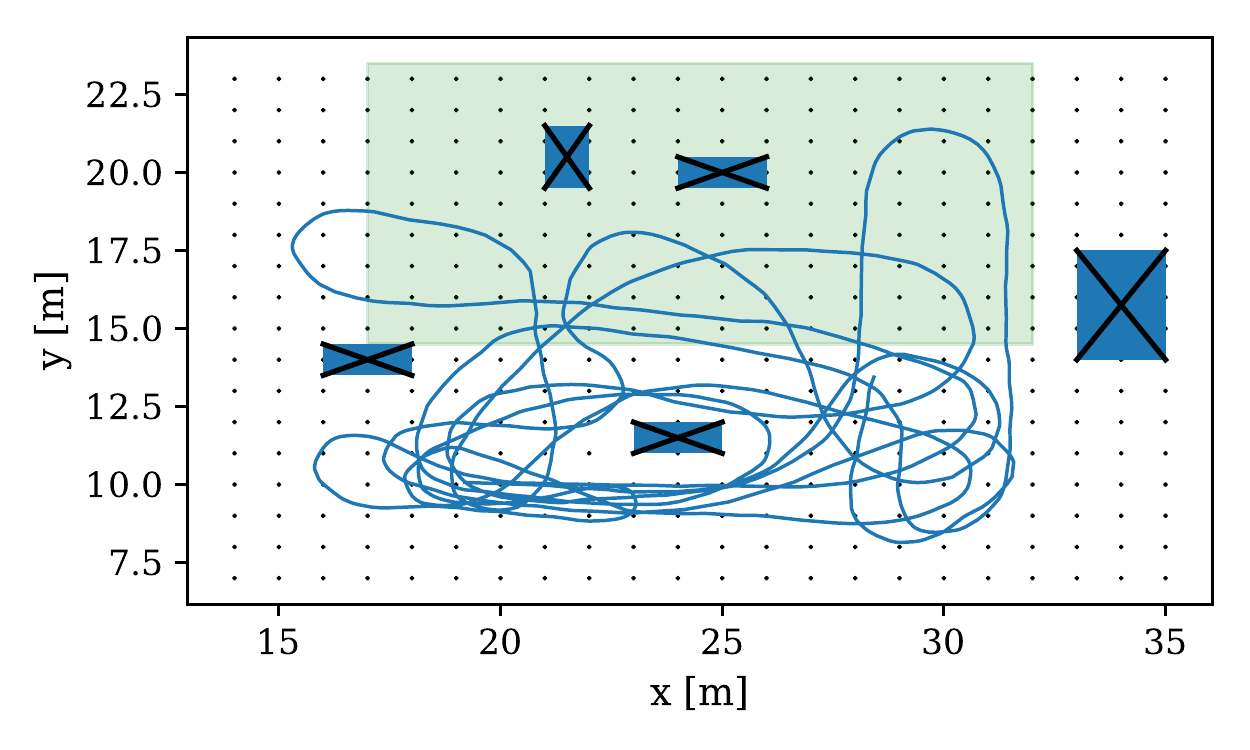}
    	\subcaption{Testing trajectory 1.}
    	\label{image_traj_test1}
    \end{minipage}
    \hfill
	\begin{minipage}[b]{0.493\linewidth}
        \centering
    	\includegraphics[trim=10 10 10 0, clip, width=1.0\linewidth]{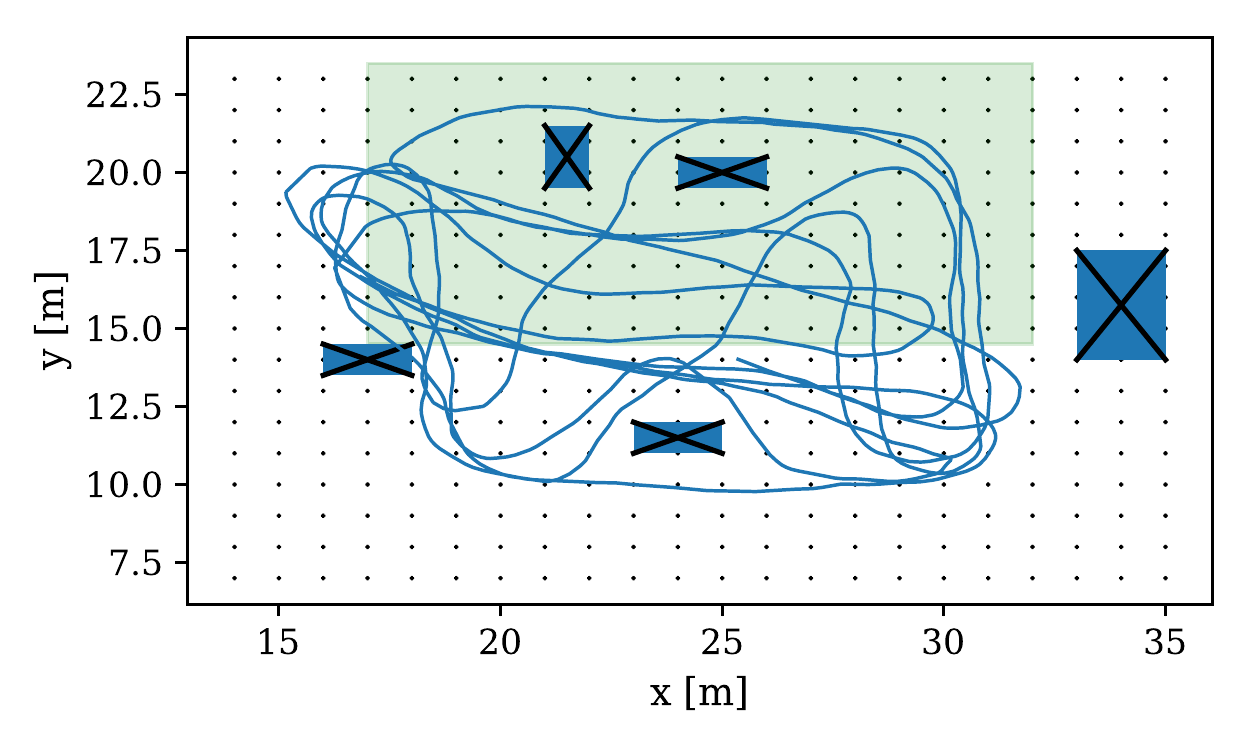}
    	\subcaption{Testing trajectory 2.}
    	\label{image_traj_test2}
    \end{minipage}
    \caption{Training and test trajectories of the IndustryVI dataset.}
    \label{image_industry_trajectories}
\end{figure}

\noindent Given that the EuRoC MAV dataset was captured in an industrial environment but its dynamics of the MAV are distinct from the dynamics of many robotic or handheld systems, and the PennCOSYVIO dataset was recorded in an environment that is different to industrial circumstances, we have recorded a novel dataset in a large-scale industrial environment. The environment is similar to \cite{ott, loeffler}. The visual-only Warehouse~\cite{loeffler} dataset (Industry scenario \#1) was captured utilizing a (robot-like) positioning system, and its eight diverse testing scenarios offer opportunities for assessments of generalizability, volatility, and scale transition. The visual-only Industry scenario \#2~\cite{ott} dataset allows an evaluation for different camera angels. The visual-only Industry scenario \#3~\cite{ott} dataset was captured on a forklift truck to evaluate for high motion dynamics. As these datasets cover only image data, we record and publish the IndustryVI (scenario \#4) dataset. Our environment covers an area of $1,320m^2$, and contains five large black absorber walls and several smaller objects (see Figure~\ref{image_industry_env}). We built a handheld device with an Orbbec3D camera with a RGB image resolution of $640 \times 480$ pixels and a recording frequency of 23\,Hz with an integrated IMU at 140\,Hz. Exemplary images are shown in Figure~\ref{image_industry_ex_images}. We use a high-precision ($< 1mm$) motion capture system for measuring reference poses at 140\,Hz. We let two persons randomly walk in the environment. Trajectories are shown in Figure~\ref{image_industry_trajectories}, where training (a) and testing (c) trajectory 1 is from person 1, and training (b) and testing (d) trajectory 2 is from person 2. This results in a total of 55,973 image and 340,620 IMU training samples and 13,990 image and 85,120 IMU test samples. We cross-validate the training and test trajectories. This allows an evaluation between different motion dynamics in a large-scale environment with texture-less ambiguous elements.

\section{Experimental Results}
\label{chap_evaluation}

\begin{table*}
\begin{center}
\setlength{\tabcolsep}{3.5pt}
    \caption{Median absolute $\mathbf{p} (m/^{\circ})$ and relative $\Delta \mathbf{p} (\Delta m/\Delta^{\circ})$ pose estimation results for the EuRoC MAV~\cite{burri} dataset.}
    \label{table_results1}
    \footnotesize \begin{tabular}{ p{0.5cm} | p{0.5cm} | p{0.5cm} | p{0.5cm} | p{0.5cm} | p{0.5cm} | p{0.5cm} | p{0.5cm} | p{0.5cm} | p{0.5cm} | p{0.5cm} | p{0.5cm} | p{0.5cm} | p{0.5cm} | p{0.5cm} | p{0.5cm} | p{0.5cm} }
    \multicolumn{1}{l|}{\textbf{Bold} are best (min) results.} & \multicolumn{4}{c|}{\textbf{MH-02-easy}} & \multicolumn{4}{c}{\textbf{MH-04-difficult}} \\
    \multicolumn{1}{l|}{\dotuline{Dotted} are worst (max) results.} & \multicolumn{2}{c|}{$\mathbf{p} (m/^{\circ})$} & \multicolumn{2}{c|}{$\Delta \mathbf{p} (\Delta m/\Delta^{\circ})$} & \multicolumn{2}{c|}{$\mathbf{p} (m/^{\circ})$} & \multicolumn{2}{c}{$\Delta \mathbf{p} (\Delta m/\Delta^{\circ})$} \\
    \multicolumn{1}{c|}{\textbf{Method}} & \multicolumn{1}{c}{$e_{\text{med,p}}$} & \multicolumn{1}{c|}{$e_{\text{med,q}}$} & \multicolumn{1}{c}{$\Delta e_{\text{med,p}}$} & \multicolumn{1}{c|}{$\Delta e_{\text{med,q}}$} & \multicolumn{1}{c}{$e_{\text{med,p}}$} & \multicolumn{1}{c|}{$e_{\text{med,q}}$} & \multicolumn{1}{c}{$\Delta e_{\text{med,p}}$} & \multicolumn{1}{c}{$\Delta e_{\text{med,q}}$} \\ \hline
    \multicolumn{1}{l|}{$\text{APR}_{\text{V}}$: PoseNet~\cite{szegedy}} & \multicolumn{1}{r}{\dotuline{0.9249}} & \multicolumn{1}{c|}{3.1718} & \multicolumn{1}{c}{-} & \multicolumn{1}{c|}{-} & \multicolumn{1}{r}{0.9405} & \multicolumn{1}{r|}{3.1086} & \multicolumn{1}{c}{-} & \multicolumn{1}{c}{-} \\
    \multicolumn{1}{l|}{$\text{RPR}_{\text{I}}$: IMUNet~\cite{silva}} & \multicolumn{1}{c}{-} & \multicolumn{1}{c|}{-} & \multicolumn{1}{r}{0.0276} & \multicolumn{1}{r|}{0.0741} & \multicolumn{1}{c}{-} & \multicolumn{1}{c|}{-} & \multicolumn{1}{r}{0.0310} & \multicolumn{1}{r}{0.1073} \\
    \multicolumn{1}{l|}{MapNet~\cite{brahmbhatt}} & \multicolumn{1}{r}{0.9859} & \multicolumn{1}{r|}{3.1879} & \multicolumn{1}{c}{-} & \multicolumn{1}{c|}{-} & \multicolumn{1}{r}{0.9905} & \multicolumn{1}{r|}{\dotuline{3.1898}} & \multicolumn{1}{c}{-} & \multicolumn{1}{c}{-} \\
    \multicolumn{1}{l|}{MapNet+PGO~\cite{brahmbhatt}} & \multicolumn{1}{r}{0.9435} & \multicolumn{1}{r|}{3.1656} & \multicolumn{1}{c}{-} & \multicolumn{1}{c|}{-} & \multicolumn{1}{r}{0.9690} & \multicolumn{1}{r|}{3.1156} & \multicolumn{1}{c}{-} & \multicolumn{1}{c}{-} \\
    \multicolumn{1}{l|}{$\text{APR}_{\text{V}}$-$\text{RPR}_{\text{I}}$+PGO~\cite{matthew}} & \multicolumn{1}{r}{\textbf{0.6914}} & \multicolumn{1}{r|}{3.1050} & \multicolumn{1}{r}{0.0350} & \multicolumn{1}{r|}{0.4124} & \multicolumn{1}{r}{\textbf{0.7211}} & \multicolumn{1}{r|}{3.1532} & \multicolumn{1}{r}{0.0352} & \multicolumn{1}{r}{0.4111} \\
    \multicolumn{1}{l|}{Late Fusion (concat)} & \multicolumn{1}{r}{0.9079} & \multicolumn{1}{r|}{3.1232} & \multicolumn{1}{r}{0.0381} & \multicolumn{1}{r|}{0.6801} & \multicolumn{1}{r}{0.9777} & \multicolumn{1}{r|}{3.1134} & \multicolumn{1}{r}{0.0483} & \multicolumn{1}{r}{0.5832} \\
    \multicolumn{1}{l|}{Late Fusion (concat) + BiLSTM} & \multicolumn{1}{r}{0.7902} & \multicolumn{1}{r|}{3.1452} & \multicolumn{1}{r}{0.0299} & \multicolumn{1}{r|}{0.5062} & \multicolumn{1}{r}{0.8391} & \multicolumn{1}{r|}{3.1164} & \multicolumn{1}{r}{0.0471} & \multicolumn{1}{r}{0.5604} \\
    \multicolumn{1}{l|}{Late Fusion (SSF) \cite{chen}} & \multicolumn{1}{r}{0.9739} & \multicolumn{1}{r|}{3.1744} & \multicolumn{1}{r}{\dotuline{0.0567}} & \multicolumn{1}{r|}{\dotuline{0.6850}} & \multicolumn{1}{r}{0.9254} & \multicolumn{1}{r|}{3.1164} & \multicolumn{1}{r}{0.0453} & \multicolumn{1}{r}{\dotuline{0.9805}} \\
    \multicolumn{1}{l|}{Late Fusion (SSF) \cite{chen} + BiLSTM} & \multicolumn{1}{r}{0.8114} & \multicolumn{1}{r|}{3.1538} & \multicolumn{1}{r}{0.0296} & \multicolumn{1}{r|}{0.5670} & \multicolumn{1}{r}{0.7862} & \multicolumn{1}{r|}{3.1277} & \multicolumn{1}{r}{0.0240} & \multicolumn{1}{r}{0.5690} \\
    \multicolumn{1}{l|}{MMTM~\cite{joze} (3 modules)} & \multicolumn{1}{r}{0.8356} & \multicolumn{1}{r|}{3.1782} & \multicolumn{1}{r}{\textbf{0.0194}} & \multicolumn{1}{r|}{\textbf{0.0601}} & \multicolumn{1}{r}{\dotuline{1.1218}} & \multicolumn{1}{r|}{3.1207} & \multicolumn{1}{r}{\textbf{0.0202}} & \multicolumn{1}{r}{\textbf{0.0800}} \\
    \multicolumn{1}{l|}{AuxiLearn (non-linear) \cite{navon_aux}} & \multicolumn{1}{r}{0.8612} & \multicolumn{1}{r|}{\textbf{3.0266}} & \multicolumn{1}{r}{0.0371} & \multicolumn{1}{r|}{0.5671} & \multicolumn{1}{r}{0.7979} & \multicolumn{1}{r|}{3.0996} & \multicolumn{1}{r}{\dotuline{0.0631}} & \multicolumn{1}{r}{0.5851} \\
    \multicolumn{1}{l|}{AuxiLearn (convolutional) \cite{navon_aux}} & \multicolumn{1}{r}{0.9050} & \multicolumn{1}{r|}{\dotuline{3.1984}} & \multicolumn{1}{r}{0.0371} & \multicolumn{1}{r|}{0.5561} & \multicolumn{1}{r}{0.8711} & \multicolumn{1}{r|}{3.1047} & \multicolumn{1}{r}{0.0612} & \multicolumn{1}{r}{0.5873} \\
    \multicolumn{1}{l|}{BNN~\cite{kendall_uncertainty} + Late Fusion} & \multicolumn{1}{r}{0.7925} & \multicolumn{1}{r|}{3.1878} & \multicolumn{1}{c}{-} & \multicolumn{1}{c|}{-} & \multicolumn{1}{r}{0.8523} & \multicolumn{1}{r|}{\textbf{3.0285}} & \multicolumn{1}{c}{-} & \multicolumn{1}{c}{-} \\
    \end{tabular}
\end{center}
\end{table*}

\begin{table*}
\begin{center}
\setlength{\tabcolsep}{3.5pt}
    \caption{Median absolute $\mathbf{p} (m/^{\circ})$ and relative $\Delta \mathbf{p} (\Delta m/\Delta^{\circ})$ pose estimation results for the PennCOSYVIO~\cite{pfrommer} dataset.}
    \label{table_results2}
    \footnotesize \begin{tabular}{ p{0.5cm} | p{0.5cm} | p{0.5cm} | p{0.5cm} | p{0.5cm} | p{0.5cm} | p{0.5cm} | p{0.5cm} | p{0.5cm} | p{0.5cm} | p{0.5cm} | p{0.5cm} | p{0.5cm} | p{0.5cm} | p{0.5cm} | p{0.5cm} | p{0.5cm} | p{0.5cm} | p{0.5cm} | p{0.5cm} | p{0.5cm} }
    \multicolumn{1}{l|}{\textbf{Bold} are best (min) results.} & \multicolumn{4}{c|}{\textbf{BF}} & \multicolumn{4}{c}{\textbf{BS}} \\
    \multicolumn{1}{l|}{\dotuline{Dotted} are worst (max) results.} & \multicolumn{2}{c|}{$\mathbf{p} (m/^{\circ})$} & \multicolumn{2}{c|}{$\Delta \mathbf{p} (\Delta m/\Delta^{\circ})$} & \multicolumn{2}{c|}{$\mathbf{p} (m/^{\circ})$} & \multicolumn{2}{c}{$\Delta \mathbf{p} (\Delta m/\Delta^{\circ})$} \\
    \multicolumn{1}{c|}{\textbf{Method}} & \multicolumn{1}{c}{$e_{\text{med,p}}$} & \multicolumn{1}{c|}{$e_{\text{med,q}}$} & \multicolumn{1}{c}{$\Delta e_{\text{med,p}}$} & \multicolumn{1}{c|}{$\Delta e_{\text{med,q}}$} & \multicolumn{1}{c}{$e_{\text{med,p}}$} & \multicolumn{1}{c|}{$e_{\text{med,q}}$} & \multicolumn{1}{c}{$\Delta e_{\text{med,p}}$} & \multicolumn{1}{c}{$\Delta e_{\text{med,q}}$} \\ \hline
    \multicolumn{1}{l|}{$\text{APR}_{\text{V}}$: PoseNet~\cite{szegedy}} & \multicolumn{1}{r}{1.8210} & \multicolumn{1}{r|}{3.1129} & \multicolumn{1}{c}{-} & \multicolumn{1}{c|}{-} & \multicolumn{1}{r}{1.4125} & \multicolumn{1}{r|}{3.1411} & \multicolumn{1}{c}{-} & \multicolumn{1}{c}{-} \\
    \multicolumn{1}{l|}{$\text{RPR}_{\text{I}}$: IMUNet~\cite{silva}} & \multicolumn{1}{c}{-} & \multicolumn{1}{c|}{-} & \multicolumn{1}{r}{0.1091} & \multicolumn{1}{r|}{\dotuline{1.0573}} & \multicolumn{1}{c}{-} & \multicolumn{1}{c|}{-} & \multicolumn{1}{r}{0.0393} & \multicolumn{1}{r}{\textbf{0.5714}} \\
    \multicolumn{1}{l|}{MapNet~\cite{brahmbhatt}} & \multicolumn{1}{r}{3.3017} & \multicolumn{1}{r|}{3.1146} & \multicolumn{1}{c}{-} & \multicolumn{1}{c|}{-} & \multicolumn{1}{r}{3.2557} & \multicolumn{1}{r|}{3.1317} & \multicolumn{1}{c}{-} & \multicolumn{1}{c}{-} \\
    \multicolumn{1}{l|}{MapNet+PGO~\cite{brahmbhatt}} & \multicolumn{1}{r}{\dotuline{3.4130}} & \multicolumn{1}{r|}{3.1211} & \multicolumn{1}{c}{-} & \multicolumn{1}{c|}{-} & \multicolumn{1}{r}{\dotuline{3.8911}} & \multicolumn{1}{r|}{3.1412} & \multicolumn{1}{c}{-} & \multicolumn{1}{c}{-} \\
    \multicolumn{1}{l|}{$\text{APR}_{\text{V}}$-$\text{RPR}_{\text{I}}$+PGO~\cite{matthew}} & \multicolumn{1}{r}{2.5563} & \multicolumn{1}{r|}{3.1016} & \multicolumn{1}{r}{0.0402} & \multicolumn{1}{r|}{0.7134} & \multicolumn{1}{r}{2.3142} & \multicolumn{1}{r|}{3.1360} & \multicolumn{1}{r}{\textbf{0.0200}} & \multicolumn{1}{r}{0.7099} \\
    \multicolumn{1}{l|}{Late Fusion (concat)} & \multicolumn{1}{r}{2.2365} & \multicolumn{1}{r|}{3.1028} & \multicolumn{1}{r}{0.0385} & \multicolumn{1}{r|}{0.8305} & \multicolumn{1}{r}{2.0696} & \multicolumn{1}{r|}{3.1390} & \multicolumn{1}{r}{\dotuline{0.1013}} & \multicolumn{1}{r}{0.9348} \\
    \multicolumn{1}{l|}{Late Fusion (concat) + BiLSTM} & \multicolumn{1}{r}{1.6543} & \multicolumn{1}{r|}{3.0962} & \multicolumn{1}{r}{0.0281} & \multicolumn{1}{r|}{0.8162} & \multicolumn{1}{r}{1.7389} & \multicolumn{1}{r|}{3.1309} & \multicolumn{1}{r}{0.0974} & \multicolumn{1}{r}{1.0773} \\
    \multicolumn{1}{l|}{Late Fusion (SSF) \cite{chen}} & \multicolumn{1}{r}{1.8693} & \multicolumn{1}{r|}{3.1021} & \multicolumn{1}{r}{0.0321} & \multicolumn{1}{r|}{0.8213} & \multicolumn{1}{r}{1.6552} & \multicolumn{1}{r|}{3.1356} & \multicolumn{1}{r}{0.0863} & \multicolumn{1}{r}{0.8762} \\
    \multicolumn{1}{l|}{Late Fusion (SSF) \cite{chen} + BiLSTM} & \multicolumn{1}{r}{1.1249} & \multicolumn{1}{r|}{3.1037} & \multicolumn{1}{r}{\textbf{0.0180}} & \multicolumn{1}{r|}{0.7571} & \multicolumn{1}{r}{1.2341} & \multicolumn{1}{r|}{3.1287} & \multicolumn{1}{r}{0.0291} & \multicolumn{1}{r}{0.8123} \\
    \multicolumn{1}{l|}{MMTM~\cite{joze} (3 modules)} & \multicolumn{1}{r}{\textbf{1.0557}} & \multicolumn{1}{r|}{\dotuline{3.1378}} & \multicolumn{1}{r}{0.0328} & \multicolumn{1}{r|}{\textbf{0.6695}} & \multicolumn{1}{r}{\textbf{1.1980}} & \multicolumn{1}{r|}{\textbf{3.1008}} & \multicolumn{1}{r}{0.0976} & \multicolumn{1}{r}{0.9073} \\
    \multicolumn{1}{l|}{AuxiLearn (non-linear) \cite{navon_aux}} & \multicolumn{1}{r}{1.5402} & \multicolumn{1}{r|}{\textbf{3.0944}} & \multicolumn{1}{r}{0.0410} & \multicolumn{1}{r|}{1.0195} & \multicolumn{1}{r}{1.3008} & \multicolumn{1}{r|}{3.1397} & \multicolumn{1}{r}{0.0525} & \multicolumn{1}{r}{1.0881} \\
    \multicolumn{1}{l|}{AuxiLearn (convolutional) \cite{navon_aux}} & \multicolumn{1}{r}{1.8931} & \multicolumn{1}{r|}{3.1098} & \multicolumn{1}{r}{0.0451} & \multicolumn{1}{r|}{1.0220} & \multicolumn{1}{r}{1.8964} & \multicolumn{1}{r|}{3.1401} & \multicolumn{1}{r}{0.1006} & \multicolumn{1}{r}{\dotuline{1.1020}} \\
    \multicolumn{1}{l|}{BNN~\cite{kendall_uncertainty} + Late Fusion} & \multicolumn{1}{r}{2.1110} & \multicolumn{1}{r|}{3.1136} & \multicolumn{1}{c}{-} & \multicolumn{1}{c|}{-} & \multicolumn{1}{r}{1.6569} & \multicolumn{1}{r|}{\dotuline{3.1450}} & \multicolumn{1}{c}{-} & \multicolumn{1}{c}{-} \\
    \end{tabular}
\end{center}
\end{table*}

\textbf{Hardware \& Training Setup.} For all experiments, we use Nvidia Tesla V100-SXM2 GPUs with 32 GB VRAM equipped with Core Xeon CPUs and 192 GB RAM. We use the Adam optimizer with a learning rate of $10^{-4}$. We run each experiment for 1,000 epochs with a batch size of 50 and report results for the best epoch.

\textbf{Evaluation Metrics.} For the evaluation of the APR, we report the median absolute position $e_{\text{med,p}}$ in $m$ and the median absolute orientation $e_{\text{med,q}}$ in $^{\circ}$, and the median relative position $\Delta e_{\text{med,p}}$ in $\Delta m$ and the median relative orientation $\Delta e_{\text{med,q}}$ in $\Delta^{\circ}$. As the global consistency of the estimated trajectory is an important quantity and to compare our relative pose prediction with state-of-the-art techniques, we additionally report the absolute trajectory error (ATE) by aligning the estimated trajectory $\mathbf{P}_{1:n}$ and the ground truth trajectory $\mathbf{Q}_{1:n}$ using the method of Horn~\cite{horn}. The ATE at time step $i$ can be computed as $\mathbf{F}_i := \mathbf{Q}_{i}^{-1}\mathbf{S}\mathbf{P}_i$ with the rigid-body transformation $\mathbf{S}$ corresponding to the least-squares solution that maps $\mathbf{P}_{1:n}$ onto $\mathbf{Q}_{1:n}$. Next, we compute the root mean squared error over all time steps of the translational components by
\begin{equation}
\label{eq_ate}
    e_{\text{ATE,p}}(\mathbf{F}_{1:n}) := \Big(\frac{1}{n}\sum_{i=1}^{n} ||\text{trans}(\mathbf{F}_i)||^2\Big)^{\frac{1}{2}}.
\end{equation}
To compare our RPR results with the results proposed by \cite{silva,chen}, we use the absolute translational error (ATLE) \cite{silva} for the position $\Delta e_{\text{ATLE,p}}$ in $m$.

\subsection{Evaluation of $\text{APR}_{\text{V}}$-$\text{RPR}_{\text{I}}$ Fusion Methods}
\label{chap_eval_apr}

\noindent We provide a quantitative evaluation of all $\text{APR}_{\text{V}}$-$\text{RPR}_{\text{I}}$ fusion methods for the EuRoC MAV (Table~\ref{table_results1}), the PennCOSYVIO (Table~\ref{table_results2}), and the IndustryVI (Table~\ref{table_results3}) datasets. For an overview of APR trajectory comparisons, see Figure~\ref{image_app_euroc_mh02} to Figure~\ref{image_app_industry2} in the appendix.

\textbf{Baseline Results, MapNet, and PGO.} We evaluate the results for MapNet~\cite{brahmbhatt}, PGO, and PGO for $\text{APR}_{\text{V}}$-$\text{RPR}_{\text{I}}$ fusion, and compare the results to the baseline methods. The $\text{APR}_{\text{V}}$ baseline yields proper results of 0.9249\textit{m} and 0.9405\textit{m} on the small-scale environment of EuRoC MAV, and $\text{RPR}_{\text{I}}$ yields small errors of 2.76\textit{cm} and 3.1\textit{cm} (even at fast dynamics of the MAV). This increases for the large-scale area of PennCOSYVIO and IndustryVI. The relative error of $\text{RPR}_{\text{I}}$ (below 3.0\textit{cm}) is low for the fast movements of the IndustryVI dataset. While MapNet and MapNet+PGO (see Section~\ref{section_pgo_for_apr}) increase the $\text{APR}_{\text{V}}$ baseline results for the EuRoC MAV and PennCOSYVIO datasets and most sequences of the IndustryVI dataset, our implementation of $\text{APR}_{\text{V}}$-$\text{RPR}_{\text{I}}$ fusion utilizing PGO (see Section~\ref{section_pgo_for_fusion}) yields notably lower results on the EuRoC MAV dataset (e.g., 0.6914\textit{m} on the MH-02 sequence compared to 0.9249\textit{m} of $\text{APR}_{\text{V}}$-only) and the train 2 dataset of IndustryVI. For the PennCOSYVIO dataset, the $\text{APR}_{\text{V}}$-$\text{RPR}_{\text{I}}$ fusion utilizing MapNet+PGO cannot outperform PoseNet. This contradicts the results of MapNet and MapNet+PGO on the 7-Scenes~\cite{shotton_scene} dataset proposed in \cite{brahmbhatt}, where MapNet+PGO significantly improves the PoseNet results. Given that MapNet and MapNet+PGO are time-distributed networks, it is possible to enhance their performance by increasing the training time steps and incorporating larger skip sizes between consecutive images.

\begin{table*}
\begin{center}
\setlength{\tabcolsep}{1.5pt}
    \caption{Median absolute $\mathbf{p} (m/^{\circ})$ and relative $\Delta \mathbf{p} (\Delta m/\Delta^{\circ})$ pose estimation results for the IndustryVI dataset.}
    \label{table_results3}
    \footnotesize \begin{tabular}{ p{3.4cm} | p{0.5cm} | p{0.5cm} | p{0.5cm} | p{0.5cm} | p{0.5cm} | p{0.5cm} | p{0.5cm} | p{0.5cm} | p{0.5cm} | p{0.5cm} | p{0.5cm} | p{0.5cm} | p{0.5cm} | p{0.5cm} | p{0.5cm} | p{0.5cm} | p{0.5cm} | p{0.5cm} | p{0.5cm} | p{0.5cm} | p{0.5cm} | p{0.5cm} | p{0.5cm} | p{0.5cm} | p{0.5cm} | p{0.5cm} | p{0.5cm} | p{0.5cm} }
    \multicolumn{1}{l|}{\textbf{Bold} are best (min) results.} & \multicolumn{4}{c|}{\textbf{Train 1, Test 1}} & \multicolumn{4}{c|}{\textbf{Train 1, Test 2}} & \multicolumn{4}{c|}{\textbf{Train 2, Test 1}} & \multicolumn{4}{c}{\textbf{Train 2, Test 2}} \\
    \multicolumn{1}{l|}{\dotuline{Dotted} are worst (max) results.} & \multicolumn{2}{c|}{$\mathbf{p} (m/^{\circ})$} & \multicolumn{2}{c|}{$\Delta \mathbf{p} (\Delta m/\Delta^{\circ})$} & \multicolumn{2}{c|}{$\mathbf{p} (m/^{\circ})$} & \multicolumn{2}{c|}{$\Delta \mathbf{p} (\Delta m/\Delta^{\circ})$} & \multicolumn{2}{c|}{$\mathbf{p} (m/^{\circ})$} & \multicolumn{2}{c|}{$\Delta \mathbf{p} (\Delta m/\Delta^{\circ})$} & \multicolumn{2}{c|}{$\mathbf{p} (m/^{\circ})$} & \multicolumn{2}{c}{$\Delta \mathbf{p} (\Delta m/\Delta^{\circ})$} \\
    \multicolumn{1}{c|}{\textbf{Method}} & \multicolumn{1}{c}{\scriptsize$e_{\text{med,p}}$} & \multicolumn{1}{c|}{\scriptsize$e_{\text{med,q}}$} & \multicolumn{1}{c}{\scriptsize$\Delta e_{\text{med,p}}$} & \multicolumn{1}{c|}{\scriptsize$\Delta e_{\text{med,q}}$} & \multicolumn{1}{c}{\scriptsize$e_{\text{med,p}}$} & \multicolumn{1}{c|}{\scriptsize$e_{\text{med,q}}$} & \multicolumn{1}{c}{\scriptsize$\Delta e_{\text{med,p}}$} & \multicolumn{1}{c|}{\scriptsize$\Delta e_{\text{med,q}}$} & \multicolumn{1}{c}{\scriptsize$e_{\text{med,p}}$} & \multicolumn{1}{c|}{\scriptsize$e_{\text{med,q}}$} & \multicolumn{1}{c}{\scriptsize$\Delta e_{\text{med,p}}$} & \multicolumn{1}{c|}{\scriptsize$\Delta e_{\text{med,q}}$} & \multicolumn{1}{c}{\scriptsize$e_{\text{med,p}}$} & \multicolumn{1}{c|}{\scriptsize$e_{\text{med,q}}$} & \multicolumn{1}{c}{\scriptsize$\Delta e_{\text{med,p}}$} & \multicolumn{1}{c}{\scriptsize$\Delta e_{\text{med,q}}$} \\ \hline
    $\text{APR}_{\text{V}}$: PoseNet~\cite{szegedy} & \multicolumn{1}{r}{1.8231} & \multicolumn{1}{r|}{9.6742} & \multicolumn{1}{c}{-} & \multicolumn{1}{c|}{-} & \multicolumn{1}{r}{1.6429} & \multicolumn{1}{r|}{7.8564} & \multicolumn{1}{c}{-} & \multicolumn{1}{c|}{-} & \multicolumn{1}{r}{\dotuline{1.9345}} & \multicolumn{1}{r|}{6.2314} & \multicolumn{1}{c}{-} & \multicolumn{1}{c|}{-} & \multicolumn{1}{r}{1.6438} & \multicolumn{1}{r|}{7.3452} & \multicolumn{1}{c}{-} & \multicolumn{1}{c}{-} \\
    $\text{RPR}_{\text{I}}$: IMUNet~\cite{silva} & \multicolumn{1}{c}{-} & \multicolumn{1}{c|}{-} & \multicolumn{1}{r}{0.0300} & \multicolumn{1}{r|}{0.8867} & \multicolumn{1}{c}{-} & \multicolumn{1}{c|}{-} & \multicolumn{1}{r}{0.0295} & \multicolumn{1}{r|}{\textbf{0.7743}} & \multicolumn{1}{c}{-} & \multicolumn{1}{c|}{-} & \multicolumn{1}{r}{0.0278} & \multicolumn{1}{r|}{\textbf{0.6134}} & \multicolumn{1}{c}{-} & \multicolumn{1}{c|}{-} & \multicolumn{1}{r}{0.0265} & \multicolumn{1}{r}{0.9641} \\
    MapNet~\cite{brahmbhatt} & \multicolumn{1}{r}{\dotuline{1.9012}} & \multicolumn{1}{r|}{\dotuline{10.120}} & \multicolumn{1}{c}{-} & \multicolumn{1}{c|}{-} & \multicolumn{1}{r}{1.8990} & \multicolumn{1}{r|}{8.1803} & \multicolumn{1}{c}{-} & \multicolumn{1}{c|}{-} & \multicolumn{1}{r}{1.9019} & \multicolumn{1}{r|}{6.9678} & \multicolumn{1}{c}{-} & \multicolumn{1}{c|}{-} & \multicolumn{1}{r}{\dotuline{1.6891}} & \multicolumn{1}{r|}{8.4012} & \multicolumn{1}{c}{-} & \multicolumn{1}{c}{-} \\
    MapNet+PGO~\cite{brahmbhatt} & \multicolumn{1}{r}{1.8865} & \multicolumn{1}{r|}{9.9905} & \multicolumn{1}{c}{-} & \multicolumn{1}{c|}{-} & \multicolumn{1}{r}{1.8126} & \multicolumn{1}{r|}{7.9078} & \multicolumn{1}{c}{-} & \multicolumn{1}{c|}{-} & \multicolumn{1}{r}{1.8991} & \multicolumn{1}{r|}{6.9120} & \multicolumn{1}{c}{-} & \multicolumn{1}{c|}{-} & \multicolumn{1}{r}{1.6694} & \multicolumn{1}{r|}{8.1021} & \multicolumn{1}{c}{-} & \multicolumn{1}{c}{-} \\
    $\text{APR}_{\text{V}}$-$\text{RPR}_{\text{I}}$+PGO~\cite{matthew} & \multicolumn{1}{r}{1.8681} & \multicolumn{1}{r|}{9.8901} & \multicolumn{1}{r}{\dotuline{0.0481}} & \multicolumn{1}{r|}{0.9841} & \multicolumn{1}{r}{1.7106} & \multicolumn{1}{r|}{7.1193} & \multicolumn{1}{r}{\dotuline{0.0458}} & \multicolumn{1}{r|}{0.7925} & \multicolumn{1}{r}{1.8379} & \multicolumn{1}{r|}{6.2731} & \multicolumn{1}{r}{\dotuline{0.0419}} & \multicolumn{1}{r|}{0.9251} & \multicolumn{1}{r}{1.6234} & \multicolumn{1}{r|}{7.8351} & \multicolumn{1}{r}{0.0413} & \multicolumn{1}{r}{0.9811} \\
    Late Fusion (concat) & \multicolumn{1}{r}{1.8945} & \multicolumn{1}{r|}{9.8976} & \multicolumn{1}{r}{0.0342} & \multicolumn{1}{r|}{\dotuline{0.9995}} & \multicolumn{1}{r}{\dotuline{1.9016}} & \multicolumn{1}{r|}{\dotuline{8.2087}} & \multicolumn{1}{r}{0.0335} & \multicolumn{1}{r|}{0.8121} & \multicolumn{1}{r}{1.8154} & \multicolumn{1}{r|}{7.7210} & \multicolumn{1}{r}{0.0400} & \multicolumn{1}{r|}{1.1301} & \multicolumn{1}{r}{1.5342} & \multicolumn{1}{r|}{\dotuline{8.6910}} & \multicolumn{1}{r}{0.2107} & \multicolumn{1}{r}{1.2004}\\
    \multicolumn{1}{r|}{+ BiLSTM} & \multicolumn{1}{r}{\textbf{1.6014}} & \multicolumn{1}{r|}{\textbf{5.1032}} & \multicolumn{1}{r}{0.0301} & \multicolumn{1}{r|}{0.9012} & \multicolumn{1}{r}{1.6521} & \multicolumn{1}{r|}{5.8724} & \multicolumn{1}{r}{0.0271} & \multicolumn{1}{r|}{0.8610} & \multicolumn{1}{r}{\textbf{1.6231}} & \multicolumn{1}{r|}{6.0013} & \multicolumn{1}{r}{\textbf{0.0242}} & \multicolumn{1}{r|}{0.9221} & \multicolumn{1}{r}{\textbf{1.3092}} & \multicolumn{1}{r|}{6.0123} & \multicolumn{1}{r}{0.0254} & \multicolumn{1}{r}{\textbf{0.7449}} \\
    Late Fusion (SSF)~\cite{chen} & \multicolumn{1}{r}{1.7613} & \multicolumn{1}{r|}{9.5848} & \multicolumn{1}{r}{0.0315} & \multicolumn{1}{r|}{0.9491} & \multicolumn{1}{r}{1.8616} & \multicolumn{1}{r|}{7.1069} & \multicolumn{1}{r}{0.0321} & \multicolumn{1}{r|}{0.7867} & \multicolumn{1}{r}{1.7823} & \multicolumn{1}{r|}{7.3210} & \multicolumn{1}{r}{0.0381} & \multicolumn{1}{r|}{\dotuline{1.2031}} & \multicolumn{1}{r}{1.5005} & \multicolumn{1}{r|}{8.2451} & \multicolumn{1}{r}{0.1097} & \multicolumn{1}{r}{\dotuline{1.3164}} \\
    \multicolumn{1}{r|}{+ BiLSTM} & \multicolumn{1}{r}{1.5875} & \multicolumn{1}{r|}{5.2016} & \multicolumn{1}{r}{\textbf{0.0214}} & \multicolumn{1}{r|}{0.9823} & \multicolumn{1}{r}{\textbf{1.5216}} & \multicolumn{1}{r|}{\textbf{5.7630}} & \multicolumn{1}{r}{\textbf{0.0260}} & \multicolumn{1}{r|}{0.7967} & \multicolumn{1}{r}{1.6823} & \multicolumn{1}{r|}{5.4601} & \multicolumn{1}{r}{0.0278} & \multicolumn{1}{r|}{0.9231} & \multicolumn{1}{r}{1.3215} & \multicolumn{1}{r|}{5.0291} & \multicolumn{1}{r}{0.0257} & \multicolumn{1}{r}{0.9164} \\
    MMTM~\cite{joze} (3 modules) & \multicolumn{1}{r}{1.6550} & \multicolumn{1}{r|}{5.2045} & \multicolumn{1}{r}{0.0374} & \multicolumn{1}{r|}{\textbf{0.8179}} & \multicolumn{1}{r}{1.5775} & \multicolumn{1}{r|}{5.8832} & \multicolumn{1}{r}{0.0443} & \multicolumn{1}{r|}{\dotuline{1.0797}} & \multicolumn{1}{r}{1.7836} & \multicolumn{1}{r|}{\textbf{5.6034}} & \multicolumn{1}{r}{0.0378} & \multicolumn{1}{r|}{0.9709} & \multicolumn{1}{r}{1.4840} & \multicolumn{1}{r|}{5.2451} & \multicolumn{1}{r}{0.0401} & \multicolumn{1}{r}{0.8726} \\
    AuxiLearn (non-linear)~\cite{navon_aux} & \multicolumn{1}{r}{1.6845} & \multicolumn{1}{r|}{6.2312} & \multicolumn{1}{r}{0.0310} & \multicolumn{1}{r|}{0.9931} & \multicolumn{1}{r}{1.6140} & \multicolumn{1}{r|}{5.8834} & \multicolumn{1}{r}{0.0312} & \multicolumn{1}{r|}{0.8127} & \multicolumn{1}{r}{1.6941} & \multicolumn{1}{r|}{5.8848} & \multicolumn{1}{r}{0.0305} & \multicolumn{1}{r|}{1.0867} & \multicolumn{1}{r}{1.3341} & \multicolumn{1}{r|}{\textbf{5.0221}} & \multicolumn{1}{r}{\textbf{0.0213}} & \multicolumn{1}{r}{0.8934} \\
    AuxiLearn (convol.)~\cite{navon_aux} & \multicolumn{1}{r}{1.8180} & \multicolumn{1}{r|}{9.7841} & \multicolumn{1}{r}{0.0332} & \multicolumn{1}{r|}{0.9836} & \multicolumn{1}{r}{1.9011} & \multicolumn{1}{r|}{8.3024} & \multicolumn{1}{r}{0.0367} & \multicolumn{1}{r|}{0.8651} & \multicolumn{1}{r}{1.7983} & \multicolumn{1}{r|}{6.9812} & \multicolumn{1}{r}{0.0361} & \multicolumn{1}{r|}{1.1331} & \multicolumn{1}{r}{1.5670} & \multicolumn{1}{r|}{7.8938} & \multicolumn{1}{r}{\dotuline{0.2207}} & \multicolumn{1}{r}{1.1956} \\
    BNN~\cite{kendall_uncertainty} + Late Fusion & \multicolumn{1}{r}{1.7956} & \multicolumn{1}{r|}{9.7547} & \multicolumn{1}{c}{-} & \multicolumn{1}{c|}{-} & \multicolumn{1}{r}{1.7691} & \multicolumn{1}{r|}{8.1268} & \multicolumn{1}{c}{-} & \multicolumn{1}{c|}{-} & \multicolumn{1}{r}{1.8251} & \multicolumn{1}{r|}{\dotuline{7.3289}} & \multicolumn{1}{c}{-} & \multicolumn{1}{c|}{-} & \multicolumn{1}{r}{1.5476} & \multicolumn{1}{r|}{7.8751} & \multicolumn{1}{c}{-} & \multicolumn{1}{c}{-} \\
    \end{tabular}
\end{center}
\end{table*}

\begin{table*}
\begin{center}
\setlength{\tabcolsep}{2.4pt}
    \caption{Intermediate $\text{APR}_{\text{V}}$-$\text{RPR}_{\text{I}}$ fusion using MMTM~\cite{joze} with separate model optimization for the EuRoC MAV~\cite{burri} dataset. While ``F'' indicates the layer with MMTM fusion (see Section~\ref{section_intermediate_fusion}), "--" indicates no fusion. For a visualization, see Figure~\ref{image_mmtm_comb} in the appendix. \textbf{Bold} are best results. \underline{Underlined} are second best results.}
    \label{table_results_mmtm_modules}
    \footnotesize \begin{tabular}{ p{1.1cm} | p{0.5cm} | p{0.5cm} | p{0.5cm} | p{0.5cm} | p{0.5cm} | p{0.5cm} | p{0.5cm} | p{0.5cm} | p{0.5cm} | p{0.5cm} | p{0.5cm} | p{0.5cm} | p{0.5cm} | p{0.5cm} | p{0.5cm} | p{0.5cm} | p{0.5cm} | p{0.5cm} | p{0.5cm} | p{0.5cm} }
    & \multicolumn{8}{c|}{\textbf{MH-02-easy}} & \multicolumn{8}{c}{\textbf{MH-04-difficult}} \\
    & \multicolumn{4}{c|}{$\mathbf{p} (m/^{\circ})$} & \multicolumn{4}{c|}{$\Delta \mathbf{p} (\Delta m/\Delta^{\circ})$} & \multicolumn{4}{c|}{$\mathbf{p} (m/^{\circ})$} & \multicolumn{4}{c}{$\Delta \mathbf{p} (\Delta m/\Delta^{\circ})$} \\
    \multicolumn{1}{c|}{\textbf{MMTM}} & \multicolumn{1}{c}{$e_{\text{med,p}}$} & \multicolumn{1}{c}{$e_{\text{ATE}}$} & \multicolumn{1}{c}{$e_{\text{ATLE}}$} & \multicolumn{1}{c|}{$e_{\text{med,q}}$} & \multicolumn{1}{c}{$\Delta e_{\text{med,p}}$} & \multicolumn{1}{c}{$e_{\text{ATE}}$} & \multicolumn{1}{c}{$e_{\text{ATLE}}$} & \multicolumn{1}{c|}{$\Delta e_{\text{med,q}}$} & \multicolumn{1}{c}{$e_{\text{med,p}}$} & \multicolumn{1}{c}{$e_{\text{ATE}}$} & \multicolumn{1}{c}{$e_{\text{ATLE}}$} & \multicolumn{1}{c|}{$e_{\text{med,q}}$} & \multicolumn{1}{c}{$\Delta e_{\text{med,p}}$} & \multicolumn{1}{c}{$e_{\text{ATE}}$} & \multicolumn{1}{c}{$e_{\text{ATLE}}$} & \multicolumn{1}{c}{$\Delta e_{\text{med,q}}$} \\ \hline
    \multicolumn{1}{c|}{(F -- --)} & \multicolumn{1}{c}{\underline{0.8399}} & \multicolumn{1}{c}{3.2603} & \multicolumn{1}{c}{0.3111} & \multicolumn{1}{c|}{3.1980} & \multicolumn{1}{c}{0.0205} & \multicolumn{1}{c}{1.7572} & \multicolumn{1}{c}{0.0691} & \multicolumn{1}{c|}{0.0661} & \multicolumn{1}{c}{1.1084} & \multicolumn{1}{c}{3.9618} & \multicolumn{1}{c}{0.4497} & \multicolumn{1}{c|}{\underline{3.1156}} & \multicolumn{1}{c}{0.0249} & \multicolumn{1}{c}{3.7785} & \multicolumn{1}{c}{0.0661} & \multicolumn{1}{c}{0.0818} \\
    \multicolumn{1}{c|}{(-- F --)} & \multicolumn{1}{c}{0.9512} & \multicolumn{1}{c}{3.2789} & \multicolumn{1}{c}{0.3114} & \multicolumn{1}{c|}{3.1940} & \multicolumn{1}{c}{0.0219} & \multicolumn{1}{c}{2.0637} & \multicolumn{1}{c}{0.0705} & \multicolumn{1}{c|}{0.0645} & \multicolumn{1}{c}{1.1098} & \multicolumn{1}{c}{4.3766} & \multicolumn{1}{c}{0.4439} & \multicolumn{1}{c|}{3.1937} & \multicolumn{1}{c}{0.0253} & \multicolumn{1}{c}{\underline{3.2841}} & \multicolumn{1}{c}{0.0649} & \multicolumn{1}{c}{0.0751} \\
    \multicolumn{1}{c|}{(-- -- F)} & \multicolumn{1}{c}{0.8743} & \multicolumn{1}{c}{\textbf{2.5593}} & \multicolumn{1}{c}{0.5797} & \multicolumn{1}{c|}{\underline{3.1795}} & \multicolumn{1}{c}{0.0207} & \multicolumn{1}{c}{2.0813} & \multicolumn{1}{c}{\underline{0.0693}} & \multicolumn{1}{c|}{0.0708} & \multicolumn{1}{c}{\underline{1.0619}} & \multicolumn{1}{c}{3.8419} & \multicolumn{1}{c}{0.7398} & \multicolumn{1}{c|}{\textbf{3.1047}} & \multicolumn{1}{c}{0.0244} & \multicolumn{1}{c}{3.6690} & \multicolumn{1}{c}{0.0658} & \multicolumn{1}{c}{\textbf{0.0686}} \\
    \multicolumn{1}{c|}{(F F --)} & \multicolumn{1}{c}{0.9199} & \multicolumn{1}{c}{3.3279} & \multicolumn{1}{c}{\textbf{0.2792}} & \multicolumn{1}{c|}{3.1689} & \multicolumn{1}{c}{0.0213} & \multicolumn{1}{c}{\textbf{1.3248}} & \multicolumn{1}{c}{0.0696} & \multicolumn{1}{c|}{0.0661} & \multicolumn{1}{c}{\textbf{1.0592}} & \multicolumn{1}{c}{4.1632} & \multicolumn{1}{c}{0.4205} & \multicolumn{1}{c|}{3.1400} & \multicolumn{1}{c}{0.0250} & \multicolumn{1}{c}{3.4385} & \multicolumn{1}{c}{0.0661} & \multicolumn{1}{c}{0.0782} \\
    \multicolumn{1}{c|}{(F -- F)} & \multicolumn{1}{c}{0.9601} & \multicolumn{1}{c}{3.3041} & \multicolumn{1}{c}{0.3090} & \multicolumn{1}{c|}{\textbf{3.1483}} & \multicolumn{1}{c}{\underline{0.0204}} & \multicolumn{1}{c}{\underline{1.5077}} & \multicolumn{1}{c}{0.0710} & \multicolumn{1}{c|}{0.0648} & \multicolumn{1}{c}{1.0994} & \multicolumn{1}{c}{4.3420} & \multicolumn{1}{c}{\textbf{0.3548}} & \multicolumn{1}{c|}{3.1468} & \multicolumn{1}{c}{\underline{0.0227}} & \multicolumn{1}{c}{4.3420} & \multicolumn{1}{c}{0.0658} & \multicolumn{1}{c}{0.0771} \\
    \multicolumn{1}{c|}{(-- F F)} & \multicolumn{1}{c}{0.8609} & \multicolumn{1}{c}{3.3352} & \multicolumn{1}{c}{\underline{0.2897}} & \multicolumn{1}{c|}{3.1997} & \multicolumn{1}{c}{0.0210} & \multicolumn{1}{c}{1.8346} & \multicolumn{1}{c}{0.0699} & \multicolumn{1}{c|}{\underline{0.0644}} & \multicolumn{1}{c}{1.0996} & \multicolumn{1}{c}{\textbf{3.4948}} & \multicolumn{1}{c}{\underline{0.3714}} & \multicolumn{1}{c|}{3.1664} & \multicolumn{1}{c}{0.0243} & \multicolumn{1}{c}{3.4948} & \multicolumn{1}{c}{\textbf{0.0638}} & \multicolumn{1}{c}{\underline{0.0707}} \\
    \multicolumn{1}{c|}{(F F F)} & \multicolumn{1}{c}{\textbf{0.8356}} & \multicolumn{1}{c}{\underline{2.9978}} & \multicolumn{1}{c}{0.5182} & \multicolumn{1}{c|}{3.1812} & \multicolumn{1}{c}{\textbf{0.0194}} & \multicolumn{1}{c}{1.9925} & \multicolumn{1}{c}{\textbf{0.0680}} & \multicolumn{1}{c|}{\textbf{0.0601}} & \multicolumn{1}{c}{1.0997} & \multicolumn{1}{c}{\underline{3.5267}} & \multicolumn{1}{c}{0.7560} & \multicolumn{1}{c|}{3.1607} & \multicolumn{1}{c}{\textbf{0.0202}} & \multicolumn{1}{c}{\textbf{2.9733}} & \multicolumn{1}{c}{\underline{0.0642}} & \multicolumn{1}{c}{0.0800} \\
    \end{tabular}
\end{center}
\end{table*}

\textbf{Late Fusion (Concatenation).} Next, we evaluate the late fusion of $\text{APR}_{\text{V}}$-$\text{RPR}_{\text{I}}$ utilizing concatenation with and without BiLSTM layers (see Section~\ref{section_concatenation}). For the EuRoC MAV dataset, the concatenation improves the $\text{APR}_{\text{V}}$-only model for the MH-02 sequence, but decreases for the MH-04 sequence. The concatenation with BiLSTM layers can notably reduce the absolute pose results, but cannot outperform the $\text{APR}_{\text{V}}$-$\text{RPR}_{\text{I}}$+PGO~\cite{matthew} fusion, while the relative position results marginally decrease. For the PennCOSYVIO dataset, the late fusion decreases the model performance for the BF and BS sequences, while adding the BiLSTM layers improves the performance for the BF sequence, and hence, outperform the $\text{APR}_{\text{V}}$-only baseline model. Concatenation with BiLSTM layers proves to be effective for the IndustryVI datasets and outperforms all fusion techniques in terms of performance on the train 1 and test 1, train 2 and test 1, and train 2 and test 2 sequences. This demonstration reveals that the straightforward method of concatenating the high-level features does not prove effective in acquiring a meaningful representation between the APR and RPR tasks. The enhancement in performance resulting from the addition of BiLSTM layers underscores the importance of modeling temporal dependencies in achieving successful outcomes for the pose regression tasks. Overall, the error increases for the IndustryVI dataset when training on person 2 and testing on person 1 (increase of position error) and vice versa (increase of orientation error). A good fusion technique can accomodate this, e.g., here, concatenation with BiLSTM.

\textbf{Late Fusion (SSF).} We evaluate the soft fusion approach of SSF~\cite{silva} as a late fusion method (see Section~\ref{section_soft_fusion}). Similar for the late fusion method with concatenation, the BiLSTM layers are a crucial part for the SSF approach. The adoption of SSF and BiLSTM layers leads to a substantial reduction of the aboslute position error compared to the $\text{APR}_{\text{V}}$-only baseline model for all three datasets. For example for the PennCOSYVIO BF dataset, SSF with BiLSTM yields a low absolute position error of 1.1249\textit{m}, but can still significantly reduce the relative position error from 10.9\textit{cm} to 1.8\textit{cm} and the relative orientation error from 1.057$^{\circ}$ to 0.757$^{\circ}$. This approach demonstrates superior or comparable results compared to the fusion technique based on concatenation. This highlights the significance of proper feature selection, rather than merely concatenating high-level features.

\textbf{MMTM.} The intermediate fusion method based on MMTM~\cite{joze} learns a joint representation between $\text{APR}_{\text{V}}$ and $\text{RPR}_{\text{I}}$. We train the fusion architecture with seven different combinations of MMTM modules (for details on the fusion layers, see Section~\ref{section_intermediate_fusion}). For the number of trainable model parameters, see Table~\ref{table_no_of_params} in the appendix. It is evident that the number of trainable parameters increases as we increase the number of MMTM modules. Table~\ref{table_results_mmtm_modules} summarizes the results for the seven combinations on the EuRoC MAV dataset. The combinations (-- -- F), (F F --), (F -- F), and (F F F) yield comparable results on the EuRoC MAV dataset, while the best and consistent model performances on all three datasets are achieved using the (F F F) combination of three MMTM modules. Consequently, we select the combination of three MMTM modules for the results presented in Tables~\ref{table_results1} to \ref{table_results3}. For the EuRoC MAV dataset, MMTM (3 modules) decreases the APR results while yielding the best RPR results on the MH-02 sequence. At the expense of the APR error, the RPR error also decreases for MH-04 against the $\text{RPR}_{\text{I}}$ baseline. In addition, MMTM yields the best results for the PennCOSYVIO dataset. Our conclusion is that despite being developed for hand gesture recognition, human activity recognition, and audio-visual speech fusion, MMTM proves to be an effective module for learning a joint representation between networks even in the context of the challenging task of fusing APR and RPR.

\textbf{Auxiliary Learning.} Next, we evaluate the AuxiLearn~\cite{navon_aux} framework (see Section~\ref{section_auxiliary_learning}), i.e., the non-linear and convolutional variants. We use the SSF architecture as the main network to optimize the main $\text{APR}_{\text{V}}$ task. For all datasets, the non-linear variant yields better results than the convolutional variant. Therefore, it is crucial to model the intricate connections between the $\text{APR}_{\text{V}}$ and $\text{RPR}_{\text{I}}$ loss functions utilizing a non-linear layer instead of the spatial relationship, which is modeled through the convolutional layer of the auxiliary network. The efficacy of the AuxiLearn model in comparison to the baseline models is contingent upon the sequences. While for MH-02, MH-04, and BS, the APR results increase, the RPR results decrease. The trend is comparable for the IndustryVI dataset. It can be deduced that the $\text{RPR}_{\text{I}}$ task serves as a suitable auxiliary task for enhancing the main $\text{APR}_{\text{V}}$ task, albeit at the cost of a decreased performance on the $\text{RPR}_{\text{I}}$ task. Conversely, the main task does not have a positive impact on the auxiliary task. Therefore, AuxiLearn may prove beneficial for specific self-localization applications that place a significant emphasis on the absolute pose.

\begin{table*}
\begin{center}
\setlength{\tabcolsep}{0.5pt}
    \caption{$\text{RPR}_{\text{V}}$-$\text{RPR}_{\text{I}}$ results given in $\Delta m$ and $\Delta^{\circ}$. \textbf{Bold} are best results. \underline{Underlined} are the best RPR baseline results.}
    \label{table_results4}
    \footnotesize \begin{tabular}{ p{0.5cm} | p{0.5cm} | p{0.5cm} | p{0.5cm} | p{0.5cm} | p{0.5cm} | p{0.5cm} | p{0.5cm} | p{0.5cm} | p{0.5cm} | p{0.5cm} | p{0.5cm} | p{0.5cm} | p{0.5cm} | p{0.5cm} | p{0.5cm} | p{0.5cm} | p{0.5cm} | p{0.5cm} | p{0.5cm} | p{0.5cm} |  p{0.5cm} | p{0.5cm} | p{0.5cm} | p{0.5cm} }
    \multicolumn{1}{c|}{} & \multicolumn{10}{c|}{\textbf{EuRoC MAV} \cite{burri}} & \multicolumn{4}{c|}{\textbf{PennCOSYVIO} \cite{pfrommer}} & \multicolumn{4}{c}{\textbf{IndustryVI}} \\
    & \multicolumn{2}{c|}{\textbf{MH-02}} & \multicolumn{2}{c|}{\textbf{MH-04}} & \multicolumn{2}{c|}{\textbf{V1-03}} & \multicolumn{2}{c|}{\textbf{V2-02}} & \multicolumn{2}{c|}{\textbf{V1-01}} & \multicolumn{2}{c|}{\textbf{BF}} & \multicolumn{2}{c|}{\textbf{BS}} & \multicolumn{2}{c|}{\textbf{Test 1}} & \multicolumn{2}{c}{\textbf{Test 2}} \\
    \multicolumn{1}{c|}{\textbf{Method}} & \multicolumn{1}{c}{\scriptsize$\Delta e_{\text{med,p}}$} & \multicolumn{1}{c|}{\scriptsize$\Delta e_{\text{med,q}}$} & \multicolumn{1}{c}{\scriptsize$\Delta e_{\text{med,p}}$} & \multicolumn{1}{c|}{\scriptsize$\Delta e_{\text{med,q}}$} & \multicolumn{1}{c}{\scriptsize$\Delta e_{\text{med,p}}$} & \multicolumn{1}{c|}{\scriptsize$\Delta e_{\text{med,q}}$} & \multicolumn{1}{c}{\scriptsize$\Delta e_{\text{med,p}}$} & \multicolumn{1}{c|}{\scriptsize$\Delta e_{\text{med,q}}$}& \multicolumn{1}{c}{\scriptsize$\Delta e_{\text{med,p}}$} & \multicolumn{1}{c|}{\scriptsize$\Delta e_{\text{med,q}}$} & \multicolumn{1}{c}{\scriptsize$\Delta e_{\text{med,p}}$} & \multicolumn{1}{c|}{\scriptsize$\Delta e_{\text{med,q}}$} & \multicolumn{1}{c}{\scriptsize$\Delta e_{\text{med,p}}$} & \multicolumn{1}{c|}{\scriptsize$\Delta e_{\text{med,q}}$} & \multicolumn{1}{c}{\scriptsize$\Delta e_{\text{med,p}}$} & \multicolumn{1}{c|}{\scriptsize$\Delta e_{\text{med,q}}$} & \multicolumn{1}{c}{\scriptsize$\Delta e_{\text{med,p}}$} & \multicolumn{1}{c}{\scriptsize$\Delta e_{\text{med,q}}$}\\ \hline
    \multicolumn{1}{l|}{$\text{RPR}_{\text{V}}$: FlowNet~\cite{dosovitskiy}} & \multicolumn{1}{c}{\underline{0.0155}} & \multicolumn{1}{c|}{\underline{\textbf{0.013}}} & \multicolumn{1}{c}{\underline{0.0223}} & \multicolumn{1}{c|}{0.114} & \multicolumn{1}{c}{\underline{0.0289}} & \multicolumn{1}{c|}{0.359} & \multicolumn{1}{c}{\underline{0.0277}} & \multicolumn{1}{c|}{0.360} & \multicolumn{1}{c}{\underline{0.0171}} & \multicolumn{1}{c|}{0.215} & \multicolumn{1}{c}{\underline{0.0256}} & \multicolumn{1}{c|}{\underline{0.336}} & \multicolumn{1}{c}{\underline{\textbf{0.0322}}} & \multicolumn{1}{c|}{\underline{0.505}} & \multicolumn{1}{c}{\underline{0.0261}} & \multicolumn{1}{c|}{\underline{0.801}} & \multicolumn{1}{c}{\underline{0.0254}} & \multicolumn{1}{c}{0.864} \\
    \multicolumn{1}{l|}{$\text{RPR}_{\text{I}}$: IMUNet~\cite{silva}} & \multicolumn{1}{c}{0.0222} & \multicolumn{1}{c|}{0.069} & \multicolumn{1}{c}{0.0261} & \multicolumn{1}{c|}{\underline{0.084}} & \multicolumn{1}{c}{0.0306} & \multicolumn{1}{c|}{\underline{\textbf{0.145}}} & \multicolumn{1}{c}{0.0320} & \multicolumn{1}{c|}{\underline{\textbf{0.140}}} & \multicolumn{1}{c}{0.0212} & \multicolumn{1}{c|}{\underline{\textbf{0.082}}} & \multicolumn{1}{c}{\dotuline{0.1091}} & \multicolumn{1}{c|}{1.057} & \multicolumn{1}{c}{0.0393} & \multicolumn{1}{c|}{0.571} & \multicolumn{1}{c}{0.0295} & \multicolumn{1}{c|}{0.810} & \multicolumn{1}{c}{0.0290} & \multicolumn{1}{c}{\underline{0.861}}\\
    \multicolumn{1}{l|}{Late Fusion (concat)} & \multicolumn{1}{c}{0.0161} & \multicolumn{1}{c|}{0.103} & \multicolumn{1}{c}{0.0247} & \multicolumn{1}{c|}{0.099} & \multicolumn{1}{c}{0.0284} & \multicolumn{1}{c|}{0.211} & \multicolumn{1}{c}{0.0286} & \multicolumn{1}{c|}{0.206} & \multicolumn{1}{c}{0.0171} & \multicolumn{1}{c|}{0.121} & \multicolumn{1}{c}{0.0257} & \multicolumn{1}{c|}{0.348} & \multicolumn{1}{c}{0.0384} & \multicolumn{1}{c|}{0.479} & \multicolumn{1}{c}{0.0278} & \multicolumn{1}{c|}{0.829} & \multicolumn{1}{c}{0.0255} & \multicolumn{1}{c}{0.865} \\
    \multicolumn{1}{r|}{+ BiLSTM} & \multicolumn{1}{c}{0.0137} & \multicolumn{1}{c|}{0.123} & \multicolumn{1}{c}{0.0189} & \multicolumn{1}{c|}{0.122} & \multicolumn{1}{c}{\textbf{0.0261}} & \multicolumn{1}{c|}{0.325} & \multicolumn{1}{c}{0.0254} & \multicolumn{1}{c|}{0.330} & \multicolumn{1}{c}{0.0159} & \multicolumn{1}{c|}{0.187} & \multicolumn{1}{c}{\textbf{0.0249}} & \multicolumn{1}{c|}{0.328} & \multicolumn{1}{c}{0.0353} & \multicolumn{1}{c|}{0.464} & \multicolumn{1}{c}{\textbf{0.0231}} & \multicolumn{1}{c|}{0.779} & \multicolumn{1}{c}{\textbf{0.0201}} & \multicolumn{1}{c}{\textbf{0.787}}\\
    \multicolumn{1}{l|}{Late Fusion (SSF)~\cite{chen}} & \multicolumn{1}{c}{0.0166} & \multicolumn{1}{c|}{0.079} & \multicolumn{1}{c}{0.0234} & \multicolumn{1}{c|}{0.088} & \multicolumn{1}{c}{0.0276} & \multicolumn{1}{c|}{0.200} & \multicolumn{1}{c}{0.0273} & \multicolumn{1}{c|}{0.215} & \multicolumn{1}{c}{0.0160} & \multicolumn{1}{c|}{0.116} & \multicolumn{1}{c}{0.0261} & \multicolumn{1}{c|}{0.343} & \multicolumn{1}{c}{0.0371} & \multicolumn{1}{c|}{0.498} & \multicolumn{1}{c}{0.0294} & \multicolumn{1}{c|}{0.860} & \multicolumn{1}{c}{0.0310} & \multicolumn{1}{c}{0.886} \\
    \multicolumn{1}{r|}{+ BiLSTM} & \multicolumn{1}{c}{0.0137} & \multicolumn{1}{c|}{0.090} & \multicolumn{1}{c}{0.0193} & \multicolumn{1}{c|}{0.088} & \multicolumn{1}{c}{0.0271} & \multicolumn{1}{c|}{0.276} & \multicolumn{1}{c}{0.0245} & \multicolumn{1}{c|}{0.280} & \multicolumn{1}{c}{0.0152} & \multicolumn{1}{c|}{0.134} & \multicolumn{1}{c}{0.0260} & \multicolumn{1}{c|}{\textbf{0.319}} & \multicolumn{1}{c}{0.0355} & \multicolumn{1}{c|}{\textbf{0.442}} & \multicolumn{1}{c}{0.0282} & \multicolumn{1}{c|}{0.795} & \multicolumn{1}{c}{0.0284} & \multicolumn{1}{c}{0.783} \\
    \multicolumn{1}{l|}{MMTM~\cite{joze}} & \multicolumn{1}{c}{\textbf{0.0121}} & \multicolumn{1}{c|}{0.073} & \multicolumn{1}{c}{\textbf{0.0181}} & \multicolumn{1}{c|}{\textbf{0.083}} & \multicolumn{1}{c}{0.0268} & \multicolumn{1}{c|}{0.185} & \multicolumn{1}{c}{\textbf{0.0222}} & \multicolumn{1}{c|}{0.179} & \multicolumn{1}{c}{\textbf{0.0138}} & \multicolumn{1}{c|}{0.107} & \multicolumn{1}{c}{0.0255} & \multicolumn{1}{c|}{0.458} & \multicolumn{1}{c}{0.0367} & \multicolumn{1}{c|}{0.697} & \multicolumn{1}{c}{0.0256} & \multicolumn{1}{c|}{\textbf{0.758}} & \multicolumn{1}{c}{0.0248} & \multicolumn{1}{c}{0.860} \\
    \end{tabular}
\end{center}
\end{table*}

\begin{figure*}[!t]
	\centering
	\begin{minipage}[b]{0.495\linewidth}
        \centering
    	\includegraphics[trim=40 40 40 40, clip, width=1.0\linewidth]{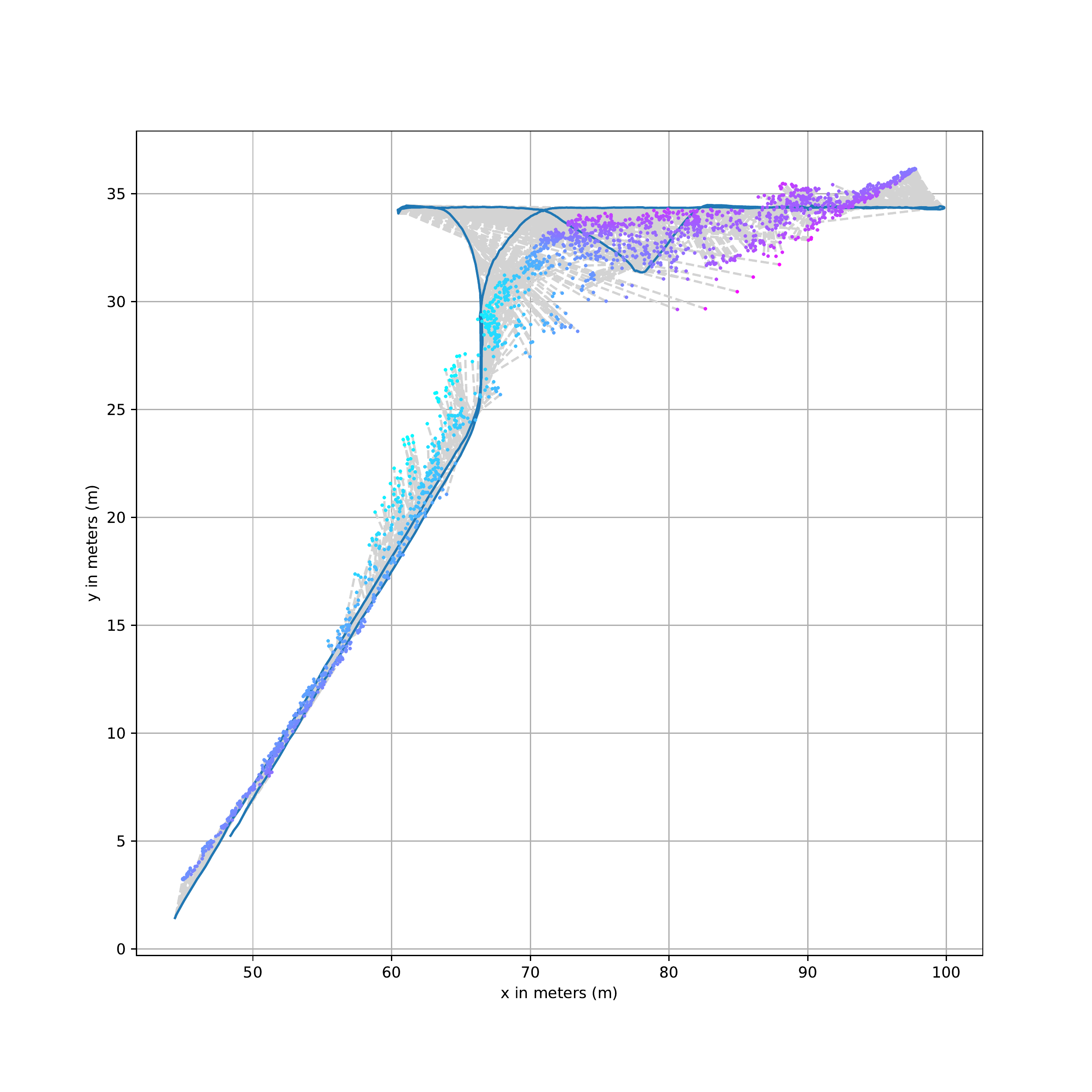}
    \end{minipage}
    \hfill
	\begin{minipage}[b]{0.495\linewidth}
        \centering
    	\includegraphics[trim=40 40 40 40, clip, width=1.0\linewidth]{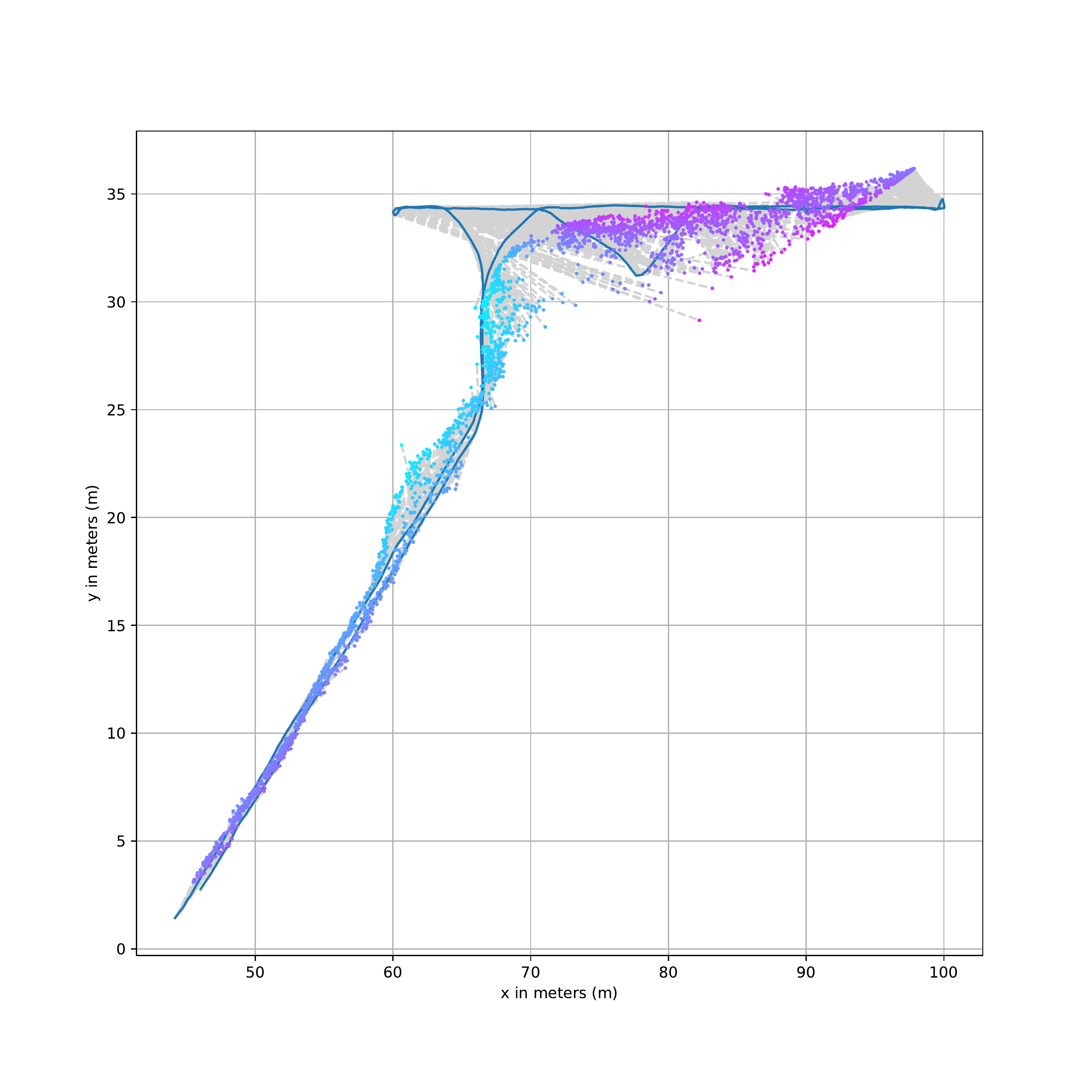}
    \end{minipage}
	\begin{minipage}[b]{0.495\linewidth}
        \centering
    	\includegraphics[width=1.0\linewidth]{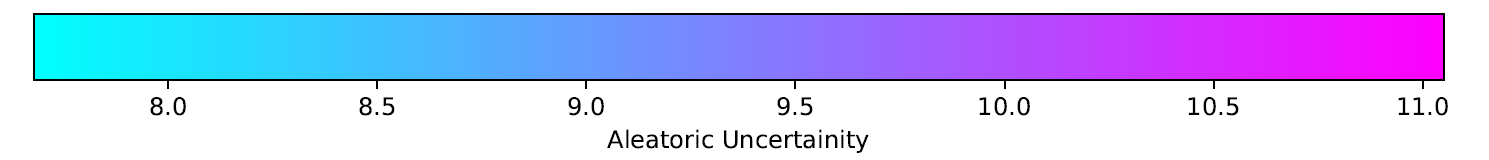}
    	\subcaption{PennCOSYVIO~\cite{pfrommer}: BF. \cite{nisha_master}}
    	\label{image_penn_uncertainty1}
    \end{minipage}
    \hfill
	\begin{minipage}[b]{0.495\linewidth}
        \centering
    	\includegraphics[width=1.0\linewidth]{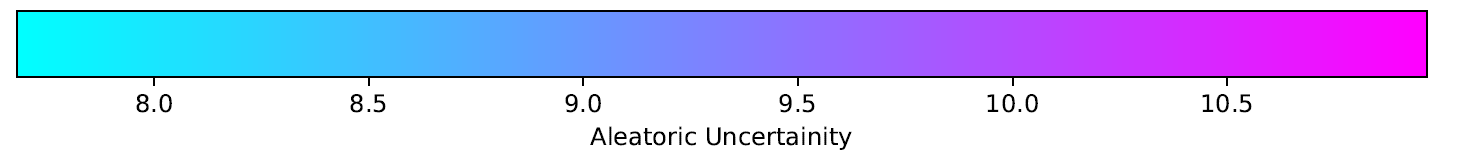}
    	\subcaption{PennCOSYVIO~\cite{pfrommer}: BS. \cite{nisha_master}}
    	\label{image_penn_uncertainty2}
    \end{minipage}
    \caption{Plot of the aleatoric uncertainty for the absolute pose prediction. Note the increased uncertainty for the top right part.}
    \label{image_penn_uncertainty}
\end{figure*}

\begin{figure*}[!t]
	\centering
	\begin{minipage}[b]{1.0\linewidth}
        \centering
    	\includegraphics[width=1.0\linewidth]{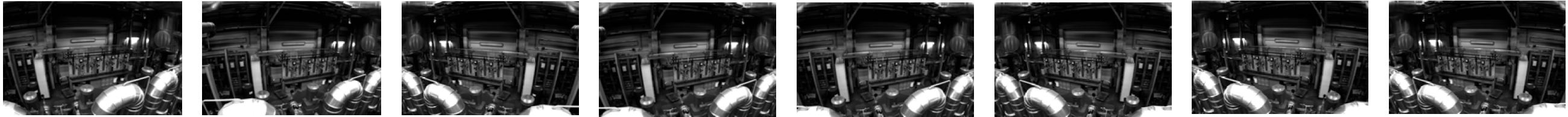}
    	\subcaption{EuRoC MAV~\cite{burri}: MH-02-easy.}
    	\label{image_euroc_uncertainty_images1}
    \end{minipage}
    \hfill
	\begin{minipage}[b]{1.0\linewidth}
        \centering
    	\includegraphics[width=1.0\linewidth]{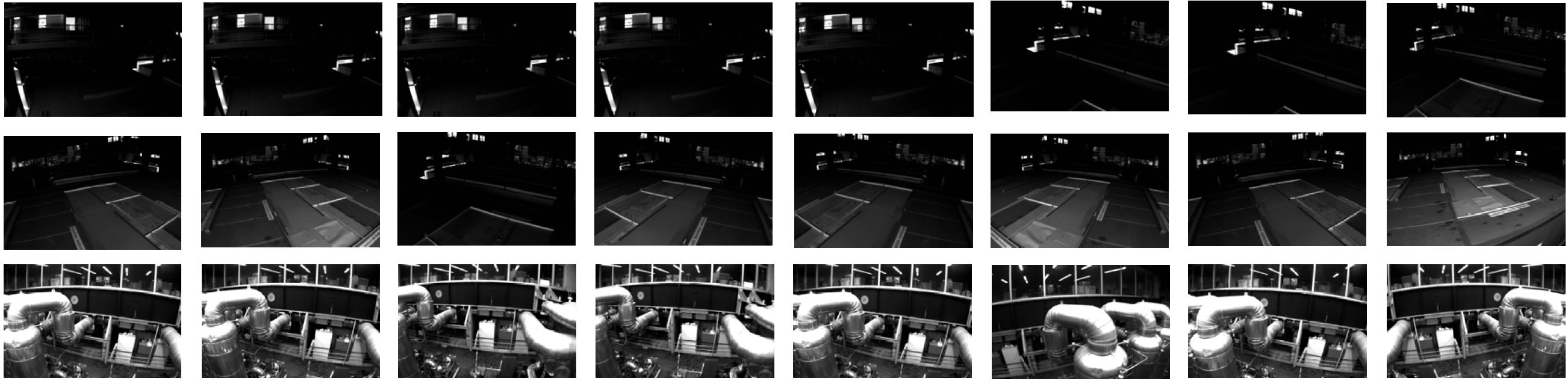}
    	\subcaption{EuRoC MAV~\cite{burri}: MH-04-difficult.}
    	\label{image_euroc_uncertainty_images2}
    \end{minipage}
    \caption{Exemplary images with a corresponding high uncertainty computed with the Bayesian model \cite{kendall_uncertainty}. From \cite{nisha_master}.}
    \label{image_euroc_uncertainty_images}
\end{figure*}

\begin{figure*}[!t]
	\centering
	\begin{minipage}[b]{1.0\linewidth}
        \centering
    	\includegraphics[width=1.0\linewidth]{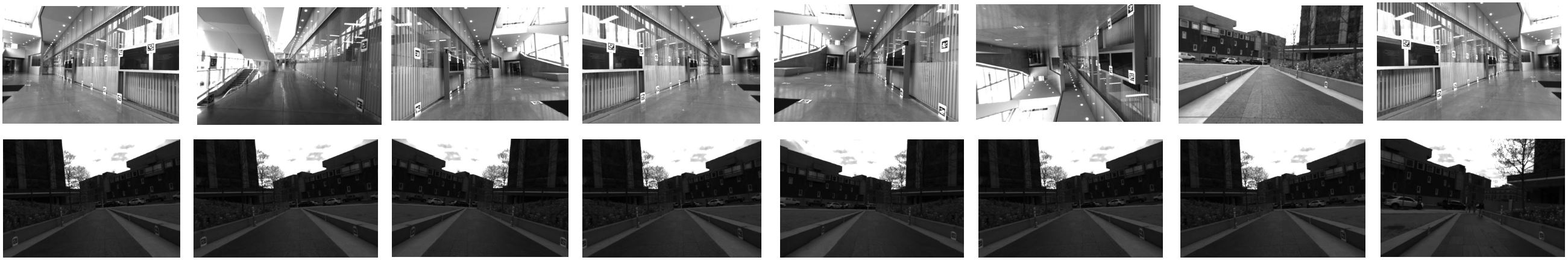}
    	\subcaption{PennCOSYVIO~\cite{pfrommer}: BF.}
    	\label{image_penn_uncertainty_images1}
    \end{minipage}
    \hfill
	\begin{minipage}[b]{1.0\linewidth}
        \centering
    	\includegraphics[width=1.0\linewidth]{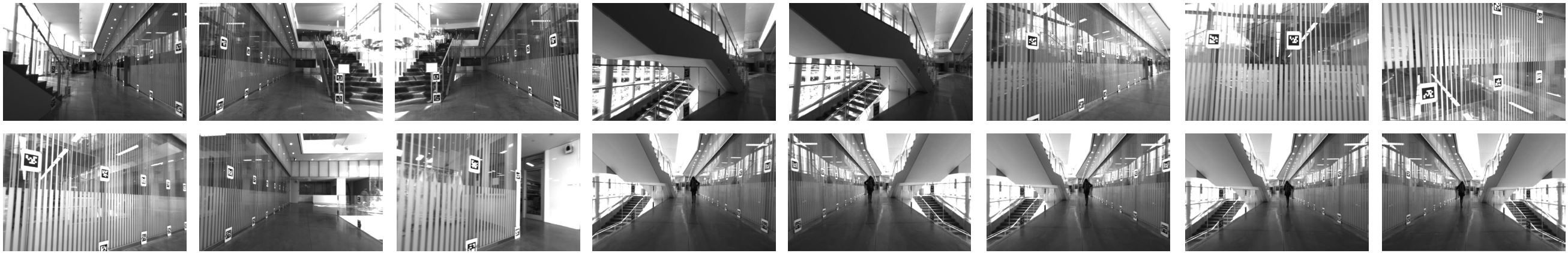}
    	\subcaption{PennCOSYVIO~\cite{pfrommer}: BS.}
    	\label{image_penn_uncertainty_images2}
    \end{minipage}
    \caption{Exemplary images with a corresponding high uncertainty computed with the Bayesian model \cite{kendall_uncertainty}. From \cite{nisha_master}.}
    \label{image_penn_uncertainty_images}
\end{figure*}

\textbf{Bayesian Neural Networks.} The BNN~\cite{kendall_uncertainty} models the aleatoric uncertainty for the $\text{APR}_{\text{V}}$ task (see Section~\ref{setcion_bayesian_learning}). We train the fusion model utilizing the modified loss function in Equation~\eqref{eq_aleatoric}. The performance of the BNN is superior to the baseline $\text{APR}_{\text{V}}$ model for the EuRoC MAV dataset, evidenced by a reduction in error from 0.9249\textit{m} to 0.7925\textit{m} and from 0.9405\textit{m} to 0.8523\textit{m}. Additionally, the BNN demonstrated improved performance for the majority of evaluation sequences of the IndustryVI dataset. However, its performance deteriorates for the PennCOSYVIO dataset. Interestingly, the predicted trajectories are unique and smoother for both the EuRoC MAV dataset (see Figure~\ref{image_app_euroc_mh02_6} and Figure~\ref{image_app_euroc_mh04_6} in the appendix) and for the IndustryVI dataset. This indicates that the BNN has learned to reduce the variance of the mean prediction values. For the PennCOSYVIO dataset, we observe that the prediction are worse inside the building (as seen in the upper-right part of the Figure~\ref{image_app_penncosy_bf_6} and Figure~\ref{image_app_penncosy_bs_6}), particularly in regions where the images feature repetitive patterns of extensive glass walls. This phenomenon is also substantiated by the high levels of aleatoric uncertainty present in these areas (see Figure~\ref{image_penn_uncertainty}). As a result, Bayesian learning may serve as a tool for interpreting complex images, for example, images with difficult illuminations (see Figure~\ref{image_euroc_uncertainty_images}) or reflective pattern (see Figure~\ref{image_penn_uncertainty_images}) can be detected.

\subsection{Evaluation of $\text{RPR}_{\text{V}}$-$\text{RPR}_{\text{I}}$ Fusion Methods}
\label{chap_eval_rpr}

\noindent We provide quantitative results for the $\text{RPR}_{\text{V}}$-$\text{RPR}_{\text{I}}$ fusion task on the EuRoC MAV, PennCOSYVIO, and IndustrialVI datasets in Table~\ref{table_results4}. For an overview of RPR trajectory comparisons, see Figure~\ref{image_app_rpr_euroc_mh04} to Figure~\ref{image_app_rpr_industry2} in the appendix. As the RPR task is independent of the scene geometry, we utilize all training sequences from both scenes (MH and V) of the EuRoC MAV dataset (i.e., MH-01, MH-03, MH-05, V1-02, V2-01, and V2-03) and test on the MH-02, MH-04, V1-01, V1-03, and V2-02 sequences. Hence, the training dataset is large to cover all movement patterns and dynamics of the MAV, while the testing datasets cover the large machine hall and the small living room with different object configurations. First, we evaluate the baseline models FlowNet~\cite{dosovitskiy} for the $\text{RPR}_{\text{V}}$ task and IMUNet~\cite{silva} for the $\text{RPR}_{\text{I}}$ task. It is noteworthy that the $\text{RPR}_{\text{V}}$ model produces superior relative translational results on the EuRoC MAV dataset, while the $\text{RPR}_{\text{I}}$ model yields superior relative orientational results. On the PennCOSYVIO dataset, the $\text{RPR}_{\text{V}}$ model outperforms the $\text{RPR}_{\text{I}}$ model. This shows that the IMU measurements contain a high sensor noise, while the $\text{RPR}_{\text{V}}$ model is robust to fast movement changes (of the MAV and of the handheld system). The objective is to merge the advantageous translational predictions of the $\text{RPR}_{\text{V}}$ model with the advantageous rotational prediction of the $\text{RPR}_{\text{I}}$ model (for the EuRoC MAV dataset), or to selectively choose favorable predictions from the $\text{RPR}_{\text{I}}$ model (for the PennCOSYVIO dataset).

\textbf{Late Fusion (Concatenation).} The combination of concatenation with BiLSTM layers can partially enhance the RPR results, particularly for the Vicon datasets within the EuRoC MAV dataset and for the IndustryVI datasets. In contrast, the performance of the late fusion model decreases on the machine hall sequences. Consequently, similar to the $\text{APR}_{\text{V}}$-$\text{RPR}_{\text{I}}$ fusion task of the model with concatenation and BiLSTM layers, the performance of the late fusion model is also improved by incorporating BiLSTM layers after fusion to capture temporal dependencies.

\textbf{Late Fusion (SSF).} Soft fusion of $\text{RPR}_{\text{V}}$ and $\text{RPR}_{\text{I}}$ (see Section~\ref{section_soft_fusion}) is only marginally different from the late fusion model with cancatenation with a small improvement in the RPR errors. As previously, the performance of the SSF model improves for all datasets by incorporating BiLSTM layers after the fusion. For the EuRoC MAV and IndustrialVI datasets, SSF with BiLSTM layers outperforms both baseline models on all sequences. In the context of PennCOSYVIO, it exhibits superior performance compared to the $\text{RPR}_{\text{I}}$ baseline model. However, it is unable to surpass the translational performance of the $\text{RPR}_{\text{V}}$ model for both sequences.

\textbf{MMTM.} As for the $\text{APR}_{\text{V}}$-$\text{RPR}_{\text{I}}$ fusion, we assess the seven different layer combinations of MMTM for the $\text{RPR}_{\text{V}}$-$\text{RPR}_{\text{I}}$ fusion (see Table~\ref{table_results_mmtm_modules}). The model featuring three MMTM modules (F F F) for the joint representation exhibits the optimal performance. These results are consistent with the experiments and findings presented by \cite{joze}. We note that the fusion with three MMTM modules reduces the translational error of the baseline and fusion techniques on the EuRoC MAV datasets. However, there is not a substantial improvement for the PennCOSYVIO and IndustrialVI datasets.

\begin{figure*}[t!]
	\centering
	\begin{minipage}[b]{0.135\linewidth}
        \centering
    	\includegraphics[width=1.0\linewidth]{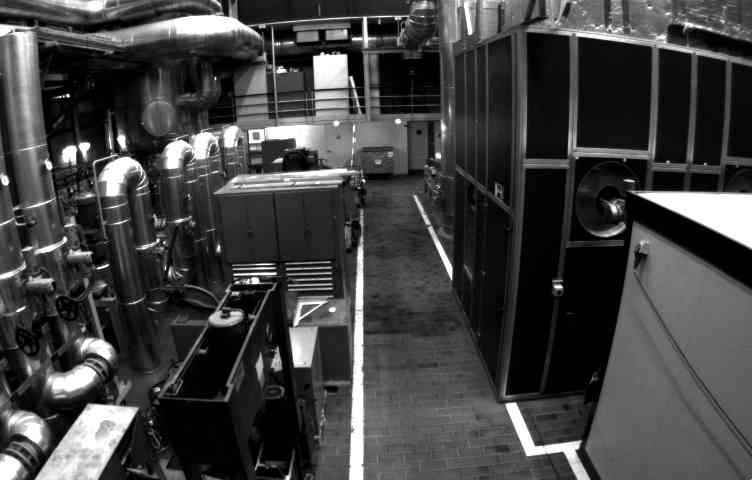}
    	\subcaption{Uncorrupted image.}
        \label{image_corr1}
    \end{minipage}
    \hfill
	\begin{minipage}[b]{0.135\linewidth}
        \centering
    	\includegraphics[width=1.0\linewidth]{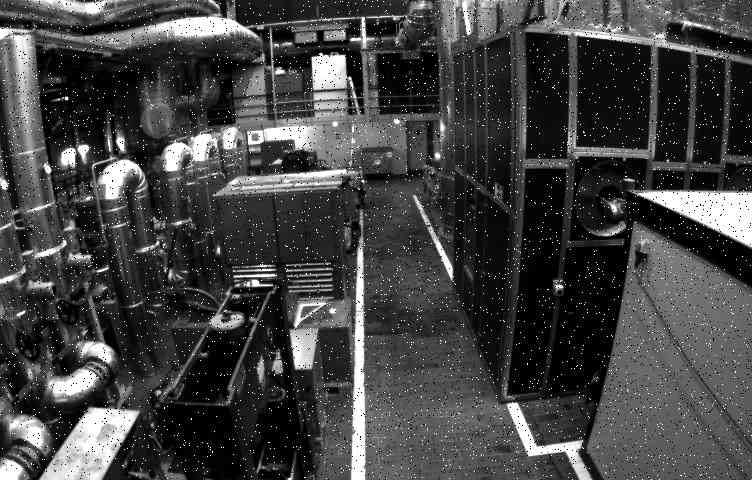}
    	\subcaption{Image with added noise.}
        \label{image_corr2}
    \end{minipage}
    \hfill
	\begin{minipage}[b]{0.135\linewidth}
        \centering
    	\includegraphics[width=1.0\linewidth]{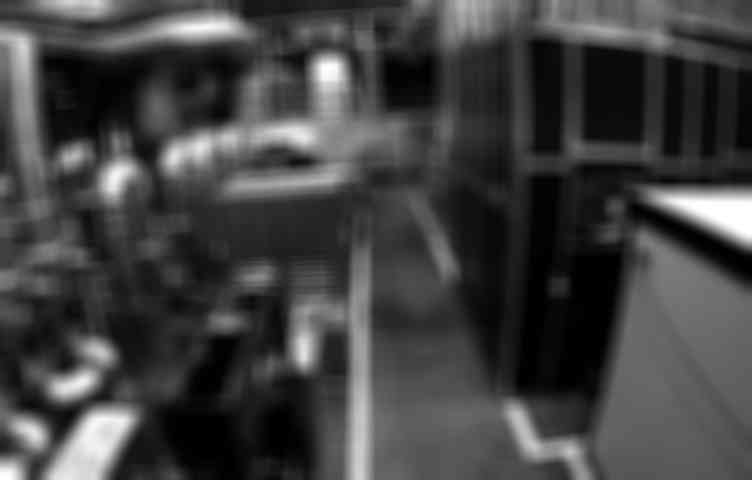}
    	\subcaption{Image with blur.}
        \label{image_corr3}
    \end{minipage}
    \hfill
	\begin{minipage}[b]{0.135\linewidth}
        \centering
    	\includegraphics[width=1.0\linewidth]{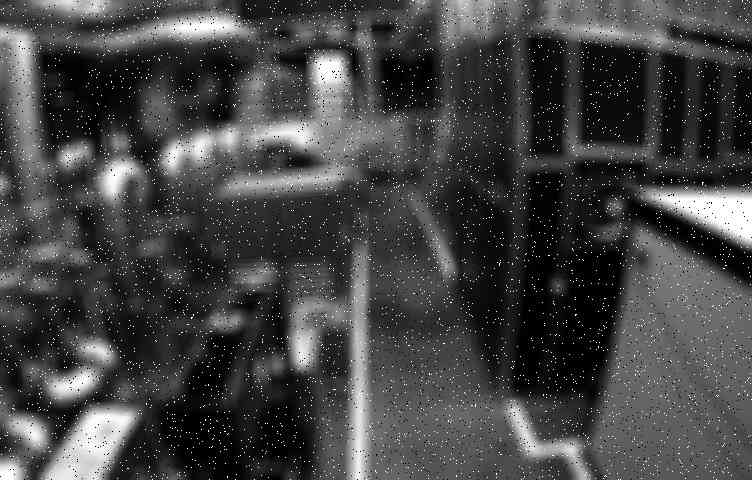}
    	\subcaption{Image with noise and blur.}
        \label{image_corr4}
    \end{minipage}
    \hfill
	\begin{minipage}[b]{0.135\linewidth}
        \centering
    	\includegraphics[width=1.0\linewidth]{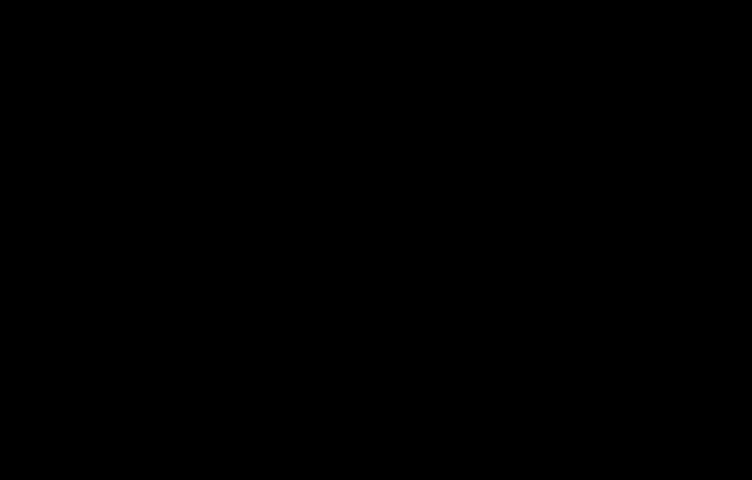}
    	\subcaption{Image with full patch.}
        \label{image_corr5}
    \end{minipage}
    \hfill
	\begin{minipage}[b]{0.135\linewidth}
        \centering
    	\includegraphics[width=1.0\linewidth]{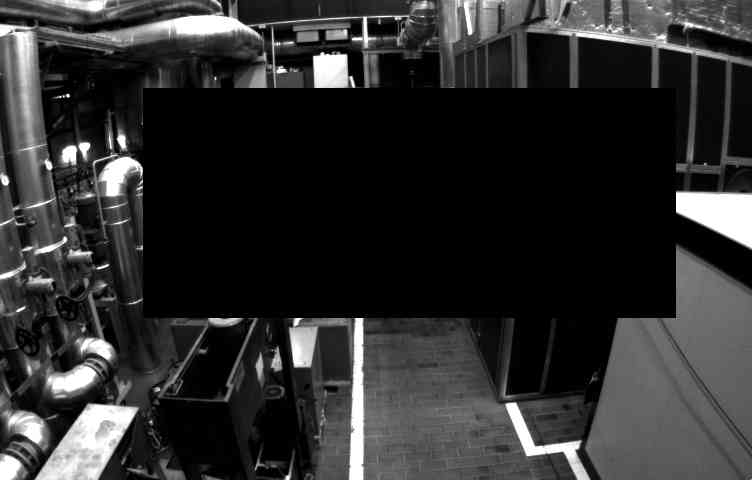}
    	\subcaption{Image with rectangular patch.}
        \label{image_corr6}
    \end{minipage}
    \hfill
	\begin{minipage}[b]{0.135\linewidth}
        \centering
    	\includegraphics[width=1.0\linewidth]{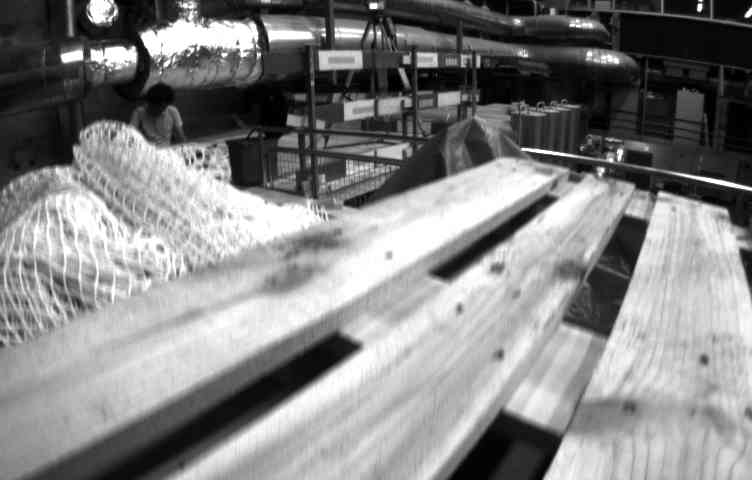}
    	\subcaption{Random corresponding image.}
        \label{image_corr7}
    \end{minipage}
    \caption{Image corruption techniques. The original image of the EuRoC MAV \cite{burri} dataset is shown in (a).}
    \label{image_corr}
\end{figure*}

\begin{figure*}[t!]
	\centering
	\begin{minipage}[b]{0.48\linewidth}
        \centering
    	\includegraphics[trim=0 10 0 10, clip, width=1.0\linewidth]{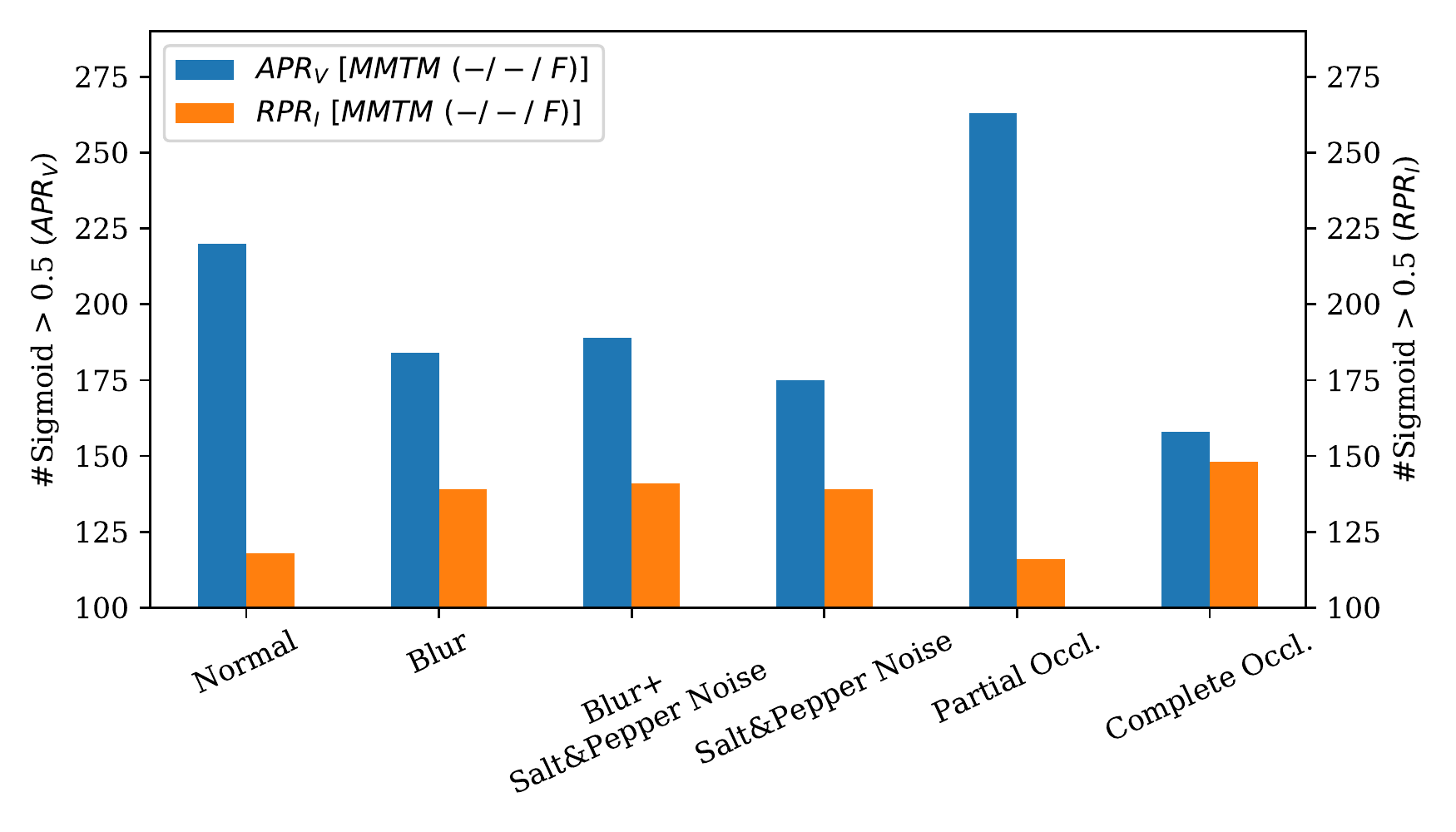}
    	\subcaption{Fusion of $\text{APR}_{\text{V}}$ and $\text{RPR}_{\text{I}}$.}
    	\label{image_masks1}
    \end{minipage}
    \hfill
	\begin{minipage}[b]{0.48\linewidth}
        \centering
    	\includegraphics[trim=0 10 0 10, clip, width=1.0\linewidth]{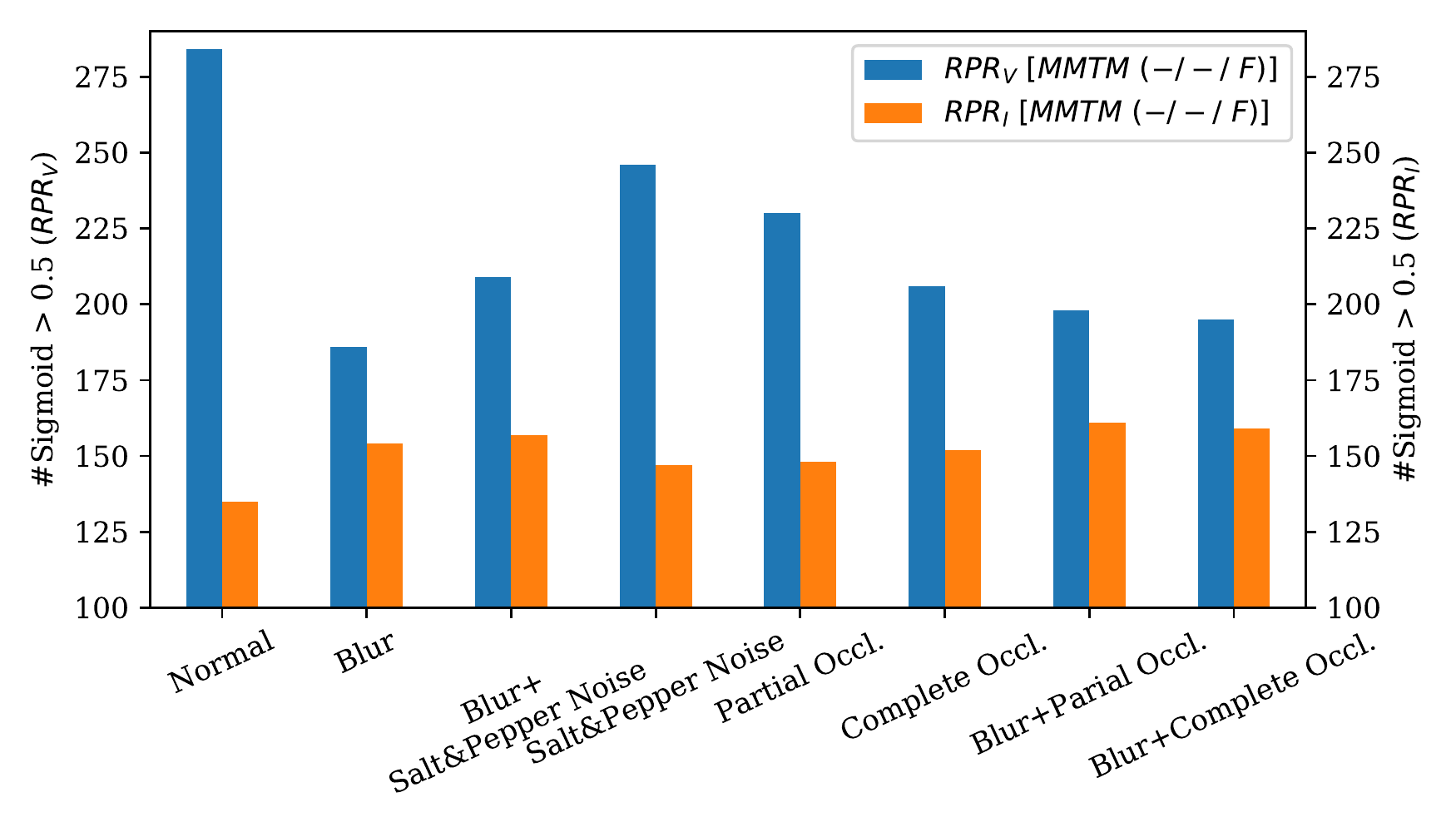}
    	\subcaption{Fusion of $\text{RPR}_{\text{V}}$ and $\text{RPR}_{\text{I}}$.}
    	\label{image_masks2}
    \end{minipage}
    \caption{Number of sigmoid activations higher than 0.5 for $\text{APR}_{\text{V}}$ and $\text{RPR}_{\text{V}}$ methods (blue) and $\text{RPR}_{\text{I}}$ methods (orange) evaluating the third layer for the MMTM module for different image corruption techniques.}
    \label{image_masks}
\end{figure*}

\subsection{Network Activation Mask Evaluation}
\label{chap_network_mask}

\begin{figure}[t!]
	\centering
	\begin{minipage}[b]{1.0\linewidth}
    	\begin{minipage}[b]{0.48\linewidth}
            \centering
        	\includegraphics[width=1.0\linewidth]{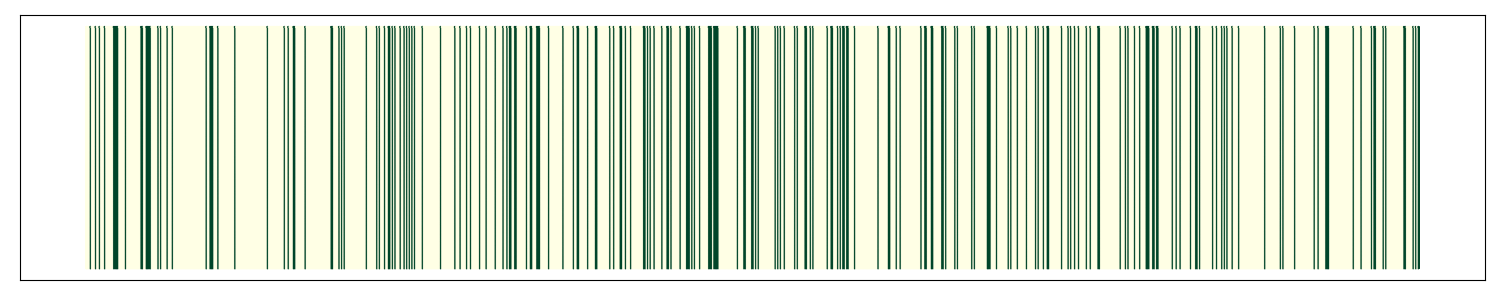}
        \end{minipage}
        \hfill
    	\begin{minipage}[b]{0.48\linewidth}
            \centering
        	\includegraphics[width=1.0\linewidth]{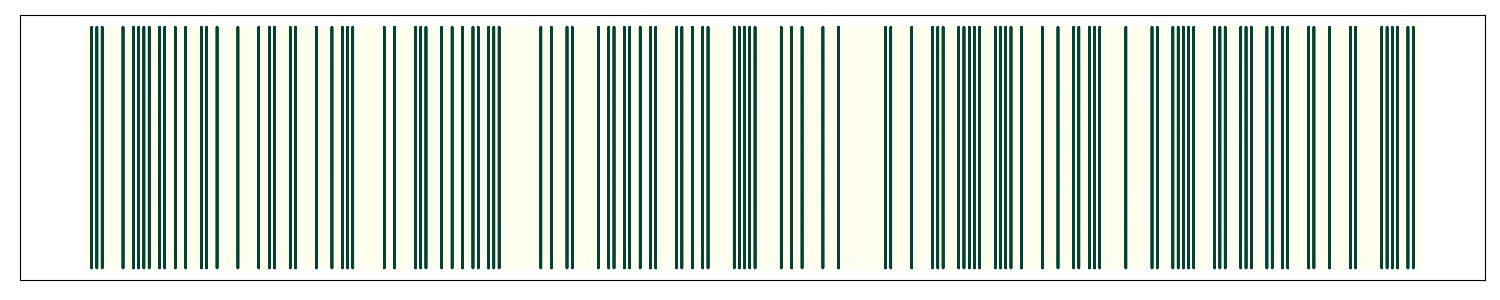}
        \end{minipage}
        \subcaption{Without disruption.}
        \label{image_act_apr_rpr1}
    \end{minipage}
	\begin{minipage}[b]{1.0\linewidth}
    	\begin{minipage}[b]{0.48\linewidth}
            \centering
        	\includegraphics[width=1.0\linewidth]{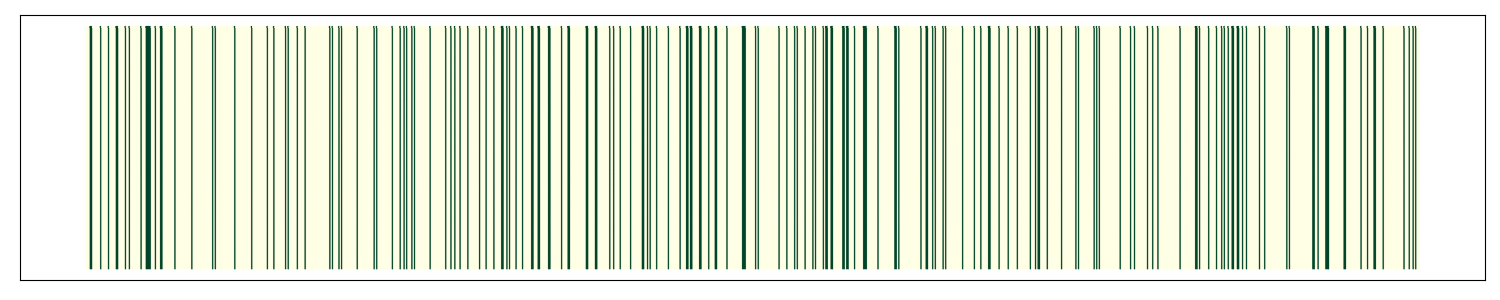}
        \end{minipage}
        \hfill
    	\begin{minipage}[b]{0.48\linewidth}
            \centering
        	\includegraphics[width=1.0\linewidth]{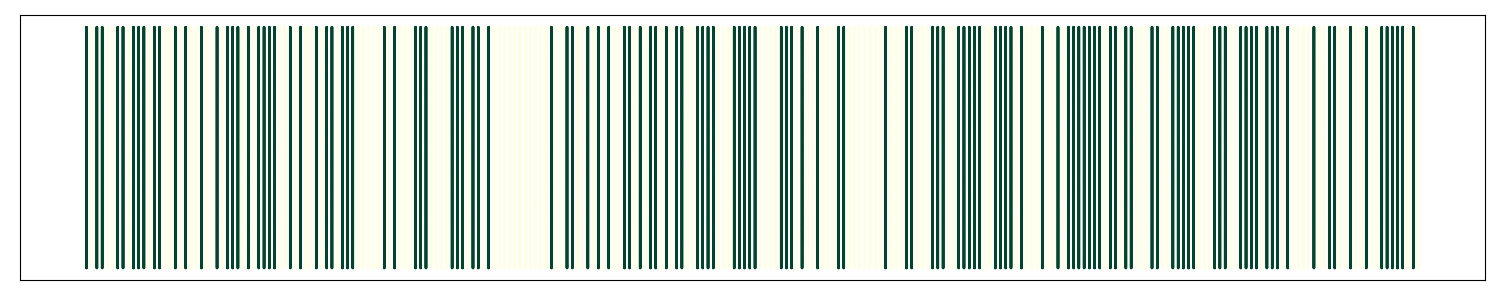}
        \end{minipage}
        \subcaption{Blur.}
        \label{image_act_apr_rpr2}
    \end{minipage}
	\begin{minipage}[b]{1.0\linewidth}
    	\begin{minipage}[b]{0.48\linewidth}
            \centering
        	\includegraphics[width=1.0\linewidth]{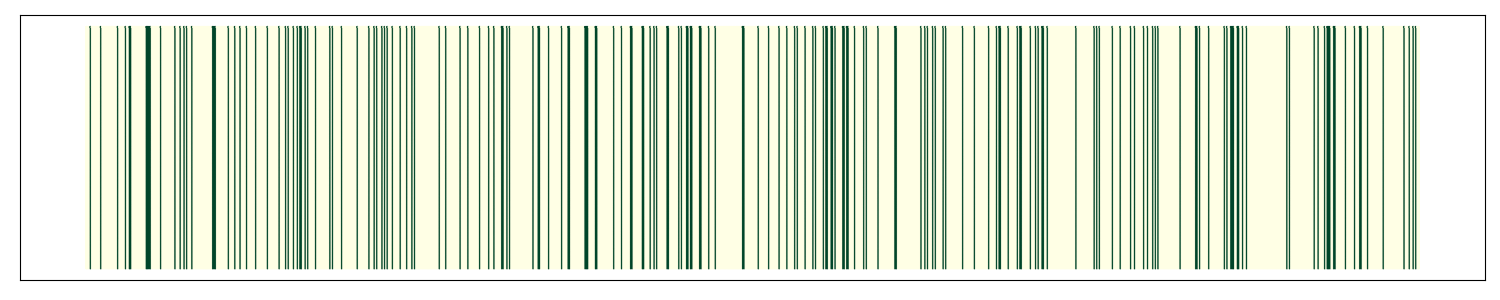}
        \end{minipage}
        \hfill
    	\begin{minipage}[b]{0.48\linewidth}
            \centering
        	\includegraphics[width=1.0\linewidth]{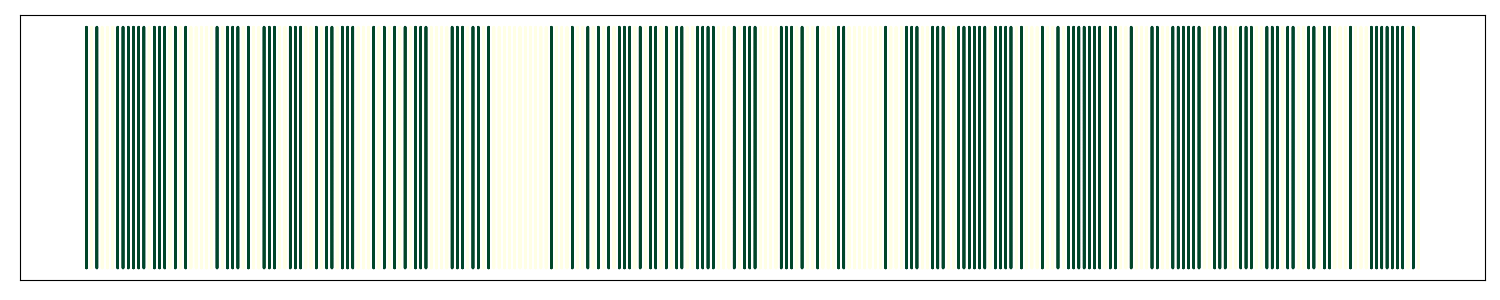}
        \end{minipage}
        \subcaption{Blur and Salt\&Pepper noise.}
        \label{image_act_apr_rpr3}
    \end{minipage}
	\begin{minipage}[b]{1.0\linewidth}
    	\begin{minipage}[b]{0.48\linewidth}
            \centering
        	\includegraphics[width=1.0\linewidth]{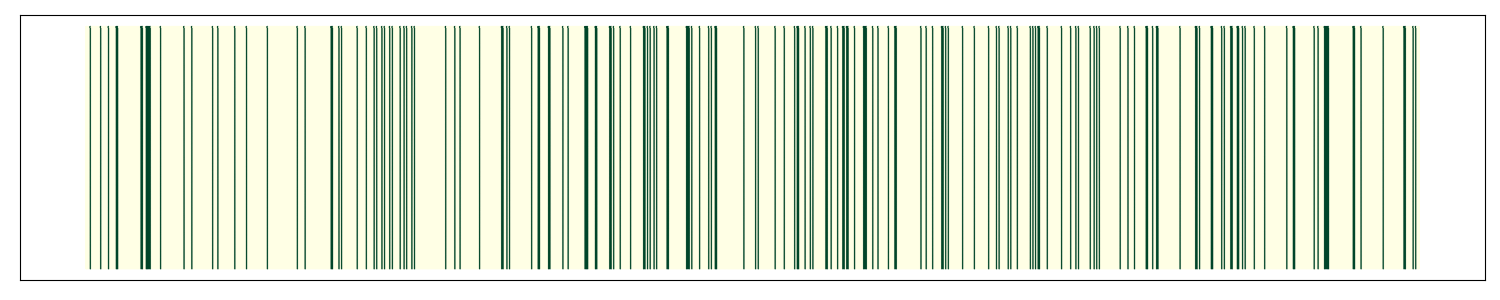}
        \end{minipage}
        \hfill
    	\begin{minipage}[b]{0.48\linewidth}
            \centering
        	\includegraphics[width=1.0\linewidth]{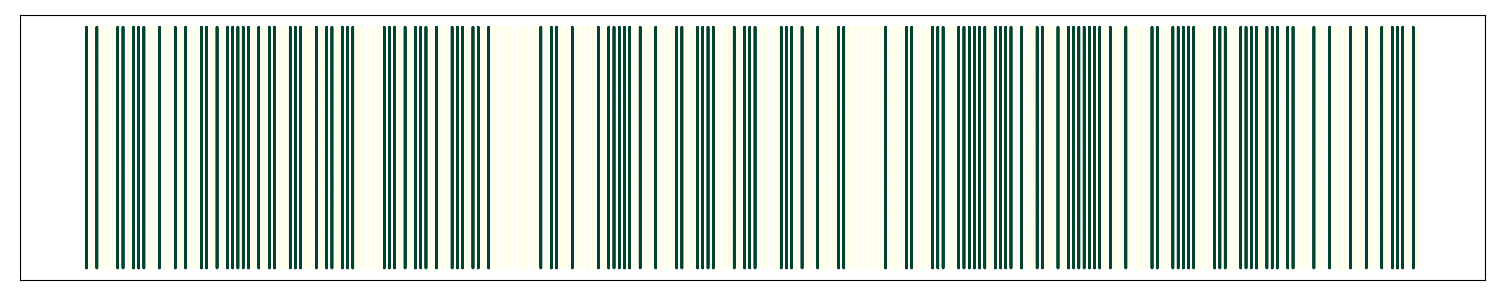}
        \end{minipage}
        \subcaption{Salt\&Pepper noise.}
        \label{image_act_apr_rpr4}
    \end{minipage}
	\begin{minipage}[b]{1.0\linewidth}
    	\begin{minipage}[b]{0.48\linewidth}
            \centering
        	\includegraphics[width=1.0\linewidth]{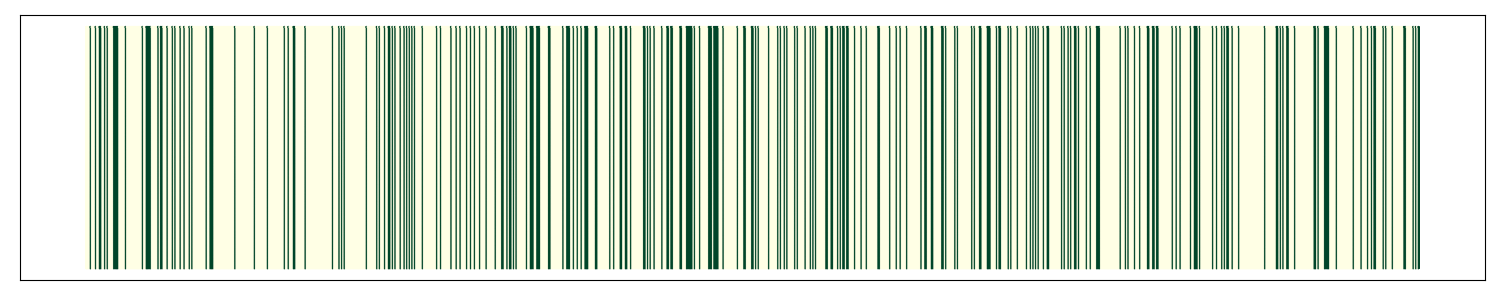}
        \end{minipage}
        \hfill
    	\begin{minipage}[b]{0.48\linewidth}
            \centering
        	\includegraphics[width=1.0\linewidth]{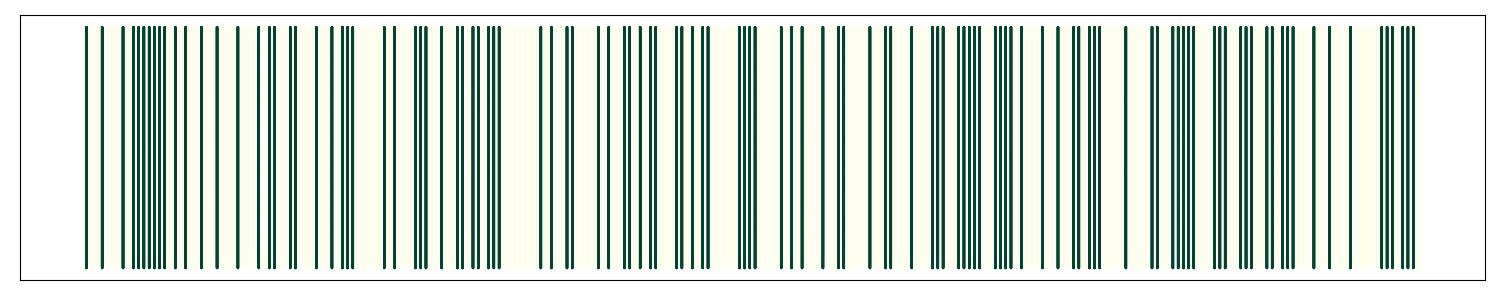}
        \end{minipage}
        \subcaption{Partial occlusion.}
        \label{image_act_apr_rpr5}
    \end{minipage}
	\begin{minipage}[b]{1.0\linewidth}
    	\begin{minipage}[b]{0.48\linewidth}
            \centering
        	\includegraphics[width=1.0\linewidth]{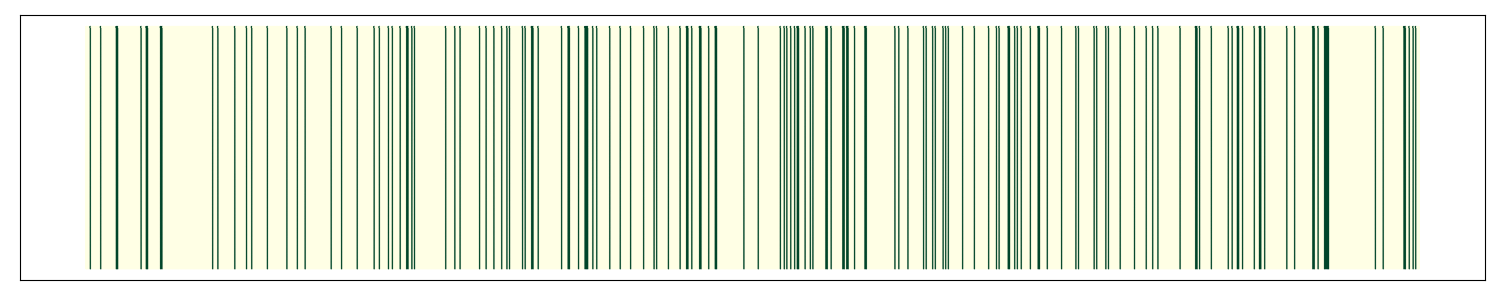}
        \end{minipage}
        \hfill
    	\begin{minipage}[b]{0.48\linewidth}
            \centering
        	\includegraphics[width=1.0\linewidth]{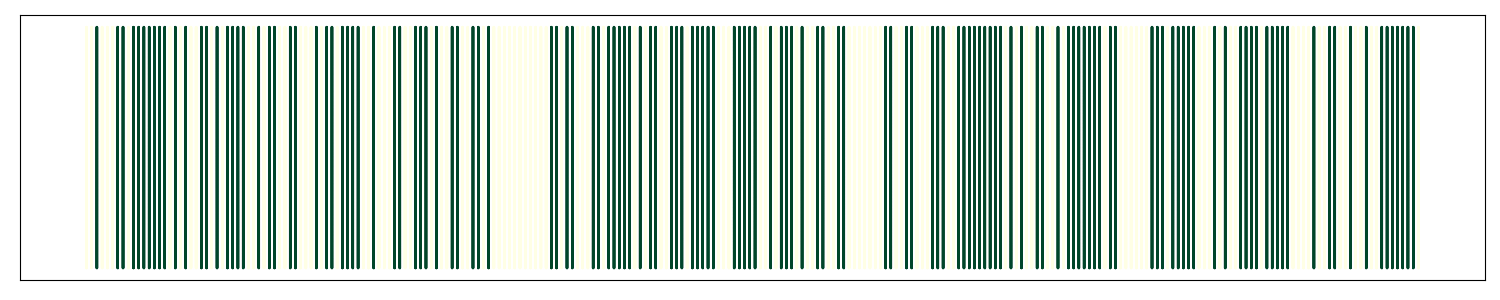}
        \end{minipage}
        \subcaption{Complete occlusion.}
        \label{image_act_apr_rpr6}
    \end{minipage}
    \caption{Activations for the third MMTM module for different image corruption techniques for $\text{APR}_{\text{V}}$ (left) and $\text{RPR}_{\text{I}}$ (right).}
    \label{image_act_apr_rpr}
\end{figure}

\begin{figure}[t!]
	\centering
	\begin{minipage}[b]{1.0\linewidth}
    	\begin{minipage}[b]{0.48\linewidth}
            \centering
        	\includegraphics[width=1.0\linewidth]{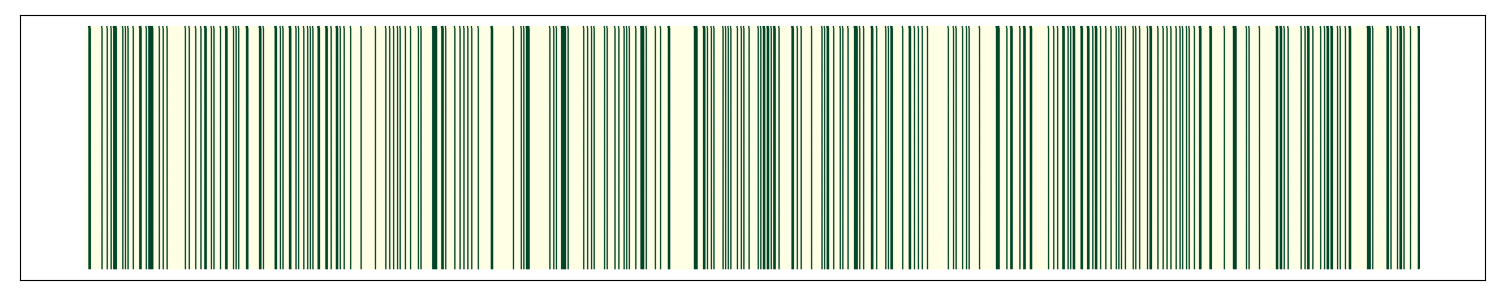}
        \end{minipage}
        \hfill
    	\begin{minipage}[b]{0.48\linewidth}
            \centering
        	\includegraphics[width=1.0\linewidth]{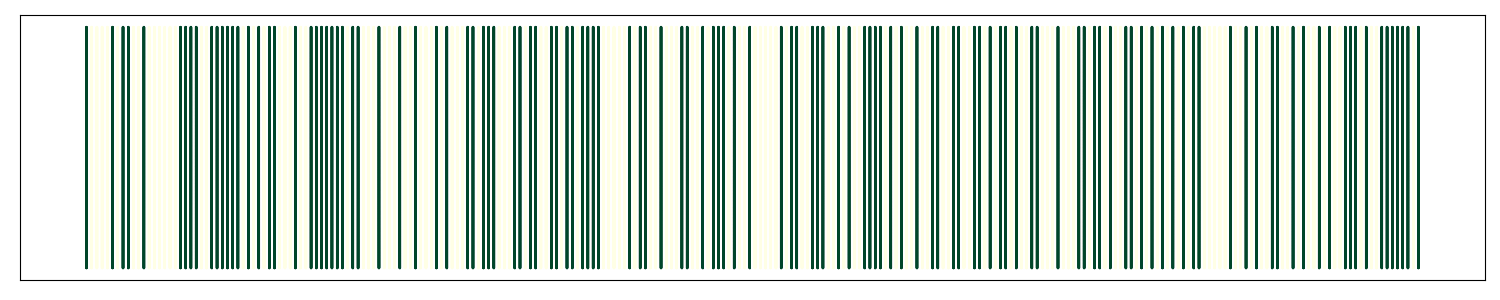}
        \end{minipage}
        \subcaption{Without disruption.}
        \label{image_act_rpr_rpr1}
    \end{minipage}
	\begin{minipage}[b]{1.0\linewidth}
    	\begin{minipage}[b]{0.48\linewidth}
            \centering
        	\includegraphics[width=1.0\linewidth]{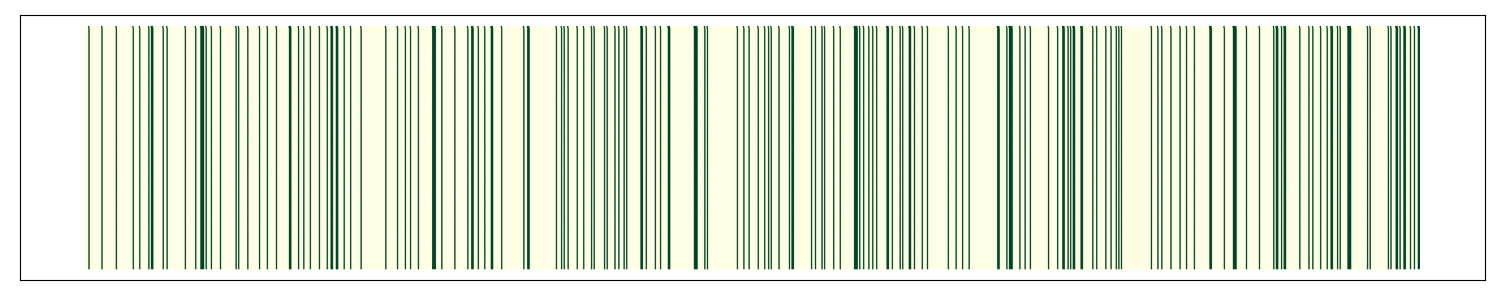}
        \end{minipage}
        \hfill
    	\begin{minipage}[b]{0.48\linewidth}
            \centering
        	\includegraphics[width=1.0\linewidth]{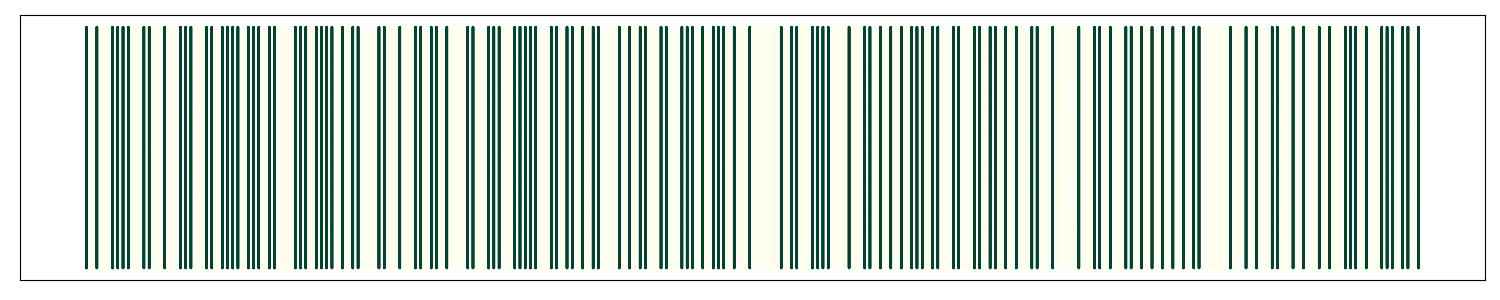}
        \end{minipage}
        \subcaption{Blur.}
        \label{image_act_rpr_rpr2}
    \end{minipage}
	\begin{minipage}[b]{1.0\linewidth}
    	\begin{minipage}[b]{0.48\linewidth}
            \centering
        	\includegraphics[width=1.0\linewidth]{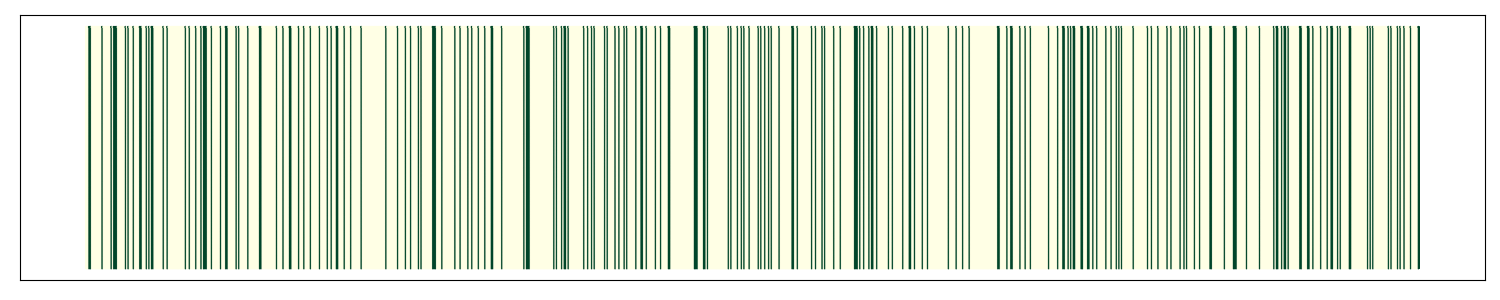}
        \end{minipage}
        \hfill
    	\begin{minipage}[b]{0.48\linewidth}
            \centering
        	\includegraphics[width=1.0\linewidth]{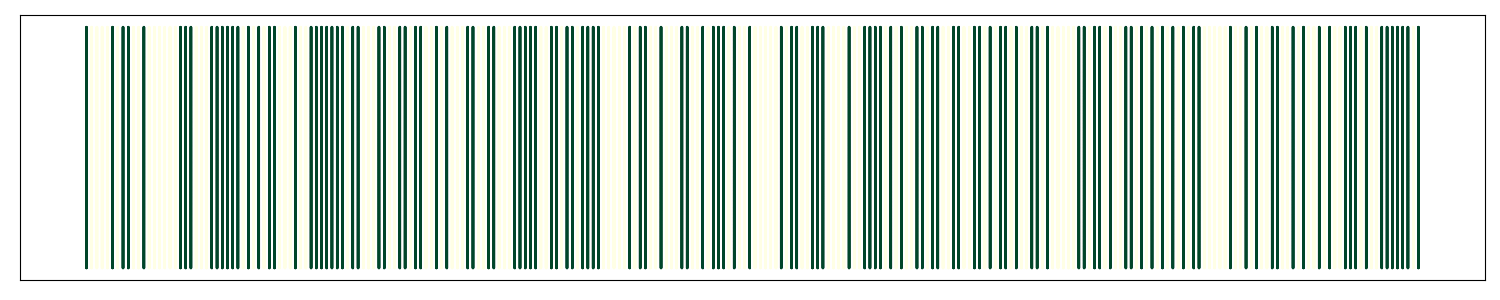}
        \end{minipage}
        \subcaption{Blur and Salt\&Pepper noise.}
        \label{image_act_rpr_rpr3}
    \end{minipage}
	\begin{minipage}[b]{1.0\linewidth}
    	\begin{minipage}[b]{0.48\linewidth}
            \centering
        	\includegraphics[width=1.0\linewidth]{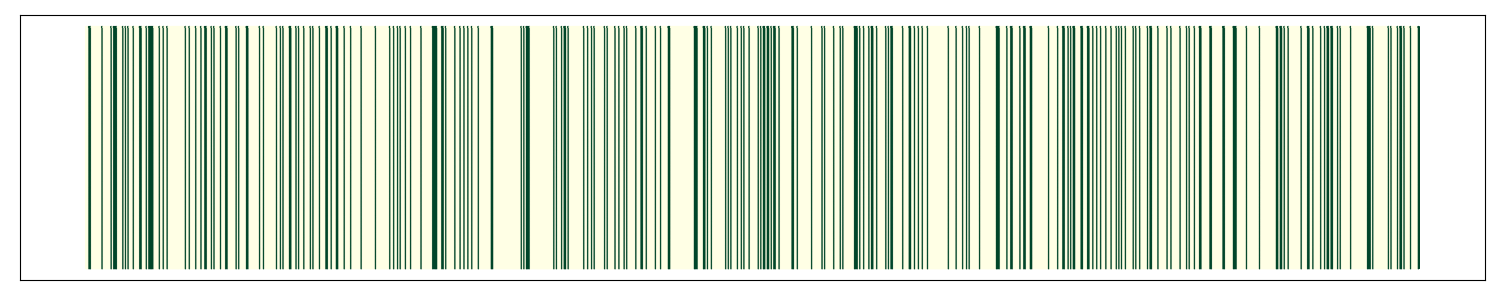}
        \end{minipage}
        \hfill
    	\begin{minipage}[b]{0.48\linewidth}
            \centering
        	\includegraphics[width=1.0\linewidth]{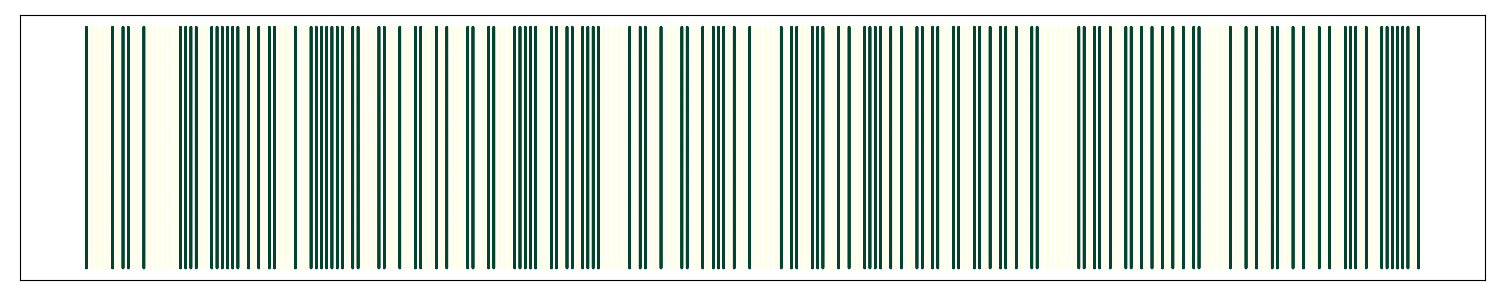}
        \end{minipage}
        \subcaption{Salt\&Pepper noise.}
        \label{image_act_rpr_rpr4}
    \end{minipage}
	\begin{minipage}[b]{1.0\linewidth}
    	\begin{minipage}[b]{0.48\linewidth}
            \centering
        	\includegraphics[width=1.0\linewidth]{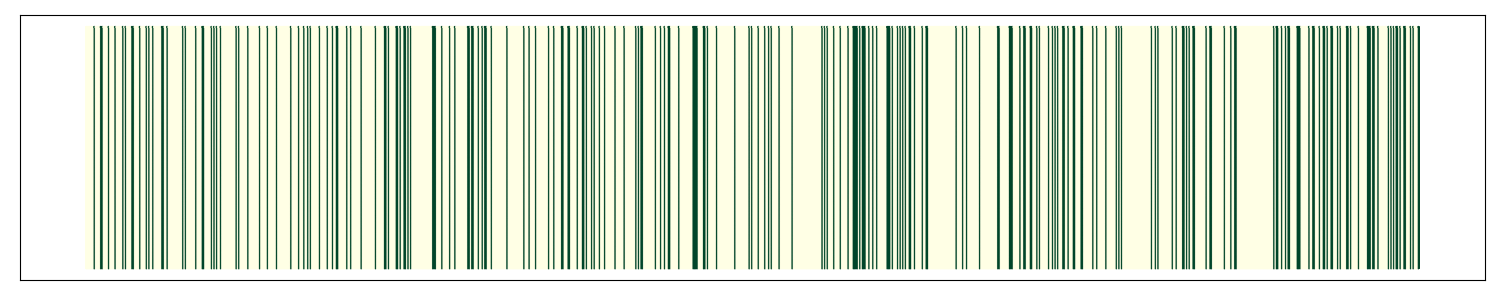}
        \end{minipage}
        \hfill
    	\begin{minipage}[b]{0.48\linewidth}
            \centering
        	\includegraphics[width=1.0\linewidth]{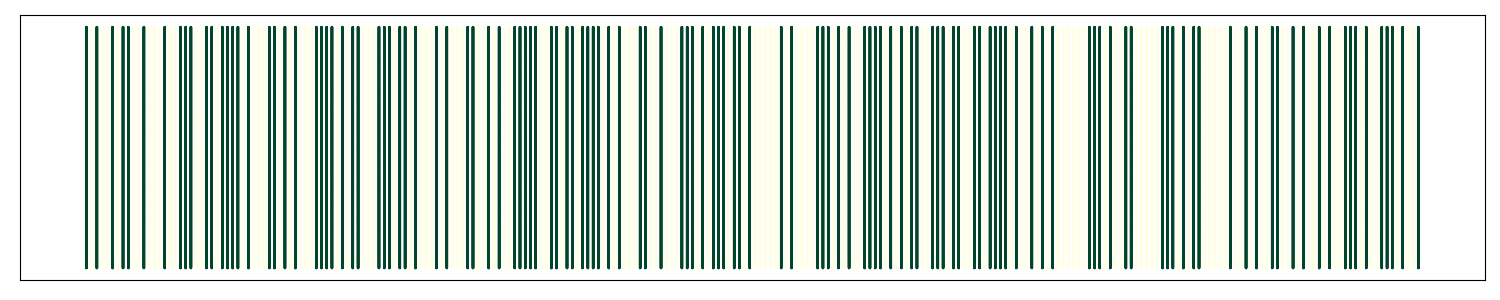}
        \end{minipage}
        \subcaption{Partial occlusion.}
        \label{image_act_rpr_rpr5}
    \end{minipage}
	\begin{minipage}[b]{1.0\linewidth}
    	\begin{minipage}[b]{0.48\linewidth}
            \centering
        	\includegraphics[width=1.0\linewidth]{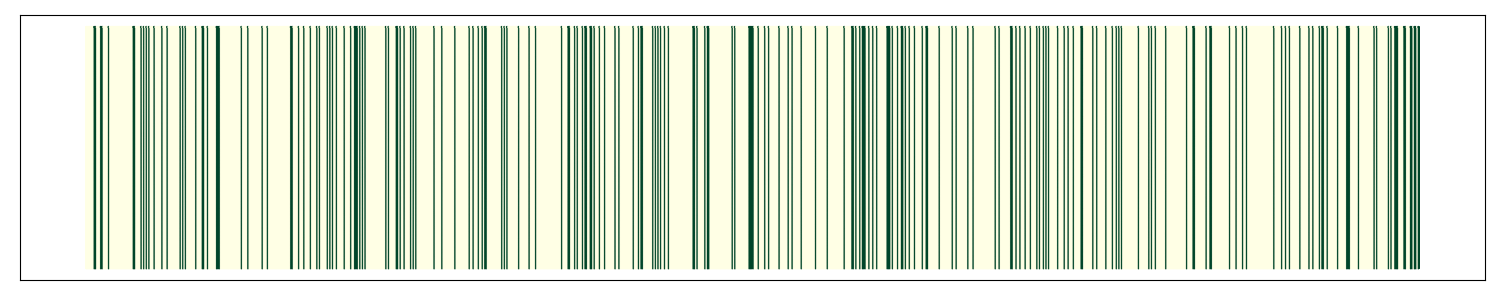}
        \end{minipage}
        \hfill
    	\begin{minipage}[b]{0.48\linewidth}
            \centering
        	\includegraphics[width=1.0\linewidth]{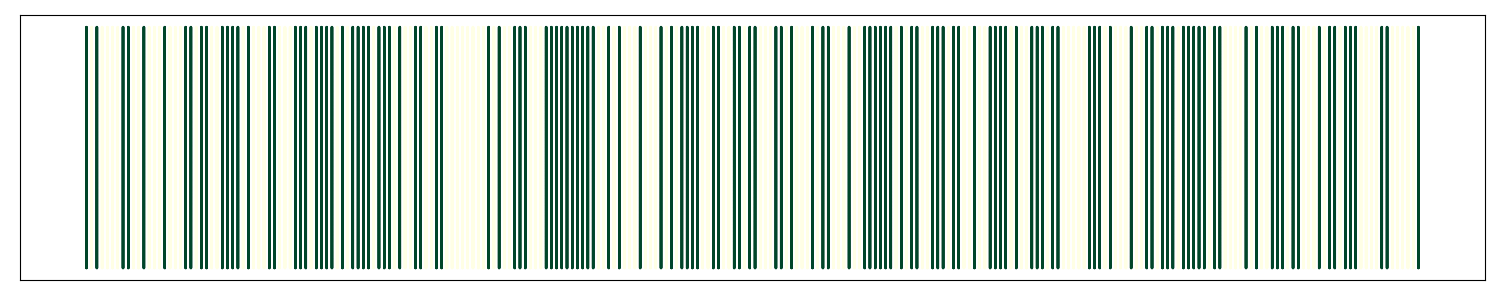}
        \end{minipage}
        \subcaption{Complete occlusion.}
        \label{image_act_rpr_rpr6}
    \end{minipage}
	\begin{minipage}[b]{1.0\linewidth}
    	\begin{minipage}[b]{0.48\linewidth}
            \centering
        	\includegraphics[width=1.0\linewidth]{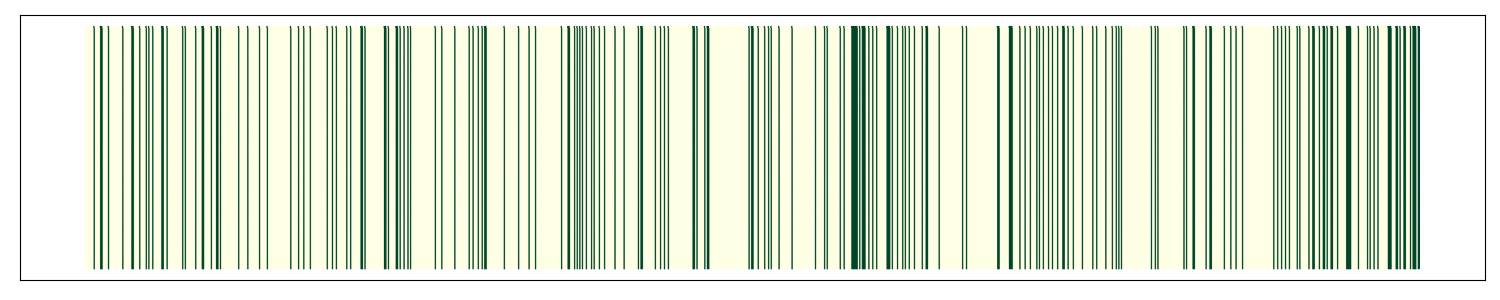}
        \end{minipage}
        \hfill
    	\begin{minipage}[b]{0.48\linewidth}
            \centering
        	\includegraphics[width=1.0\linewidth]{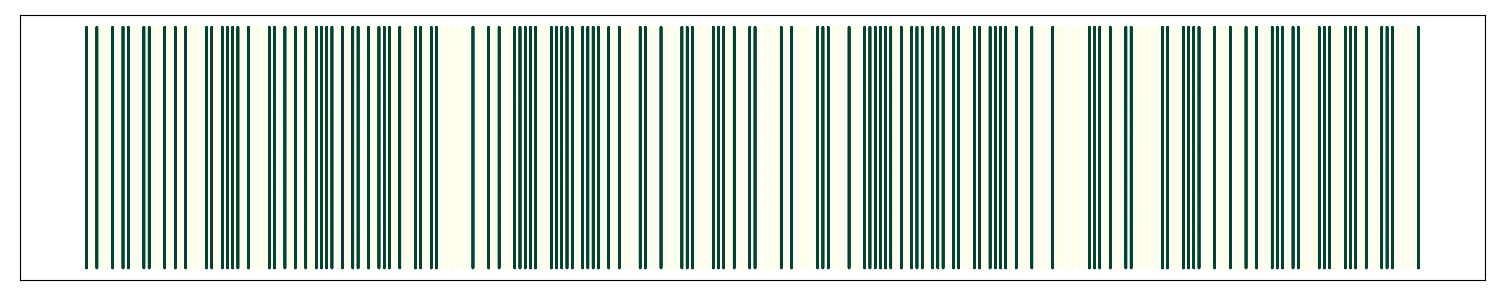}
        \end{minipage}
        \subcaption{Blur and partial occlusion.}
        \label{image_act_rpr_rpr7}
    \end{minipage}
	\begin{minipage}[b]{1.0\linewidth}
    	\begin{minipage}[b]{0.48\linewidth}
            \centering
        	\includegraphics[width=1.0\linewidth]{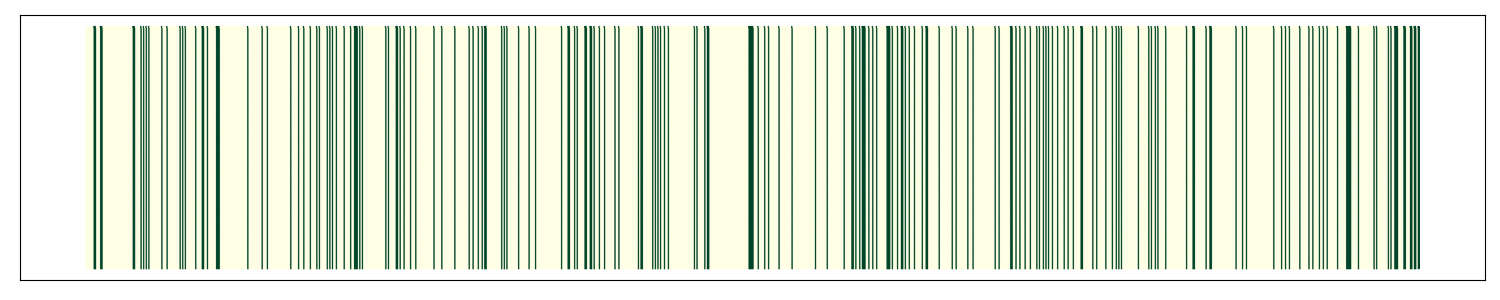}
        \end{minipage}
        \hfill
    	\begin{minipage}[b]{0.48\linewidth}
            \centering
        	\includegraphics[width=1.0\linewidth]{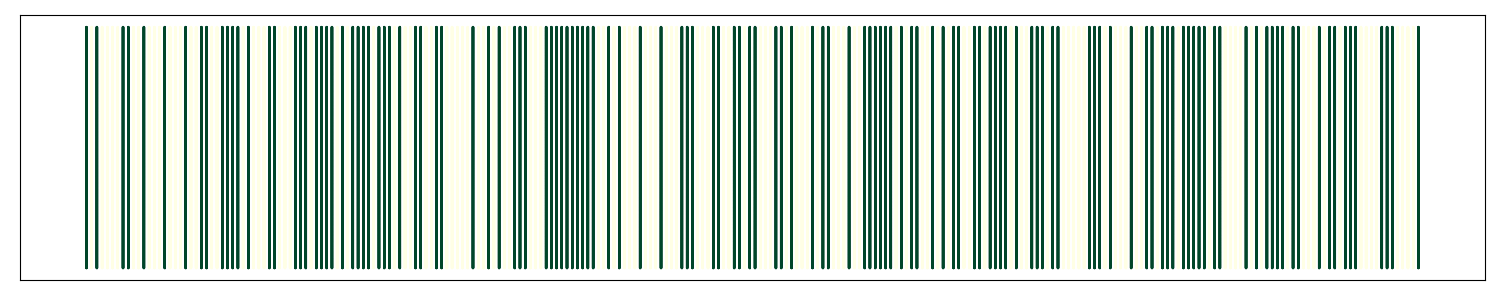}
        \end{minipage}
        \subcaption{Blur and complete occlusion.}
        \label{image_act_rpr_rpr8}
    \end{minipage}
    \caption{Activations for the third MMTM module for different image corruption techniques for $\text{RPR}_{\text{V}}$ (left) and $\text{RPR}_{\text{I}}$ (right).}
    \label{image_act_rpr_rpr}
\end{figure}

\noindent In this section, we evaluate the network masks (i.e., the sigmoid activation function outputs) of the MMTM modules in the $\text{APR}_{\text{V}}$-$\text{RPR}_{\text{I}}$ and $\text{RPR}_{\text{V}}$-$\text{RPR}_{\text{I}}$ fusion tasks. We visualize pairs of the continuous masks $E_\mathbf{A}$ and $E_\mathbf{B}$ as shown in Figure~\ref{image_mmtm}, which depict the feature selection mechanism of the features extracted from the visual and inertial encoders prior to their transmission to the temporal modeling and pose regression. The sigmoid activations ensures that each feature channel is re-weighted within the range of $[0, 1]$ based on its significance at a particular time step. To accomplish this, we corrupt the image input, as depicted in Figure~\ref{image_corr1}, and apply five distinct corruptions: image with Salt\&Pepper noise (Figure~\ref{image_corr2}), blurred image (Figure~\ref{image_corr3}), full occlusion (Figure~\ref{image_corr5}), partial occlusion with a rectangular patch (Figure~\ref{image_corr6}), and selecting a random corresponding image for $\text{RPR}_{\text{V}}$ (Figure~\ref{image_corr7}). This encompasses a range of real-time scenarios, including fast rotations, occlusions, and low-light conditions. Figure~\ref{image_masks} provides the number of sigmoid activations that are higher than 0.5 for the $\text{APR}_{\text{V}}$, $\text{RPR}_{\text{V}}$, and $\text{RPR}_{\text{I}}$ regression tasks for all image corruptions. With regards to the $\text{APR}_{\text{V}}$-$\text{RPR}_{\text{I}}$ fusion task (see Figure~\ref{image_masks1}), the number of sigmoid activations decreases for the image encoder and increases for the inertial encoder for varying image corruptions. This demonstrates that the fusion model progressively relies on the inertial data with an increasing level of image corruption. For the $\text{RPR}_{\text{V}}$-$\text{RPR}_{\text{I}}$ fusion model, we corrupt one of the two input images that correspond to each other. We provide the number of sigmoid activations in Figure~\ref{image_masks2}. The number of activations of the $\text{RPR}_{\text{V}}$ model decreases for all image corruptions, while the number of activations increases of the $\text{RPR}_{\text{I}}$ model. Particularly, as the level of noise increases (e.g., compare the number of activations for Salt\&Pepper noise with the number of activations for the combination of image blur and Salt\&Pepper noise), the fusion model becomes more reliant on the inertial encoder. This phenomenon can also be observed when comparing the combination of image blur and partial occlusion or for the combination of image blur and complete occlusion to complete occlusion alone. Upon the comparison of the $\text{APR}_{\text{V}}$-$\text{RPR}_{\text{I}}$ fusion with the $\text{RPR}_{\text{V}}$-$\text{RPR}_{\text{I}}$ fusion, it can be observed that the $\text{RPR}_{\text{V}}$ model contains higher activations compared to the $\text{APR}_{\text{V}}$ model. This is due to the increased reliability of the model resulting from the use of two consecutive images, rather than just a single image. Subsequently, we directly visualize the feature selection masks of the MMTM module employed in the $\text{APR}_{\text{V}}$-$\text{RPR}_{\text{I}}$ (see Figure~\ref{image_act_apr_rpr}) and in the $\text{RPR}_{\text{V}}$-$\text{RPR}_{\text{I}}$ (see Figure~\ref{image_act_rpr_rpr}) for various image corruptions. The fusion networks learn to assign a greater weight to the inertial features when the images are degraded (as evidenced by the thick green lines of the $\text{RPR}_{\text{I}}$ model compared to the $\text{APR}_{\text{V}}$ and $\text{RPR}_{\text{V}}$ models). This highlights that the networks have learned to place more importance on a complementary sensor in order to perform the regression task in the presence of a challenging image input.

\section{Conclusion}
\label{chap_conclusion}

\noindent We investigated deep multimodal fusion between the visual APR task supported with inertial RPR and between the visual-inertial RPR tasks. As baseline models, we utilized PoseNet for $\text{APR}_{\text{V}}$, IMUNet for $\text{RPR}_{\text{I}}$, and FlowNet for $\text{RPR}_{\text{V}}$. In ordr to attain globally consistent pose predictions during interference, we analyzed and compared various techniques including MapNet, pose graph optimization, late fusion techniques such as concatenation and selective sensor fusion with BiLSTM layers, intermediate fusion with transfer modules, auxiliary learning, and Bayesian learning. Our assessments on the EuRoC MAV aerial vehicle dataset, the handheld PennCOSYVIO dataset, and our novel large-scale IndustryVI indoor dataset serve as a comprehensive benchmark for the robustness of fusion techniques across various challenging environments and motion dynamics. In conclusion, the results and key findings can be succinctly summarized as follows: (1) The $\text{APR}_{\text{V}}$-$\text{RPR}_{\text{I}}$+PGO approach and the intermediate fusion with the MMTM technique demonstrate superiority over other techniques for the $\text{APR}_{\text{V}}$-$\text{RPR}_{\text{I}}$ task on the EuRoC MAV dataset. These methods exhibit an improved capacity for generalization on the dataset with smoother predicted trajectories. (2) Selective sensor fusion and fusion with MMTM exhibit superiority on the PennCOSYVIO and IndustryVI datasets. (3) In addition, the MMTM fusion technique yields the highest performance on the $\text{RPR}_{\text{V}}$-$\text{RPR}_{\text{I}}$ task. (4) Fusing three MMTM modules is more advantageous than fusing one or two MMTM modules as it results in a more generalized representation between both modalities. (5) For all the datasets, the non-linear-based auxiliary learning approach enhances the performances of the main task. (6) The estimation of aleatoric uncertainty using Bayesian networks provides valuable insights into the model's robustness against challenging images. (7) We examined the network activations by subjecting the image inputs to various disruption techniques. Upon increasing the image corruption, the number of high softmax activations in the visual model increased, whereas the number in the inertial model decreased, indicating higher reliability on the inertial model for challenging images. (8) The IMU bias has a considerable impact on the performance of RPR-only methods, as evidenced by the deviation of the trajectories in the appendix, while the relative pose -- even with its inherent noise -- can still be effectively utilized to smooth the absolute trajectory.

\section{Acknowledgments}
\noindent This work was supported by the Federal Ministry of Education and Research (BMBF) of Germany by Grant No. 01IS18036A (David R\"ugamer) and by the Bavarian Ministry for Economic Affairs, Infrastructure, Transport and Technology through the Center for Analytics-Data-Applications within the framework of ``BAYERN DIGITAL II''.

\bibliographystyle{plain}
\bibliography{T-RO2023}

\begin{thebibliography}{100}

\bibitem{almalioglu}
Yasin Almalioglu, Mehmet Turan, Alp~Eren Sari, Muhamad Risqi~U. Saputra, Pedro
  P.~B. de~Gusm\~{a}o, Andrew Markham, and Niki Trigoni.
\newblock {SelfVIO: Self-Supervised Deep Monocular Visual-Inertial Odometry and
  Depth Estimation}.
\newblock In {\em \href{https://arxiv.org/abs/1911.09968}{arXiv:1911.09968}},
  November 2019.

\bibitem{majdik}
{Andr\'{a}s L. Majdik and Charles Till and Davide Scaramuzza}.
\newblock {The Zurich Urban Micro Aerial vehicle Dataset}.
\newblock In {\em \href{https://doi.org/10.1177/0278364917702237}{IJRR}},
  volume {36(3)}, April 2017.

\bibitem{anoosheh}
Asha Anoosheh, Torsten Sattler, Radu Rimofte, Marc Pollefeys, and Luc~Van Gool.
\newblock {Night-to-Day Image Translation for Retrieval-based Localization}.
\newblock In {\em \href{https://ieeexplore.ieee.org/document/8794387}{ICRA}},
  Montreal, QC, May 2019.

\bibitem{relocnet}
Vassileios Balntas, Shuda Li, and Victor Prisacariu.
\newblock {RelocNet: Continuous Metric Learning Relocalisation Using Neural
  Nets}.
\newblock In {\em
  \href{https://openaccess.thecvf.com/content_ECCV_2018/html/Vassileios_Balntas_RelocNet_Continous_Metric_ECCV_2018_paper.html}{ECCV}},
  pages 751--767, Munich, Germany, September 2018.

\bibitem{beauregard}
St\'{e}phane Beauregard and Harald Haas.
\newblock {Pedestrian Dead Reckoning: A Basis for Personal Positioning}.
\newblock In {\em
  \href{https://www.semanticscholar.org/paper/Pedestrian-Dead-Reckoning-\%3A-A-Basis-for-Personal-Beauregard-Haas/f5ad3ca7e33c8cd3277d33f76a2d4fc3c1114cbf}{Positioning,
  Navigation and Communication}}, pages 27--35, 2006.

\bibitem{blanco}
Jos{\'e}-Luis Blanco-Claraco, Francisco \'{A}ngel Moreno-Duenas, and Javier
  Gonz{\'a}lez-Jim{\'e}nez.
\newblock {The M{\'a}laga Urban Dataset: High-Rate Stereo and LiDAR in a
  Realistic Urban Scenario}.
\newblock In {\em
  \href{https://journals.sagepub.com/doi/10.1177/0278364913507326}{IJRR}},
  volume {33(2)}, pages 207--214, October 2014.

\bibitem{bloesch}
Michael Bloesch, Michael Burri, Sammy Omari, Marco Hutter, and Roland Siegwart.
\newblock {Iterated Extended Kalman Filter Based Visual-Inertial Odometry Using
  Direct Photometric Feedback}.
\newblock In {\em \href{https://doi.org/10.1177/0278364917728574}{IJRR}},
  volume {36(10)}, pages 1053--1072, September 2017.

\bibitem{bloesch_omari}
Michael Bloesch, Sammy Omari, Marco Hutter, and Roland Siegwart.
\newblock {Robust Visual Inertial Odometry Using a Direct EKF-Based Approach}.
\newblock In {\em \href{https://ieeexplore.ieee.org/document/7353389}{IROS}},
  pages 298--304, Hamburg, Germany, October 2015.

\bibitem{bower}
Matt Bower, Cathie Howe, Nerida McCredie, Austin Robinson, and David Grover.
\newblock {Augmented Reality in Education — Cases, Places, and Potentials}.
\newblock In {\em
  \href{https://ieeexplore.ieee.org/document/6820176?reload=true}{ICEM}},
  Singapore, Singapore, October 2002.

\bibitem{brachmann_humenberger}
Eric Brachmann, Martin Humenberger, Carsten Rother, and Torsten Sattler.
\newblock {On the Limits of Pseudo Ground Truth in Visual Camera
  Re-Localisation}.
\newblock In {\em
  \href{https://openaccess.thecvf.com/content/ICCV2021/html/Brachmann_On_the_Limits_of_Pseudo_Ground_Truth_in_Visual_Camera_ICCV_2021_paper.html}{CVPR}},
  pages 6218--6228, September 2021.

\bibitem{brachmann_dsac}
Eric Brachmann, Alexander Krull, Sebastian Nowozin, Jamie Shotton, Frank
  Michel~Stefan Gumhold, and Carsten Rother.
\newblock {DSAC — Differentiable RANSAC for Camera Localization}.
\newblock In {\em \href{https://ieeexplore.ieee.org/document/8099750}{CVPR}},
  pages 2492--2500, Honolulu, HI, 2017.

\bibitem{brachmann_michel}
Eric Brachmann, Frank Michel, Alexander Krull, Michael~Ying Yang, Stefan
  Gumhold, and Carsten Rother.
\newblock {Uncertainty-Driven 6D Pose Estimation of Objects and Scenes from a
  Single RGB Image}.
\newblock In {\em \href{https://ieeexplore.ieee.org/document/7780735}{CVPR}},
  pages 3364--3372, Las Vegas, NV, June 2016.

\bibitem{brachmann}
Eric Brachmann and Carsten Rother.
\newblock {Learning Less is More - 6D Camera Localization via 3D Surface
  Regression}.
\newblock In {\em \href{https://ieeexplore.ieee.org/document/8578587}{CVPR}},
  pages 4654--4662, Salt Lake City, UT, June 2018.

\bibitem{brahmbhatt}
Samarth Brahmbhatt, Jinwei Gu, Kihwan Kim, James Hays, and Jan Kautz.
\newblock {Geometry-Aware Learning of Maps for Camera Localization}.
\newblock In {\em \href{https://ieeexplore.ieee.org/document/8578375}{CVPR}},
  pages 2616--2625, Salt Lake City, UT, June 2018.

\bibitem{brajdic}
Agata Brajdic and Robert Harle.
\newblock {Walk Detection and Step Counting on Unconstrained Smartphones}.
\newblock In {\em
  \href{https://dl.acm.org/doi/10.1145/2493432.2493449}{IJCPUC}}, pages
  225--234, September 2013.

\bibitem{brieger_ion_gnss}
Tobias Brieger, Nisha~Lakshmana Raichur, Dorsaf Jdidi, Felix Ott, Tobias Feigl,
  Johannes~Rossouw van~der Merwe, Alexander Rügamer, and Wolfgang Felber.
\newblock {Multimodal Learning for Reliable Interference Classification in GNSS
  Signals}.
\newblock In {\em
  \href{https://www.ion.org/gnss/abstracts.cfm?paperID=11411}{ION GNSS+}},
  pages 3210--3234, Denver, CO, September 2022.

\bibitem{brox}
Thomas Brox, Andr\'{e}s Bruhn, Nils Papenberg, and Joachim Weickert.
\newblock {High Accuracy Optical Flow Estimation Based on a Theory for
  Warping}.
\newblock In {\em
  \href{https://link.springer.com/chapter/10.1007/978-3-540-24673-2_3}{ECCV}},
  volume 3024, pages 25--36, 2004.

\bibitem{burri}
Michael Burri, Janosch Nikolic, Pascal Gohl, Thomas Schneider, Joern Rehder,
  Sammy Omari, Markus~W. Achtelik, and Roland Siegwart.
\newblock {The EuRoC Micro Aerial Vehicle Datasets}.
\newblock In {\em
  \href{https://journals.sagepub.com/doi/full/10.1177/0278364915620033}{IJRR}},
  volume {35(10)}, pages 1157--1163, January 2016.

\bibitem{butyrev2019}
Leonid Butyrev, Thorsten Edelh\"au{\ss}er, and Christopher Mutschler.
\newblock {Deep Reinforcement Learning for Motion Planning of Mobile Robots}.
\newblock In {\em \href{https://arxiv.org/abs/1912.09260}{arXiv:1912.09260}},
  December 2019.

\bibitem{campos_elvira}
Carlos Campos, Richard Elvira, Juan J.~G\'{o}mez Rodr\'{i}guez, Jos\'{e} M.~M.
  Montiel, and Juan~D. Tard\'{o}s.
\newblock {ORB-SLAM3: An Accurate Open-Source Library for Visual,
  Visual-Inertial, and Multimap SLAM}.
\newblock In {\em \href{https://ieeexplore.ieee.org/document/9440682}{T-RO}},
  volume {37(6)}, pages 1874--1890, May 2021.

\bibitem{carlevaris}
Nicholas Carlevaris-Bianco, Arash~K. Ushani, and Ryan~M. Eustice.
\newblock {University of Michigan North Campus Long-Term Vision and Lidar
  Dataset}.
\newblock In {\em
  \href{https://journals.sagepub.com/doi/10.1177/0278364915614638}{IJRR}},
  volume {35(9)}, December 2015.

\bibitem{castle}
Robert~Oliver Castle, Darren~J. Gawley, Georg Klein, and David~William Murray.
\newblock {Towards Simultaneous Recognition, Localization and Mapping for
  Hand-Held and Wearable Cameras}.
\newblock In {\em \href{https://ieeexplore.ieee.org/document/4209727}{ICRA}},
  pages 4102--4107, Rome, Italy, April 2007.

\bibitem{inertial_navigation}
Moises~J. Castro-Toscano, Julio C. Rodr\'{i}guez-Qui\ {n}onez, Daniel
  Hern\'{a}ndez-Balbuena, Lars Lindner, Oleg Sergiyenko, Moises Rivas-Lopez,
  and Wendy Flores-Fuentes.
\newblock {A Methodological use of Inertial Navigation Systems for Strapdown
  Navigation Task}.
\newblock In {\em
  \href{https://ieeexplore.ieee.org/abstract/document/8001484}{ISIE}}, pages
  1589--1595, Edinburgh, UK, June 2017.

\bibitem{Lu}
Changhao Chen, Xiaoxuan Lu, Andrew Markham, and Niki Trigoni.
\newblock {IONet: Learning to Cure the Curse of Drift in Inertial Odometry}.
\newblock In {\em
  \href{https://ojs.aaai.org/index.php/AAAI/article/view/12102}{AAAI Technical
  Tracks: Robotics}}, volume {32(1)}, January 2018.

\bibitem{chen}
Changhao Chen, Stefano Rosa, Yishu Miao, Chris~Xiaoxuan Lu, Wei Wu, Andrew
  Markham, and Niki Trigoni.
\newblock {Selective Sensor Fusion for Neural Visual-Inertial Odometry}.
\newblock In {\em
  \href{https://openaccess.thecvf.com/content_CVPR_2019/html/Chen_Selective_Sensor_Fusion_for_Neural_Visual-Inertial_Odometry_CVPR_2019_paper.html}{CVPR}},
  pages 10542--10551, Long Beach, CA, June 2019.

\bibitem{chen_vijay}
Zhao Chen, Vijay Badrinarayanan, Chen-Yu Lee, and Andrew Rabinovich.
\newblock {Gradnorm: Gradient Normalization for Adaptive Loss Balancing in Deep
  Multitask Networks}.
\newblock In {\em \href{http://proceedings.mlr.press/v80/chen18a.html}{ICML}},
  volume~80, pages 794--803, 2018.

\bibitem{gposenet}
Ming Chi, Chunhua Shen, and Ian Reid.
\newblock {A Hybrid Probabilistic Model for Camera Relocalization}.
\newblock In {\em \href{http://bmvc2018.org/contents/papers/0799.pdf}{BMVC}},
  York, UK, 2018.

\bibitem{clark_vidloc}
Ronald Clark, Sen Wang, Andrew Markham, Niki Trigoni, and Hongkai Wen.
\newblock {VidLoc: A Deep Spatio-Temporal Model for 6-DoF Video-Clip
  Relocalization}.
\newblock In {\em \href{https://ieeexplore.ieee.org/document/8099767}{CVPR}},
  pages 2652--2660, Honolulu, HI, July 2017.

\bibitem{clark}
Ronald Clark, Sen Wang, Hongkai Wen, Andrew Markham, and Niki Trigoni.
\newblock {VINet: Visual-Inertial Odometry as a Sequence-to-Sequence Learning
  Problem}.
\newblock In {\em \href{https://dl.acm.org/doi/10.5555/3298023.3298149}{AAAI}},
  pages 3995--4001, February 2017.

\bibitem{constante}
Gabriele Costante and Thomas~Alessandro Ciarfuglia.
\newblock {LS-VO: Learning Dense Optical Subspace for Robust Visual Odometry
  Estimation}.
\newblock In {\em
  \href{https://ieeexplore.ieee.org/abstract/document/8283696}{RA-L}}, volume
  {3(3)}, pages 1735--1742, February 2018.

\bibitem{pcnn}
Gabriele Costante, Michele Mancini, Paolo Valigi, and Thomas~A. Ciarfuglia.
\newblock {Exploring Representation Learning With CNNs for Frame-to-Frame
  Ego-Motion Estimation}.
\newblock In {\em
  \href{https://ieeexplore.ieee.org/document/7347378?reload=true}{RA-L}},
  volume {1(1)}, pages 18--25, December 2015.

\bibitem{das}
Anweshan Das and Gijs Dubbelman.
\newblock {An Experimental Study on Relative and Absolute Pose Graph Fusion for
  Vehicle Localization}.
\newblock In {\em
  \href{https://ieeexplore.ieee.org/document/8500512}{Intelligent Vehicles
  Symposium (IV)}}, pages 630--635, Changshu, China, June 2018.

\bibitem{github_decayale}
DecaYale.
\newblock {GitHub VLocNet Implementation}.
\newblock In {\em \url{https://github.com/DecaYale/VLocNet}}, 2019.

\bibitem{dengt}
Jia Deng, Wei Dong, Richard Socher, Li-Jia Li, Kai Li, and Li~Fei-Fei.
\newblock {ImageNet: A Large-Scale Hierarchical Image Database}.
\newblock In {\em
  \href{https://ieeexplore.ieee.org/document/5206848?arnumber=5206848&tag=1}{CVPR}},
  pages 248--255, Miami, FL, June 2009.

\bibitem{camnet_ding}
Mingyu Ding, Zhe Wang, Jiankai Sun, Jianping Shi, and Ping Luo.
\newblock {CamNet: Coarse-to-Fine Retrieval for Camera Re-Localization}.
\newblock In {\em \href{https://ieeexplore.ieee.org/document/9008579}{ICCV}},
  pages 2871--2880, Seoul, South Korea, October 2019.

\bibitem{silva}
Jo{\~a}o Paulo~Silva do~Monte~Lima, Hideaki Uchiyama, and Rin ichiro Taniguchi.
\newblock {End-to-End Learning Framework for IMU-based 6-DOF Odometry}.
\newblock In {\em \href{https://www.mdpi.com/1424-8220/19/17/3777}{MDPI
  Sensors}}, volume {19(17)}, page 3777, August 2019.

\bibitem{dosovitskiy}
Alexey Dosovitskiy, Philipp Fischer, Eddy Ilg, Philipp H{\"a}usser, Caner
  Hazirbas, Vladimir Golkov, Patrick van~der Smagt, Daniel Cremers, and Thomas
  Brox.
\newblock {FlowNet: Learning Optical Flow with Convolutional Networks}.
\newblock In {\em \href{https://ieeexplore.ieee.org/document/7410673}{ICCV}},
  pages 2758--2766, Santiago de Chile, Chile, December 2015.

\bibitem{el}
Naser El-Sheimy, Haiying Hou, and Xiaoji Niu.
\newblock {Analysis and Modeling of Inertial Sensors Using Allan Variance}.
\newblock In {\em \href{https://ieeexplore.ieee.org/document/4404126}{Trans. on
  Instrumentation and Measurement}}, volume {57(1)}, pages 140--149, January
  2008.

\bibitem{engel_dso}
Jakob Engel, Vladlen Koltun, and Daniel Cremers.
\newblock {Direct Sparse Odometry}.
\newblock In {\em \href{https://ieeexplore.ieee.org/document/7898369}{TPAMI}},
  volume {40(3)}, pages 611--625, April 2017.

\bibitem{engel}
Jakob Engel, Thomas Sch{\"o}ps, and Daniel Cremers.
\newblock {LSD-SLAM: Large-Scale Cirect Monocular SLAM}.
\newblock In {\em
  \href{https://link.springer.com/chapter/10.1007/978-3-319-10605-2_54}{ECCV}},
  volume 8690, pages 834--849, 2014.

\bibitem{faessler}
Matthias Faessler, Flavio Fontana, Christian Forster, Elias Mueggler, Matia
  Pizzoli, and Davide Scaramuzza.
\newblock {Autonomous, Vision-based Flight and Live Dense 3D Mapping with a
  Quadrotor Micro Aerial Vehicle}.
\newblock In {\em \href{https://dl.acm.org/doi/10.1002/rob.21581}{Journal Field
  Robotics}}, volume {33(4)}, pages 431--450, June 2016.

\bibitem{feng}
Zheyu Feng, Jianwen Li, Lundong Zhang, and Chen Chen.
\newblock {Online Spatial and Temporal Calibration for Monocular Direct
  Visual-Inertial Odometry}.
\newblock In {\em \href{https://www.mdpi.com/1424-8220/19/10/2273}{MDPI
  Sensors}}, volume {19(10)}, May 2019.

\bibitem{forster}
Christian Forster, Zichao Zhang, Michael Gassner, Manuel Werlberger, and Davide
  Scaramuzza.
\newblock {SVO: Semidirect Visual Odometry for Monocular and Multicamera
  Systems}.
\newblock In {\em \href{https://ieeexplore.ieee.org/document/7782863}{T-RO}},
  volume {33(2)}, pages 249--265, April 2017.

\bibitem{gal}
Yarin Gal and Zoubin Ghahramani.
\newblock {Bayesian Convolutional Neural Networks with Bernoulli Approximate
  Variational Inference}.
\newblock In {\em \href{https://arxiv.org/abs/1506.02158}{arXiv:1506.02158}},
  June 2015.

\bibitem{gao_hou}
Xiao-Shan Gao, Xiao-Rong Hou, Jianliang Tang, and Hang-Fei Cheng.
\newblock {Complete Solution Classification for the Perspective-Three-Point
  Problem}.
\newblock In {\em \href{https://ieeexplore.ieee.org/document/1217599}{TPAMI}},
  pages 930--943, August 2003.

\bibitem{geiger}
Andreas Geiger, Philip Lenz, Christoph Stiller, and Raquel Urtasun.
\newblock {Vision Meets Robotics: The KITTI Dataset}.
\newblock In {\em
  \href{https://journals.sagepub.com/doi/full/10.1177/0278364913491297}{IJRR}},
  volume {32(11)}, pages 1231--1237, August 2013.

\bibitem{geiger_kitti}
Andreas Geiger, Philip Lenz, and Raquel Urtasun.
\newblock {Are we Ready for Autonomous Driving? The KITTI Vision Benchmark
  Suite}.
\newblock In {\em \href{https://ieeexplore.ieee.org/document/6248074_}{CVPR}},
  pages 3354--3361, Providence, RI, June 2012.

\bibitem{goel}
Puneet Goel, Stergios~I. Roumeliotis, and Gaurav~S. Sukhatme.
\newblock {Robust Localization using Relative and Absolute Position Estimates}.
\newblock In {\em \href{https://ieeexplore.ieee.org/document/812832}{Intl.
  Conf. on Intelligent Robots and Systems. Human and Environment Friendly
  Robots with High Intelligence and Emotional Quotients}}, volume~2, pages
  1134--1140, Kyongju, South Korea, October 1999.

\bibitem{graves_liwicki}
Alex Graves, Marcus Liwicki, Santiago Fern\'{a}ndez, Roman Bertolami, Horst
  Bunke, and Jürgen Schmidhuber.
\newblock {A Novel Connectionist System for Unconstrained Handwriting
  Recognition}.
\newblock In {\em \href{https://ieeexplore.ieee.org/document/4531750}{TPAMI}},
  volume {31(5)}, pages 855--868, May 2009.

\bibitem{matthew}
Matthew~Koichi Grimes, Dragomir Anguelov, and Yann LeCun.
\newblock {Hybrid Hessians for Flexible Optimization of Pose Graphs}.
\newblock In {\em \href{https://ieeexplore.ieee.org/document/5650091}{Intl.
  Conf. on Intelligent Robots and Systems}}, Taipei, Taiwan, October 2010.

\bibitem{han}
Dai-In Han, Timothy Jung, and Alex Gibson.
\newblock {Dublin AR: Implementing Augmented Reality in Tourism}.
\newblock In {\em
  \href{https://www.springerprofessional.de/dublin-ar-implementing-augmented-reality-in-tourism/2076568}{ICTT}},
  pages 511--523, 2013.

\bibitem{han_lim}
Liming Han, Yimin Lin, Guoguang Du, and Shiguo Lian.
\newblock {DeepVIO: Self-supervised Deep Learning of Monocular Visual Inertial
  Odometry using 3D Geometric Constraints}.
\newblock In {\em \href{https://ieeexplore.ieee.org/document/8968467}{IROS}},
  Macau, China, November 2019.

\bibitem{hochreiter}
Sepp Hochreiter and Jürgen Schmidhuber.
\newblock {Long Short-Term Memory}.
\newblock In {\em \href{https://ieeexplore.ieee.org/document/6795963}{Neural
  Computation}}, volume {9(8)}, pages 1735--1780, November 1997.

\bibitem{hong}
Euntae Hong and Jongwoo Lim.
\newblock {Visual-Inertial Odometry with Robust Initialization and Online Scale
  Estimation}.
\newblock In {\em
  \href{https://www.ncbi.nlm.nih.gov/pmc/articles/PMC6308559/}{MDPI Sensors}},
  volume {18(12)}, page 4287, December 2018.

\bibitem{horn}
Berthold K.~P. Horn.
\newblock {Closed-form Solution of Absolute Orientation Using Unit
  Quaternions}.
\newblock In {\em
  \href{https://www.osapublishing.org/josaa/abstract.cfm?uri=josaa-4-4-629}{Journal
  of the Optical Society of America}}, volume {4(4)}, pages 629--642, April
  1987.

\bibitem{hu}
Di~Hu, Chengze Wang, Feiping Nie, and Xuelong Li.
\newblock {Dense Multimodal Fusion for Hierarchically Joint Representation}.
\newblock In {\em \href{https://ieeexplore.ieee.org/document/8683898}{ICASSP}},
  pages 3941--3945, Brighton, UK, May 2019.

\bibitem{hu2018squeeze}
Jie Hu, Li~Shen, and Gang Sun.
\newblock {Squeeze-and-excitation networks, 7132--7141}.
\newblock In {\em \href{https://ieeexplore.ieee.org/document/8578843}{(CVPR)}},
  2018.

\bibitem{hu_imu}
Jwu-Sheng Hu and Ming-Yuan Chen.
\newblock {A Sliding-Window Visual-IMU Odometer Based on Tri-Focal Tensor
  Geometry}.
\newblock In {\em
  \href{https://ieeexplore.ieee.org/abstract/document/6907434}{ICRA}}, pages
  3963--3968, Hong Kong, China, May 2014.

\bibitem{huang_xu}
Zhaoyang Huang, Yan Xu, Jianping Shi, Xiaowei Zhou, Hujun Bao, and Guofeng
  Zhang.
\newblock {Prior Guided Dropout for Robust Visual Localization in Dynamic
  Environments}.
\newblock In {\em \href{https://ieeexplore.ieee.org/document/9008288}{ICCV}},
  pages 2791--2800, Seoul, Korea, October 2019.

\bibitem{lim}
{Hyon Lim and Sudipta N. Sinha and Michael F. Cohen and Matthew Uyttendaele}.
\newblock Real-time monocular image-based 6-dof localization in large-scale
  environments.
\newblock In {\em
  \href{https://ieeexplore.ieee.org/abstract/document/6247782}{CVPR}}, pages
  476--492, Providence, RI, June 2015.

\bibitem{flownet2}
Eddy Ilg, Nikolaus Mayer, Tonmoy Saikia, Margret Keuper, Alexey Dosovitskiy,
  and Thomas Brox.
\newblock {FlowNet 2.0: Evolution of Optical Flow Estimation with Deep
  Networks}.
\newblock In {\em \href{https://ieeexplore.ieee.org/document/8099662}{CVPR}},
  pages 1647--1655, Honolulu, HI, July 2017.

\bibitem{ctcnet}
Ganesh Iyer, J.~Krishna Murthy, Gunshi Gupta, K.~Madhava Krishna, and Liam
  Paull.
\newblock {Geometric Consistency for Self-Supervised End-to-End Visual
  Odometry}.
\newblock In {\em \href{https://ieeexplore.ieee.org/document/8575525}{CVPRW}},
  Salt Lake City, UT, June 2018.

\bibitem{jaderberg}
Max Jaderberg, Volodymyr Mnih, Wojciech~Marian Czarnecki, Tom Schaul, Joel~Z.
  Leibo, David Silver, and Koray Kavukcuoglu.
\newblock {Reinforcement Learning with Unsupervised Auxiliary Tasks}.
\newblock In {\em \href{https://arxiv.org/abs/1611.05397}{arXiv:1611.05397}},
  November 2016.

\bibitem{marker2}
Gijeong Jang, Sungho Lee, and Inso Kweon.
\newblock {Color Landmark Based Self-Localization for Indoor Mobile Robots}.
\newblock In {\em \href{https://ieeexplore.ieee.org/document/1013492}{ICRA}},
  pages 1037--1042, Washington, DC, May 2002.

\bibitem{joze}
Hamid Reza~Vaezi Joze, Amirreza Shaban, Michael~L. Iuzzolino, and Kazuhito
  Koishida.
\newblock {MMTM: Multimodal Transfer Module for CNN Fusion}.
\newblock In {\em \href{https://ieeexplore.ieee.org/document/9156844}{CVPR}},
  pages 13289--13299, Seattle, WA, June 2020.

\bibitem{kaess}
Michael Kaess, Hordur Johannsson, Richard Roberts, Viorela Ila, John Leonard,
  and Frank Dellaert.
\newblock {iSAM2: Incremental Smoothing and Mapping Using the Bayes Tree}.
\newblock In {\em \href{https://ieeexplore.ieee.org/document/5979641}{IJRR}},
  pages 216--235, Shanghai, China, May 2012.

\bibitem{karpathy}
Andrej Karpathy, George Toderici, Sanketh Shetty, Thomas Leung, Rahul
  Sukthankar, and Li~Fei-Fei.
\newblock {Large-Scale Video Classification with Convolutional Neural
  Networks}.
\newblock In {\em \href{https://ieeexplore.ieee.org/document/6909619}{CVPR}},
  pages 1725--1732, Columbus, OH, June 2014.

\bibitem{kasyanov}
Anton Kasyanov, Francis Engelmann, J{\"o}rg St{\"u}ckler, and Bastian Leibe.
\newblock {Keyframe-based Visual-Inertial Online SLAM with Relocalization}.
\newblock In {\em \href{https://ieeexplore.ieee.org/document/8206581}{IROS}},
  pages 6662--6669, Vancouver, Canada, September 2017.

\bibitem{kendall_modelling}
Alex Kendall and Roberto Cipolla.
\newblock {Modelling Uncertainty in Deep Learning for Camera Relocalization}.
\newblock In {\em \href{https://ieeexplore.ieee.org/document/7487679}{ICRA}},
  pages 4762--4769, Stockholm, Sweden, May 2016.

\bibitem{kendall_cipolla}
Alex Kendall and Roberto Cipolla.
\newblock {Geometric Loss Functions for Camera Pose Regression with Deep
  Learning}.
\newblock In {\em \href{https://ieeexplore.ieee.org/document/8100177/}{CVPR}},
  pages 6555--6564, Honolulu, HI, July 2017.

\bibitem{kendall_uncertainty}
Alex Kendall and Yarin Gal.
\newblock {What Uncertainties Do We Need in Bayesian Deep Learning for Computer
  Vision?}
\newblock In {\em NIPS}, pages 5580--5590, December 2017.

\bibitem{kendall}
Alex Kendall, Matthew Grimes, and Roberto Cipolla.
\newblock {PoseNet: A Convolutional Network for Real-Time 6-DOF Camera
  Relocalization}.
\newblock In {\em \href{https://ieeexplore.ieee.org/document/7410693}{ICCV}},
  pages 2938--2946, Santiago de Chile, Chile, December 2015.

\bibitem{F18}
Christian Kerl, Jurgen Sturm, and Daniel Cremers.
\newblock {Dense Visual SLAM for RGB-D Cameras}.
\newblock In {\em \href{https://ieeexplore.ieee.org/document/6696650}{IROS}},
  pages 2100--2106, Tokyo, Japan, November 2013.

\bibitem{klein}
Georg Klein and David Murray.
\newblock {Parallel Tracking and Mapping for Small AR Workspaces}.
\newblock In {\em \href{https://ieeexplore.ieee.org/document/4538852}{ISMAR}},
  pages 225--234, Nara, Japan, November 2007.

\bibitem{distancenet}
Robin Kreuzig, Matthias Ochs, and Rudolf Mester.
\newblock {DistanceNet: Estimating Traveled Distance From Monocular Images
  Using a Recurrent Convolutional Neural Network}.
\newblock In {\em \href{https://ieeexplore.ieee.org/document/9025538}{CVPRW}},
  Long Beach, CA, June 2019.

\bibitem{laskar}
Zakaria Laskar, Iaroslav Melekhov, Surya Kalia, and Juho Kannala.
\newblock {Camera Relocalization by Computing Pairwise Relative Poses Using
  Convolutional Neural Network}.
\newblock In {\em
  \href{https://openaccess.thecvf.com/content_ICCV_2017_workshops/papers/w17/Laskar_Camera_Relocalization_by_ICCV_2017_paper.pdf}{ICCVW}},
  pages 920--929, Venice, Italy, 2017.

\bibitem{marker}
Sooyong Lee and Jae-Bok Song.
\newblock {Mobile Robot Localization Using Infrared Light Reflecting
  Landmarks}.
\newblock In {\em \href{https://ieeexplore.ieee.org/document/4406984}{Intl.
  Conf. Control, Automation and Systems}}, pages 674--677, Seoul, South Korea,
  October 2007.

\bibitem{leutenegger_furgale}
Stefan Leutenegger, Paul Furgale, Vincent Rabaud, Margarita Chli, Kurt
  Konolige, and Roland Siegwart.
\newblock {Keyframe-Based Visual-Inertial SLAM Using Nonlinear Optimization}.
\newblock In {\em
  \href{https://journals.sagepub.com/doi/pdf/10.1177/0278364914554813}{IJRR}},
  volume {34(3)}, pages 314--334, 2015.

\bibitem{li_tell}
Kunpeng Li, Ziyan Wu, Kuan-Chuan Peng, Jan Ernst, and Yun Fu.
\newblock {Tell Me Where to Look: Guided Attention Inference Network}.
\newblock In {\em \href{https://ieeexplore.ieee.org/document/8579058}{CVPR}},
  pages 9215--9223, Salt Lake City, UT, June 2018.

\bibitem{F15}
Peiliang Li, Tong Qin, Botao Hu, Fengyuan Zhu, and Shaojie Shen.
\newblock {Monocular Visual-Inertial State Estimation for Mobile Augmented
  Reality}.
\newblock In {\em \href{https://ieeexplore.ieee.org/document/8115400}{ISMAR}},
  Nantes, France, October 2017.

\bibitem{F8}
Wenbin Li, Sajad Saeedi, John McCormac, Ronald Clark, Dimos Tzoumanikas, Qing
  Ye, Yuzhong Huang, Rui Tang, and Stefan Leutenegger.
\newblock {InteriorNet: Mega-Scale Multi-Sensor Photo-Realistic Indoor Scenes
  Dataset}.
\newblock In {\em \href{http://bmvc2018.org/contents/papers/0249.pdf}{BMVC}},
  volume {18(17)}, 2018.

\bibitem{li_ou_wei}
Xin Li, Xingtao Ou, Zhi Li, Henglu Wei, Wei Zhou, and Zhemin Duan.
\newblock {On-Line Temperature Estimation for Noisy Thermal Sensors Using a
  Smoothing Filter-Based Kalman Predictor}.
\newblock In {\em \href{https://www.mdpi.com/1424-8220/18/2/433}{Sensors}},
  volume {18(2)}, page 433, February 2018.

\bibitem{li}
Yunpeng Li, Noah Snavely, and Daniel~P. Huttenlocher.
\newblock {Location Recognition Using Prioritized Feature Matching}.
\newblock In {\em
  \href{https://link.springer.com/chapter/10.1007/978-3-642-15552-9_57}{ECCV}},
  pages 791--804, 2010.

\bibitem{liebel}
Lukas Liebel and Marco Körner.
\newblock {Auxiliary Tasks in Multi-Task Learning}.
\newblock In {\em \href{https://arxiv.org/abs/1805.06334}{arXiv:1805.06334}},
  May 2018.

\bibitem{lin}
Yimin Lin, Zhaoxiang Liu, Jianfeng Huang, Chaopeng Wang, Guoguang Du, Jinqiang
  Bai, Shiguo Lian, and Bill Huang.
\newblock {Deep Global-Relative Networks for End-to-End 6-DoF Visual
  Localization and Odometry}.
\newblock In {\em
  \href{https://link.springer.com/chapter/10.1007/978-3-030-29911-8_35}{PRICAI}},
  pages 454--467, Cuvu, Fiji, 2019.

\bibitem{F6}
Haomin Liu, Mingyu Chen, Guofeng Zhang, Hujun Bao, and Yingze Bao.
\newblock {ICE-BA: Incremental, Consistent and Efficient Bundle Adjustment for
  Visual-Inertial SLAM}.
\newblock In {\em \href{https://ieeexplore.ieee.org/document/8578309}{CVPR}},
  pages 1974--1982, Salt Lake City, UT, June 2018.

\bibitem{liu}
Shikun Liu, Edward Johns, and Andrew~J. Davison.
\newblock {End-to-End Multi-Task Learning with Attention}.
\newblock In {\em \href{https://ieeexplore.ieee.org/document/8954221}{CVPR}},
  pages 1871--1880, Long Beach, CA, June 2019.

\bibitem{loeffler}
Christoffer L{\"o}ffler, Sascha Riechel, Janina Fischer, and Christopher
  Mutschler.
\newblock {Evaluation Criteria for Inside-Out Indoor Positioning Systems Based
  on Machine Learning}.
\newblock In {\em \href{https://ieeexplore.ieee.org/document/8533862}{IPIN}},
  pages 1--8, Nantes, France, September 2018.

\bibitem{lynen}
Simon Lynen, Torsten Sattler, Michael Bosse, Joel~A. Hesch, Marc Pollefeys, and
  Roland Siegwart.
\newblock {Get Out of My Lab: Large-scale, Real-Time Visual-Inertial
  Localization}.
\newblock In {\em \href{https://doi.org/10.1177/0278364920931151}{Robotics:
  Science ans Systems}}, volume {39(9)}, July 2020.

\bibitem{robotcar}
Will Maddern, Geoffrey Pascoe, Chris Linegar, and Paul Newman.
\newblock {1 Year, 1000km: The Oxford RobotCar Dataset}.
\newblock In {\em \href{https://doi.org/10.1177/0278364916679498}{IJRR}},
  volume {36(1)}, pages 3--15, November 2016.

\bibitem{magnusson}
Niklas Magnusson and Tobias Odenman.
\newblock {Improving Absolute Position Estimates of an Automotive Vehicle using
  GPS in Sensor Fusion}.
\newblock In {\em
  \href{https://www.semanticscholar.org/paper/Improving-absolute-position-estimates-of-an-vehicle-Magnusson-Odenman/0f52e6941968a0aa5c7f57e0991a307b8b94b029}{Chalmers
  University of Technology, Thesis}}, G{\"{o}}teborg, Sweden, 2012.

\bibitem{mansur}
Syaiful Mansur, Muhammad Habib, Gilang Nugraha~Putu Pratama, Adha~Imam Cahyadi,
  and Igi Ardiyanto.
\newblock {Real Time Monocular Visual Odometry using Optical Flow: Study on
  Navigation of Quadrotora's UAV}.
\newblock In {\em \href{https://ieeexplore.ieee.org/document/8011864}{Intl.
  Conf. on Science and Technology - Computer (ICST)}}, pages 122--126,
  Yogyakarta, Indonesia, July 2017.

\bibitem{F14}
Eric Marchand, Hideaki Uchiyama, and Fabien Spindler.
\newblock {Pose Estimation for Augmented Reality: A Hands-On Survey}.
\newblock In {\em
  \href{https://ieeexplore.ieee.org/abstract/document/7368948}{TVCG}}, volume
  {22(12)}, pages 2633--2651, December 2016.

\bibitem{melekhov}
Iaroslav Melekhov, Juha Ylioinas, Juho Kannala, and Esa Rahtu.
\newblock {Image-Based Localization Using Hourglass Networks}.
\newblock In {\em \href{https://ieeexplore.ieee.org/document/8265316}{ICCVW}},
  pages 870--877, Venice, Italy, October 2017.

\bibitem{mohamed}
Sherif A.~S. Mohamed, Mohammad-Hashem Haghbayan, Tomi Westerlund, Jukka
  Heikkonen, Hannu Tenhunen, and Juha Plosila.
\newblock {A Survey on Odometry for Autonomous Navigation Systems}.
\newblock In {\em \href{https://ieeexplore.ieee.org/document/8764393}{IEEE
  Access}}, volume~7, pages 97466--97486, July 2019.

\bibitem{moreau}
Arthur Moreau, Nathan Piasco, Dzmitry Tsishkou, Bogdan Stanciulescu, and Arnaud
  de~La~Fortelle.
\newblock {CoordiNet: Uncertainty-aware Pose Regressor for Reliable Vehicle
  Localization}.
\newblock In {\em
  \href{https://openaccess.thecvf.com/content/WACV2022/papers/Moreau_CoordiNet_Uncertainty-Aware_Pose_Regressor_for_Reliable_Vehicle_Localization_WACV_2022_paper.pdf}{WACV}},
  January 2022.

\bibitem{mourikis}
Anastasios~I. Mourikis and Stergios~I. Roumeliotis.
\newblock {A Multi-State Constraint Kalman Filter for Vision-aided Inertial
  Navigation}.
\newblock In {\em \href{https://ieeexplore.ieee.org/document/4209642}{ICRA}},
  pages 3565--3572, Roma, Italy, April 2007.

\bibitem{mu_chen}
Xufu Mu, Jing Chen, Zixiang Zhou, Zhen Leng, and Lei Fan.
\newblock {Accurate Initial State Estimation in a Monocular Visual-Inertial
  SLAM System}.
\newblock In {\em \href{https://www.mdpi.com/1424-8220/18/2/506}{MDPI
  Sensors}}, volume {18(2)}, page 506, February 2018.

\bibitem{muller_savakis}
Peter Muller and Andreas Savakis.
\newblock {Flowdometry: An Optical Flow and Deep Learning Based Approach to
  Visual Odometry}.
\newblock In {\em
  \href{https://www.computer.org/csdl/proceedings-article/wacv/2017/07926658/12OmNxWuilO}{WACV}},
  volume~1, pages 624--631, Santa Rosa, CA, 2017.

\bibitem{muller}
Peter~M. Muller, Andreas Savakis, Raymond Ptucha, and Roy Melton.
\newblock {Optical Flow and Deep Learning Based Approach to Visual Odometry}.
\newblock In {\em
  \href{https://scholarworks.rit.edu/cgi/viewcontent.cgi?article=10455&context=theses}{Rochester
  Institute of Technology, Department of Computer Engineering}}, Rochester, NY,
  November 2016.

\bibitem{mur}
Ra\'{u}l Mur-Artal, J.~M.~M. Montiel, and Juan~D. Tard\'{o}s.
\newblock {ORB-SLAM: A Versatile and Accurate Monocular SLAM System}.
\newblock In {\em \href{https://ieeexplore.ieee.org/document/7219438}{T-RO}},
  volume {31(5)}, pages 1147--1163, 2015.

\bibitem{mur2}
Raul Mur-Artal and Juan~D. Tard{\'o}s.
\newblock {ORB-SLAM2: An Open-Source SLAM System for Monocular, Stereo, and
  RGB-D Cameras}.
\newblock In {\em \href{https://ieeexplore.ieee.org/document/7946260}{T-RO}},
  volume {33(5)}, pages 1255--1262, June 2017.

\bibitem{mur_vislam}
Ra\'{u}l Mur-Artal and Juan~D. Tard\'{o}s.
\newblock {Visual-Inertial Monocular SLAM With Map Reuse}.
\newblock In {\em \href{https://ieeexplore.ieee.org/document/7817784}{RA-L}},
  volume {2(2)}, pages 796--803, January 2017.

\bibitem{naseer}
Tayyab Naseer and Wolfram Burgard.
\newblock {Deep Regression for Monocular Camera-based DoF Global Localization
  in Outdoor Environments}.
\newblock In {\em
  \href{https://ieeexplore.ieee.org/abstract/document/8205957}{IROS}}, pages
  1525--1530, Vancouver, BC, September 2017.

\bibitem{natarajan}
Pradeep Natarajan, Shuang Wu, Shiv Vitaladevuni, Xiaodan Zhuang, Stavros
  Tsakalidis, Unsang Park, Rohit Prasad, and Premkumar Natarajan.
\newblock {Multimodal Feature Fusion for Robust Event Detection in Web Videos}.
\newblock In {\em \href{https://ieeexplore.ieee.org/document/6247814}{CVPR}},
  pages 1298--1305, Providence, RI, June 2012.

\bibitem{navon_aux}
Aviv Navon, Idan Achituve, Haggai Maron, Gal Chechik, and Ethan Fetaya.
\newblock {Auxiliary Learning by Implicit Differentiation}.
\newblock In {\em \href{https://openreview.net/forum?id=n7wIfYPdVet}{ICLR}},
  2021.

\bibitem{bergen_visual_2004}
David Nist{\'{e}}r, Oleg Naroditsky, and James Bergen.
\newblock {Visual Odometry}.
\newblock In {\em \href{https://ieeexplore.ieee.org/document/1315094}{CVPR}},
  Washington, DC, June 2004.

\bibitem{nuetzi}
Gabriel N{\"u}tzi, Stephan Weiss, Davide Scaramuzza, and Roland Siegwart.
\newblock {Fusion of IMU and Vision for Absolute Scale Estimation in Monocular
  SLAM}.
\newblock In {\em
  \href{https://link.springer.com/article/10.1007/s10846-010-9490-z}{Journal of
  Intelligent and Robotic Systems}}, volume~{61}, pages 287--–299, 2011.

\bibitem{ott}
Felix Ott, Tobias Feigl, Christoffer Löffler, and Christopher Mutschler.
\newblock {ViPR: Visual-Odometry-aided Pose Regression for 6DoF Camera
  Localization}.
\newblock In {\em \href{https://ieeexplore.ieee.org/document/9150733}{CVPRW}},
  pages 187--198, Seattle, WA, June 2020.

\bibitem{owens}
Andrew Owens and Alexei~A. Efros.
\newblock {Audio-Visual Scene Analysis with Self-Supervised Multisensory
  Features}.
\newblock In {\em
  \href{https://openaccess.thecvf.com/content_ECCV_2018/papers/Andrew_Owens_Audio-Visual_Scene_Analysis_ECCV_2018_paper.pdf}{ECCV}},
  pages 631--648, 2018.

\bibitem{contextualnet}
Mitesh Patel, Brendan Emery, and Yan-Ying Chen.
\newblock {ContextualNet: Exploiting Contextual Information using LSTMs to
  Improve Image-based Localization}.
\newblock In {\em \href{https://ieeexplore.ieee.org/document/8461124}{ICRA}},
  Brisbane, Australia, May 2018.

\bibitem{pfrommer}
Bernd Pfrommer, Nitin Sanket, Kostas Daniilidis, and Jonas Cleveland.
\newblock {PennCOSYVIO: A Challenging Visual Inertial Odometry Benchmark}.
\newblock In {\em \href{https://ieeexplore.ieee.org/document/7989443}{ICRA}},
  pages 3847--3854, Singapore, Singapore, May 2017.

\bibitem{piasco}
Nathan Piasco, D\'{e}sir\'{e} Sidib\'{e}, Val\'{e}rie Gouet-Brunet, and
  C\'{e}dric Demonceaux.
\newblock {Improving Image Description with Auxiliary Modality for Visual
  Localization in Challenging Conditions}.
\newblock In {\em \href{https://doi.org/10.1007/s11263-020-01363-6}{IJCV}},
  volume {129(1)}, pages 185--202, August 2020.

\bibitem{qin_li}
Tong Qin, Peiliang Li, and Shaojie Shen.
\newblock {Relocalization, Global Optimization and Map Merging for Monocular
  Visual-Inertial SLAM}.
\newblock In {\em \href{https://ieeexplore.ieee.org/document/8460780}{ICRA}},
  Brisbane, Australia, May 2018.

\bibitem{qin}
Tong Qin, Peiliang Li, and Shaojie Shen.
\newblock {VINS-Mono: A Robust and Versatile Monocular Visual-Inertial State
  Estimator}.
\newblock In {\em \href{https://ieeexplore.ieee.org/document/8421746}{T-RO}},
  volume {34(4)}, pages 1004--1020, July 2018.

\bibitem{qin_pan}
Tong Qin, Jie Pan, Shaozu Cao, and Shaojie Shen.
\newblock {A General Optimization-based Framework for Local Odometry Estimation
  with Multiple Sensors}.
\newblock In {\em \href{https://arxiv.org/abs/1901.03638}{arXiv:1901.03638}},
  January 2019.

\bibitem{radwan}
Noha Radwan, Abhinav Valada, and Wolfram Burgard.
\newblock {VLocNet++: Deep Multitask Learning for Semantic Visual Localization
  and Odometry}.
\newblock In {\em
  \href{http://ais.informatik.uni-freiburg.de/publications/papers/radwan18ral.pdf}{RA-L}},
  volume {3(4)}, pages 4407--4414, 2018.

\bibitem{nisha_master}
Nisha~Lakshmana Raichur.
\newblock {Image-/IMU-based Deep Multimodal Information Fusion for Pedestrian
  Self-Localization}.
\newblock In {\em Master's Thesis, University of Ulm}, April 2021.

\bibitem{russel}
Rebecca~L. Russell and Christopher Reale.
\newblock {Multivariate Uncertainty in Deep Learning}.
\newblock In {\em \href{https://ieeexplore.ieee.org/document/9457247}{Trans. on
  Neural Networks and Learning Systems}}, pages 1--7, June 2021.

\bibitem{sattler_activesearch}
Torsten Sattler, Bastian Leibe, and Leif Kobbelt.
\newblock {Efficient \& Effective Prioritized Matching for Large-Scale
  Image-Based Localization}.
\newblock In {\em \href{https://ieeexplore.ieee.org/document/7572201}{TPAMI}},
  volume {39(9)}, pages 1744--1756, September 2017.

\bibitem{sattler_cmu}
Torsten Sattler, Will Maddern, Carl Toft, Akihiko Torii, Lars Hammarstrand,
  Erik Stenborg, Daniel Safari, Masatoshi Okutomi, Marc Pollefeys, Josef Sivic,
  Fredrik Kahl, and Tomas Pajdla.
\newblock {Benchmarking 6DOF Outdoor Visual Localization in Changing
  Conditions}.
\newblock In {\em \href{https://ieeexplore.ieee.org/document/8578995}{CVPR}},
  Salt Lake City, UT, June 2018.

\bibitem{sattler_apr}
Torsten Sattler, Qunjie Zhou, Marc Pollefeys, and Laura Leal-Taix\'{e}.
\newblock {Understanding the Limitations of CNN-based Absolute Camera Pose
  Regression}.
\newblock In {\em
  \href{https://ieeexplore.ieee.org/abstract/document/8954331}{CVPR}}, pages
  3302--3312, Long Beach, CA, June 2019.

\bibitem{schubert}
David Schubert, Thore Goll, Nikolaus Demmel, Vladyslav Usenko, Jörg Stückler,
  and Daniel Cremers.
\newblock {The TUM VI Benchmark for Evaluating Visual-Inertial Odometry}.
\newblock In {\em \href{https://ieeexplore.ieee.org/document/8593419}{IROS}},
  Madrid, Spain, October 2018.

\bibitem{seifi}
Soroush Seifi and Tinne Tuytelaars.
\newblock {How to improve CNN-based 6-DoF Camera Pose Estimation}.
\newblock In {\em \href{https://ieeexplore.ieee.org/document/9022045}{ICCVW}},
  Seoul, Korea, October 2019.

\bibitem{shotton_scene}
Jamie Shotton, Ben Glocker, Christopher Zach, Shahram Izadi, Antonio Criminisi,
  and Andrew Fitzgibbon.
\newblock {Scene Coordinate Regression Forests for Camera Relocalization in
  RGB-D Images}.
\newblock In {\em \href{https://ieeexplore.ieee.org/document/6619221}{CVPR}},
  pages 2930--2937, Portland, OR, June 2013.

\bibitem{shu}
Yuanchao Shu, Kang~G. Shin, Tian He, and Jiming Chen.
\newblock {Last-Mile Navigation Using Smartphones}.
\newblock In {\em \href{https://dl.acm.org/doi/10.1145/2789168.2790099}{Intl.
  Conf. on Mobile Computing and Networking (MCN)}}, pages 512--524, September
  2015.

\bibitem{sibley}
Gabe Sibley, Larry Matthies, and Gaurav Sukhatme.
\newblock {Sliding Window Filter with Application to Planetary Landing}.
\newblock In {\em \href{https://doi.org/10.1002/rob.20360}{Journal of Field
  Robotics (JFR)}}, pages 587--608, August 2010.

\bibitem{smith}
Randall~C. Smith and Peter Cheeseman.
\newblock {On the Representation and Estimation of Spatial Uncertainty}.
\newblock In {\em \href{https://doi.org/10.1177/027836498600500404}{IJRR}},
  volume {5(4)}, pages 56--68, 1986.

\bibitem{standley}
Trevor Standley, Amir Zamir, Dawn Chen, Leonidas Guibas, Jitendra Malik, and
  Silvio Savarese.
\newblock {Which Tasks Should be Learned Together in Multi-Task Learning?}
\newblock In {\em
  \href{http://proceedings.mlr.press/v119/standley20a.html}{ICML}}, volume 119,
  pages 9120--9132, 2020.

\bibitem{stapleton}
Christopher Stapleton, Charles Hughes, Michael Moshell, Paulius Micikevicius,
  and Marty Altman.
\newblock {Applying Mixed Reality to Entertainment}.
\newblock In {\em \href{https://ieeexplore.ieee.org/document/1106186}{IEEE}},
  volume {35(12)}, pages 122--124, December 2002.

\bibitem{sturm}
Jürgen Sturm, Nikolas Engelhard, Felix Endres, Wolfram Burgard, and Daniel
  Cremers.
\newblock {A Benchmark for the Evaluation of RGB-D SLAM Systems}.
\newblock In {\em
  \href{http://ais.informatik.uni-freiburg.de/publications/papers/sturm12iros.pdf}{IROS}},
  October 2012.

\bibitem{szegedy}
Christian Szegedy, Wei Liu, Yangqing Jia, Pierre Sermanet, Scott Reed, Dragomir
  Anguelov, Dumitru Erhan, Vincent Vanhoucke, and Andrew Rabinovich.
\newblock {Going Deeper with Convolutions}.
\newblock In {\em \href{https://ieeexplore.ieee.org/document/7298594}{CVPR}},
  pages 1--9, Boston, MA, 2015.

\bibitem{F144}
Takahiro Terashima and Osamu Hasegawa.
\newblock {A Visual-SLAM for First Person Vision and Mobile Robots}.
\newblock In {\em \href{https://ieeexplore.ieee.org/document/7986779}{MVA}},
  Nagoya, Japan, May 2017.

\bibitem{valada}
Abhinav Valada, Noha Radwan, and Wolfram Burgard.
\newblock {Deep Auxiliary Learning for Visual Localization and Odometry}.
\newblock In {\em \href{https://ieeexplore.ieee.org/document/8462979}{ICRA}},
  pages 6939--6946, Brisbane, Australia, may 2018.

\bibitem{12scenes}
Julien Valentin, Angela Dai, Matthias Niessner, Pushmeet Kohli, Philip Torr,
  Shahram Izadi, and Cem Keskin.
\newblock {Learning to Navigate the Energy Landscape}.
\newblock In {\em
  \href{https://ieeexplore.ieee.org/abstract/document/7785106}{3DV}}, Stanford,
  CA, October 2016.

\bibitem{vaswani}
Ashish Vaswani, Noam Shazeer, Niki Parmar, Jakob Uszkoreit, Llion Jones,
  Aidan~N. Gomez, Lukasz Kaiser, and Illia Polosukhin.
\newblock {Attention is All You Need}.
\newblock In {\em
  \href{https://papers.nips.cc/paper/2017/hash/3f5ee243547dee91fbd053c1c4a845aa-Abstract.html}{NIPS}},
  2017.

\bibitem{resnet}
Francesco Visin, Kyle Kastner, Kyunghyun Cho, Matteo Matteucci, Aaron
  Courville, and Yoshua Bengio.
\newblock {ResNet: A Recurrent Neural Network Based Alternative to
  Convolutional Networks}.
\newblock In {\em \href{https://arxiv.org/abs/1505.00393}{arXiv:1505.00393}},
  May 2015.

\bibitem{stumberg}
Lukas von Stumberg, Vladyslav Usenko, and Daniel Cremers.
\newblock {Direct Sparse Visual-Inertial Odometry using Dynamic
  Marginalization}.
\newblock In {\em \href{https://ieeexplore.ieee.org/document/8462905}{ICRA}},
  Brisbane, Australia, May 2018.

\bibitem{stumberg_wenzel}
Lukas von Stumberg, Patrick Wenzel, Nan Yang, and Daniel Cremers.
\newblock {LM-Reloc: Levenberg-Marquardt Based Direct Visual Relocalization}.
\newblock In {\em \href{https://ieeexplore.ieee.org/document/9320444}{3DV}},
  Fukuoka, Japan, November 2020.

\bibitem{walch}
Florian Walch, Caner Hazirbas, Laura Leal-Taix\'{e}, Torsten Sattler, Sebastian
  Hilsenbeck, and Daniel Cremers.
\newblock {Image-Based Localization Using LSTMs for Structured Feature
  Correlation}.
\newblock In {\em \href{https://ieeexplore.ieee.org/document/8237337}{ICCV}},
  pages 627--637, Venice, Italy, October 2017.

\bibitem{wang_chen}
Bing Wang, Changhao Chen, Chris~Xiaoxuan Lu, Peijun Zhao, Niki Trigoni, and
  Andrew Markham.
\newblock {AtLoc: Attention Guided Camera Localization}.
\newblock In {\em
  \href{https://ojs.aaai.org//index.php/AAAI/article/view/6608}{AAAI}}, volume
  {34(06)}, pages 10393--10401, April 2020.

\bibitem{deepvo_vo}
Sen Wang, Ronald Clark, Hongkai Wen, and Niki Trigoni.
\newblock {DeepVO: Towards end-to-end Visual Odometry with deep Recurrent
  Convolutional Neural Networks}.
\newblock In {\em \href{https://ieeexplore.ieee.org/document/7989236}{ICRA}},
  pages 2043--2050, Singapore, Singapore, June 2017.

\bibitem{williams}
Brian Williams, Georg Klein, and Ian Reid.
\newblock {Real-time SLAM Relocalisation}.
\newblock In {\em \href{https://ieeexplore.ieee.org/document/4409115}{ICCV}},
  pages 1--8, Rio de Janeiro, Brazil, October 2007.

\bibitem{Liwei}
Jian Wu, Liwei Ma, and Xiaolin Hu.
\newblock {Delving Deeper Into Convolutional Neural Networks for Camera
  Relocalization}.
\newblock In {\em
  \href{https://ieeexplore.ieee.org/abstract/document/7989663}{ICRA}}, pages
  5644--5651, Singapore, Singapore, May 2017.

\bibitem{xu_davison}
Binbin Xu, Andrew~J. Davison, and Stefan Leutenegger.
\newblock {Deep Probabilistic Feature-Metric Tracking}.
\newblock In {\em \href{https://ieeexplore.ieee.org/document/9264717}{RA-L}},
  volume {6(1)}, pages 223--230, January 2021.

\bibitem{xu_show}
Kelvin Xu, Jimmy Ba, Ryan Kiros, Kyunghyun Cho, Aaron Courville, Ruslan
  Salakhudinov, Rich Zemel, and Yoshua Bengio.
\newblock {Show, Attend and Tell: Neural Image Caption Generation with Visual
  Attention}.
\newblock In {\em \href{https://proceedings.mlr.press/v37/xuc15.html}{ICML}},
  volume~37, pages 2048--2057, 2015.

\bibitem{yan}
Xuejiao Yan, Zongying Shi, and Yisheng Zhong.
\newblock {Vision-based Global Localization of Unmanned Aerial Vehicles with
  Street View Images}.
\newblock In {\em \href{https://ieeexplore.ieee.org/document/8483081}{CCC}},
  pages 4672--4678, Wuhan, China, July 2018.

\bibitem{yang_stumberg}
Nan Yang, Lukas von Stumberg, Rui Wang, and Daniel Cremers.
\newblock {D3VO: Deep Depth, Deep Pose and Deep Uncertainty for Monocular
  Visual Odometry}.
\newblock In {\em \href{https://ieeexplore.ieee.org/document/9157454}{CVPR}},
  pages 1281--1292, Seattle, WA, June 2020.

\bibitem{yang_wang}
Nan Yang, Rui Wang, Xiang Gao, and Daniel Cremers.
\newblock {Challenges in Monocular Visual Odometry: Photometric Calibration,
  Motion Bias, and Rolling Shutter Effect}.
\newblock In {\em \href{https://ieeexplore.ieee.org/document/8384000}{RA-L}},
  volume {3(4)}, pages 2878--2885, June 2018.

\bibitem{yol}
Aur\'{e}lien Yol, Bertrand Delabarre, Amaury Dame, Jean \'{E}mile Dartois, and
  Eric Marchand.
\newblock {Vision-based Absolute Localization for Unmanned Aerial Vehicles}.
\newblock In {\em \href{https://ieeexplore.ieee.org/document/6943040}{ICIRS}},
  pages 3429--3434, Chicago, IL, September 2014.

\bibitem{zeisl}
Bernhard Zeisl, Torsten Sattler, and Marc Pollefeys.
\newblock {Camera Pose Voting for Large-Scale Image-Based Localization}.
\newblock In {\em \href{https://ieeexplore.ieee.org/document/7410667}{ICCV}},
  pages 2704--2712, Santiage de Chile, Chile, December 2015.

\bibitem{zhan_weerasekera}
Huangying Zhan, Chamara~Saroj Weerasekera, Jia-Wang Bian, and Ian Reid.
\newblock {Visual Odometry Revisited: What Should be Learnt?}
\newblock In {\em \href{https://ieeexplore.ieee.org/document/9197374}{ICRA}},
  pages 4203--4210, Paris, France, September 2020.

\bibitem{zhao}
Qi~Zhao, Fangmin Li, and Xinhua Liu.
\newblock {Real-time Visual Odometry based on Optical Flow and Depth Learning}.
\newblock In {\em
  \href{https://ieeexplore.ieee.org/abstract/document/8337375}{ICMTMA}}, pages
  239--242, Changsha, China, February 2018.

\bibitem{zhou_luo}
Lei Zhou, Zixin Luo, Tianwei Shen, Jiahui Zhang, Mingmin Zhen, Yao Yao, Tian
  Fang, and Long Quan.
\newblock {KFNet: Learning Temporal Camera Relocalization using Kalman
  Filtering}.
\newblock In {\em \href{https://ieeexplore.ieee.org/document/9157119}{CVPR}},
  pages 4919--4928, Seattle, WA, June 2020.

\bibitem{zhou_barnes}
Yi~Zhou, Connelly Barnes, Jingwan Lu, Jimei Yang, and Hao Li.
\newblock {On the Continuity of Rotation Representations in Neural Networks}.
\newblock In {\em \href{https://ieeexplore.ieee.org/document/8953486}{CVPR}},
  pages 5738--5746, Long Beach, CA, June 2019.

\bibitem{zuniga_noel}
David Zu{\~{n}}iga-No{\"{e}}l, Alberto Jaenal, Ruben Gomez-Ojeda, and
  Gonzalez-Jimenez.
\newblock {The UMA-VI Dataset: Visual-Inertial Odometry in Low-Textured and
  Dynamic Illumination Environments}.
\newblock In {\em \href{https://doi.org/10.1177/0278364920938439}{IJRR}},
  volume {39(9)}, July 2020.

\end{thebibliography}

\section*{Biography}
\begin{IEEEbiography}[{\includegraphics[width=1in,height=1.25in,clip,keepaspectratio]{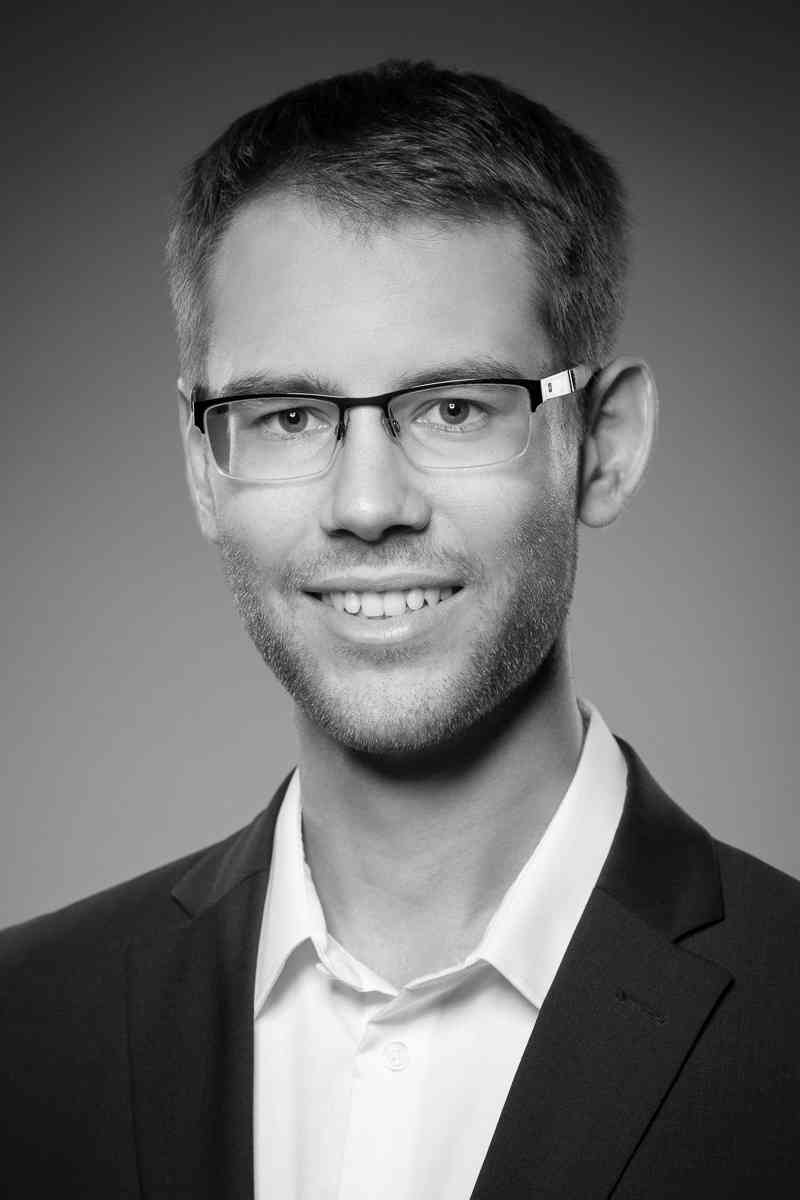}}] {Felix Ott} received his M.Sc. degree in Computational Engineering at the Friedrich-Alexander-Universität (FAU) Erlangen-N{\"u}rnberg, Germany, in 2019. He joined the Hybrid Positioning \& Information Fusion group in the Locating and Communication Systems department at Fraunhofer IIS. In 2020, he started his Ph.D. at the Ludwig-Maximilians-Universität (LMU) in Munich in the Probabilistic Machine and Deep Learning group. His research covers multimodal information fusion for self-localization.\end{IEEEbiography}
\vspace{-1.5cm}
\begin{IEEEbiography}[{\includegraphics[width=1in,height=1.25in,clip,keepaspectratio]{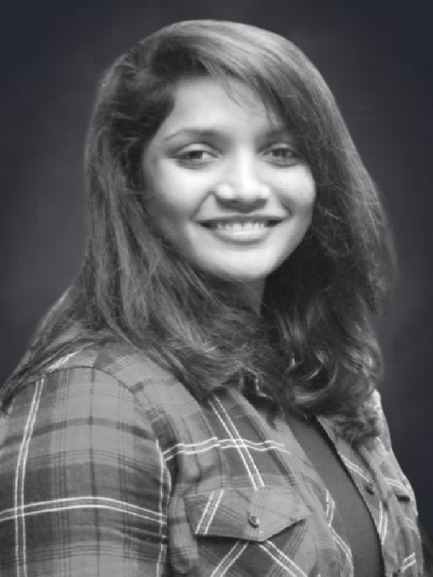}}] {Nisha Lakshmana Raichur} received her M.Sc. degree in Cognitive Systems at the Technical University of Ulm in 2021. She joined the Hybrid Positioning \& Information Fusion group in the Localization and Communication Systems department at Fraunhofer IIS. Her research covers the areas of machine learning and deep multi-modal sensor fusion. \end{IEEEbiography}
\vspace{-1.5cm}
\begin{IEEEbiography}[{\includegraphics[width=1in,height=1.25in,clip,keepaspectratio]{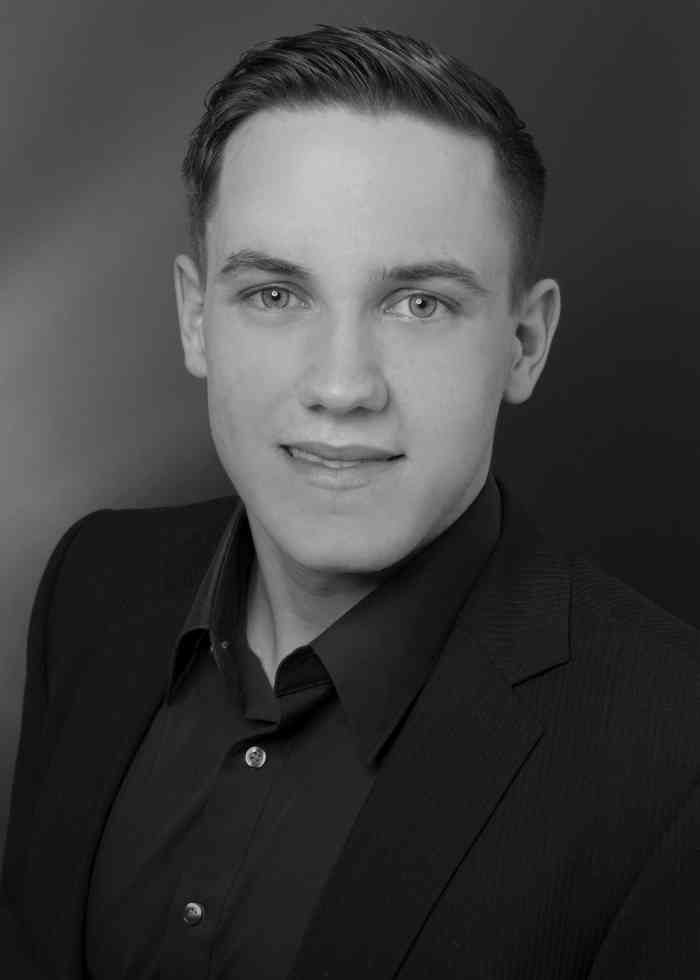}}] {David Rügamer} is an interim professor for Computational Statistics at the TU Dortmund. Before he was an interim professor at the RWTH Aachen and interim professor for Data Science at the LMU Munich, where he also received his Ph.D. in 2018. His research is concerned with scalability of statistical modeling as well as machine learning for functional and multimodal data. \end{IEEEbiography}
\vspace{-1.5cm}
\begin{IEEEbiography}[{\includegraphics[width=1in,height=1.25in,clip,keepaspectratio]{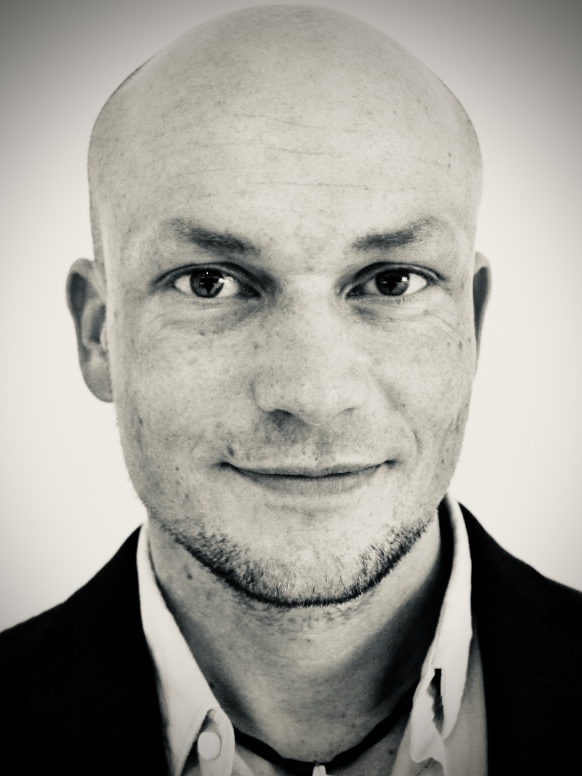}}]{Tobias Feigl} is a research associate at the Fraunhofer Institute for Integrated Circuits (IIS) in N{\"u}rnberg, and received his Ph.D. at the FAU Erlangen-N{\"u}rnberg. His research covers the areas of augmented and virtual reality, human-computer interaction, and machine learning. He improves human motion behavior in immersive virtual environments with inertial and radio sensors. \end{IEEEbiography}
\vspace{-1.5cm}
\begin{IEEEbiography}[{\includegraphics[width=1in,height=1.25in,clip,keepaspectratio]{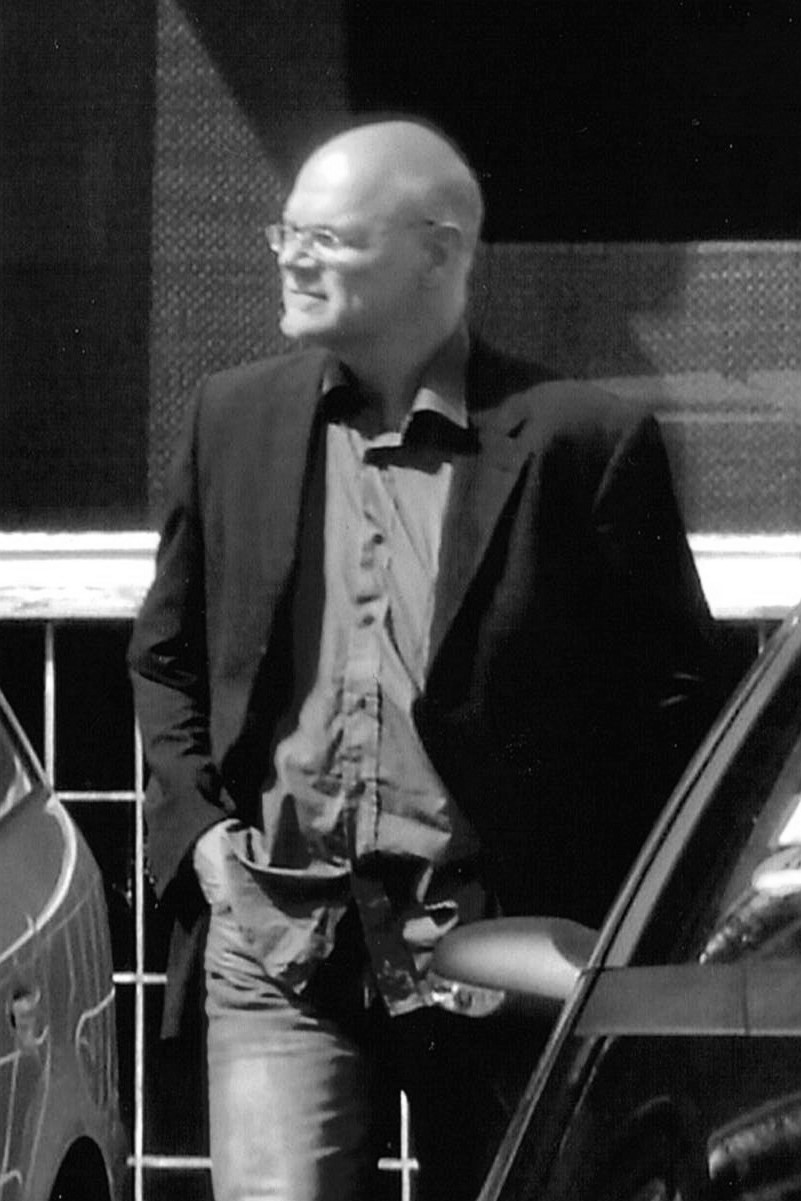}}] {Heiko Neumann} is a full professor at the unviersity of Ulm. He is heading the vision and perception science group focusing on the investigation of mechanisms and the underlying structure of visual information processing in biological and technical systems as well as their adaptation to changing environments. The topics of his investigation primarily focus on mechanisms of visual information processing, namely biological and machine vision. \end{IEEEbiography}
\vspace{-1.5cm}
\begin{IEEEbiography}[{\includegraphics[width=1in,height=1.25in,clip,keepaspectratio]{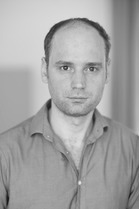}}] {Bernd Bischl} is a full professor for statistical learning and data science at the LMU Munich and a director of the Munich Center of Machine Learning. His research focuses amongst other things on AutoML, interpretable machine learning and machine learning benchmarking. \end{IEEEbiography}
\vspace{-1.5cm}
\begin{IEEEbiography}[{\includegraphics[width=1in,height=1.25in,clip,keepaspectratio]{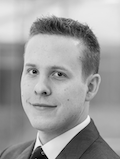}}]{Christopher Mutschler} leads the precise positioning and analytics department at Fraunhofer IIS. Prior to that, Christopher headed the Machine Learning \& Information Fusion group. He gives lectures on machine learning at the FAU Erlangen-N{\"u}rnberg, from which he also received both his Diploma and Ph.D. in 2010 and 2014 respectively. Christopher’s research combines machine learning with radio-based localization.\end{IEEEbiography}

\clearpage

\newcommand\len{0.329}
\newcommand\vspacefigure{0.4cm}
\newcommand\vspacecaption{-0.6cm}
\begin{figure*}[t!]
	\centering
	\begin{minipage}[b]{\len\linewidth}
        \centering
    	\includegraphics[width=1.0\linewidth]{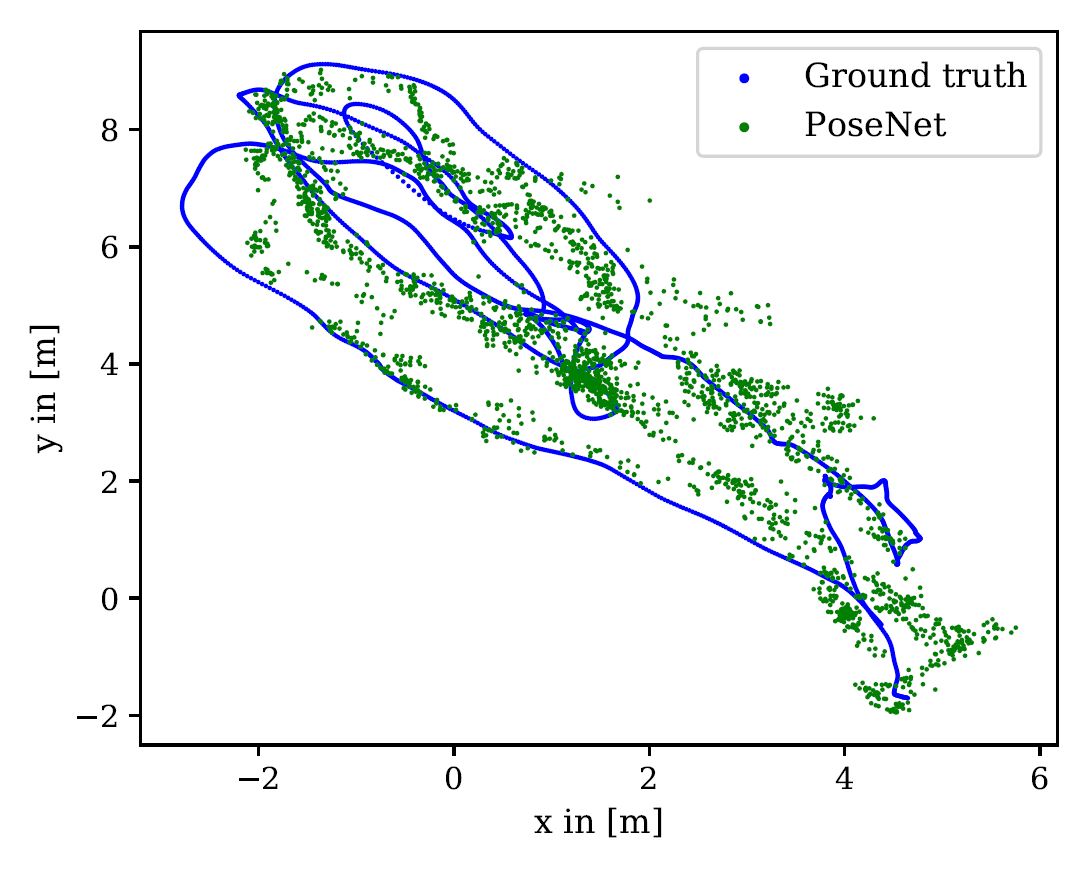}
    	\vspace{\vspacecaption}
    	\subcaption{$\text{APR}_{\text{V}}$: PoseNet~\cite{kendall}.}
    	\label{image_app_euroc_mh02_1}
        \vspace{\vspacefigure}
    \end{minipage}
    \hfill
    \begin{minipage}[b]{\len\linewidth}
        \centering
    	\includegraphics[width=1.0\linewidth]{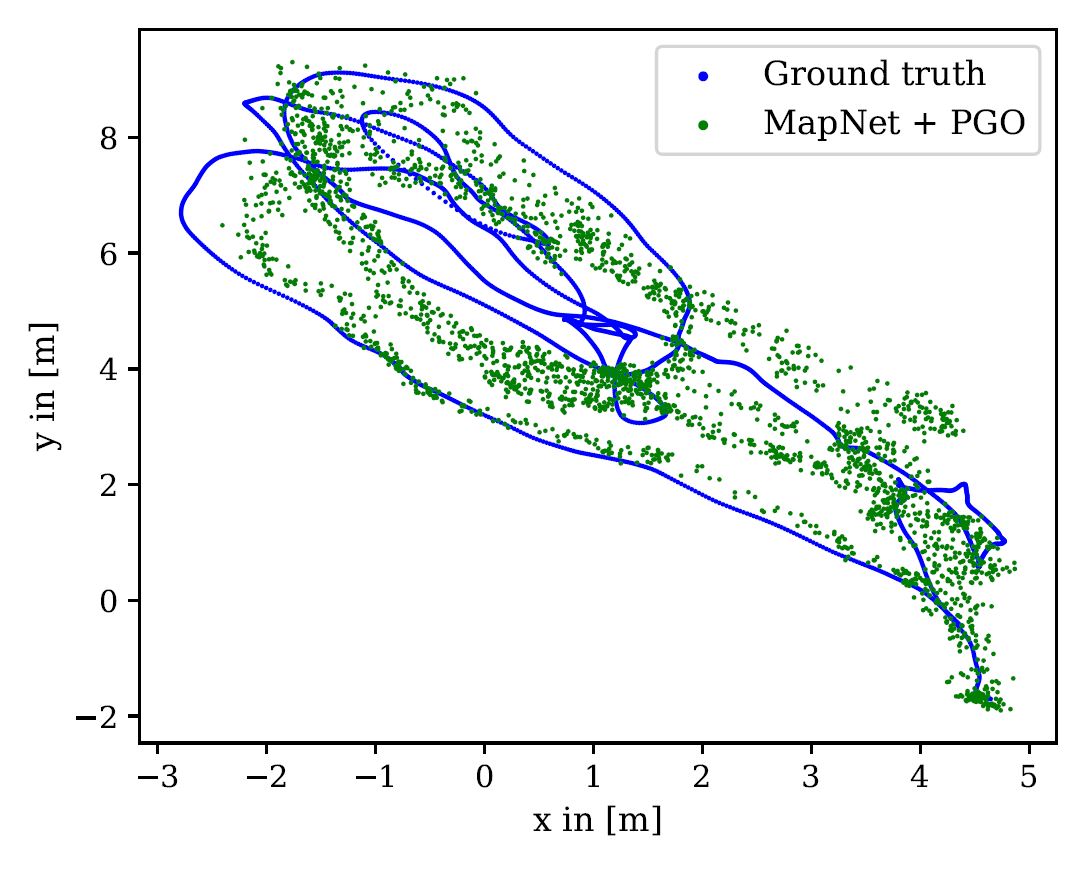}
    	\vspace{\vspacecaption}
    	\subcaption{MapNet+PGO~\cite{brahmbhatt}.}
    	\label{image_app_euroc_mh02_2}
        \vspace{\vspacefigure}
    \end{minipage}
    \hfill
    \begin{minipage}[b]{\len\linewidth}
        \centering
    	\includegraphics[width=1.0\linewidth]{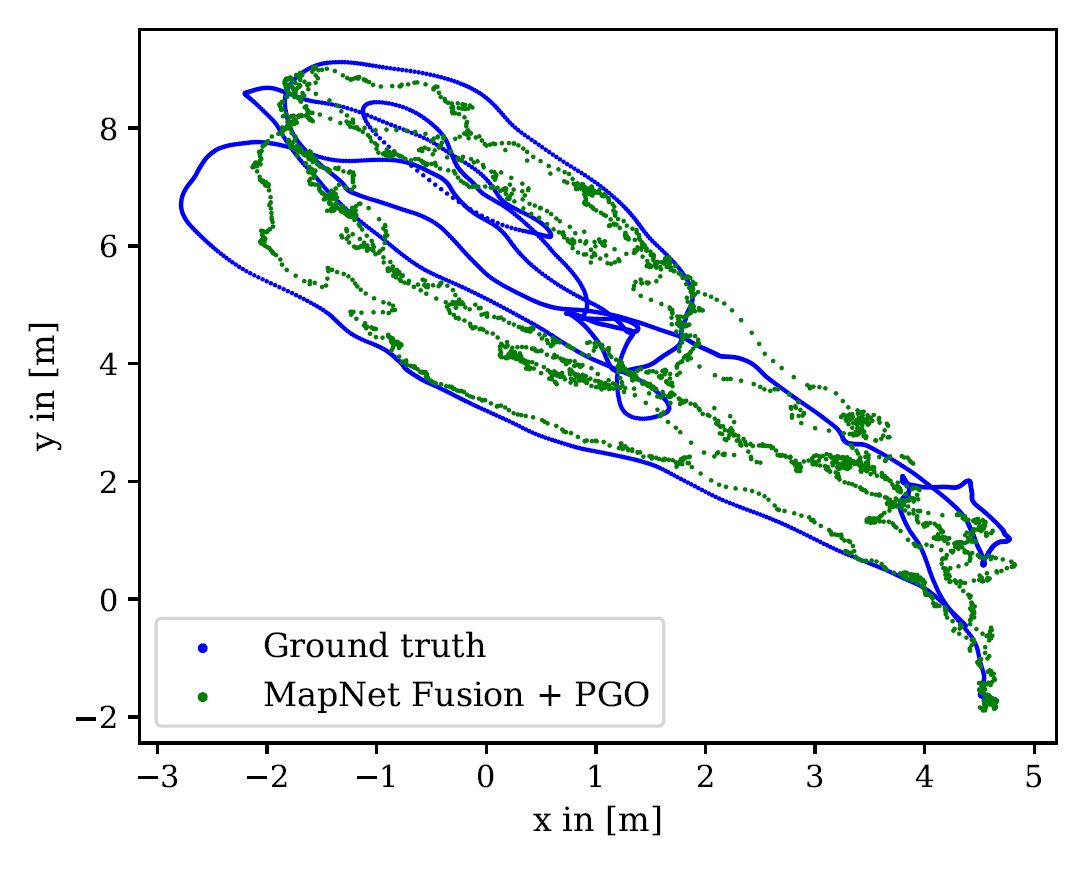}
    	\vspace{\vspacecaption}
    	\subcaption{$\text{APR}_{\text{V}}$-$\text{RPR}_{\text{I}}$+PGO.}
    	\label{image_app_euroc_mh02_3}
        \vspace{\vspacefigure}
    \end{minipage}
    \begin{minipage}[b]{\len\linewidth}
        \centering
    	\includegraphics[width=1.0\linewidth]{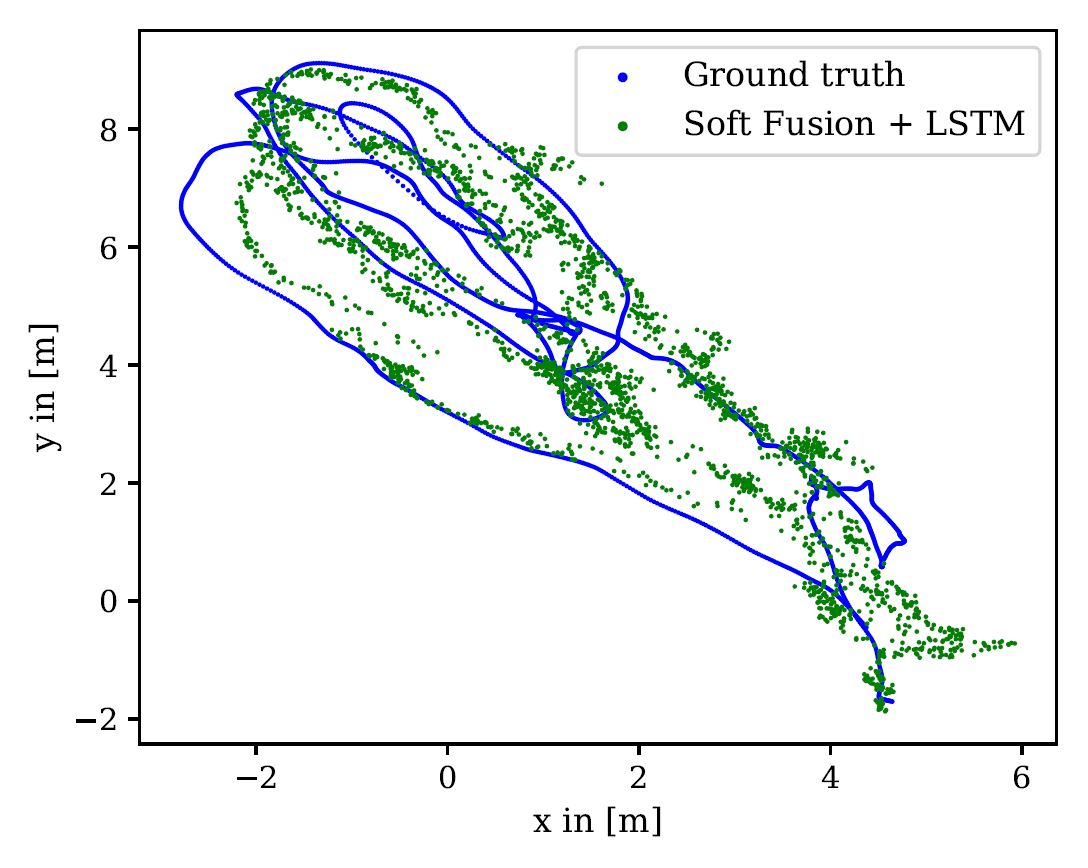}
    	\vspace{\vspacecaption}
    	\subcaption{SSF~\cite{chen} + BiLSTM.}
    	\label{image_app_euroc_mh02_4}
        \vspace{\vspacefigure}
    \end{minipage}
    \hfill
    \begin{minipage}[b]{\len\linewidth}
        \centering
    	\includegraphics[width=1.0\linewidth]{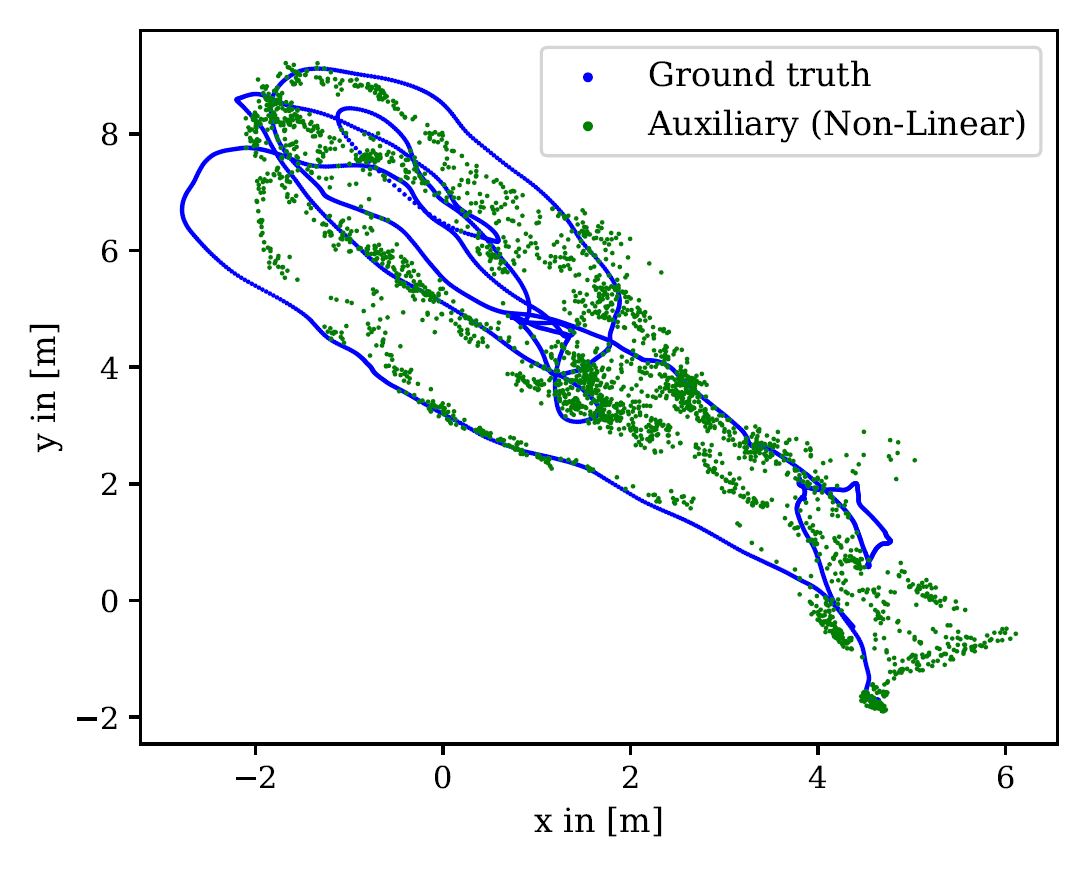}
    	\vspace{\vspacecaption}
    	\subcaption{Auxiliary learning (non-linear)~\cite{navon_aux}.}
    	\label{image_app_euroc_mh02_5}
        \vspace{\vspacefigure}
    \end{minipage}
    \hfill
	\begin{minipage}[b]{\len\linewidth}
        \centering
    	\includegraphics[width=1.0\linewidth]{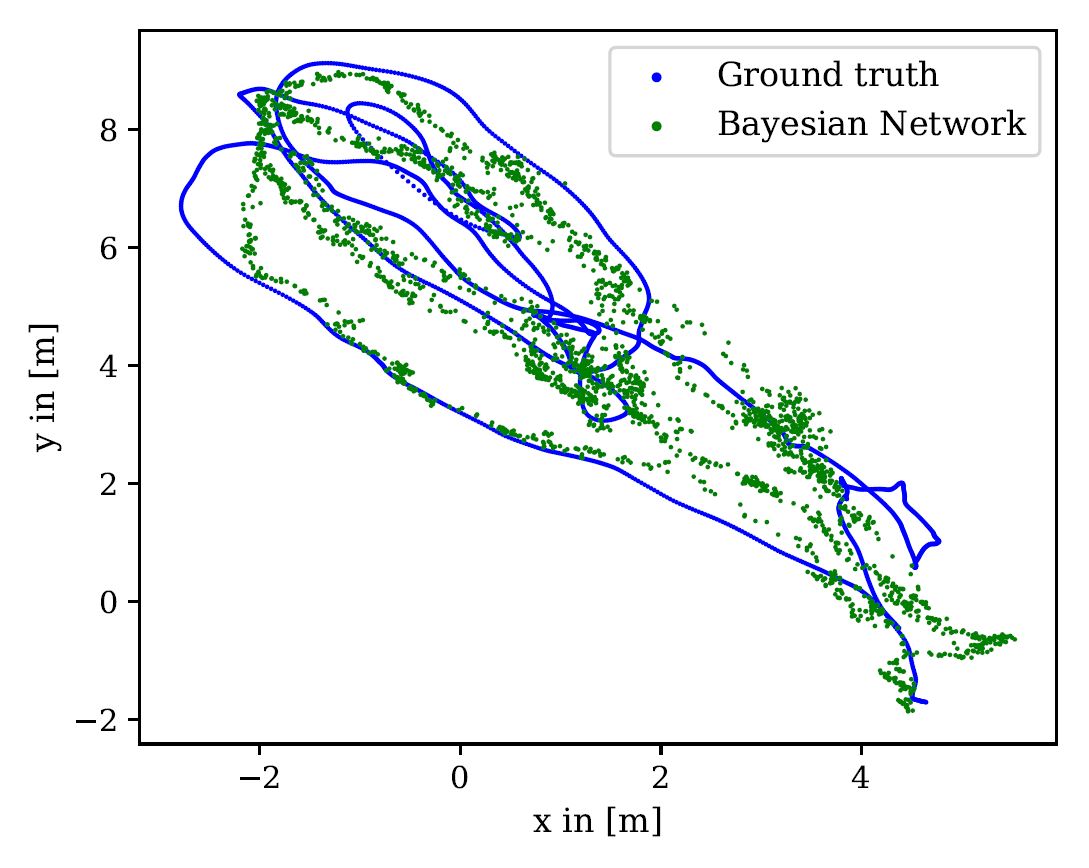}
    	\vspace{\vspacecaption}
    	\subcaption{Bayesian network~\cite{kendall_uncertainty}.}
    	\label{image_app_euroc_mh02_6}
        \vspace{\vspacefigure}
    \end{minipage}
    \caption{$\text{APR}_{\text{V}}$-$\text{RPR}_{\text{I}}$ fusion on EuRoC MAV~\cite{burri}: MH-02-easy.}
    \label{image_app_euroc_mh02}
\end{figure*}

\begin{figure*}[t!]
	\centering
	\begin{minipage}[b]{\len\linewidth}
        \centering
    	\includegraphics[width=1.0\linewidth]{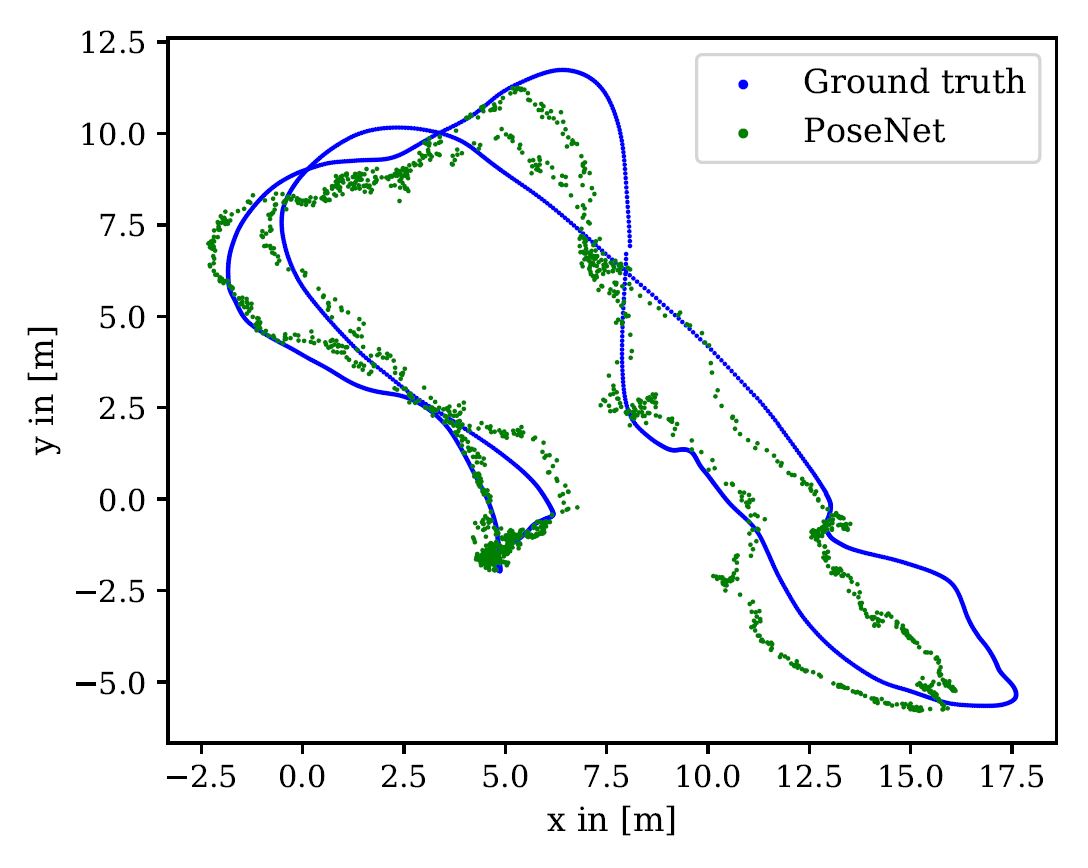}
    	\vspace{\vspacecaption}
    	\subcaption{$\text{APR}_{\text{V}}$: PoseNet~\cite{kendall}.}
    	\label{image_app_euroc_mh04_1}
        \vspace{\vspacefigure}
    \end{minipage}
    \hfill
    \begin{minipage}[b]{\len\linewidth}
        \centering
    	\includegraphics[width=1.0\linewidth]{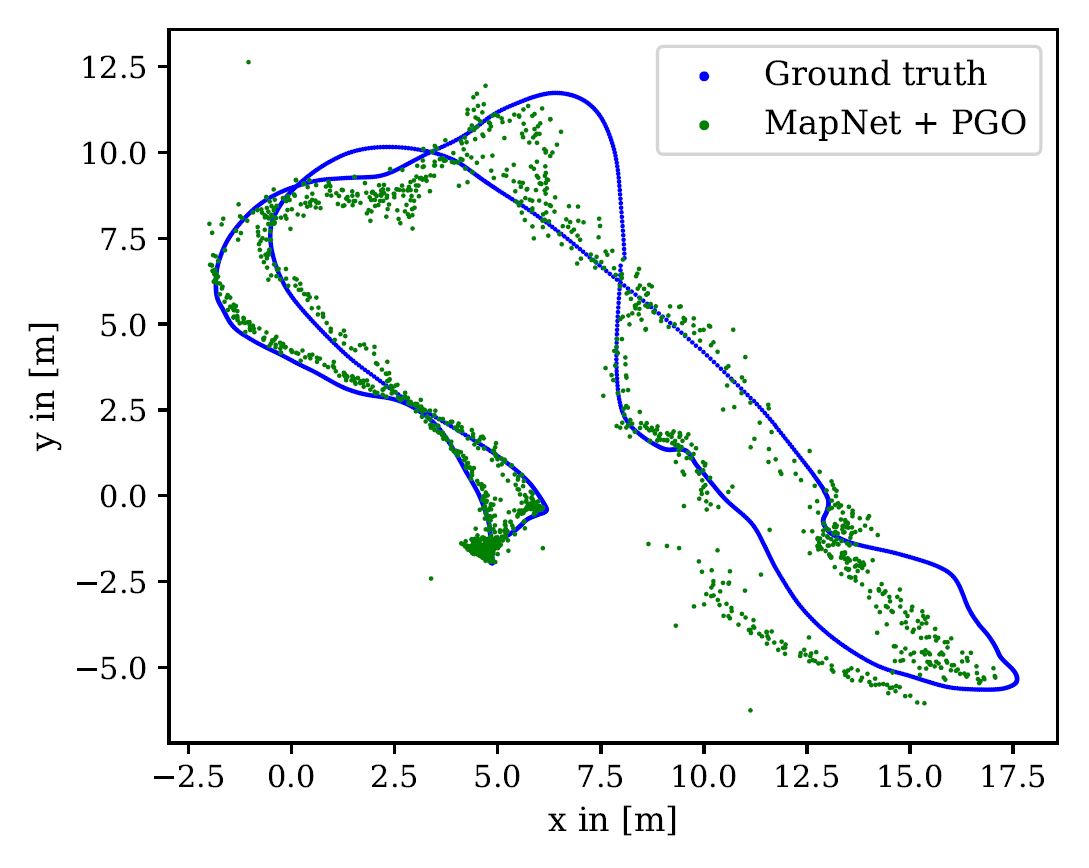}
    	\vspace{\vspacecaption}
    	\subcaption{MapNet+PGO~\cite{brahmbhatt}.}
    	\label{image_app_euroc_mh04_2}
        \vspace{\vspacefigure}
    \end{minipage}
    \hfill
    \begin{minipage}[b]{\len\linewidth}
        \centering
    	\includegraphics[width=1.0\linewidth]{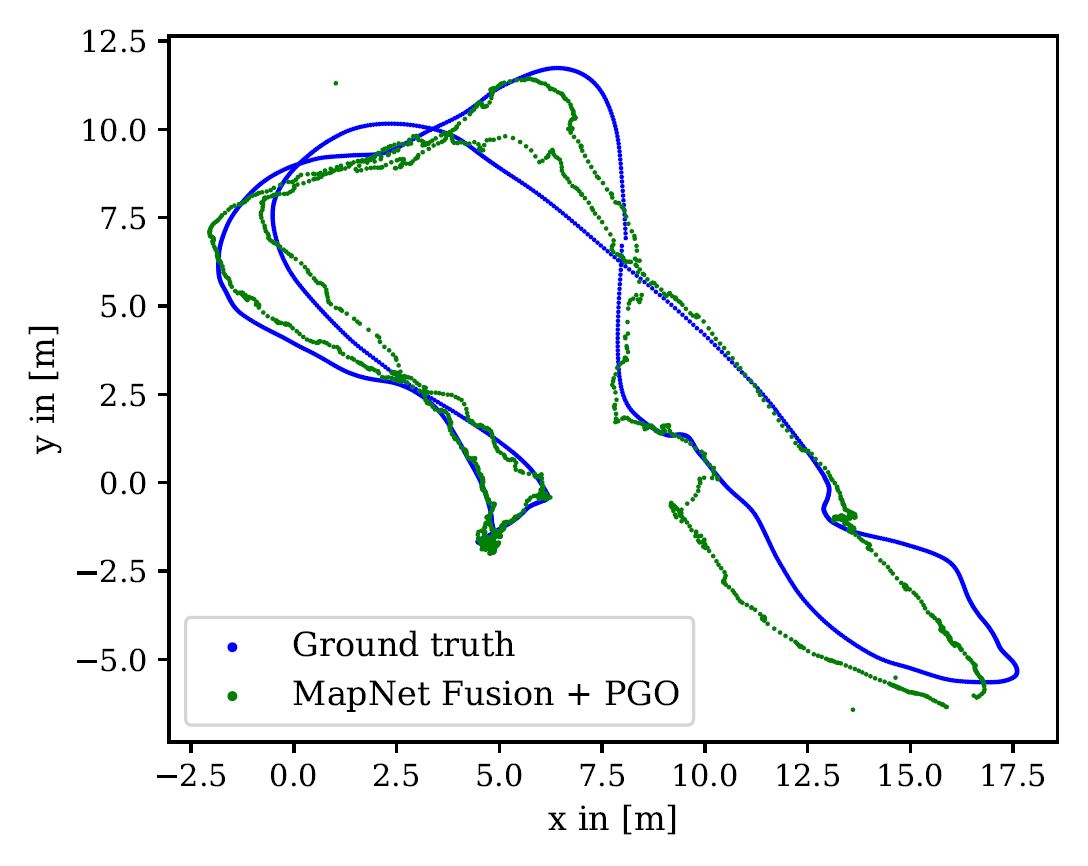}
    	\vspace{\vspacecaption}
    	\subcaption{$\text{APR}_{\text{V}}$-$\text{RPR}_{\text{I}}$+PGO.}
    	\label{image_app_euroc_mh04_3}
        \vspace{\vspacefigure}
    \end{minipage}
    \begin{minipage}[b]{\len\linewidth}
        \centering
    	\includegraphics[width=1.0\linewidth]{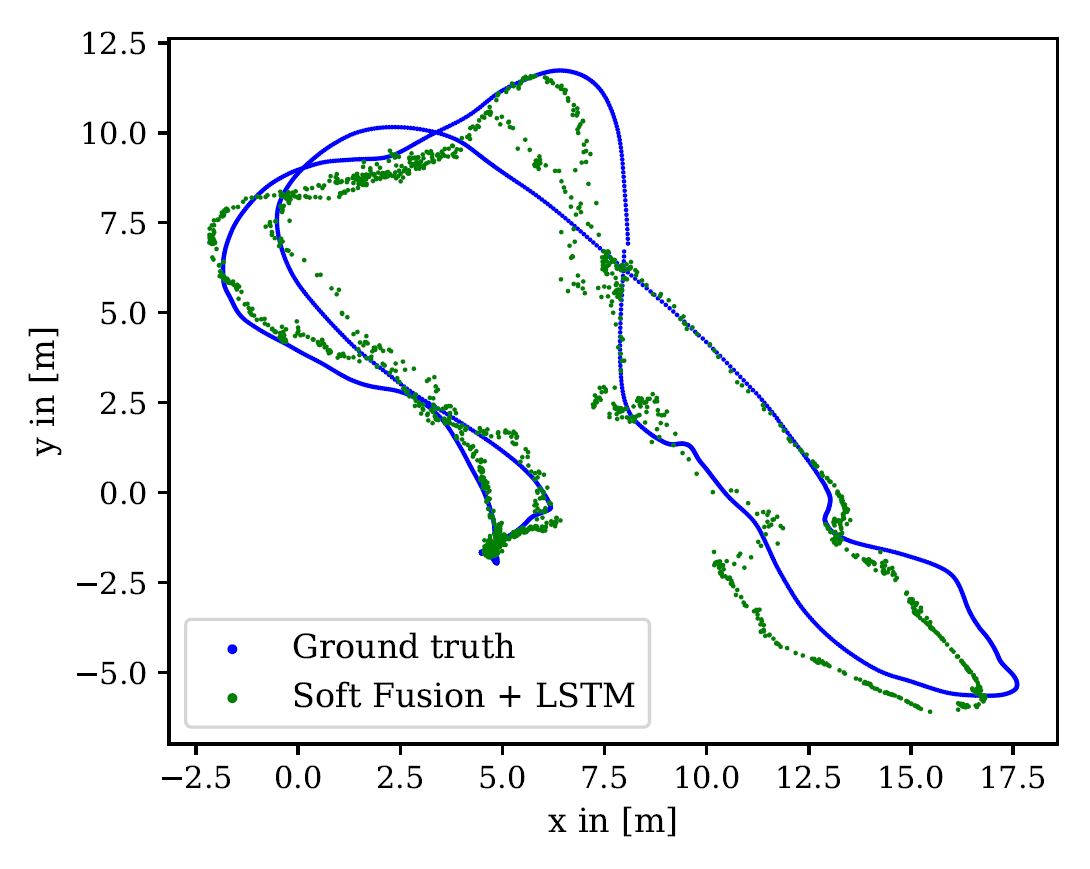}
    	\vspace{\vspacecaption}
    	\subcaption{SSF~\cite{chen} + BiLSTM.}
    	\label{image_app_euroc_mh04_4}
        \vspace{\vspacefigure}
    \end{minipage}
    \hfill
    \begin{minipage}[b]{\len\linewidth}
        \centering
    	\includegraphics[width=1.0\linewidth]{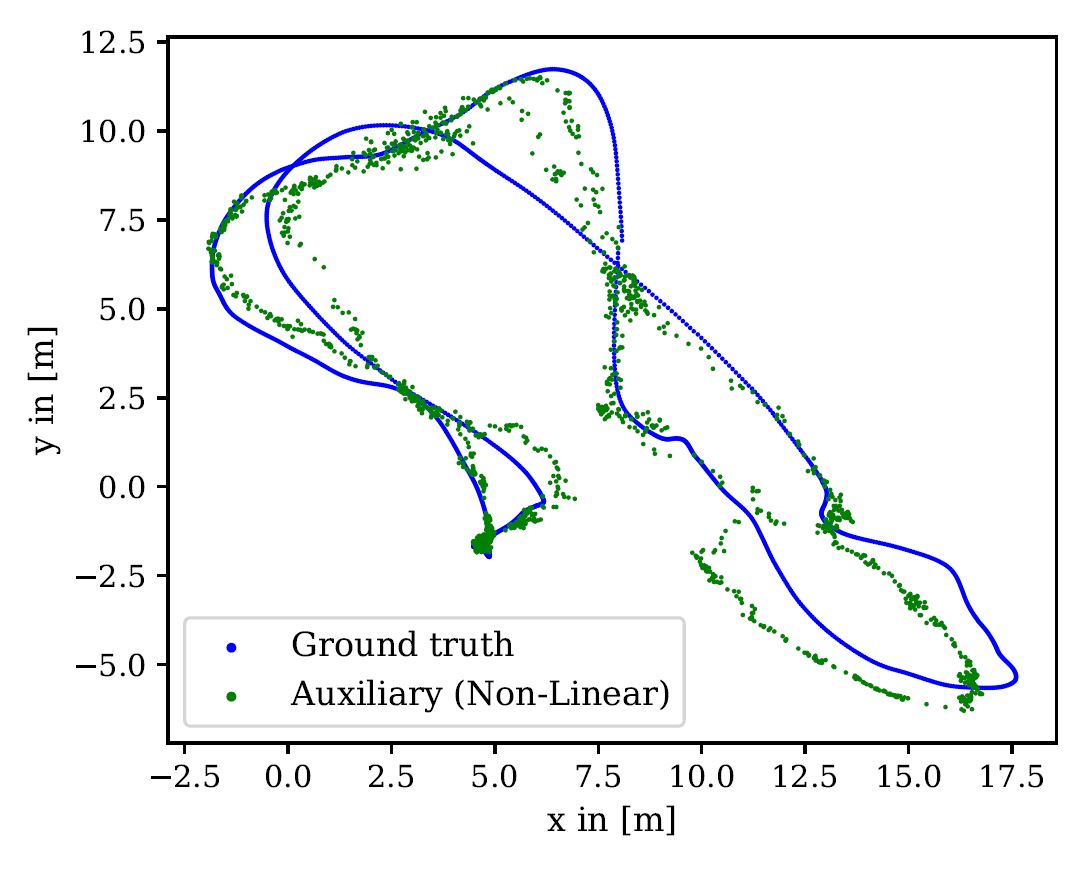}
    	\vspace{\vspacecaption}
    	\subcaption{Auxiliary learning (non-linear)~\cite{navon_aux}.}
    	\label{image_app_euroc_mh04_5}
        \vspace{\vspacefigure}
    \end{minipage}
    \hfill
	\begin{minipage}[b]{\len\linewidth}
        \centering
    	\includegraphics[width=1.0\linewidth]{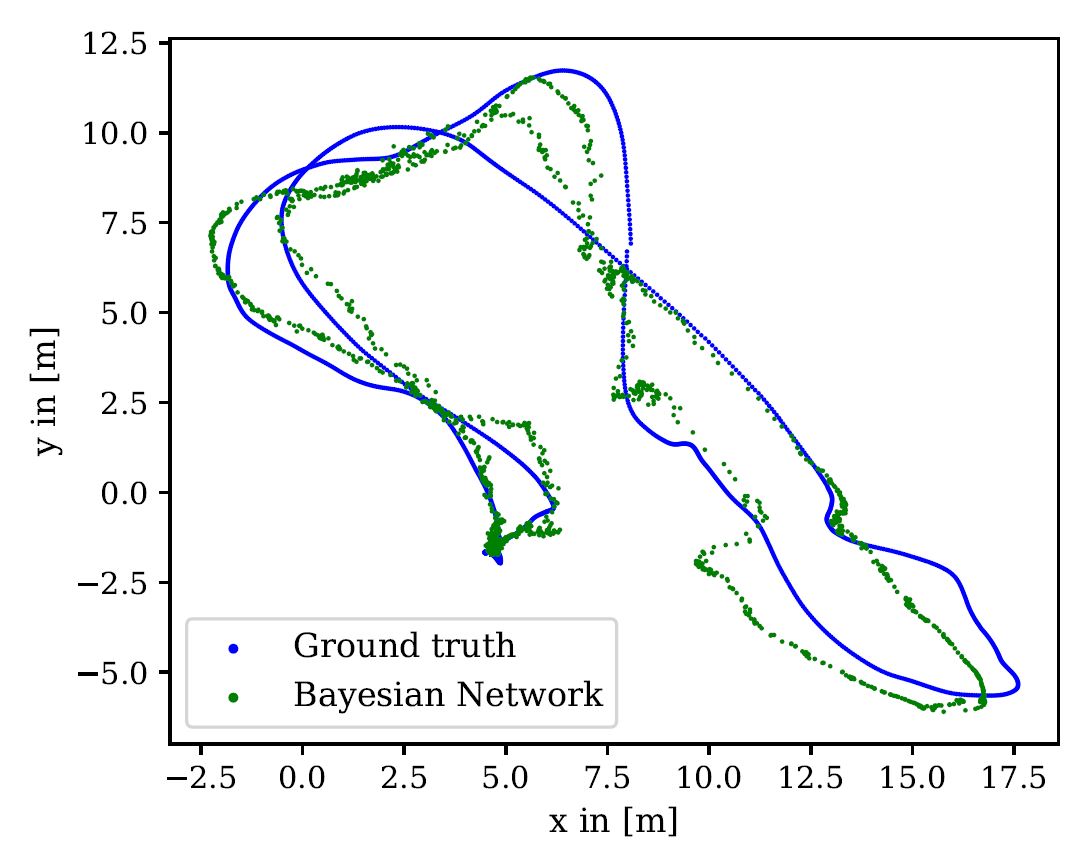}
    	\vspace{\vspacecaption}
    	\subcaption{Bayesian network~\cite{kendall_uncertainty}.}
    	\label{image_app_euroc_mh04_6}
        \vspace{\vspacefigure}
    \end{minipage}
    \caption{$\text{APR}_{\text{V}}$-$\text{RPR}_{\text{I}}$ fusion on EuRoC MAV~\cite{burri}: MH-04-difficult.}
    \label{image_app_euroc_mh04}
\end{figure*}

\begin{figure*}[t!]
	\centering
	\begin{minipage}[b]{\len\linewidth}
        \centering
    	\includegraphics[width=1.0\linewidth]{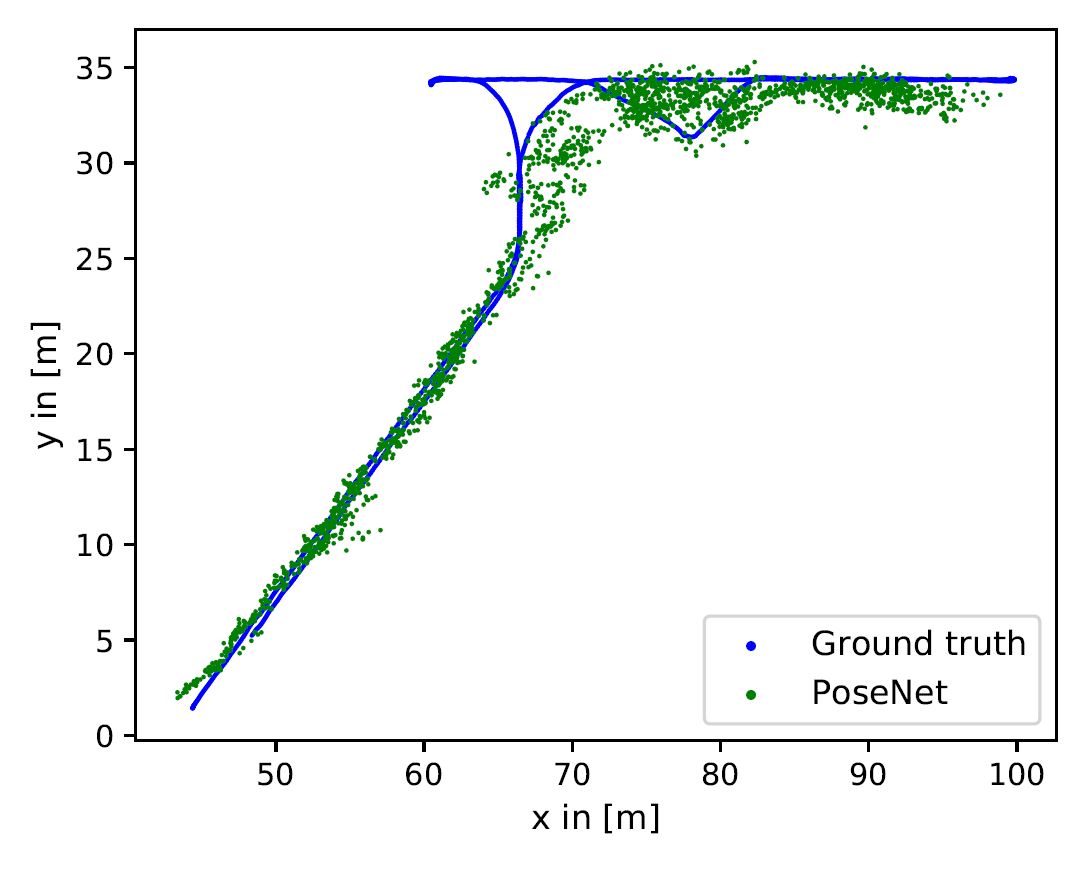}
    	\vspace{\vspacecaption}
    	\subcaption{$\text{APR}_{\text{V}}$: PoseNet~\cite{kendall}.}
    	\label{image_app_penncosy_bf_1}
        \vspace{\vspacefigure}
    \end{minipage}
    \hfill
    \begin{minipage}[b]{\len\linewidth}
        \centering
    	\includegraphics[width=1.0\linewidth]{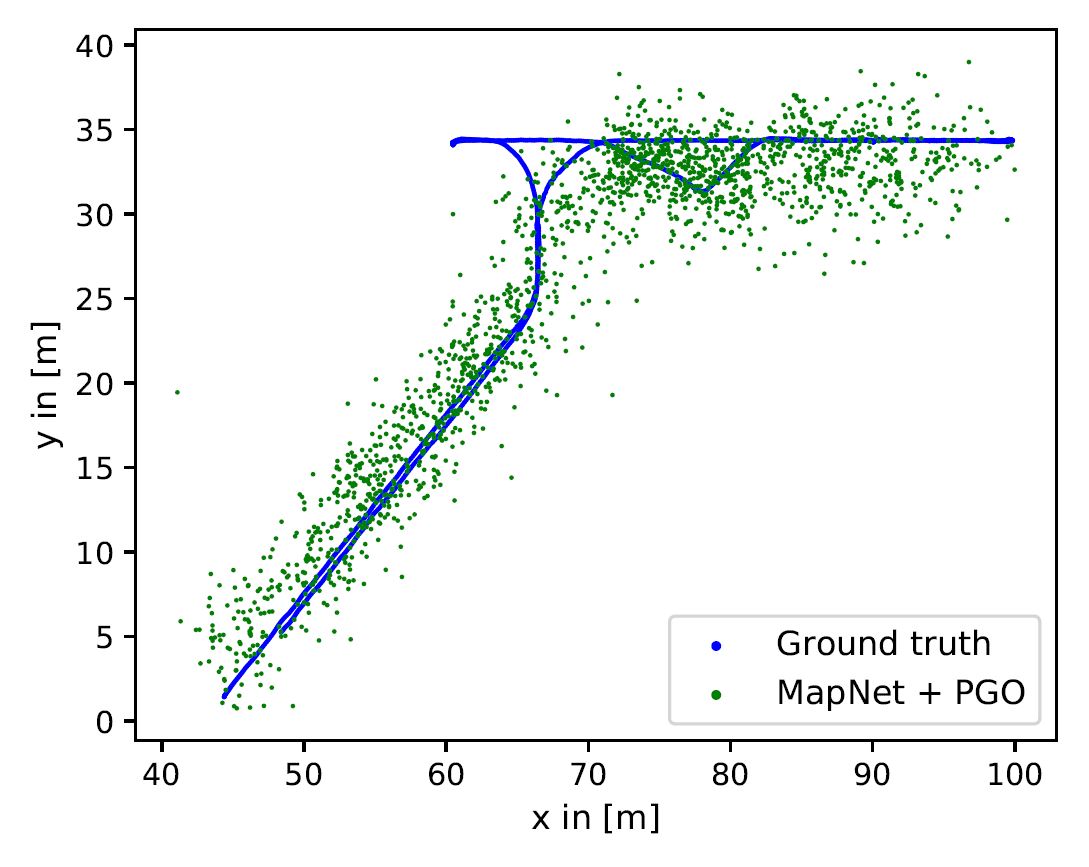}
    	\vspace{\vspacecaption}
    	\subcaption{MapNet+PGO~\cite{brahmbhatt}.}
    	\label{image_app_penncosy_bf_2}
        \vspace{\vspacefigure}
    \end{minipage}
    \hfill
    \begin{minipage}[b]{\len\linewidth}
        \centering
    	\includegraphics[width=1.0\linewidth]{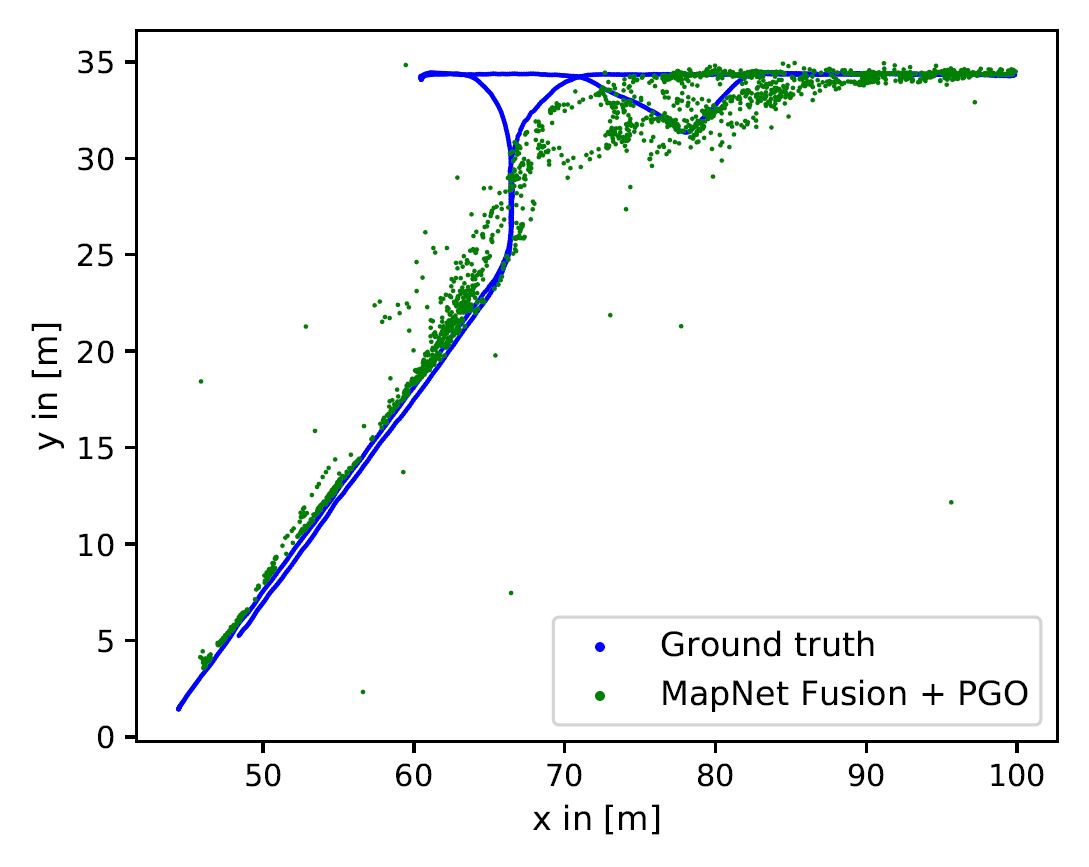}
    	\vspace{\vspacecaption}
    	\subcaption{$\text{APR}_{\text{V}}$-$\text{RPR}_{\text{I}}$+PGO.}
    	\label{image_app_penncosy_bf_3}
        \vspace{\vspacefigure}
    \end{minipage}
    \begin{minipage}[b]{\len\linewidth}
        \centering
    	\includegraphics[width=1.0\linewidth]{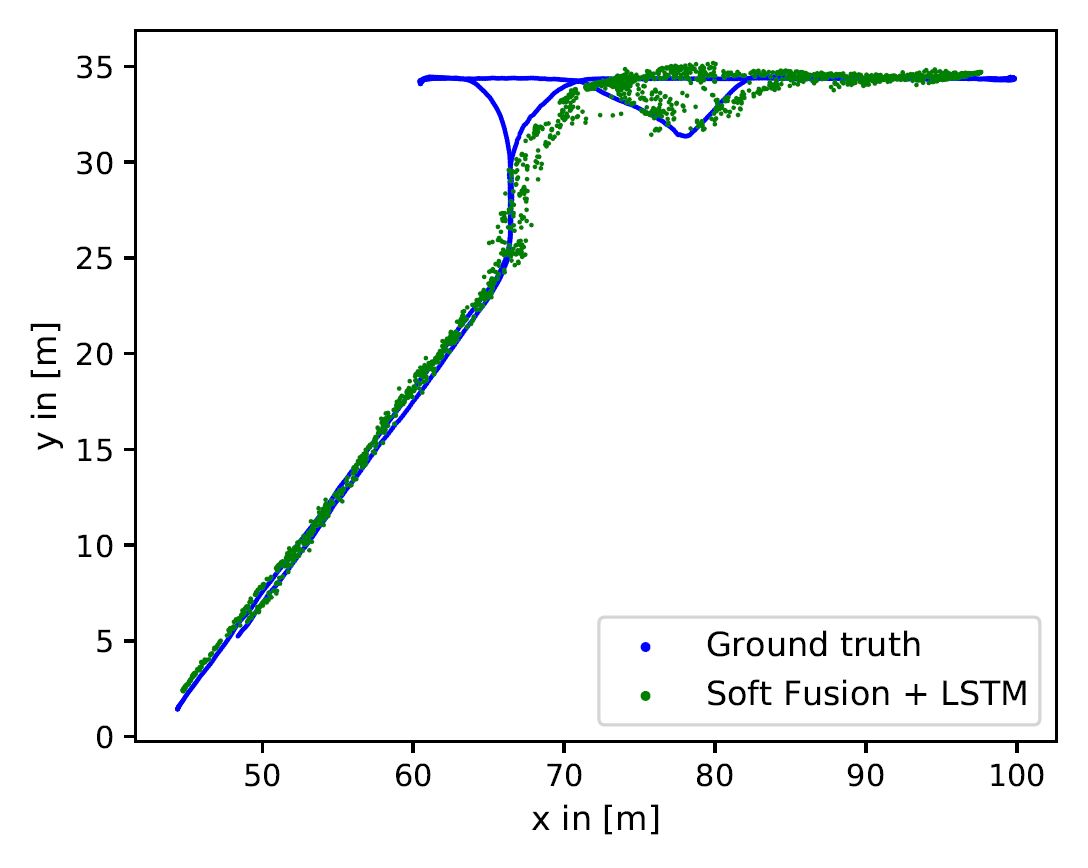}
    	\vspace{\vspacecaption}
    	\subcaption{Soft Fusion~\cite{chen}  with LSTM.}
    	\label{image_app_penncosy_bf_4}
        \vspace{\vspacefigure}
    \end{minipage}
    \hfill
    \begin{minipage}[b]{\len\linewidth}
        \centering
    	\includegraphics[width=1.0\linewidth]{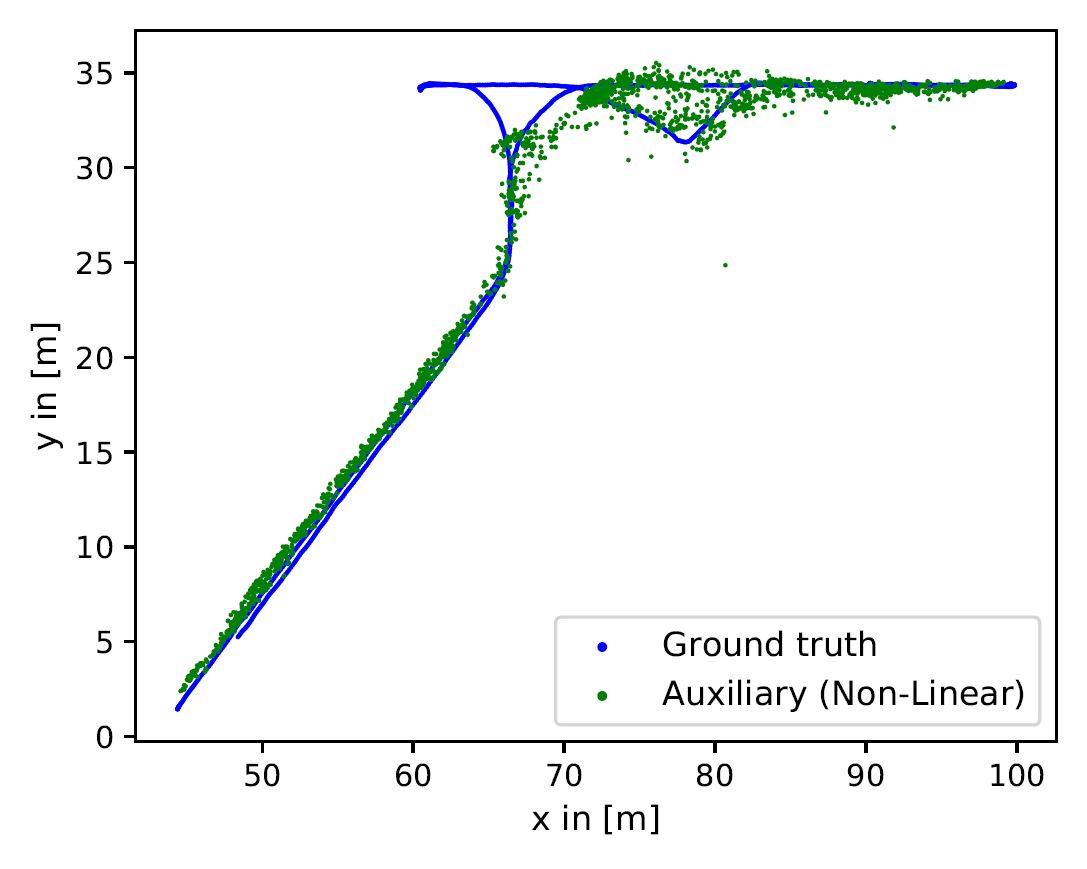}
    	\vspace{\vspacecaption}
    	\subcaption{Auxiliary learning (non-linear)~\cite{navon_aux}.}
    	\label{image_app_penncosy_bf_5}
        \vspace{\vspacefigure}
    \end{minipage}
    \hfill
	\begin{minipage}[b]{\len\linewidth}
        \centering
    	\includegraphics[width=1.0\linewidth]{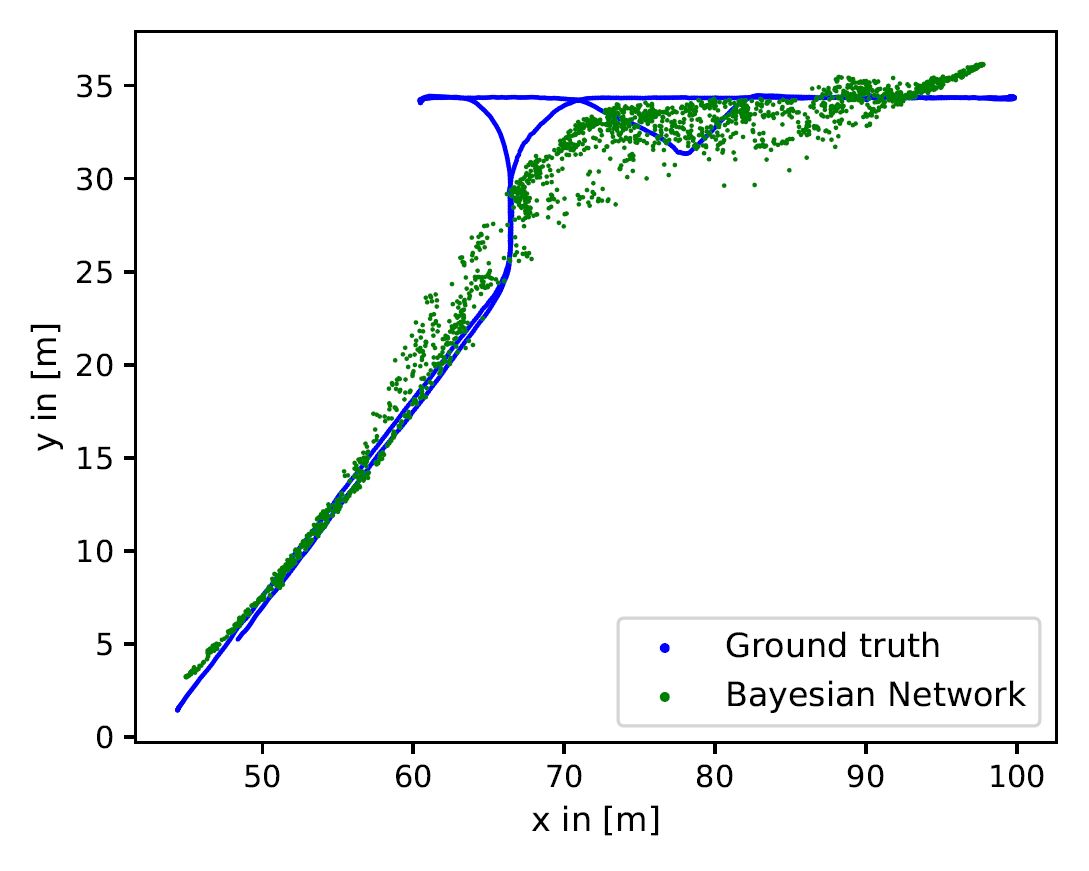}
    	\vspace{\vspacecaption}
    	\subcaption{Bayesian network~\cite{kendall_uncertainty}.}
    	\label{image_app_penncosy_bf_6}
        \vspace{\vspacefigure}
    \end{minipage}
    \caption{$\text{APR}_{\text{V}}$-$\text{RPR}_{\text{I}}$ fusion on PennCOSYVIO~\cite{pfrommer}: BF.}
    \label{image_app_penncosy_bf}
\end{figure*}

\begin{figure*}[t!]
	\centering
	\begin{minipage}[b]{\len\linewidth}
        \centering
    	\includegraphics[width=1.0\linewidth]{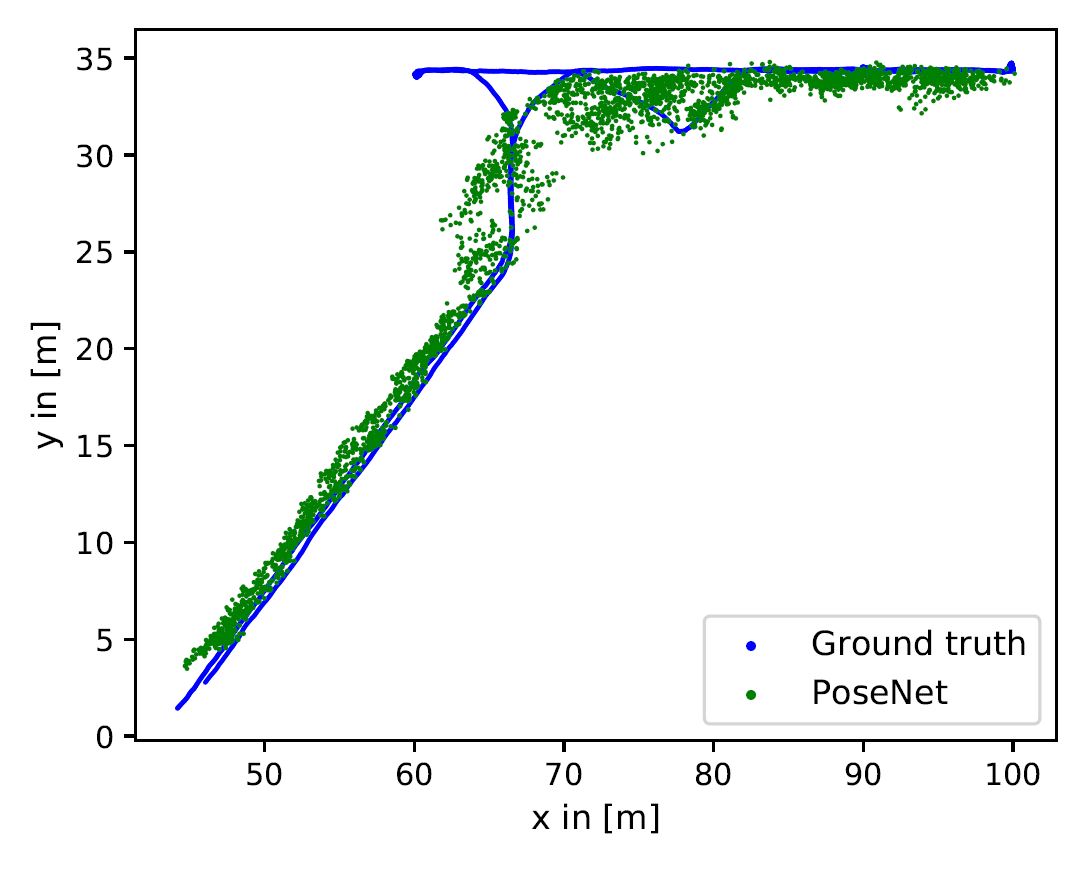}
    	\vspace{\vspacecaption}
    	\subcaption{$\text{APR}_{\text{V}}$: PoseNet~\cite{kendall}.}
    	\label{image_app_penncosy_bs_1}
        \vspace{\vspacefigure}
    \end{minipage}
    \hfill
    \begin{minipage}[b]{\len\linewidth}
        \centering
    	\includegraphics[width=1.0\linewidth]{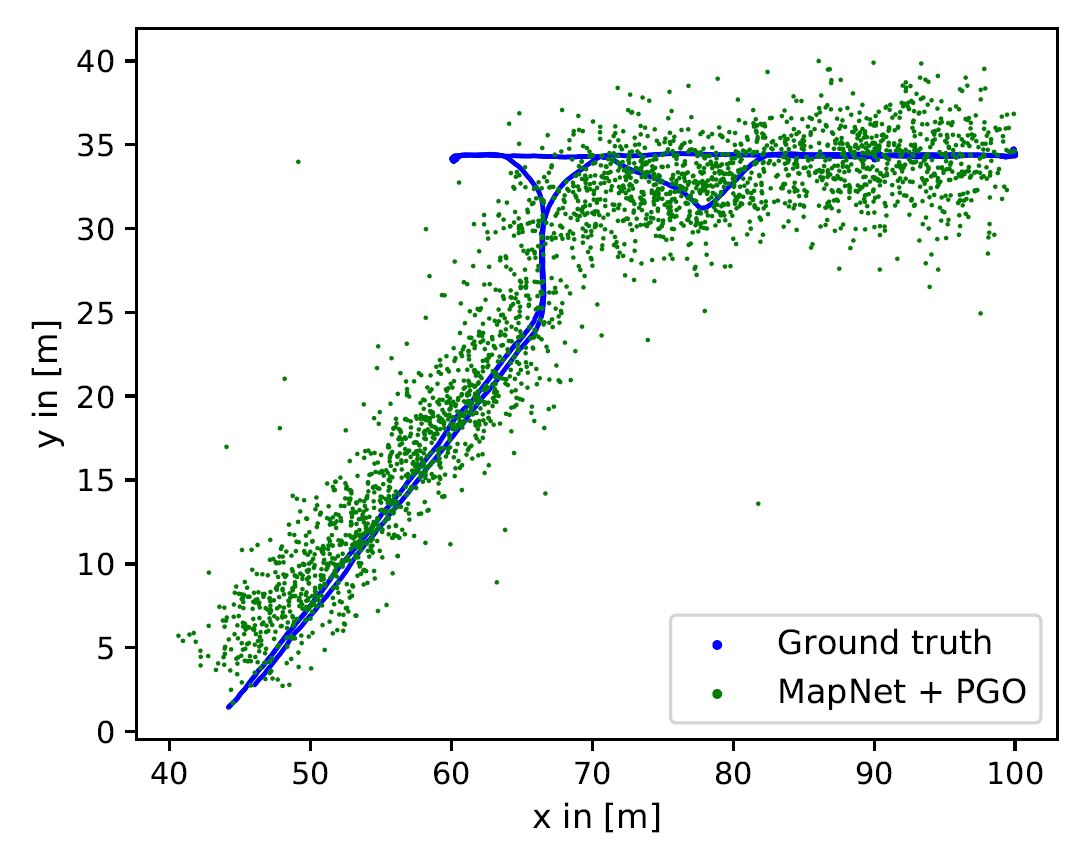}
    	\vspace{\vspacecaption}
    	\subcaption{MapNet+PGO~\cite{brahmbhatt}.}
    	\label{image_app_penncosy_bs_2}
        \vspace{\vspacefigure}
    \end{minipage}
    \hfill
    \begin{minipage}[b]{\len\linewidth}
        \centering
    	\includegraphics[width=1.0\linewidth]{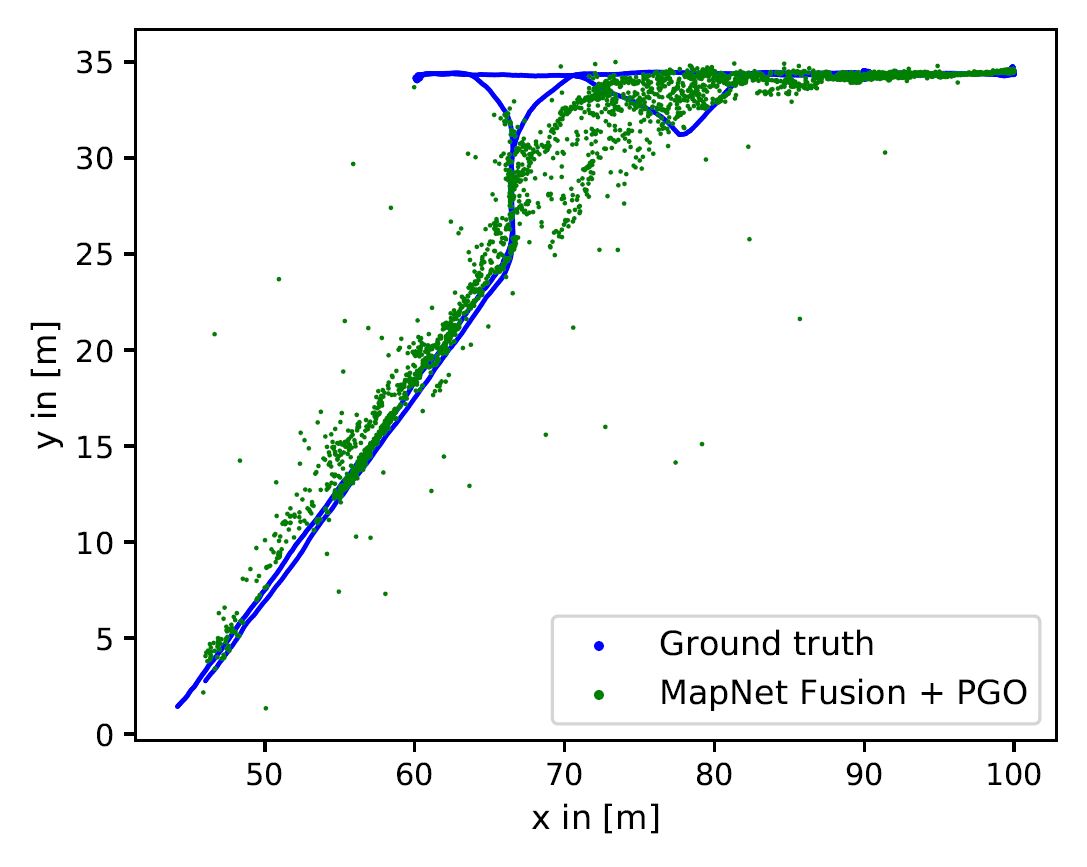}
    	\vspace{\vspacecaption}
    	\subcaption{$\text{APR}_{\text{V}}$-$\text{RPR}_{\text{I}}$+PGO.}
    	\label{image_app_penncosy_bs_3}
        \vspace{\vspacefigure}
    \end{minipage}
    \begin{minipage}[b]{\len\linewidth}
        \centering
    	\includegraphics[width=1.0\linewidth]{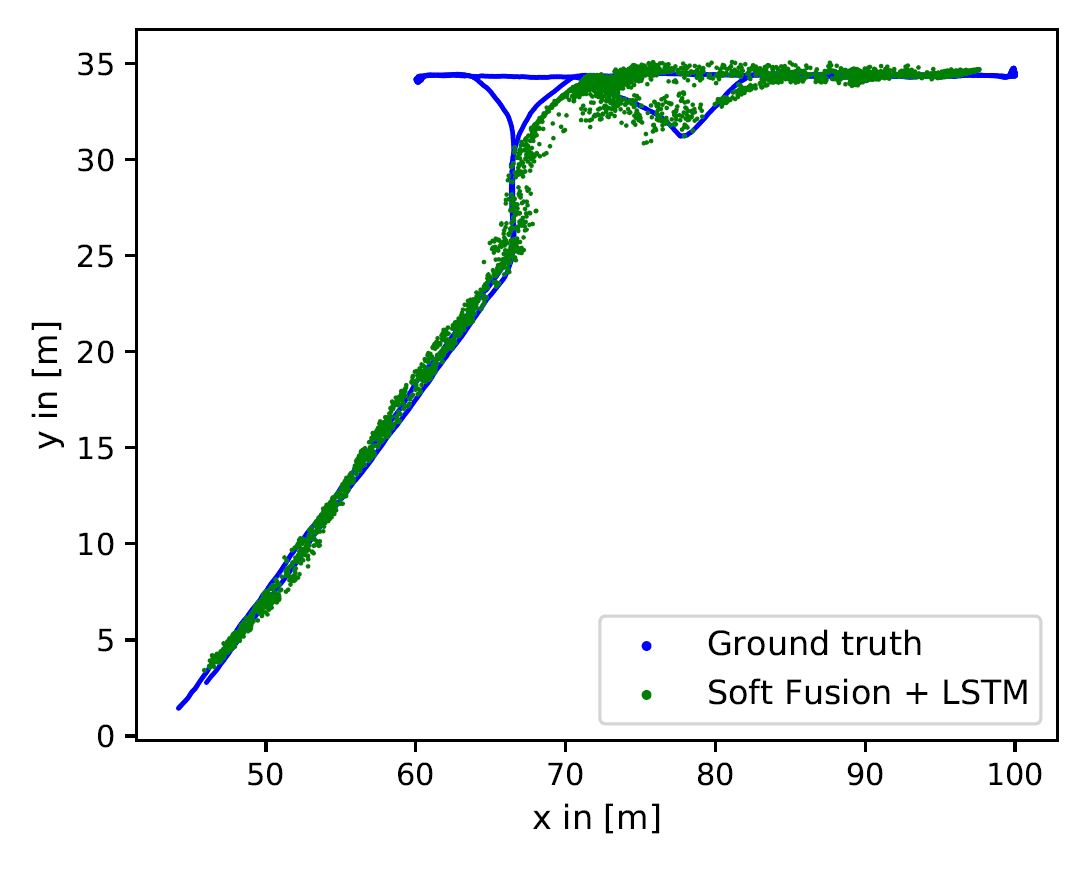}
    	\vspace{\vspacecaption}
    	\subcaption{SSF~\cite{chen} + BiLSTM.}
    	\label{image_app_penncosy_bs_4}
        \vspace{\vspacefigure}
    \end{minipage}
    \hfill
    \begin{minipage}[b]{\len\linewidth}
        \centering
    	\includegraphics[width=1.0\linewidth]{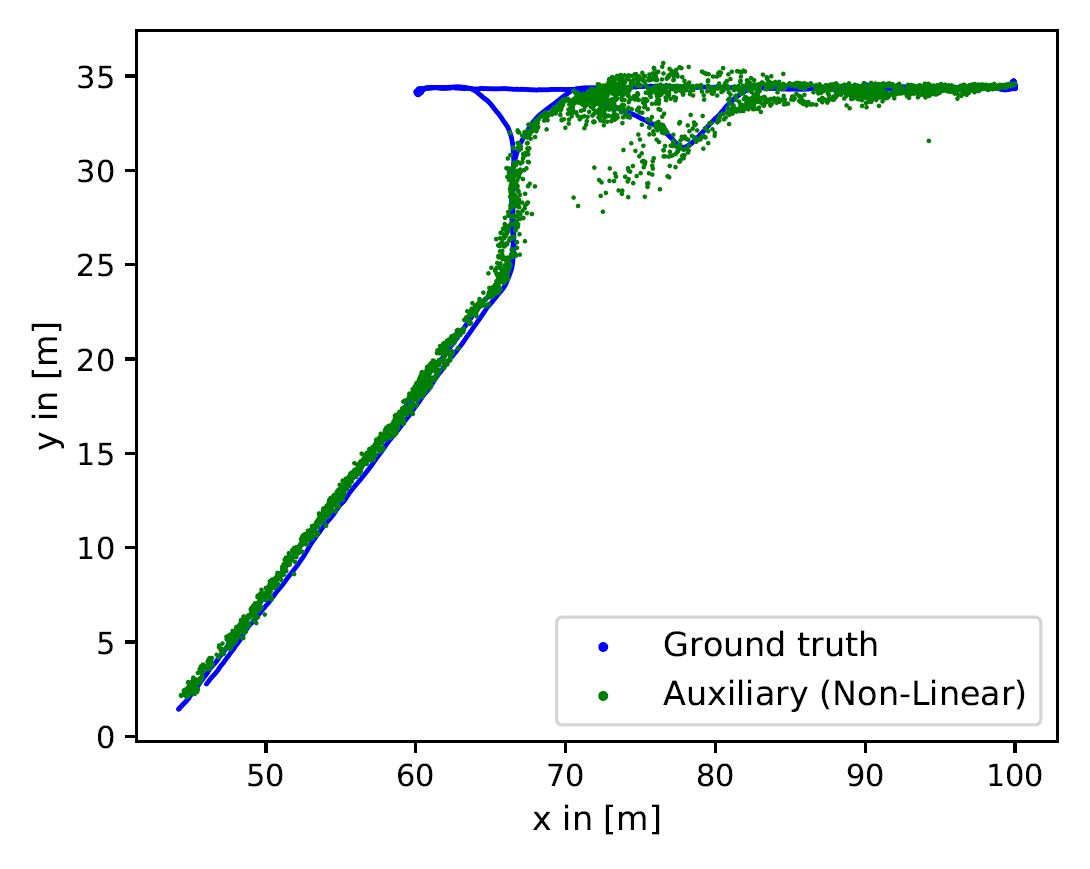}
    	\vspace{\vspacecaption}
    	\subcaption{Auxiliary learning (non-linear) \cite{navon_aux}.}
    	\label{image_app_penncosy_bs_5}
        \vspace{\vspacefigure}
    \end{minipage}
    \hfill
	\begin{minipage}[b]{\len\linewidth}
        \centering
    	\includegraphics[width=1.0\linewidth]{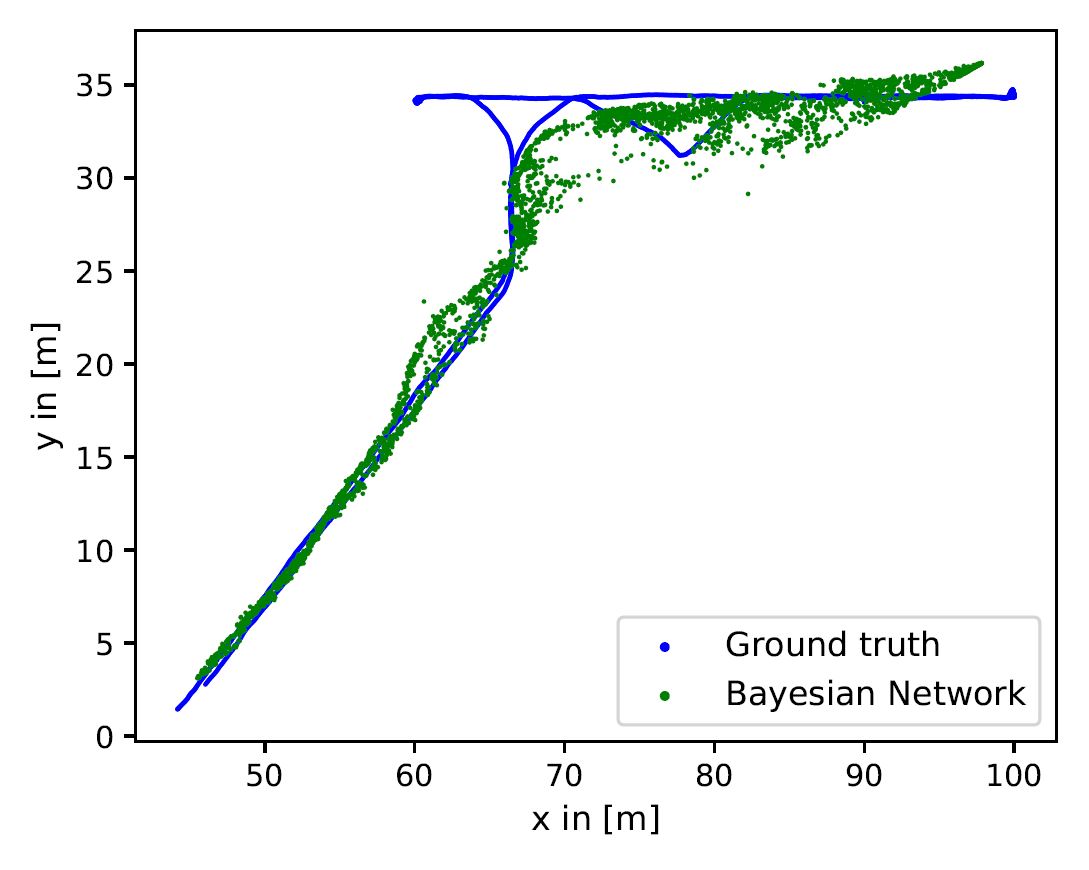}
    	\vspace{\vspacecaption}
    	\subcaption{Bayesian network~\cite{kendall_uncertainty}.}
    	\label{image_app_penncosy_bs_6}
        \vspace{\vspacefigure}
    \end{minipage}
    \caption{$\text{APR}_{\text{V}}$-$\text{RPR}_{\text{I}}$ fusion on PennCOSYVIO~\cite{pfrommer}: BS.}
    \label{image_app_penncosy_bs}
\end{figure*}

\begin{figure*}[t!]
	\centering
	\begin{minipage}[b]{\len\linewidth}
        \centering
    	\includegraphics[width=1.0\linewidth]{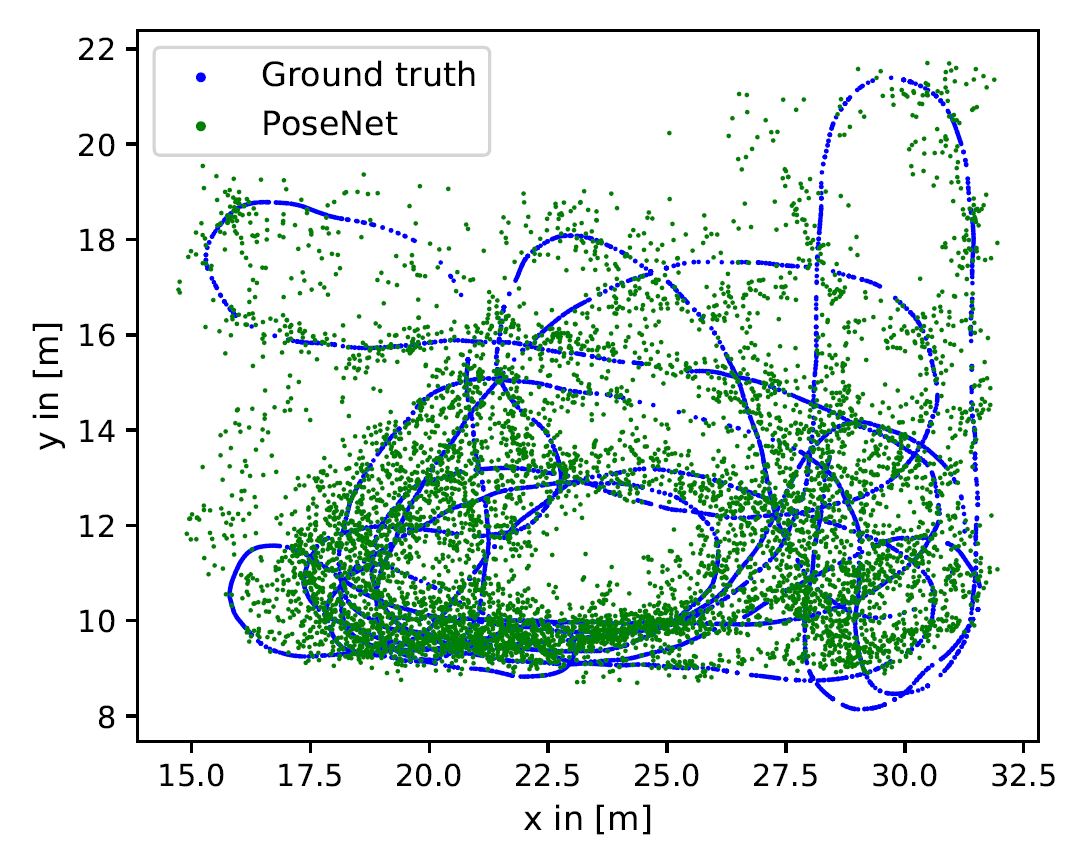}
    	\vspace{\vspacecaption}
    	\subcaption{$\text{APR}_{\text{V}}$: PoseNet~\cite{kendall}.}
    	\label{image_app_industry1_1}
        \vspace{\vspacefigure}
    \end{minipage}
    \hfill
        \begin{minipage}[b]{\len\linewidth}
        \centering
    	\includegraphics[width=1.0\linewidth]{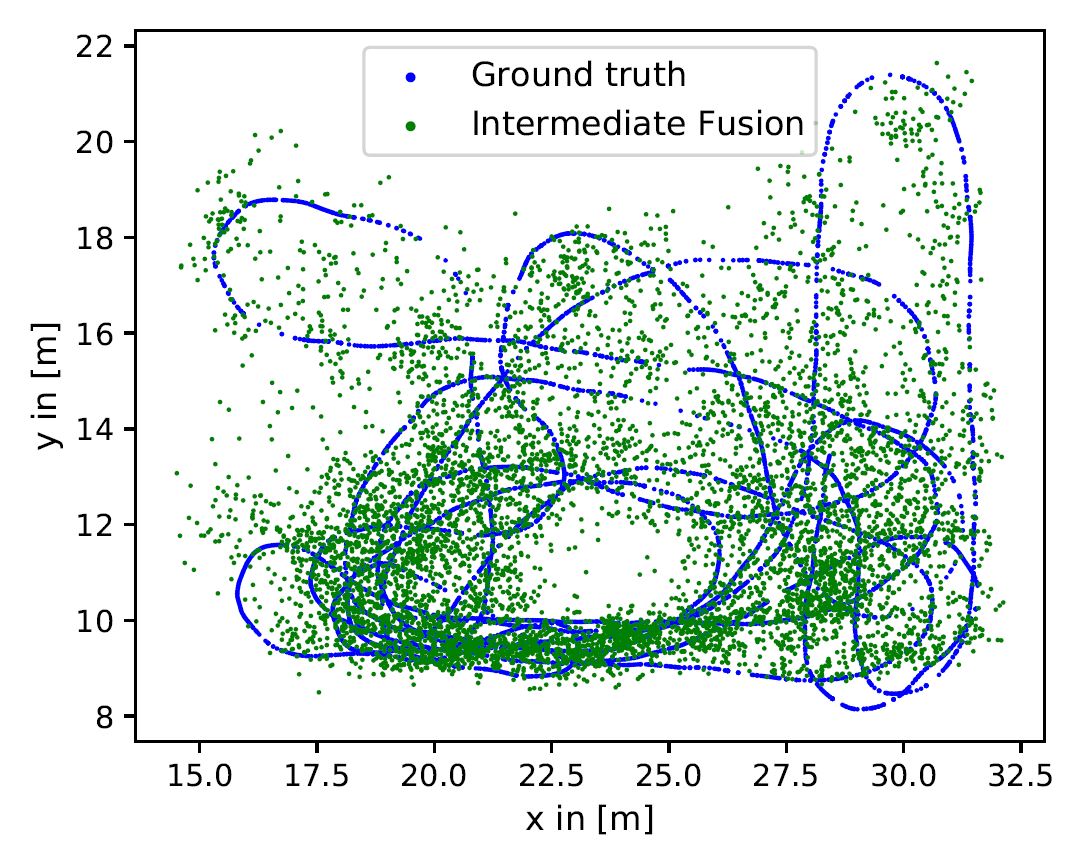}
    	\vspace{\vspacecaption}
    	\subcaption{MMTM~\cite{joze}.}
    	\label{image_app_industry1_2}
        \vspace{\vspacefigure}
    \end{minipage}
    \hfill
    \begin{minipage}[b]{\len\linewidth}
        \centering
    	\includegraphics[width=1.0\linewidth]{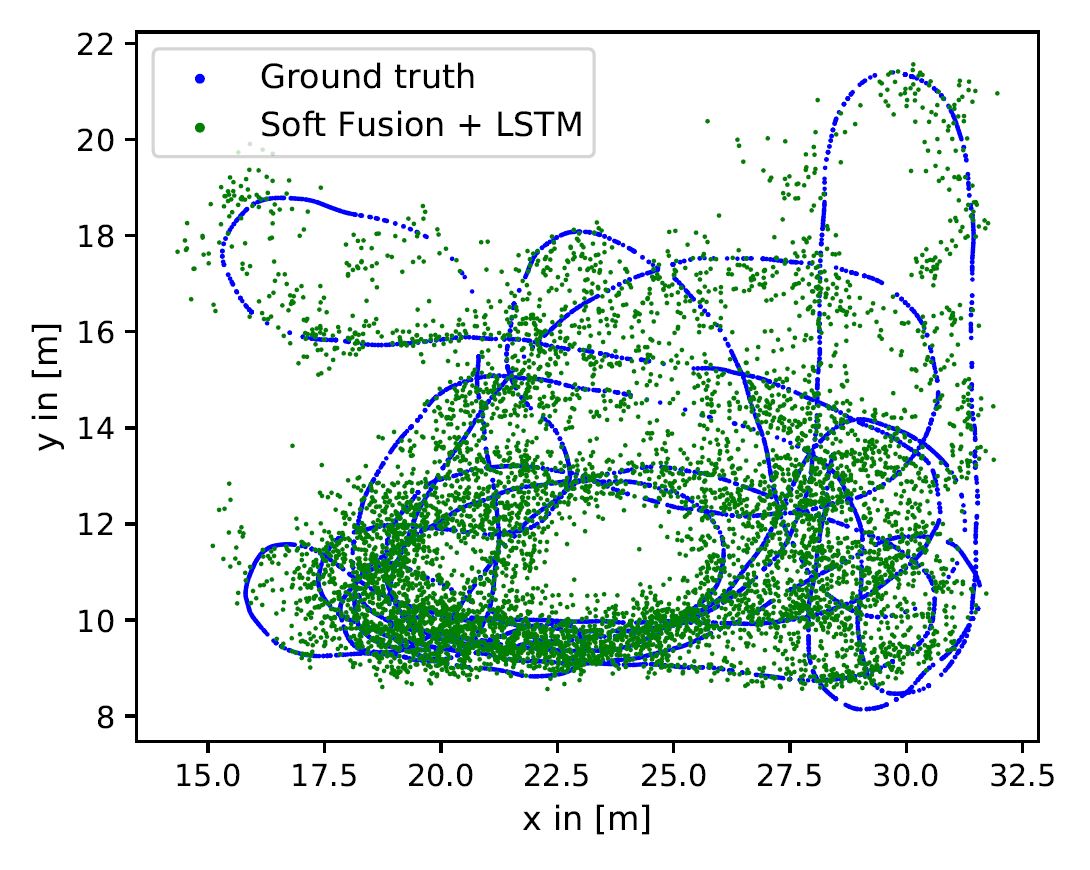}
    	\vspace{\vspacecaption}
    	\subcaption{SSF~\cite{chen} + BiLSTM.}
    	\label{image_app_industry1_3}
        \vspace{\vspacefigure}
    \end{minipage}
    \caption{$\text{APR}_{\text{V}}$-$\text{RPR}_{\text{I}}$ fusion on IndustyVI: Testing trajectory 1.}
    \label{image_app_industry1}
\end{figure*}

\begin{figure*}[th]
	\centering
	\begin{minipage}[b]{\len\linewidth}
        \centering
    	\includegraphics[width=1.0\linewidth]{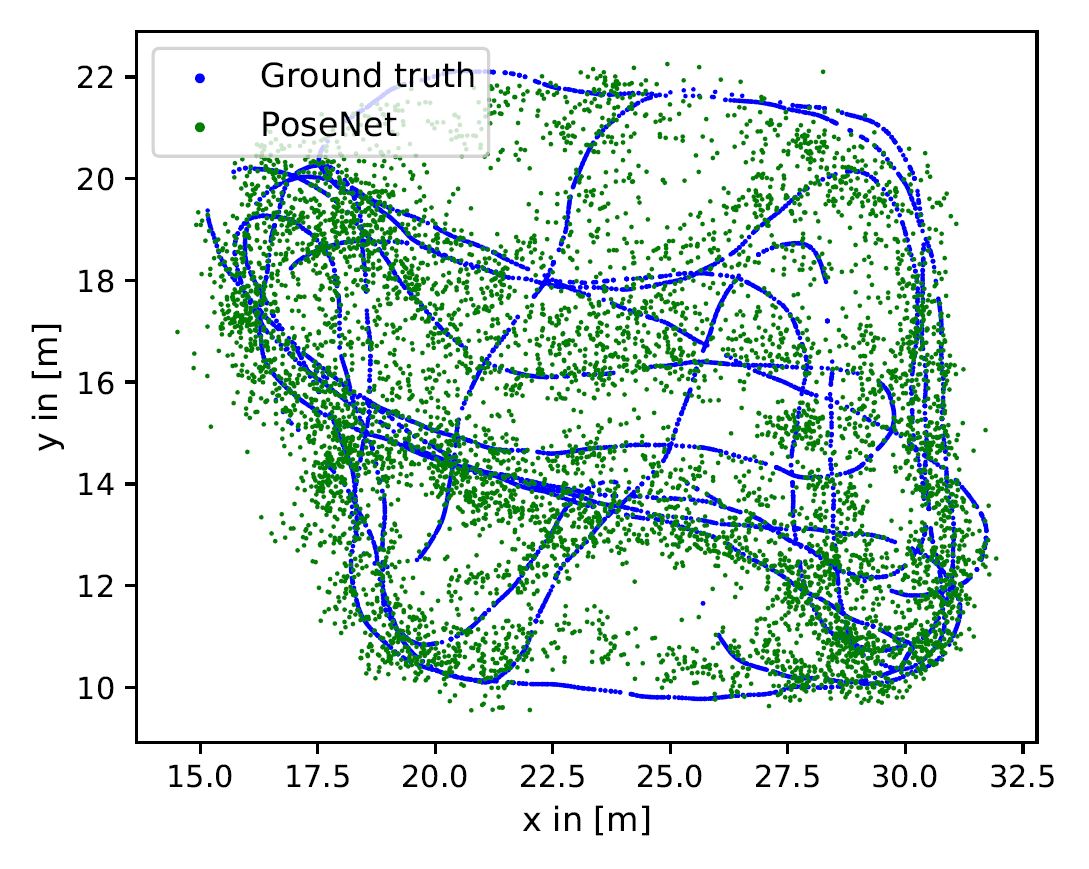}
    	\vspace{\vspacecaption}
    	\subcaption{$\text{APR}_{\text{V}}$: PoseNet~\cite{kendall}.}
    	\label{image_app_industry2_1}
        \vspace{\vspacefigure}
    \end{minipage}
    \hfill
    \begin{minipage}[b]{\len\linewidth}
        \centering
    	\includegraphics[width=1.0\linewidth]{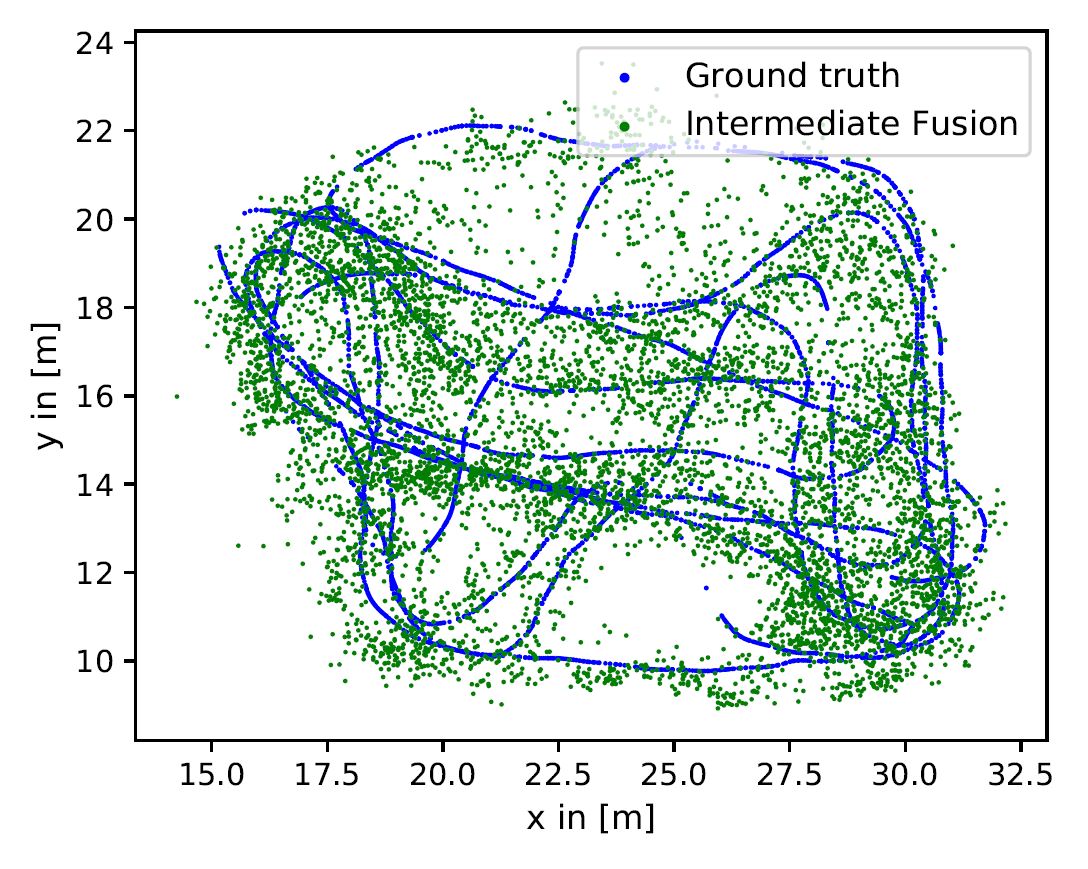}
    	\vspace{\vspacecaption}
    	\subcaption{MMTM~\cite{joze}.}
    	\label{image_app_industry2_2}
        \vspace{\vspacefigure}
    \end{minipage}
    \hfill
    \begin{minipage}[b]{\len\linewidth}
        \centering
    	\includegraphics[width=1.0\linewidth]{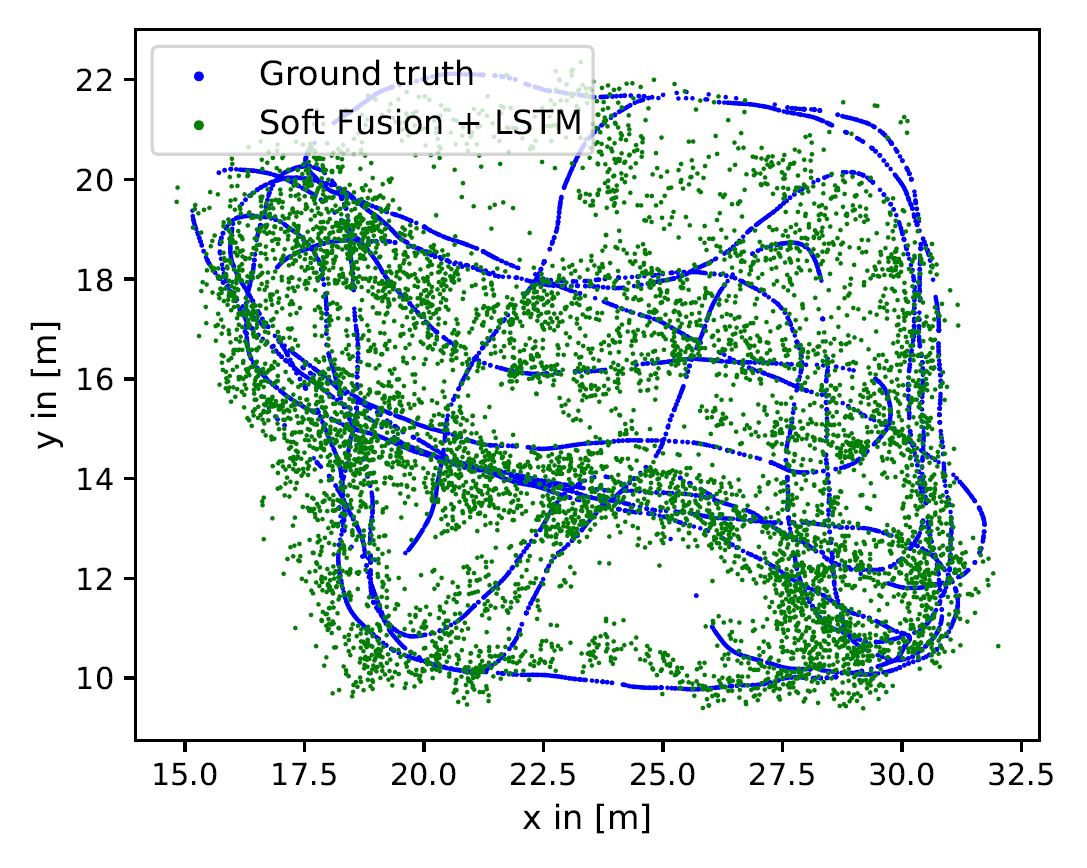}
    	\vspace{\vspacecaption}
    	\subcaption{SSF~\cite{chen} + BiLSTM.}
    	\label{image_app_industry2_3}
        \vspace{\vspacefigure}
    \end{minipage}
    \caption{$\text{APR}_{\text{V}}$-$\text{RPR}_{\text{I}}$ fusion on IndustyVI: Testing trajectory 2.}
    \label{image_app_industry2}
\end{figure*}

\newcommand\lene{0.24}
\newcommand\vspacefiguree{0.4cm}
\newcommand\vspacecaptione{-0.6cm}
\begin{figure*}[t!]
	\centering
	\begin{minipage}[b]{\lene\linewidth}
        \centering
    	\includegraphics[width=1.0\linewidth]{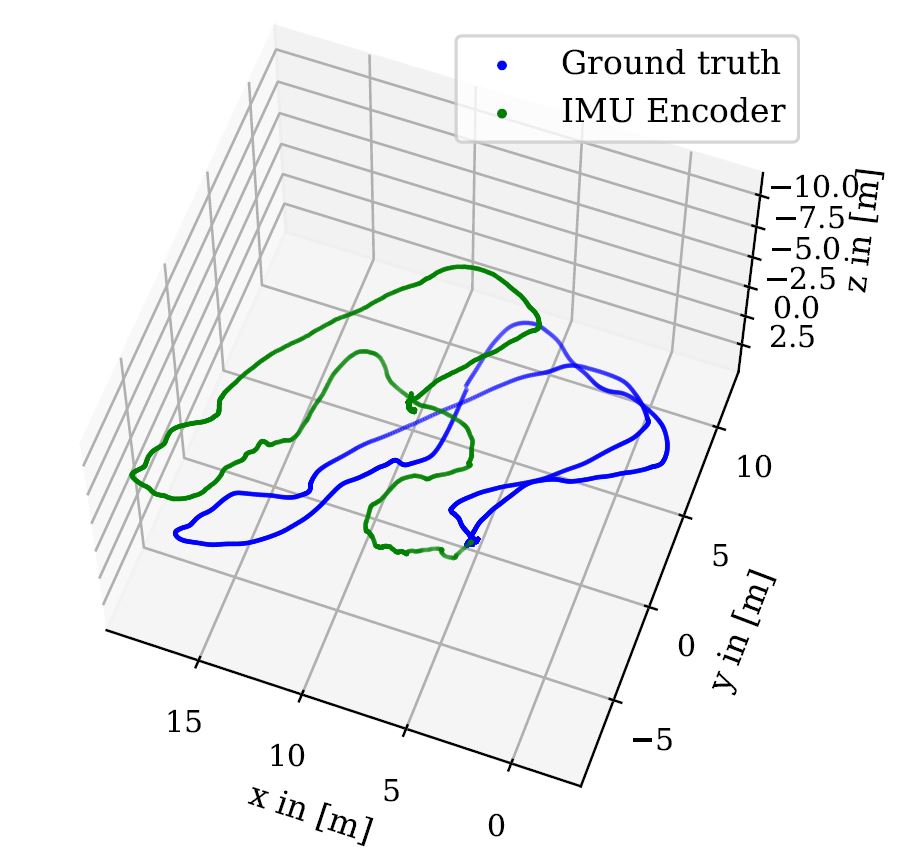}
    	\vspace{\vspacecaptione}
    	\subcaption{$\text{RPR}_{\text{I}}$: IMUNet~\cite{silva}.}
    	\label{image_app_rpr_euroc_mh04_1}
        \vspace{\vspacefiguree}
    \end{minipage}
    \hfill
    \begin{minipage}[b]{\lene\linewidth}
        \centering
    	\includegraphics[width=1.0\linewidth]{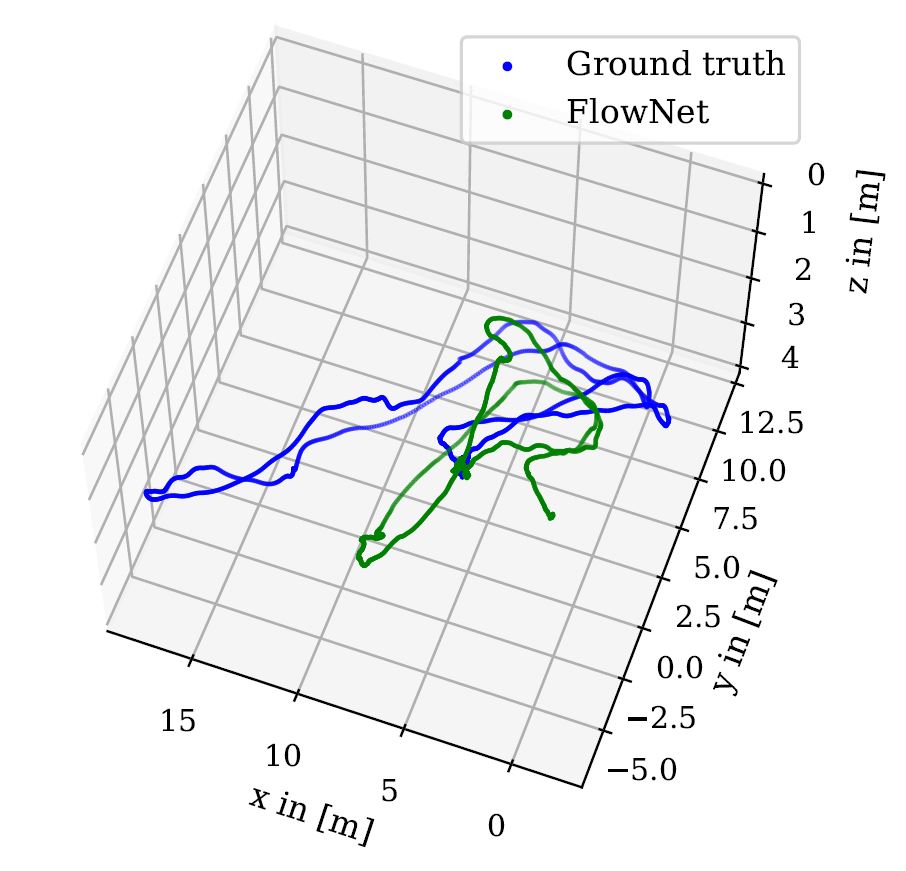}
    	\vspace{\vspacecaptione}
    	\subcaption{$\text{RPR}_{\text{V}}$: FlowNet~\cite{dosovitskiy}.}
    	\label{image_app_rpr_euroc_mh04_2}
        \vspace{\vspacefiguree}
    \end{minipage}
    \hfill
    \begin{minipage}[b]{\lene\linewidth}
        \centering
    	\includegraphics[width=1.0\linewidth]{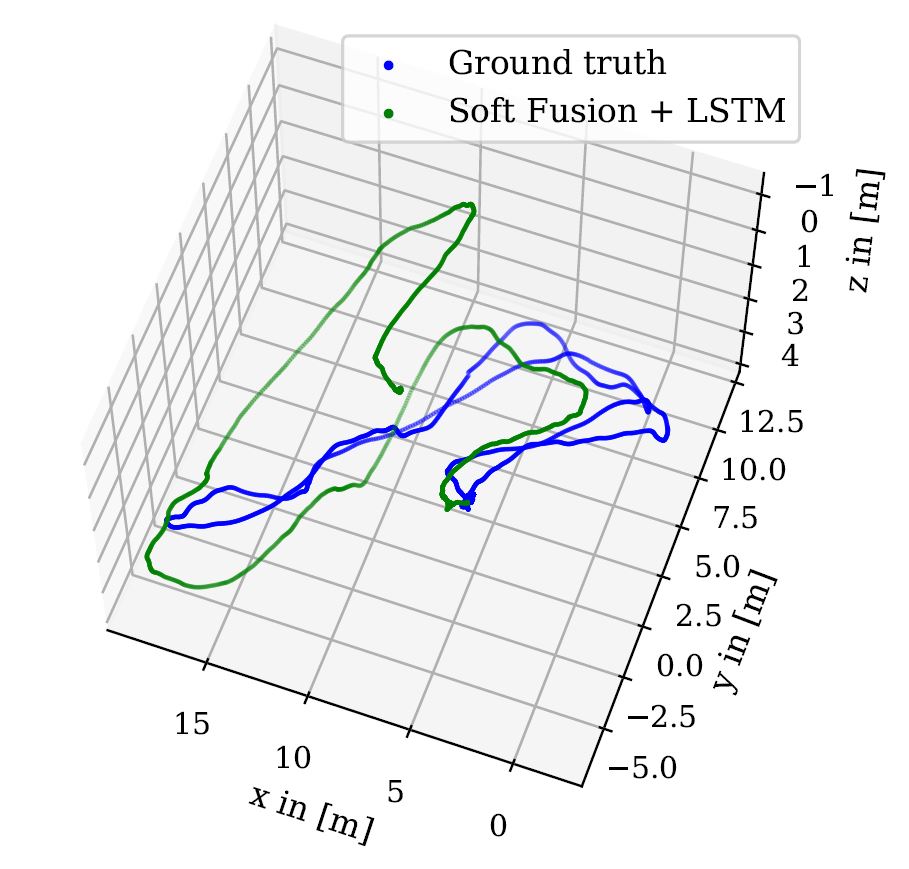}
    	\vspace{\vspacecaptione}
    	\subcaption{SSF~\cite{chen} + BiLSTM.}
    	\label{image_app_rpr_euroc_mh04_3}
        \vspace{\vspacefiguree}
    \end{minipage}
    \hfill
    \begin{minipage}[b]{\lene\linewidth}
        \centering
    	\includegraphics[width=1.0\linewidth]{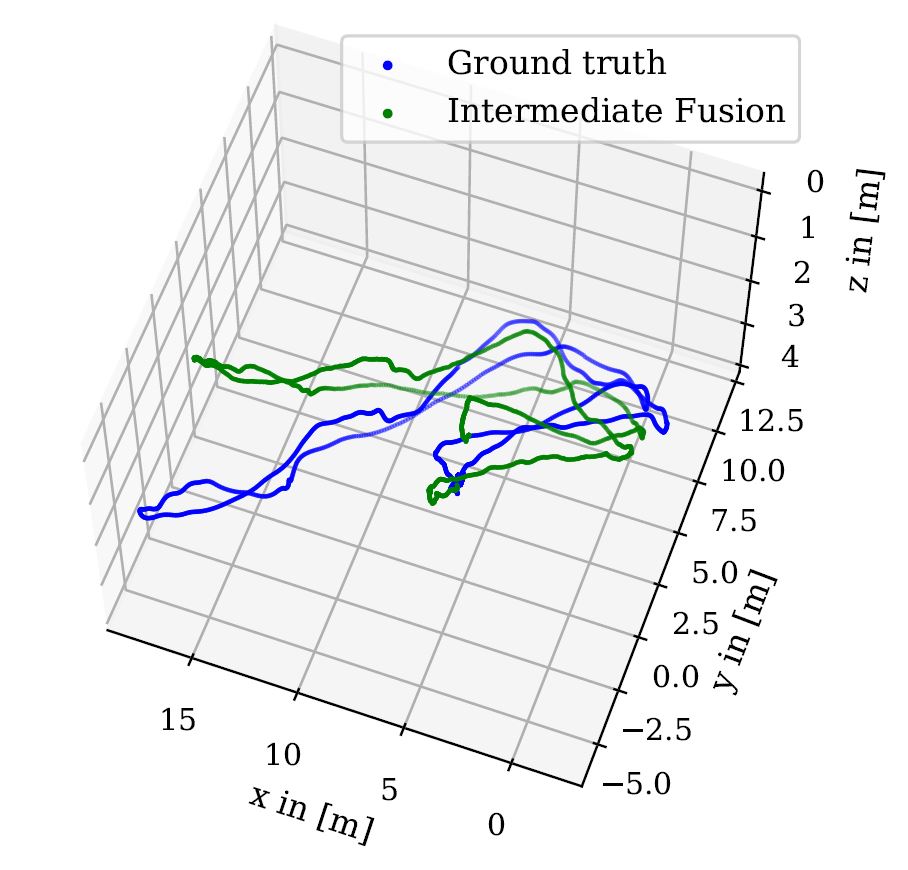}
    	\vspace{\vspacecaptione}
    	\subcaption{MMTM~\cite{joze}.}
    	\label{image_app_rpr_euroc_mh04_4}
        \vspace{\vspacefiguree}
    \end{minipage}
    \caption{$\text{RPR}_{\text{V}}$-$\text{RPR}_{\text{I}}$ fusion on EuRoC MAV~\cite{burri}: MH-04-difficult.}
    \label{image_app_rpr_euroc_mh04}
\end{figure*}

\begin{figure*}[t!]
	\centering
	\begin{minipage}[b]{\lene\linewidth}
        \centering
    	\includegraphics[width=1.0\linewidth]{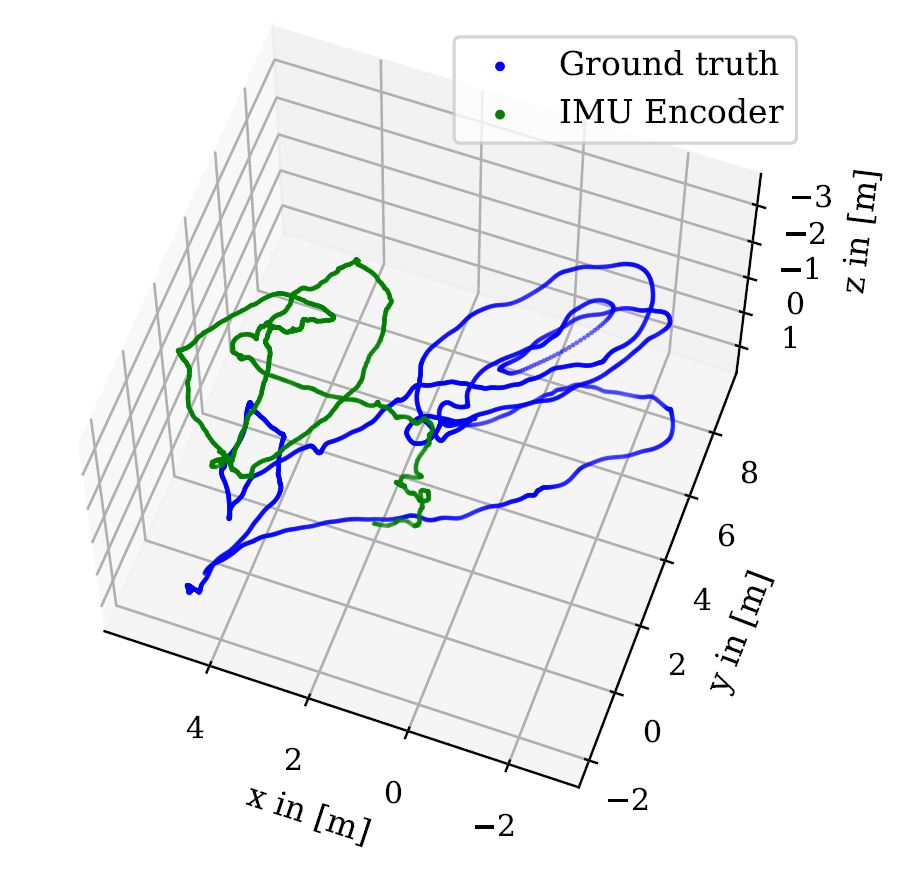}
    	\vspace{\vspacecaptione}
    	\subcaption{$\text{RPR}_{\text{I}}$: IMUNet~\cite{silva}.}
    	\label{image_app_rpr_euroc_mh01_1}
        \vspace{\vspacefiguree}
    \end{minipage}
    \hfill
    \begin{minipage}[b]{\lene\linewidth}
        \centering
    	\includegraphics[width=1.0\linewidth]{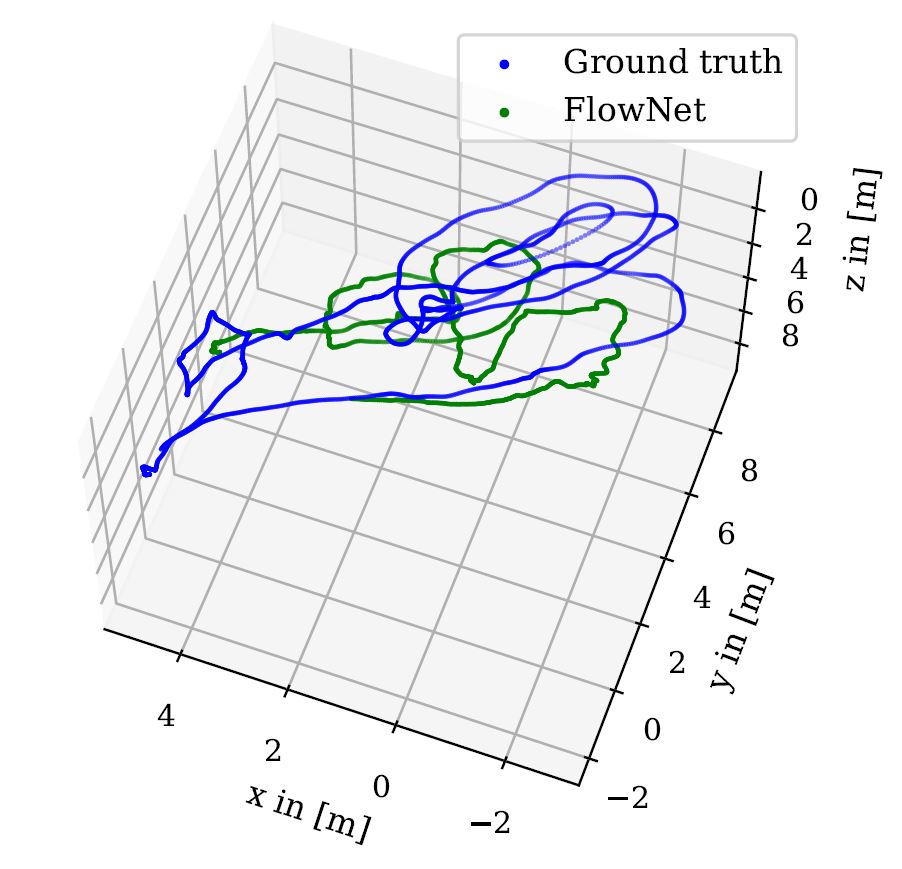}
    	\vspace{\vspacecaptione}
    	\subcaption{$\text{RPR}_{\text{V}}$: FlowNet~\cite{dosovitskiy}.}
    	\label{image_app_rpr_euroc_mh01_2}
        \vspace{\vspacefiguree}
    \end{minipage}
    \hfill
    \begin{minipage}[b]{\lene\linewidth}
        \centering
    	\includegraphics[width=1.0\linewidth]{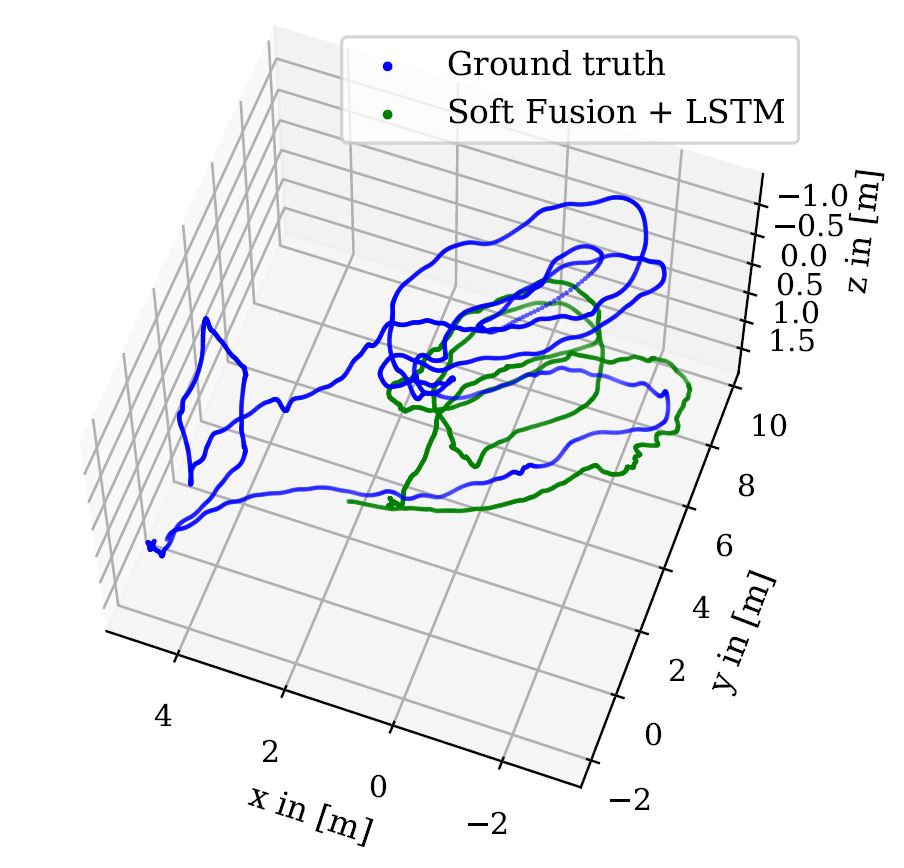}
    	\vspace{\vspacecaptione}
    	\subcaption{SSF~\cite{chen} + BiLSTM.}
    	\label{image_app_rpr_euroc_mh01_3}
        \vspace{\vspacefiguree}
    \end{minipage}
    \hfill
    \begin{minipage}[b]{\lene\linewidth}
        \centering
    	\includegraphics[width=1.0\linewidth]{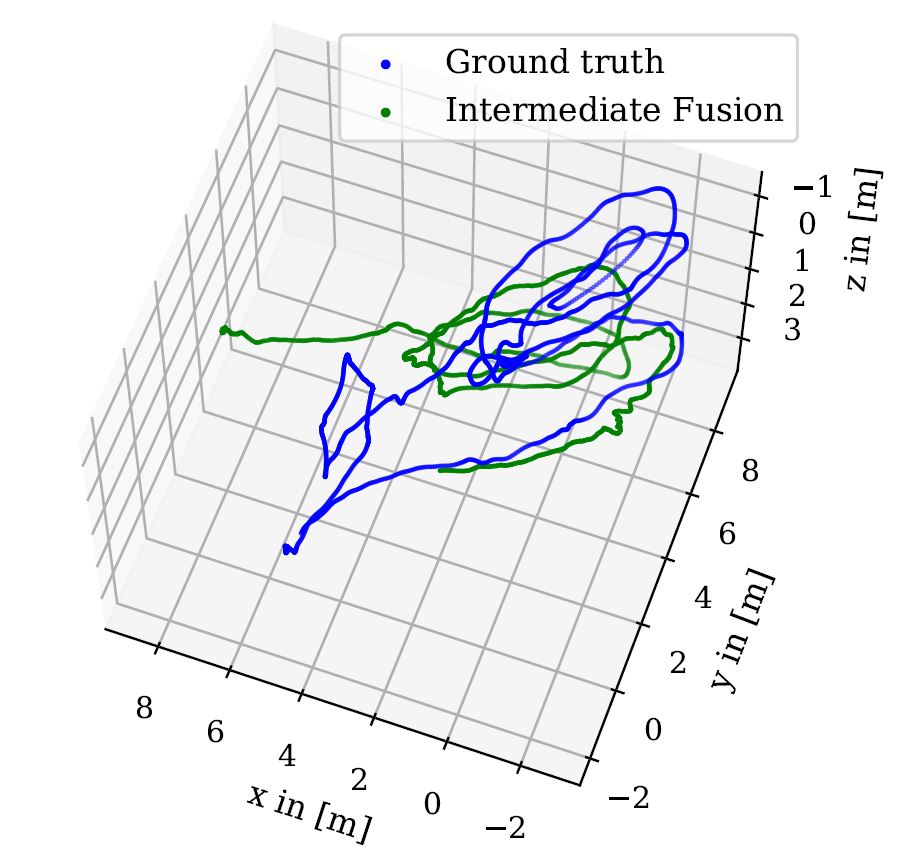}
    	\vspace{\vspacecaptione}
    	\subcaption{MMTM~\cite{joze}.}
    	\label{image_app_rpr_euroc_mh01_4}
        \vspace{\vspacefiguree}
    \end{minipage}
    \caption{$\text{RPR}_{\text{V}}$-$\text{RPR}_{\text{I}}$ fusion on EuRoC MAV~\cite{burri}: MH-01-easy.}
    \label{image_app_rpr_euroc_mh01}
\end{figure*}

\begin{figure*}[t!]
	\centering
	\begin{minipage}[b]{\lene\linewidth}
        \centering
    	\includegraphics[width=1.0\linewidth]{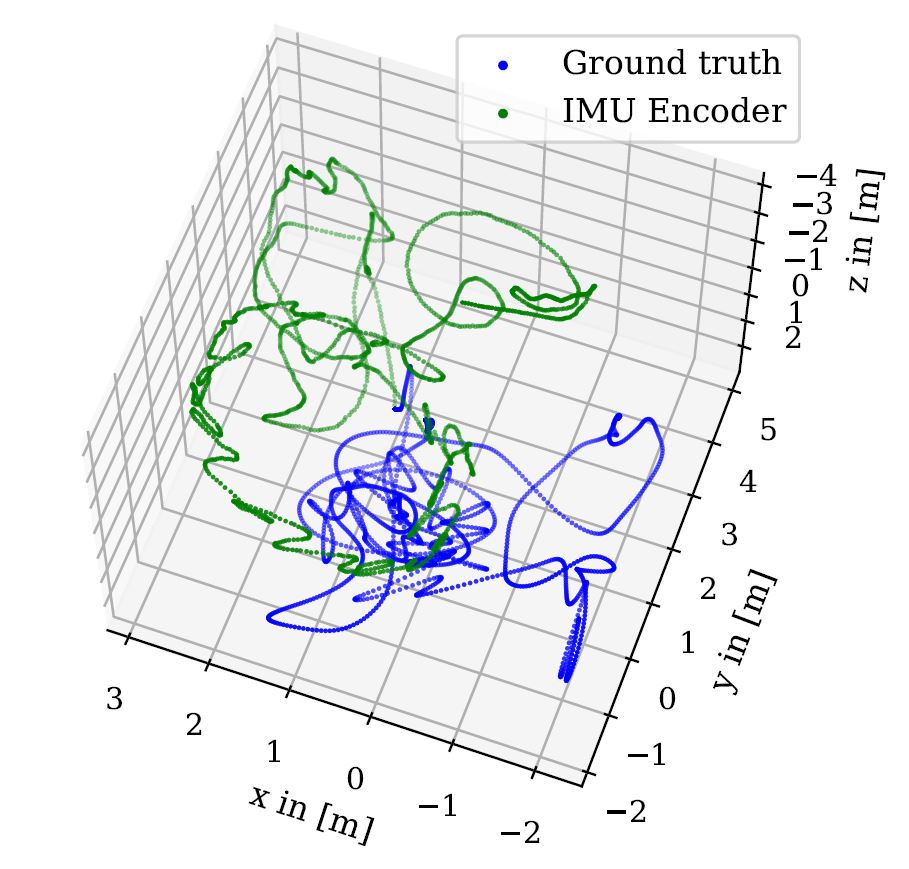}
    	\vspace{\vspacecaptione}
    	\subcaption{$\text{RPR}_{\text{I}}$: IMUNet~\cite{silva}.}
    	\label{image_app_rpr_euroc_v103_1}
        \vspace{\vspacefiguree}
    \end{minipage}
    \hfill
    \begin{minipage}[b]{\lene\linewidth}
        \centering
    	\includegraphics[width=1.0\linewidth]{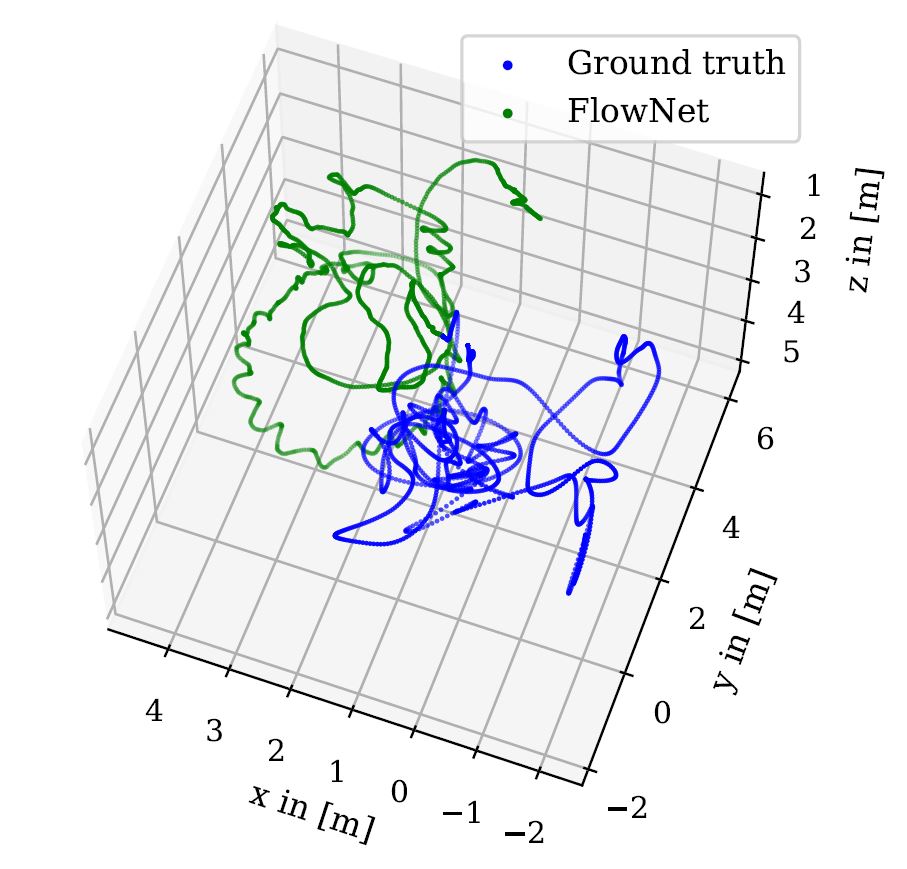}
    	\vspace{\vspacecaptione}
    	\subcaption{$\text{RPR}_{\text{V}}$: FlowNet~\cite{dosovitskiy}.}
    	\label{image_app_rpr_euroc_v103_2}
        \vspace{\vspacefiguree}
    \end{minipage}
    \hfill
    \begin{minipage}[b]{\lene\linewidth}
        \centering
    	\includegraphics[width=1.0\linewidth]{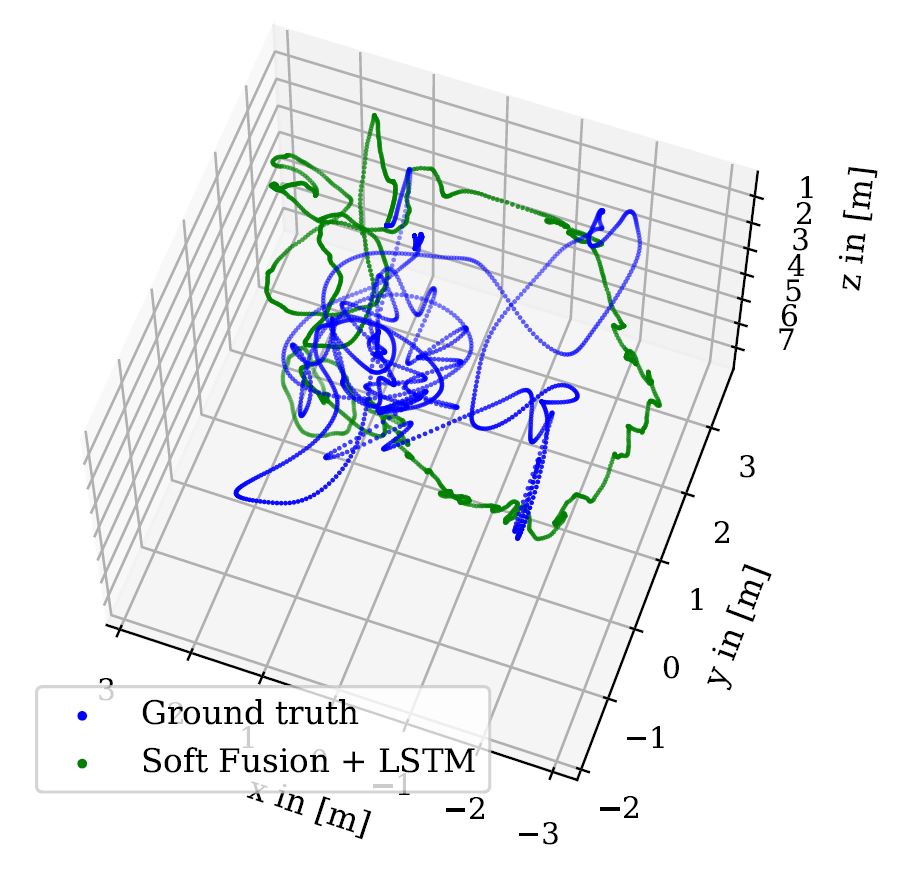}
    	\vspace{\vspacecaptione}
    	\subcaption{SSF~\cite{chen} + BiLSTM.}
    	\label{image_app_rpr_euroc_v103_3}
        \vspace{\vspacefiguree}
    \end{minipage}
    \hfill
    \begin{minipage}[b]{\lene\linewidth}
        \centering
    	\includegraphics[width=1.0\linewidth]{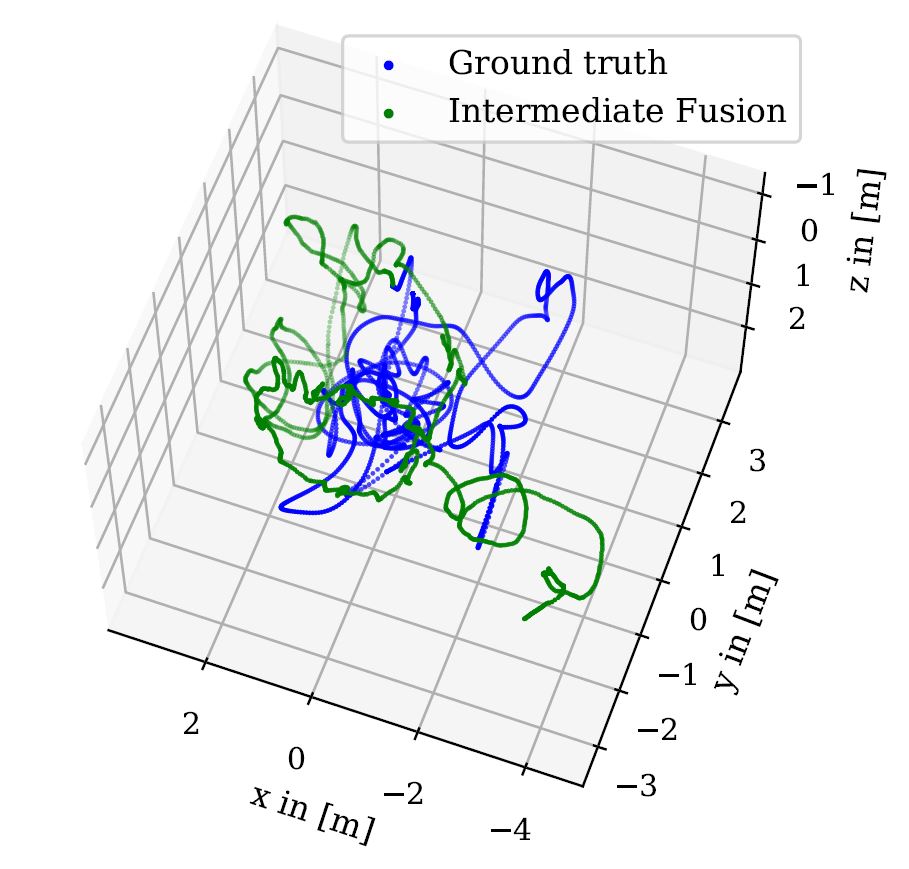}
    	\vspace{\vspacecaptione}
    	\subcaption{MMTM~\cite{joze}.}
    	\label{image_app_rpr_euroc_v103_4}
        \vspace{\vspacefiguree}
    \end{minipage}
    \caption{$\text{RPR}_{\text{V}}$-$\text{RPR}_{\text{I}}$ fusion on EuRoC MAV~\cite{burri}: V1-03-difficult.}
    \label{image_app_rpr_euroc_v103}
\end{figure*}

\begin{figure*}[t!]
	\centering
	\begin{minipage}[b]{\lene\linewidth}
        \centering
    	\includegraphics[width=1.0\linewidth]{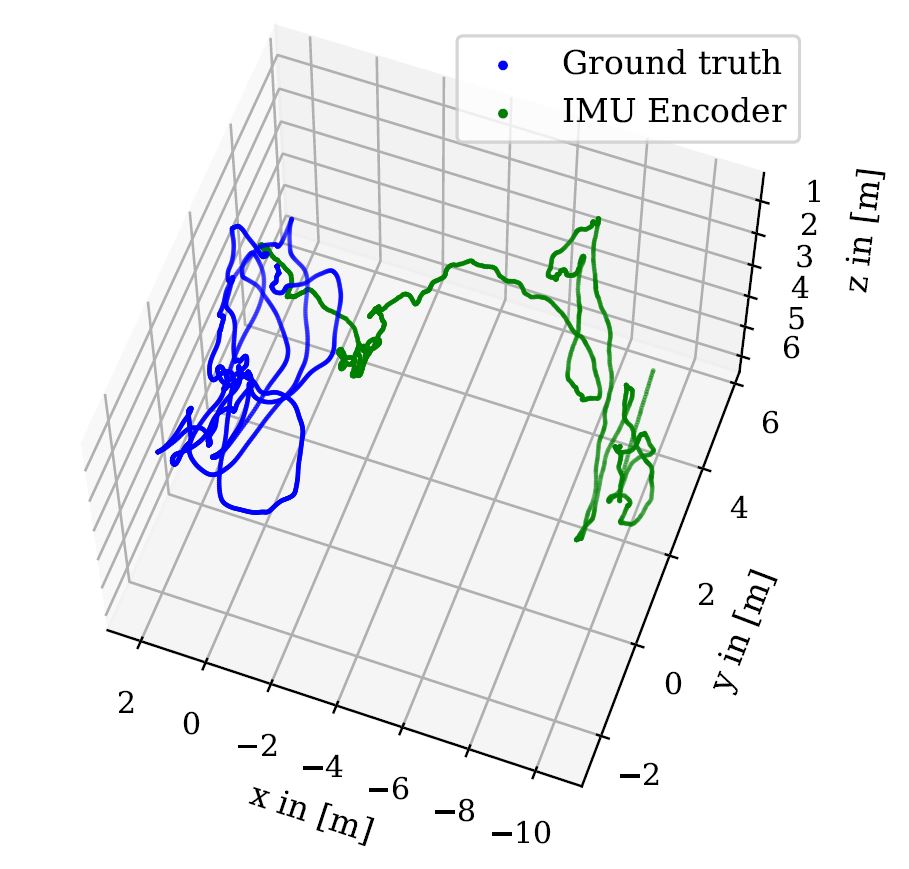}
    	\vspace{\vspacecaptione}
    	\subcaption{$\text{RPR}_{\text{I}}$: IMUNet~\cite{silva}.}
    	\label{image_app_rpr_euroc_v101_1}
        \vspace{\vspacefiguree}
    \end{minipage}
    \hfill
    \begin{minipage}[b]{\lene\linewidth}
        \centering
    	\includegraphics[width=1.0\linewidth]{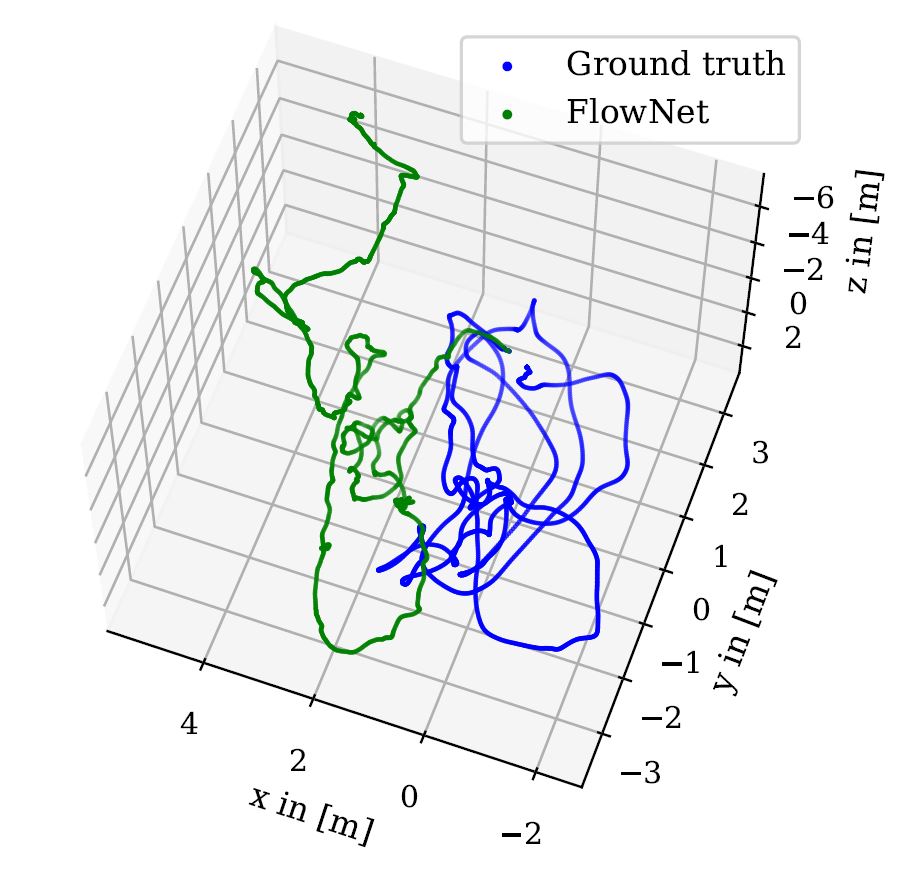}
    	\vspace{\vspacecaptione}
    	\subcaption{$\text{RPR}_{\text{V}}$: FlowNet~\cite{dosovitskiy}.}
    	\label{image_app_rpr_euroc_v101_2}
        \vspace{\vspacefiguree}
    \end{minipage}
    \hfill
    \begin{minipage}[b]{\lene\linewidth}
        \centering
    	\includegraphics[width=1.0\linewidth]{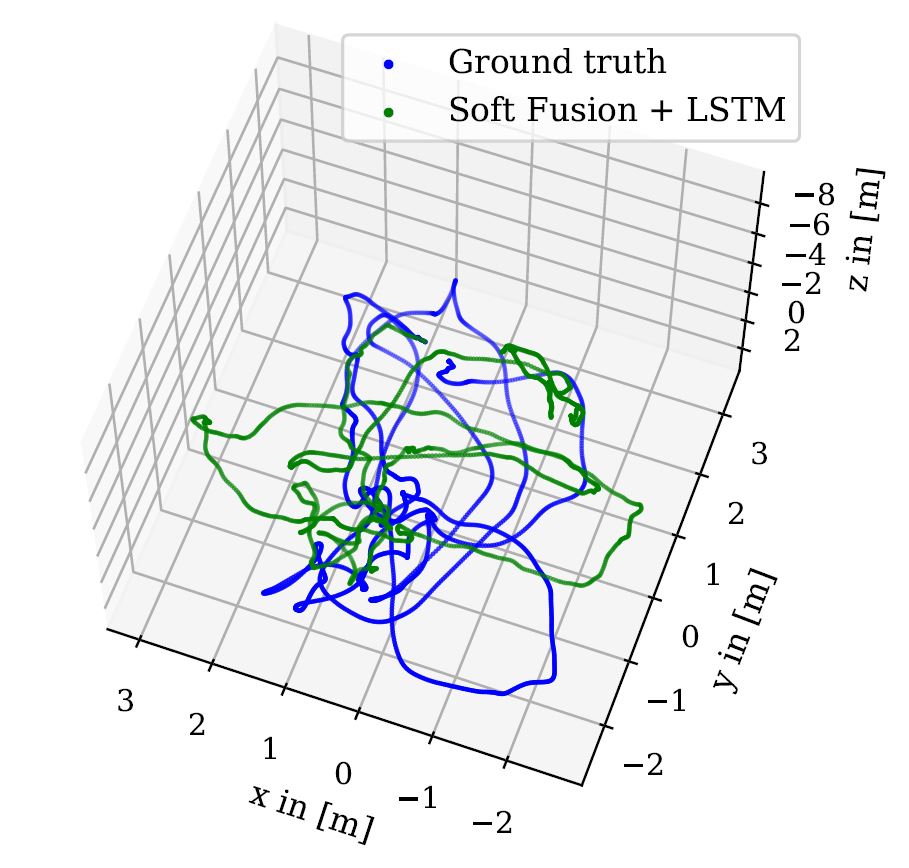}
    	\vspace{\vspacecaptione}
    	\subcaption{SSF~\cite{chen} + BiLSTM.}
    	\label{image_app_rpr_euroc_v101_3}
        \vspace{\vspacefiguree}
    \end{minipage}
    \hfill
    \begin{minipage}[b]{\lene\linewidth}
        \centering
    	\includegraphics[width=1.0\linewidth]{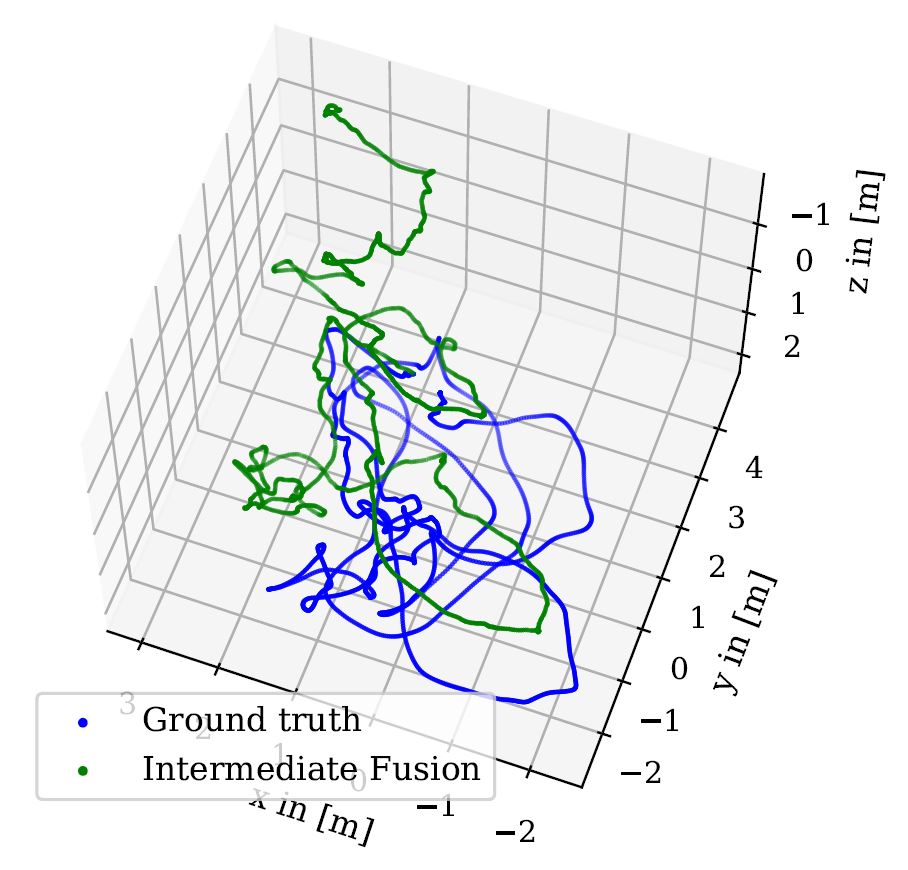}
    	\vspace{\vspacecaptione}
    	\subcaption{MMTM~\cite{joze}.}
    	\label{image_app_rpr_euroc_v101_4}
        \vspace{\vspacefiguree}
    \end{minipage}
    \caption{$\text{RPR}_{\text{V}}$-$\text{RPR}_{\text{I}}$ fusion on EuRoC MAV~\cite{burri}: V1-01-easy.}
    \label{image_app_rpr_euroc_v101}
\end{figure*}

\begin{figure*}[t!]
	\centering
	\begin{minipage}[b]{\lene\linewidth}
        \centering
    	\includegraphics[width=1.0\linewidth]{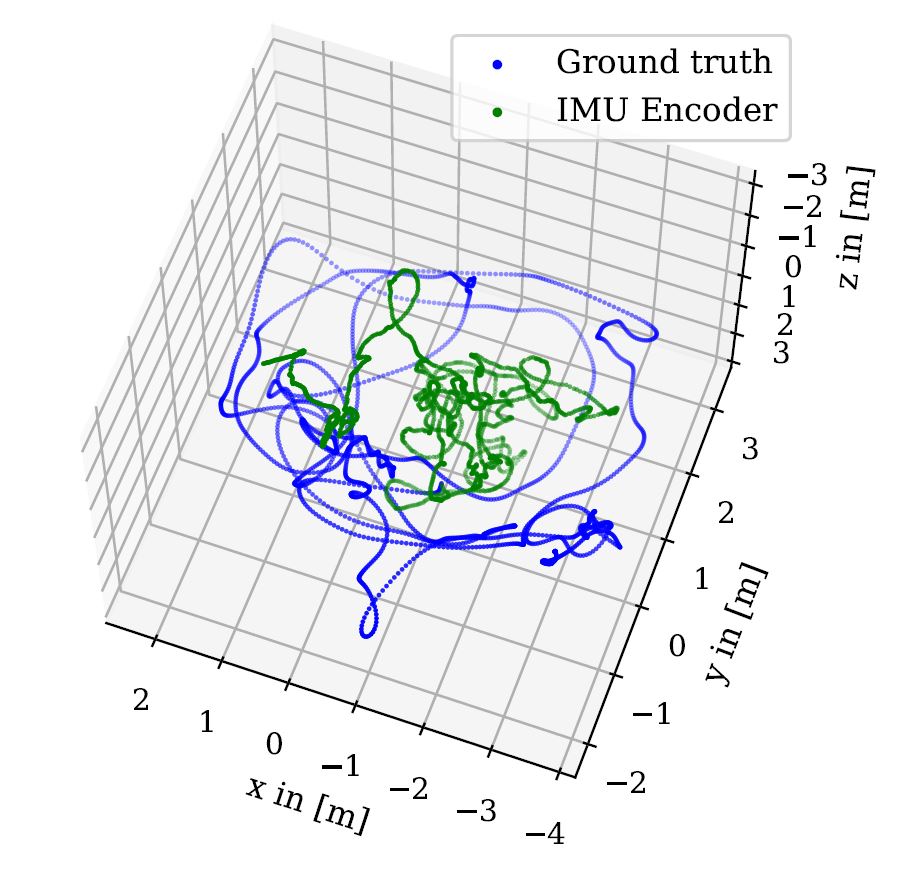}
    	\vspace{\vspacecaptione}
    	\subcaption{$\text{RPR}_{\text{I}}$: IMUNet~\cite{silva}.}
    	\label{image_app_rpr_euroc_v202_1}
        \vspace{\vspacefiguree}
    \end{minipage}
    \hfill
    \begin{minipage}[b]{\lene\linewidth}
        \centering
    	\includegraphics[width=1.0\linewidth]{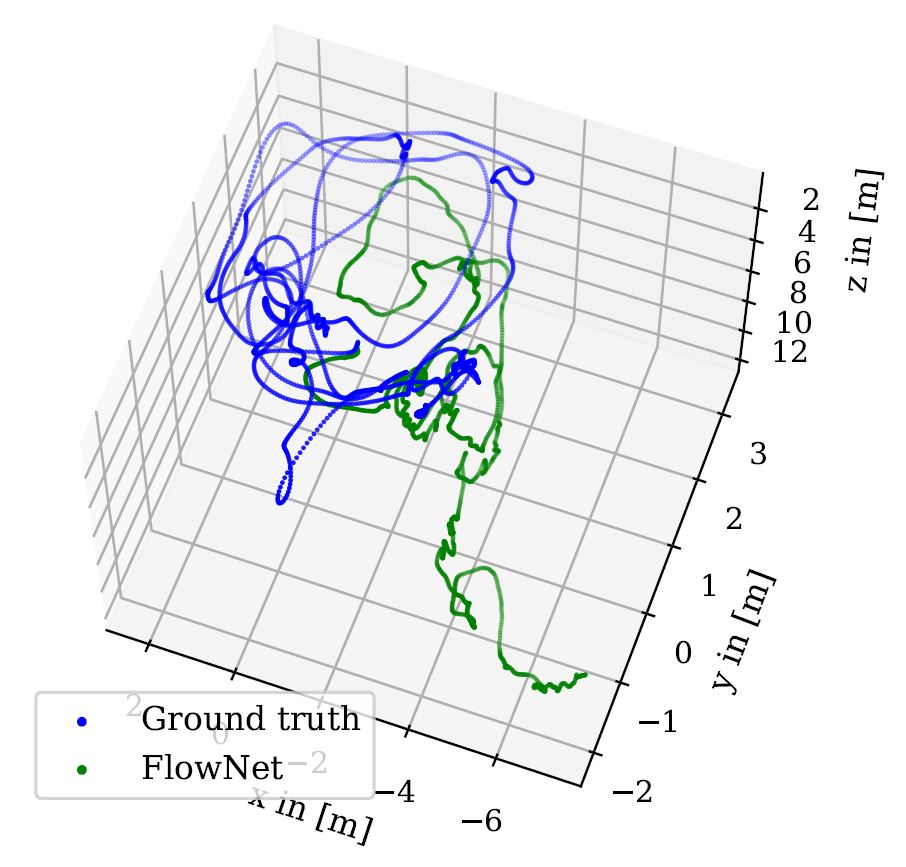}
    	\vspace{\vspacecaptione}
    	\subcaption{$\text{RPR}_{\text{V}}$: FlowNet~\cite{dosovitskiy}.}
    	\label{image_app_rpr_euroc_v202_2}
        \vspace{\vspacefiguree}
    \end{minipage}
    \hfill
    \begin{minipage}[b]{\lene\linewidth}
        \centering
    	\includegraphics[width=1.0\linewidth]{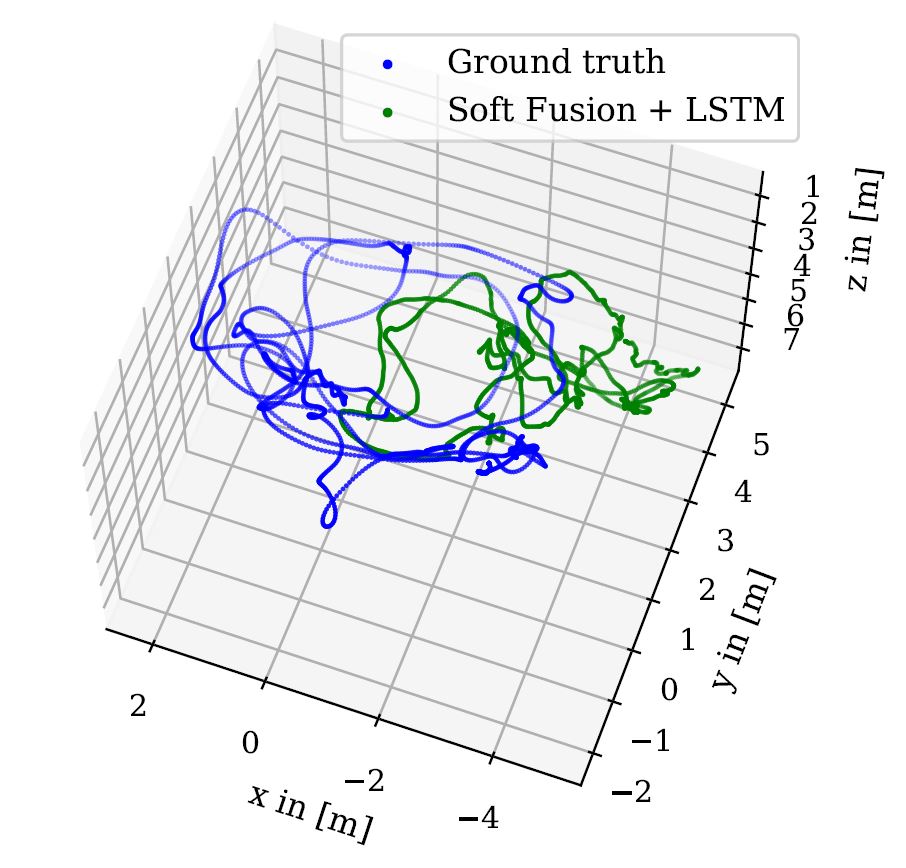}
    	\vspace{\vspacecaptione}
    	\subcaption{SSF~\cite{chen} + BiLSTM.}
    	\label{image_app_rpr_euroc_v202_3}
        \vspace{\vspacefiguree}
    \end{minipage}
    \hfill
    \begin{minipage}[b]{\lene\linewidth}
        \centering
    	\includegraphics[width=1.0\linewidth]{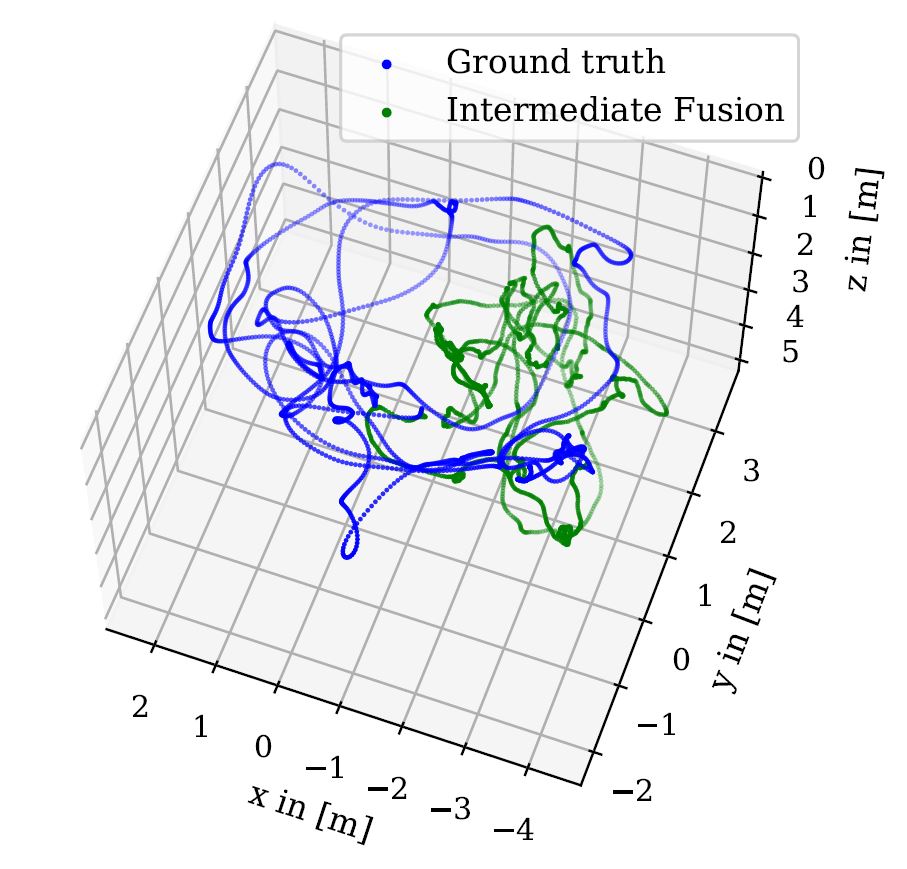}
    	\vspace{\vspacecaptione}
    	\subcaption{MMTM~\cite{joze}.}
    	\label{image_app_rpr_euroc_v202_4}
        \vspace{\vspacefiguree}
    \end{minipage}
    \caption{$\text{RPR}_{\text{V}}$-$\text{RPR}_{\text{I}}$ fusion on EuRoC MAV~\cite{burri}: V2-02-Medium.}
    \label{image_app_rpr_euroc_v202}
\end{figure*}

\begin{figure*}[t!]
	\centering
	\begin{minipage}[b]{\lene\linewidth}
        \centering
    	\includegraphics[width=1.0\linewidth]{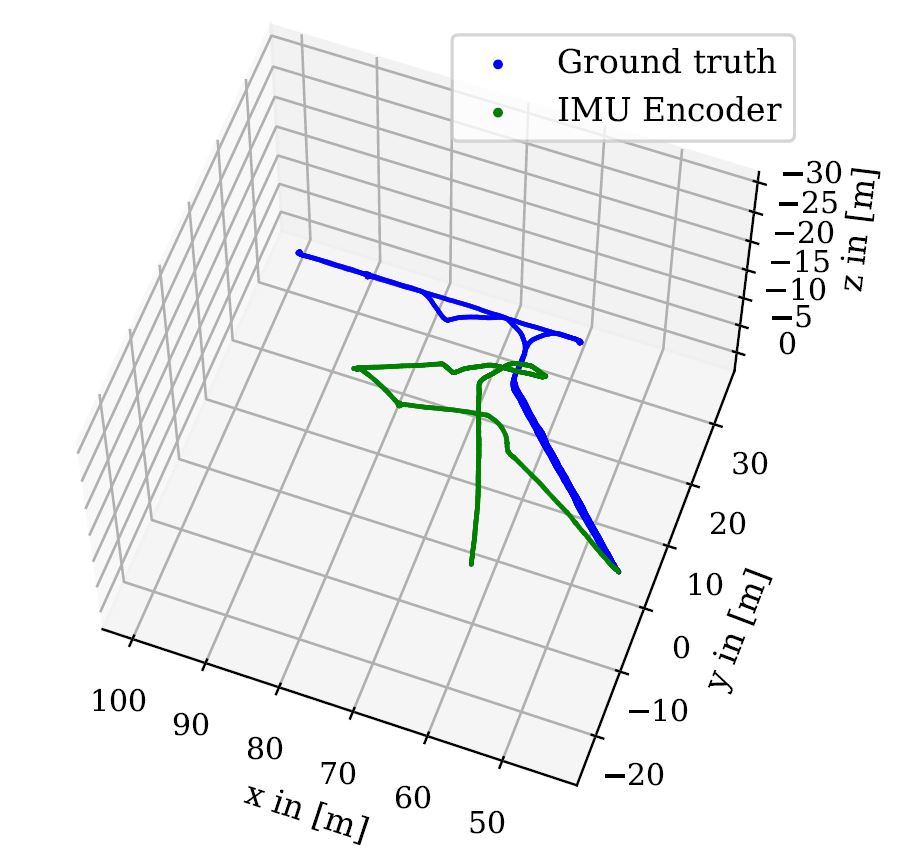}
    	\vspace{\vspacecaptione}
    	\subcaption{$\text{RPR}_{\text{I}}$: IMUNet~\cite{silva}.}
    	\label{image_app_rpr_penncosy_bf_1}
        \vspace{\vspacefiguree}
    \end{minipage}
    \hfill
    \begin{minipage}[b]{\lene\linewidth}
        \centering
    	\includegraphics[width=1.0\linewidth]{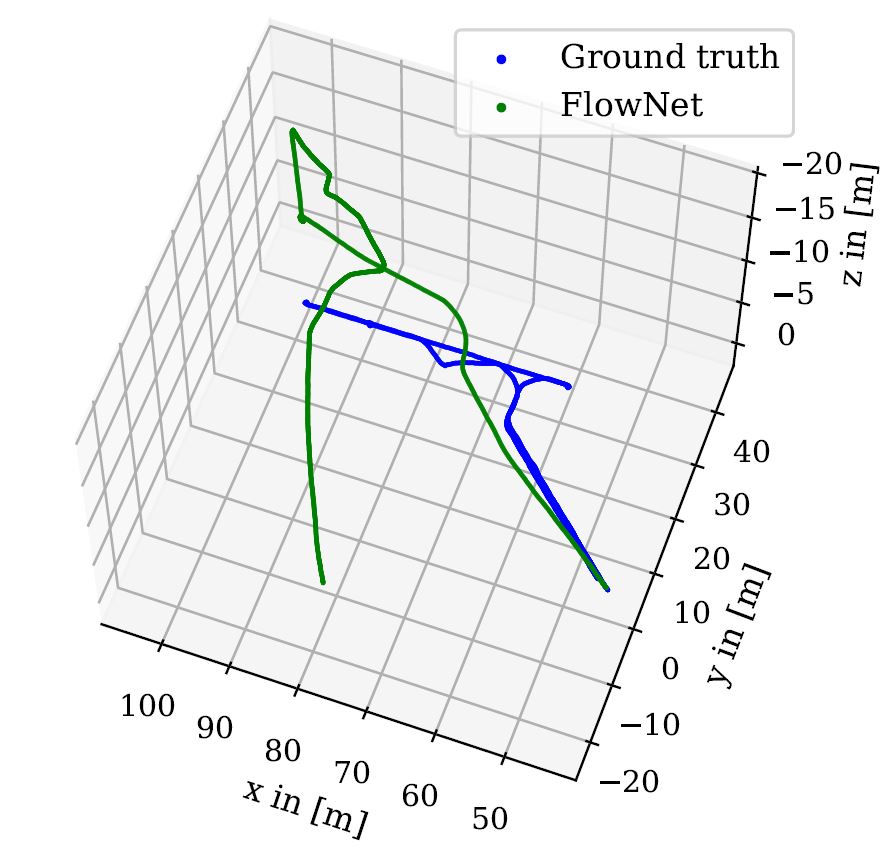}
    	\vspace{\vspacecaptione}
    	\subcaption{$\text{RPR}_{\text{V}}$: FlowNet~\cite{dosovitskiy}.}
    	\label{image_app_rpr_penncosy_bf_2}
        \vspace{\vspacefiguree}
    \end{minipage}
    \hfill
    \begin{minipage}[b]{\lene\linewidth}
        \centering
    	\includegraphics[width=1.0\linewidth]{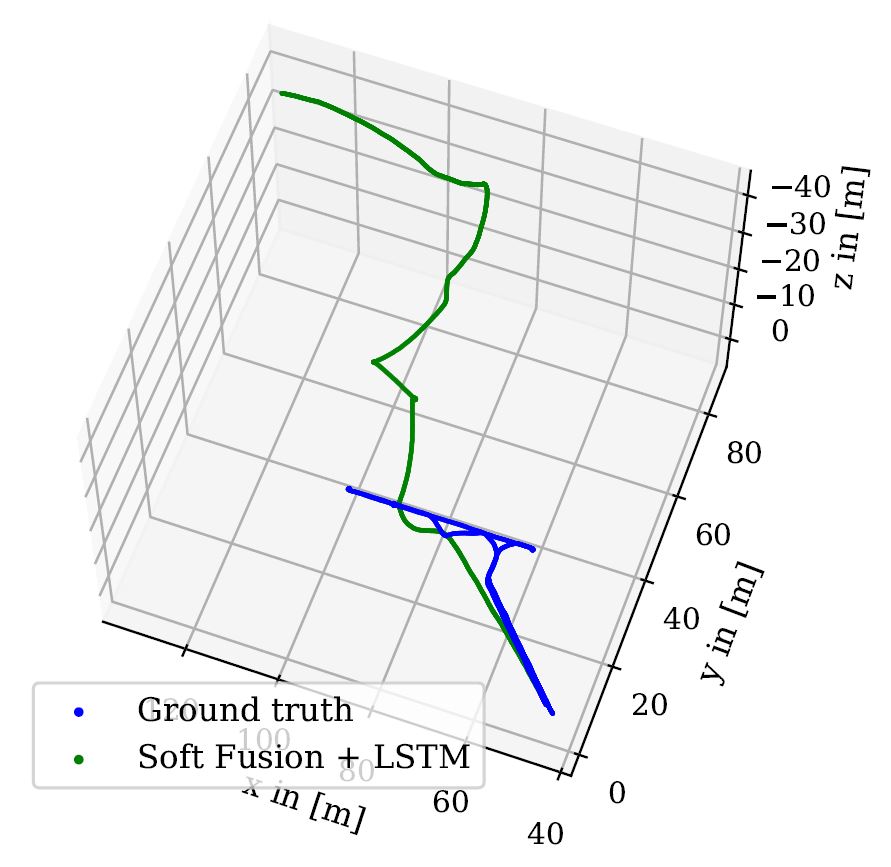}
    	\vspace{\vspacecaptione}
    	\subcaption{SSF~\cite{chen} + BiLSTM.}
    	\label{image_app_rpr_penncosy_bf_3}
        \vspace{\vspacefiguree}
    \end{minipage}
    \hfill
    \begin{minipage}[b]{\lene\linewidth}
        \centering
    	\includegraphics[width=1.0\linewidth]{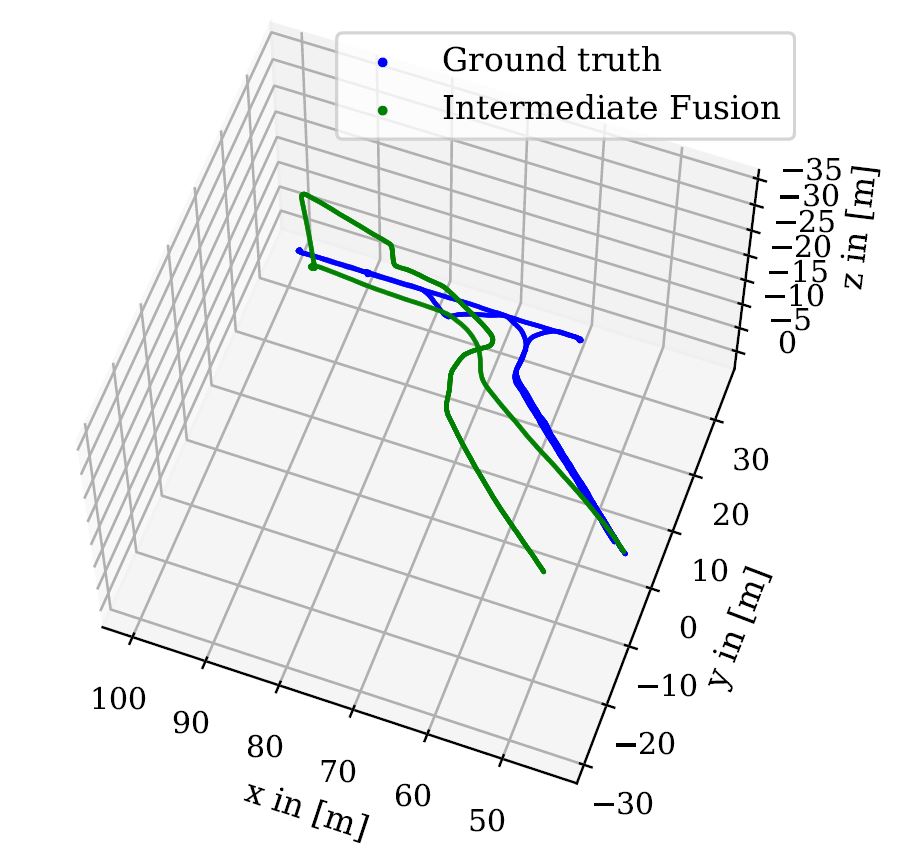}
    	\vspace{\vspacecaptione}
    	\subcaption{MMTM~\cite{joze}.}
    	\label{image_app_rpr_penncosy_bf_4}
        \vspace{\vspacefiguree}
    \end{minipage}
    \caption{$\text{RPR}_{\text{V}}$-$\text{RPR}_{\text{I}}$ fusion on PennCOSYVIO~\cite{pfrommer}: BF.}
    \label{image_app_rpr_penncosy_bf}
\end{figure*}

\begin{figure*}[t!]
	\centering
	\begin{minipage}[b]{\lene\linewidth}
        \centering
    	\includegraphics[width=1.0\linewidth]{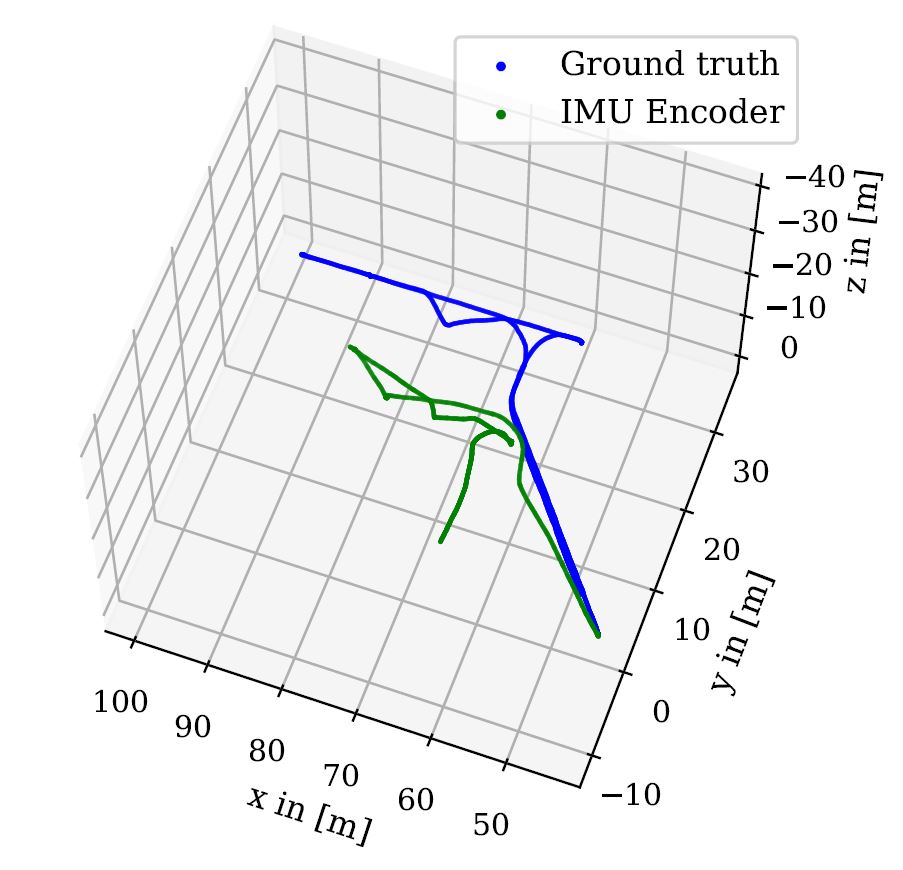}
    	\vspace{\vspacecaptione}
    	\subcaption{$\text{RPR}_{\text{I}}$: IMUNet~\cite{silva}.}
    	\label{image_app_rpr_penncosy_bs_1}
        \vspace{\vspacefiguree}
    \end{minipage}
    \hfill
    \begin{minipage}[b]{\lene\linewidth}
        \centering
    	\includegraphics[width=1.0\linewidth]{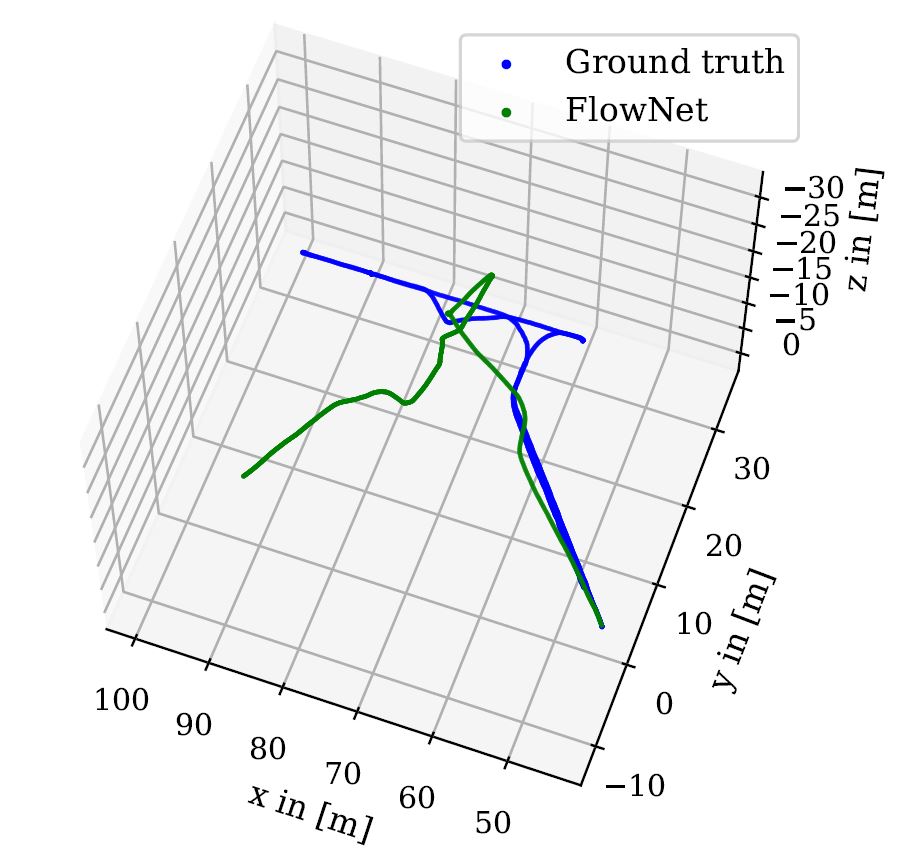}
    	\vspace{\vspacecaptione}
    	\subcaption{$\text{RPR}_{\text{V}}$: FlowNet~\cite{dosovitskiy}.}
    	\label{image_app_rpr_penncosy_bs_2}
        \vspace{\vspacefiguree}
    \end{minipage}
    \hfill
    \begin{minipage}[b]{\lene\linewidth}
        \centering
    	\includegraphics[width=1.0\linewidth]{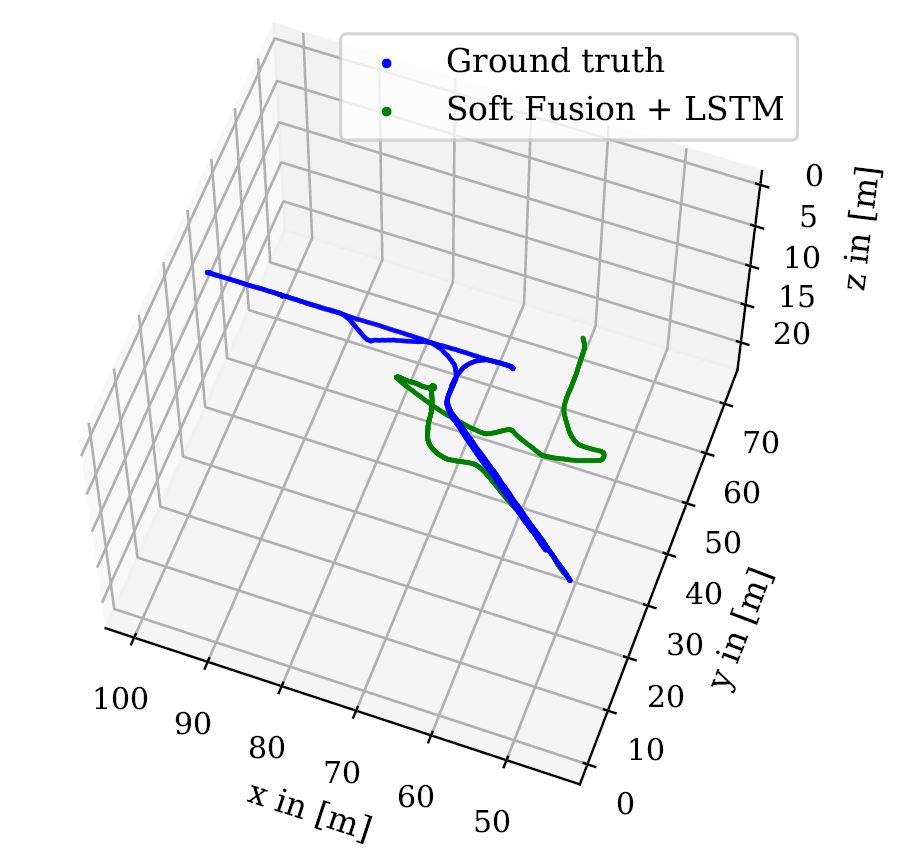}
    	\vspace{\vspacecaptione}
    	\subcaption{SSF~\cite{chen} + BiLSTM.}
    	\label{image_app_rpr_penncosy_bs_3}
        \vspace{\vspacefiguree}
    \end{minipage}
    \hfill
    \begin{minipage}[b]{\lene\linewidth}
        \centering
    	\includegraphics[width=1.0\linewidth]{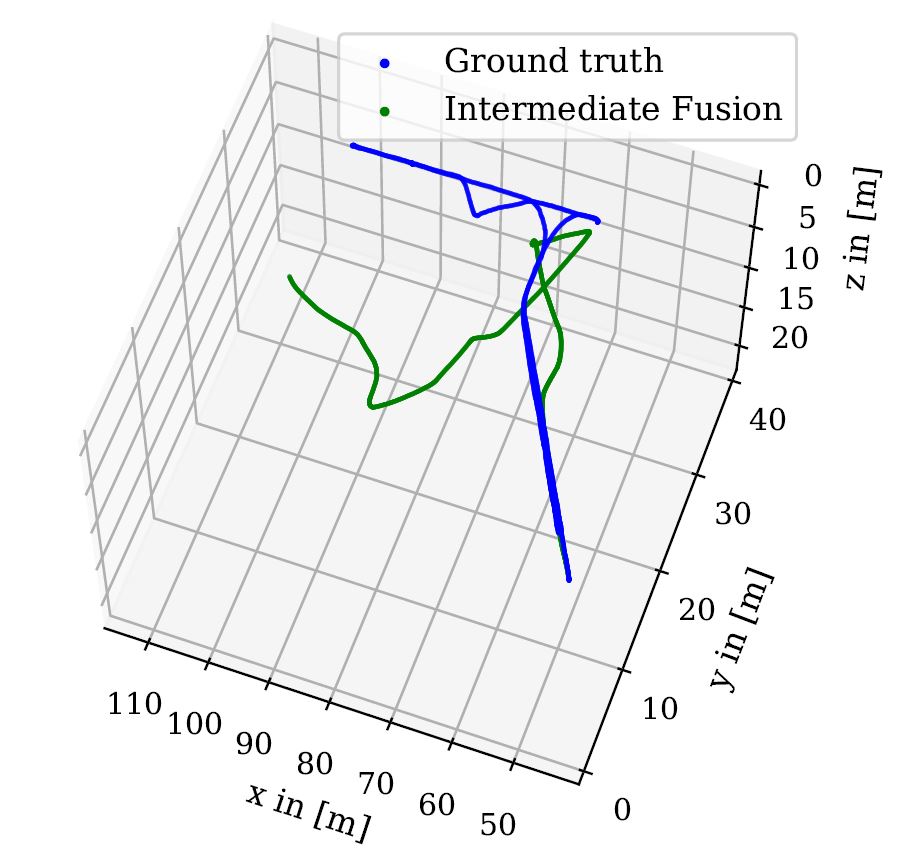}
    	\vspace{\vspacecaptione}
    	\subcaption{MMTM~\cite{joze}.}
    	\label{image_app_rpr_penncosy_bs_4}
        \vspace{\vspacefiguree}
    \end{minipage}
    \caption{$\text{RPR}_{\text{V}}$-$\text{RPR}_{\text{I}}$ fusion on PennCOSYVIO~\cite{pfrommer}: BS.}
    \label{image_app_rpr_penncosy_bs}
\end{figure*}

\begin{figure*}[t!]
	\centering
	\begin{minipage}[b]{\lene\linewidth}
        \centering
    	\includegraphics[width=1.0\linewidth]{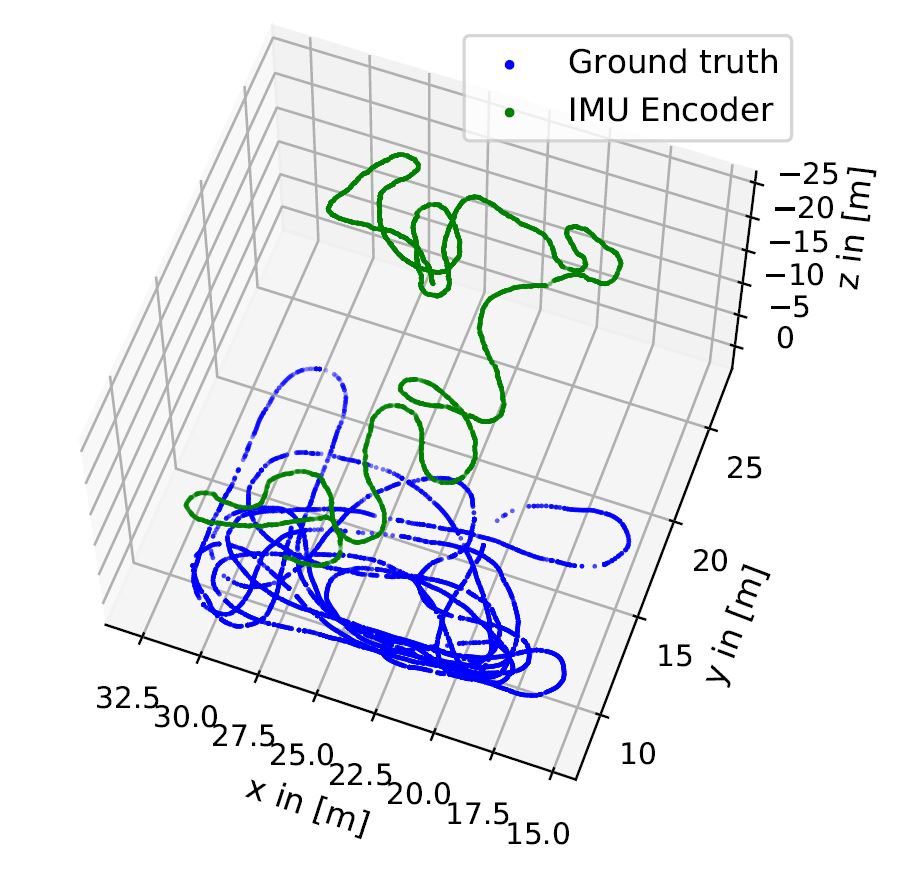}
    	\vspace{\vspacecaptione}
    	\subcaption{$\text{RPR}_{\text{I}}$: IMUNet~\cite{silva}.}
    	\label{image_app_rpr_industry1_1}
        \vspace{\vspacefiguree}
    \end{minipage}
    \hfill
    \begin{minipage}[b]{\lene\linewidth}
        \centering
    	\includegraphics[width=1.0\linewidth]{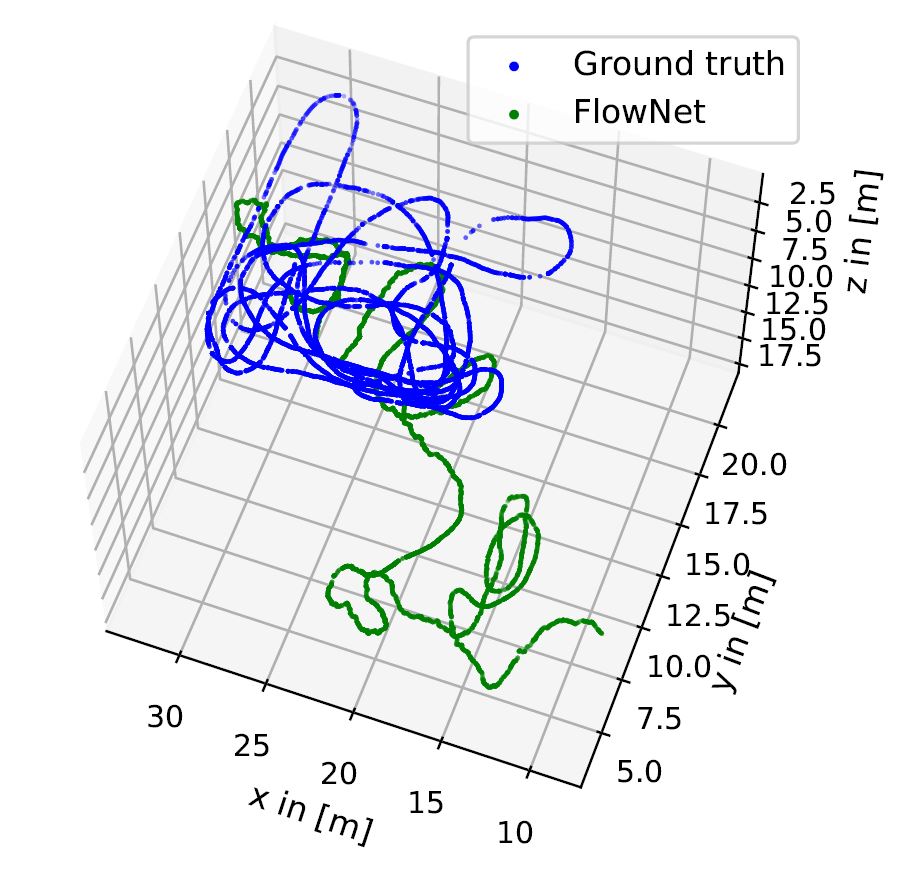}
    	\vspace{\vspacecaptione}
    	\subcaption{$\text{RPR}_{\text{V}}$: FlowNet~\cite{dosovitskiy}.}
    	\label{image_app_rpr_industry1_2}
        \vspace{\vspacefiguree}
    \end{minipage}
    \hfill
    \begin{minipage}[b]{\lene\linewidth}
        \centering
    	\includegraphics[width=1.0\linewidth]{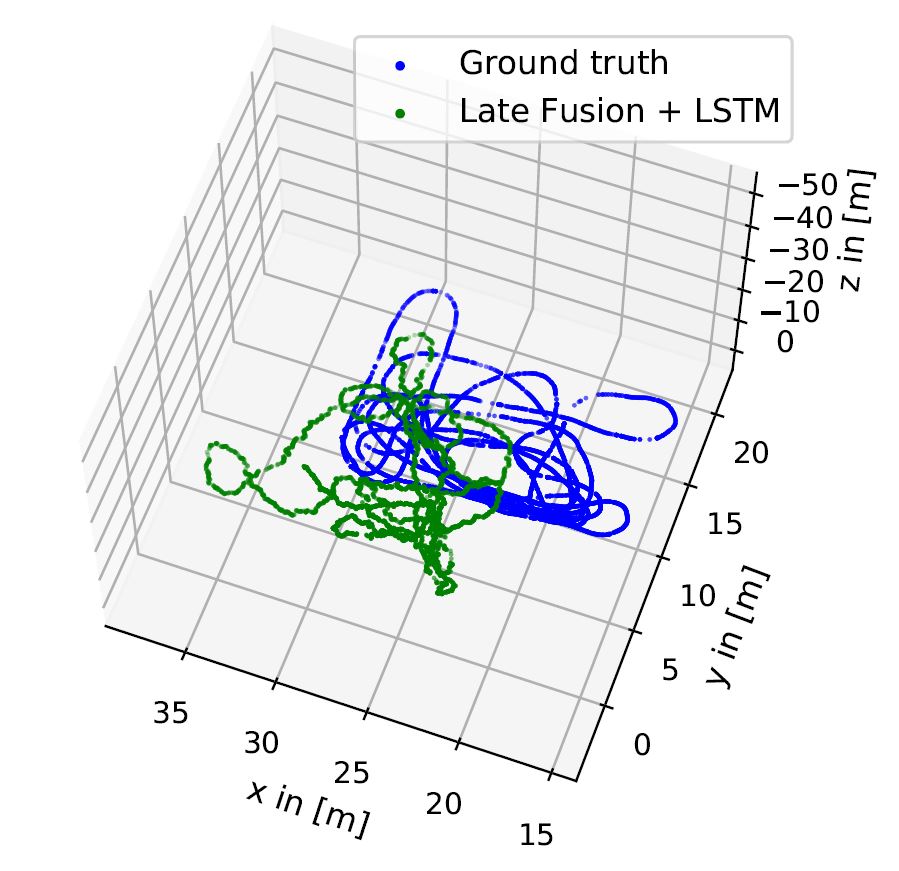}
    	\vspace{\vspacecaptione}
    	\subcaption{SSF~\cite{chen} + BiLSTM.}
    	\label{image_app_rpr_industry1_3}
        \vspace{\vspacefiguree}
    \end{minipage}
    \hfill
    \begin{minipage}[b]{\lene\linewidth}
        \centering
    	\includegraphics[width=1.0\linewidth]{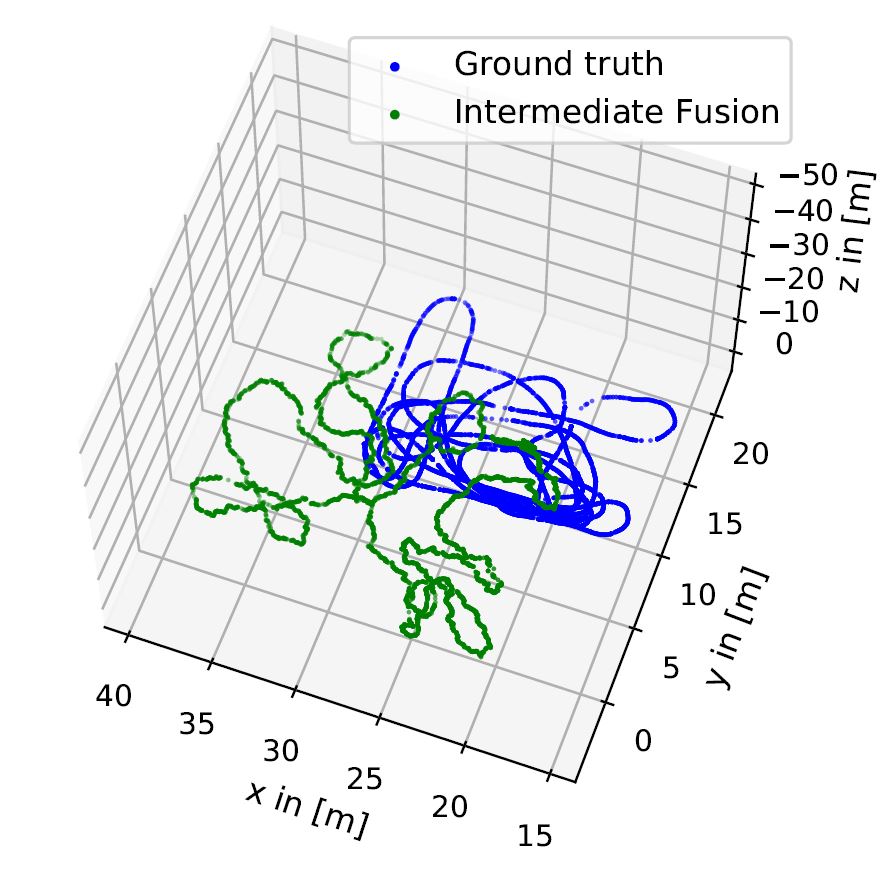}
    	\vspace{\vspacecaptione}
    	\subcaption{MMTM~\cite{joze}.}
    	\label{image_app_rpr_industry1_4}
        \vspace{\vspacefiguree}
    \end{minipage}
    \caption{$\text{RPR}_{\text{V}}$-$\text{RPR}_{\text{I}}$ fusion on IndustryVI: Testing Sequence 1.}
    \label{image_app_rpr_industry1}
\end{figure*}

\begin{figure*}[t!]
	\centering
	\begin{minipage}[b]{\lene\linewidth}
        \centering
    	\includegraphics[width=1.0\linewidth]{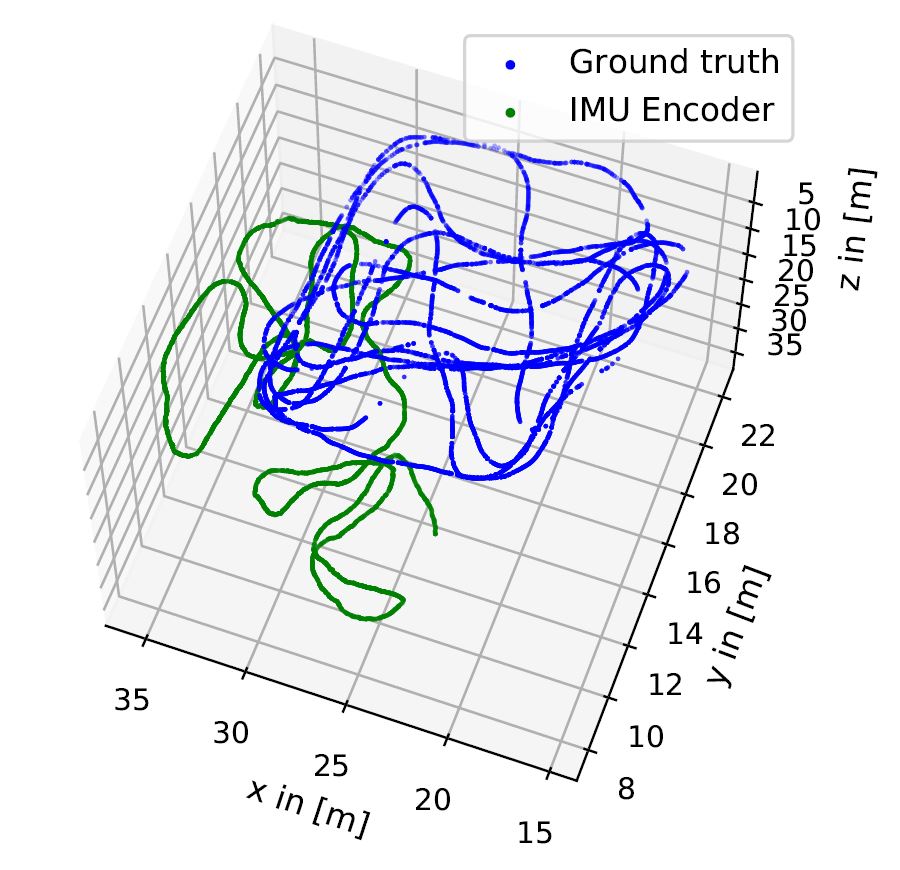}
    	\vspace{\vspacecaptione}
    	\subcaption{$\text{RPR}_{\text{I}}$: IMUNet~\cite{silva}.}
    	\label{image_app_rpr_industry2_1}
        \vspace{\vspacefiguree}
    \end{minipage}
    \hfill
    \begin{minipage}[b]{\lene\linewidth}
        \centering
    	\includegraphics[width=1.0\linewidth]{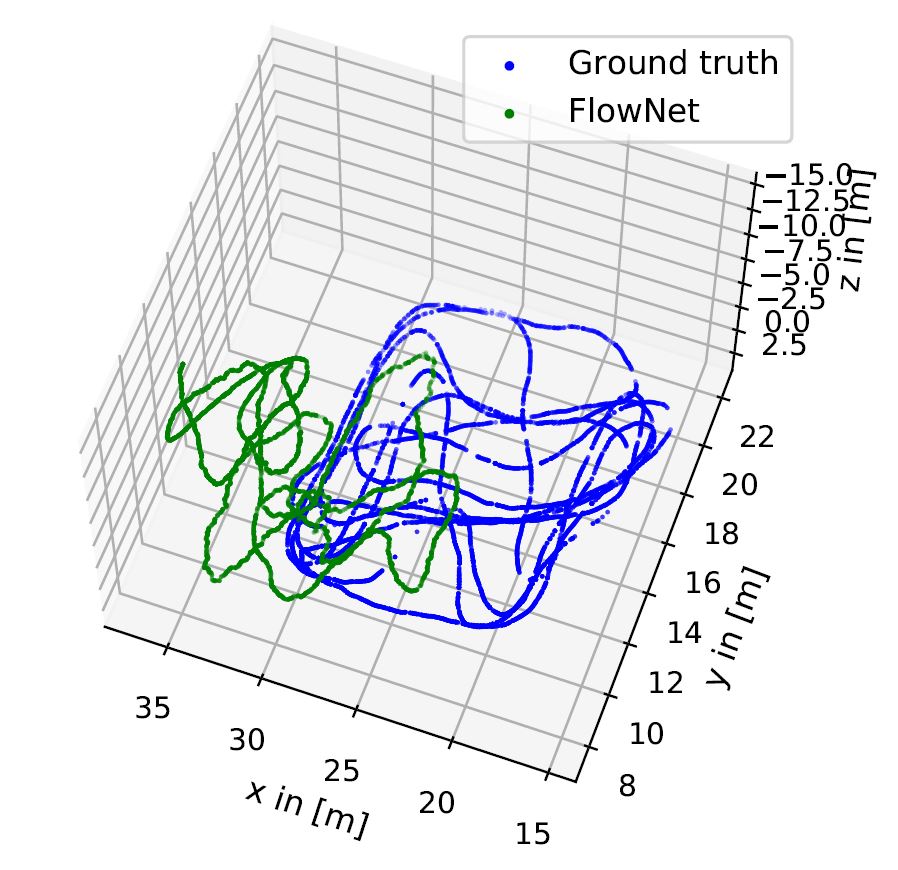}
    	\vspace{\vspacecaptione}
    	\subcaption{$\text{RPR}_{\text{V}}$: FlowNet~\cite{dosovitskiy}.}
    	\label{image_app_rpr_industry2_2}
        \vspace{\vspacefiguree}
    \end{minipage}
    \hfill
    \begin{minipage}[b]{\lene\linewidth}
        \centering
    	\includegraphics[width=1.0\linewidth]{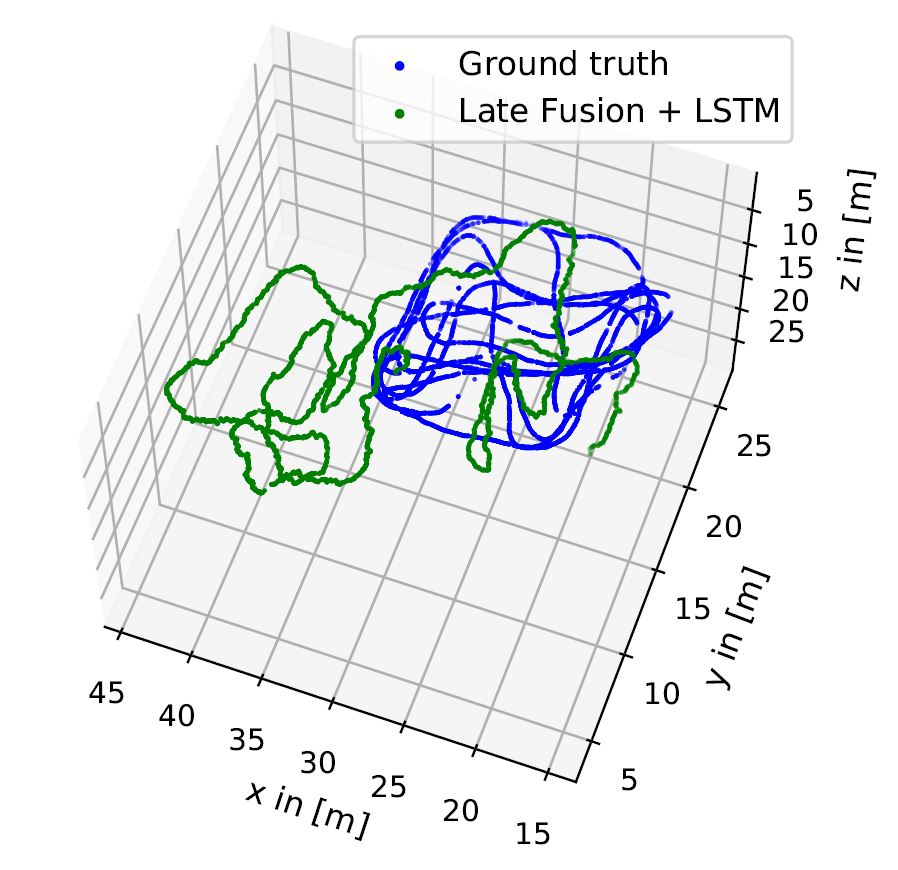}
    	\vspace{\vspacecaptione}
    	\subcaption{SSF~\cite{chen} + BiLSTM.}
    	\label{image_app_rpr_industry2_3}
        \vspace{\vspacefiguree}
    \end{minipage}
    \hfill
    \begin{minipage}[b]{\lene\linewidth}
        \centering
    	\includegraphics[width=1.0\linewidth]{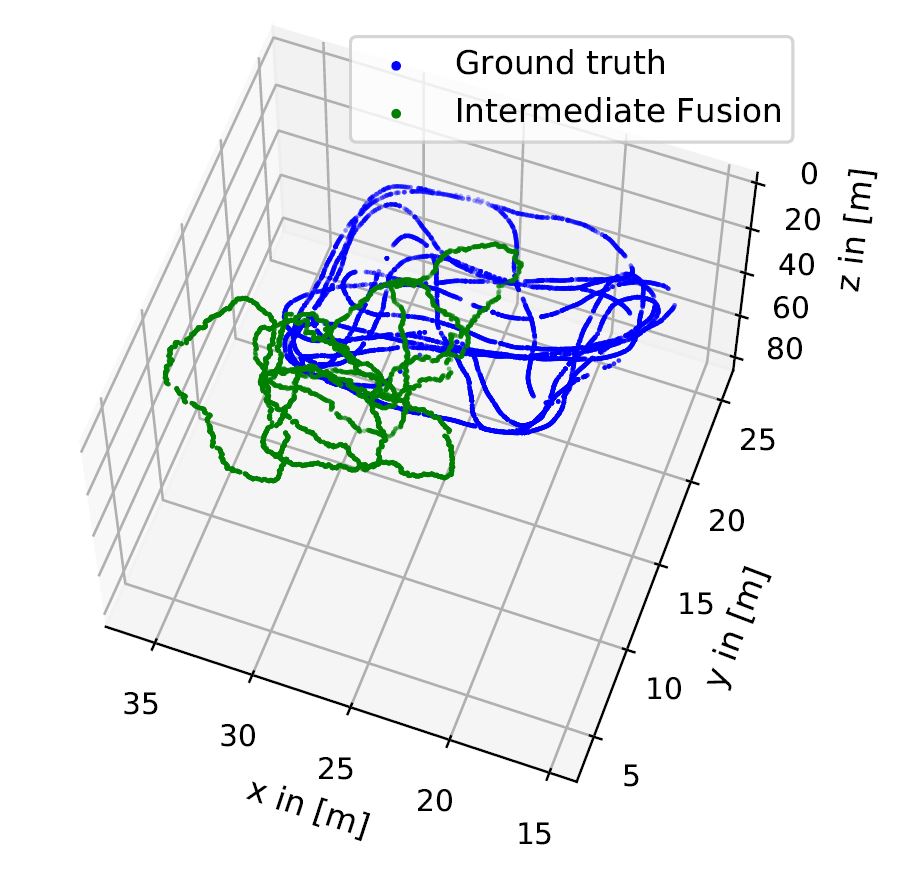}
    	\vspace{\vspacecaptione}
    	\subcaption{MMTM~\cite{joze}.}
    	\label{image_app_rpr_industry2_4}
        \vspace{\vspacefiguree}
    \end{minipage}
    \caption{$\text{RPR}_{\text{V}}$-$\text{RPR}_{\text{I}}$ fusion on IndustryVI: Testing Sequence 2.}
    \label{image_app_rpr_industry2}
\end{figure*}

\clearpage

\begin{table*}[t!]
\begin{center}
    \caption{We summarize the number of trainable parameters for the baseline and fusion models. While ``F'' indicates the layer with MMTM fusion (see Section~\ref{section_intermediate_fusion}), "--" indicates no fusion. While the IMUNet is a small network with 961,031 parameters, PoseNet (7.7 million) and FlowNet (16.7 million) are significantly larger in size and require more computing time. The number of parameters decreases by combining the $\text{APR}_{\text{V}}$ or $\text{RPR}_{\text{V}}$ models with the $\text{RPR}_{\text{I}}$ model for the late fusion, as the last model layers are removed. By adding BiLSTM layers, the model size increases.}
    \label{table_no_of_params}
    \begin{tabular}{ p{3.7cm} | p{1.1cm} | p{1.0cm} }
    \multicolumn{1}{c|}{\textbf{Method}} & \multicolumn{1}{c|}{\textbf{\# Parameters}} &  \multicolumn{1}{c}{\textbf{\# Parameters}} \\ \hline
    \textbf{Baseline Models} & & \\
    $\text{APR}_{\text{V}}$ (PoseNet) & \multicolumn{1}{r|}{7,713,447} & \multicolumn{1}{c}{-} \\
    $\text{RPR}_{\text{I}}$ (IMUNet) & \multicolumn{1}{r|}{961,031} & \multicolumn{1}{r}{961,031} \\ 
    $\text{RPR}_{\text{V}}$ (FlowNet) & \multicolumn{1}{c|}{-} & \multicolumn{1}{r}{16,715,520} \\ \hline \hline
    & \multicolumn{1}{c|}{\textbf{\# Parameters}} & \multicolumn{1}{c}{\textbf{\# Parameters}} \\ 
    \multicolumn{1}{c|}{\textbf{Method}} & \multicolumn{1}{c|}{\textbf{$\text{APR}_{\text{V}}$-$\text{RPR}_{\text{I}}$}} & \multicolumn{1}{c}{\textbf{$\text{RPR}_{\text{V}}$-$\text{RPR}_{\text{I}}$}} \\ \hline
    \textbf{Fusion Models} & & \\
    MapNet+PGO & \multicolumn{1}{r|}{7,713,447} & \multicolumn{1}{c}{-} \\
    $\text{APR}_{\text{V}}$-$\text{RPR}_{\text{I}}$+PGO & \multicolumn{1}{r|}{7,304,238} & \multicolumn{1}{c}{-} \\
    Late Fusion (concat) & \multicolumn{1}{r|}{6,726,830} & \multicolumn{1}{r}{15,654,727} \\
    Late Fusion (concat) + BiLSTM & \multicolumn{1}{r|}{7,304,238} & \multicolumn{1}{r}{16,315,079} \\
    Late Fusion (SSF) & \multicolumn{1}{r|}{6,792,622} & \multicolumn{1}{r}{15,671,239}\\ 
    Late Fusion (SSF) + BiLSTM & \multicolumn{1}{r|}{7,320,750} & \multicolumn{1}{r}{16,331,591} \\ 
    \textbf{Intermediate Fusion:} & & \\
    \multicolumn{1}{r|}{MMTM (F/--/--)} & \multicolumn{1}{r|}{8,489,614} & \multicolumn{1}{r}{16,906,055} \\
    \multicolumn{1}{r|}{MMTM (--/F/--)} & \multicolumn{1}{r|}{8,489,614} & \multicolumn{1}{r}{16,906,055} \\
    \multicolumn{1}{r|}{MMTM (--/--/F)} & \multicolumn{1}{r|}{8,944,558} & \multicolumn{1}{r}{17,955,399} \\
    \multicolumn{1}{r|}{MMTM (F/F/--)} & \multicolumn{1}{r|}{9,674,990} & \multicolumn{1}{r}{17,497,031} \\
    \multicolumn{1}{r|}{MMTM (--/F/F)} & \multicolumn{1}{r|}{10,129,934} & \multicolumn{1}{r}{18,546,375} \\
    \multicolumn{1}{r|}{MMTM (F/--/F)} & \multicolumn{1}{r|}{10,129,934} & \multicolumn{1}{r}{18,546,375} \\
    \multicolumn{1}{r|}{MMTM (F/F/F)} & \multicolumn{1}{r|}{11,315,310} & \multicolumn{1}{r}{19,137,351} \\
    Auxiliary (non-linear) &  \multicolumn{1}{r|}{7,320,769} & \multicolumn{1}{c}{-} \\
    Auxiliary (convolutional) & \multicolumn{1}{r|}{7,320,948} & \multicolumn{1}{c}{-} \\
    Bayesian network &  \multicolumn{1}{r|}{7,320,750} &  \multicolumn{1}{r}{16,322,453} \\
    \end{tabular}
\end{center}
\end{table*}

\begin{figure*}[!t]
	\centering
	\begin{minipage}[b]{0.495\linewidth}
        \centering
    	\includegraphics[width=1.0\linewidth]{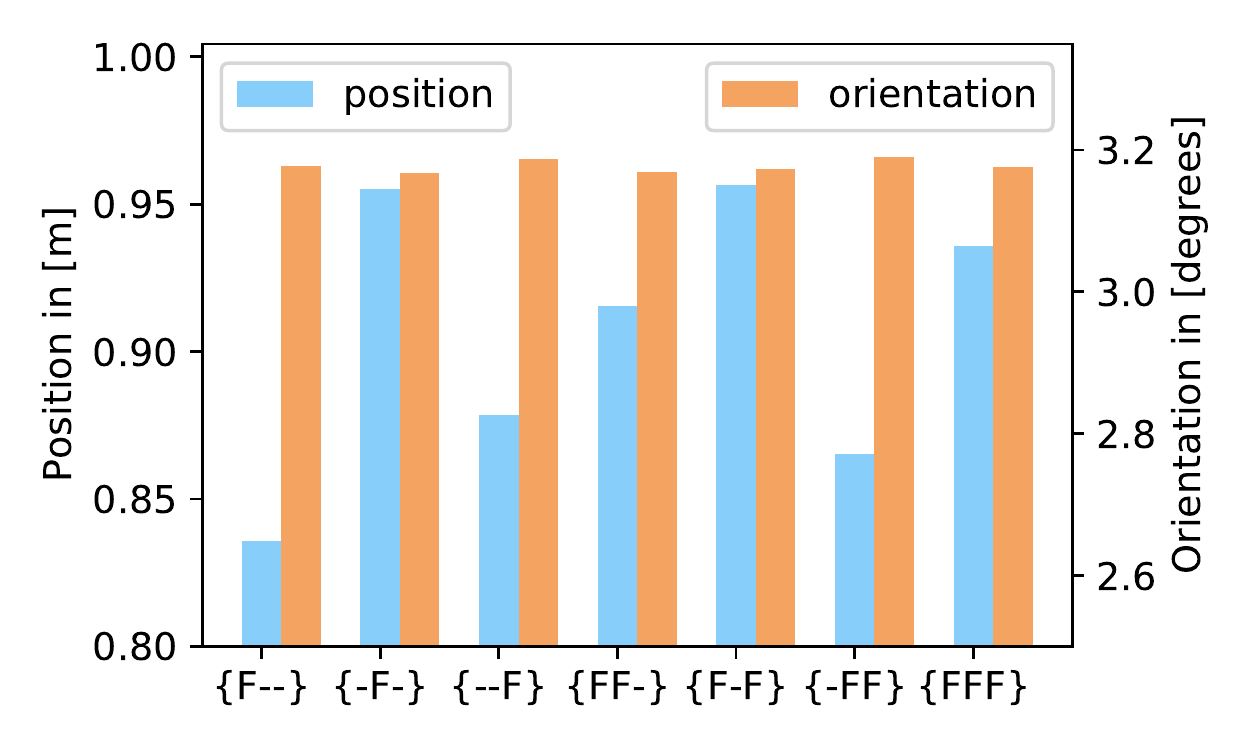}
    	\subcaption{MH-02-easy.}
    	\label{image_mmtm_comb1}
    \end{minipage}
    \hfill
	\begin{minipage}[b]{0.495\linewidth}
        \centering
    	\includegraphics[width=1.0\linewidth]{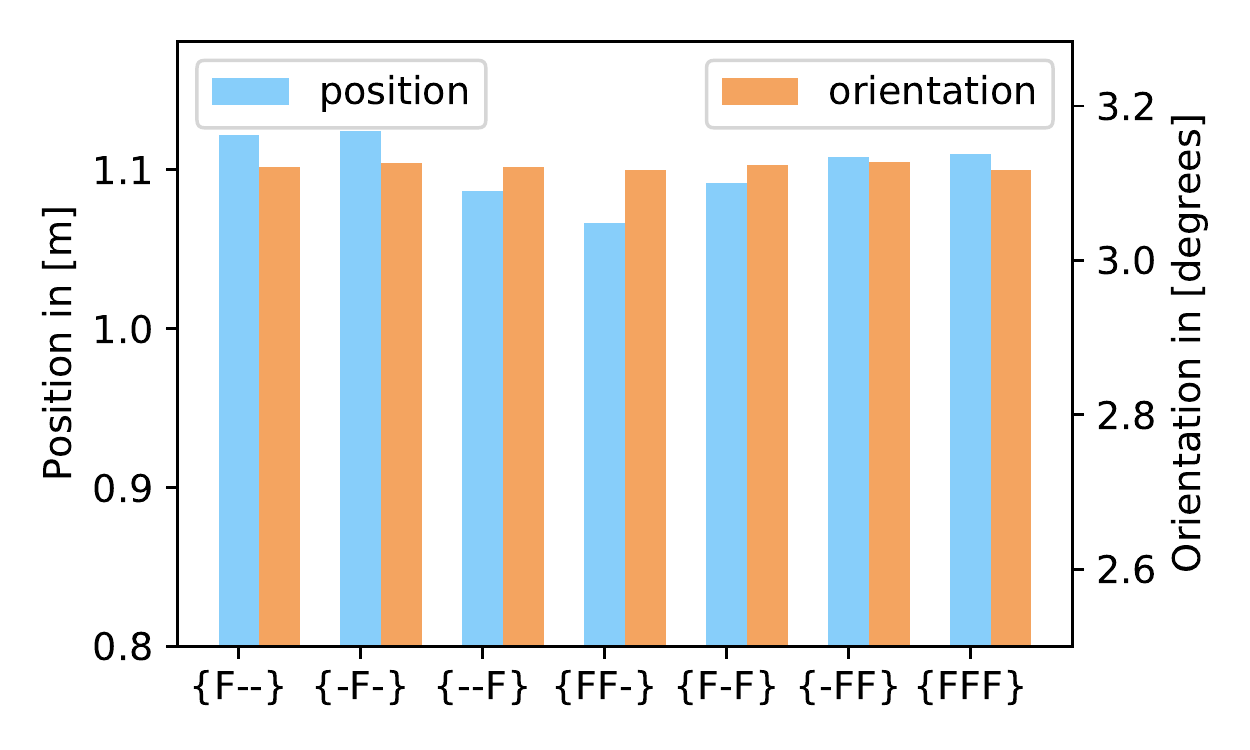}
    	\subcaption{MH-04-difficult.}
    	\label{image_mmtm_comb2}
    \end{minipage}
    \caption{Evaluation of various fusion combinations of MMTM modules for the EuRoC MAV~\cite{burri} dataset. While ``F'' indicates the layer with MMTM fusion (see Section~\ref{section_intermediate_fusion}), "--" indicates no fusion.}
    \label{image_mmtm_comb}
\end{figure*}

\clearpage

\begin{figure*}[!t]
	\centering
	\begin{minipage}[b]{0.495\linewidth}
        \centering
    	\includegraphics[trim=40 40 40 40, clip, width=1.0\linewidth]{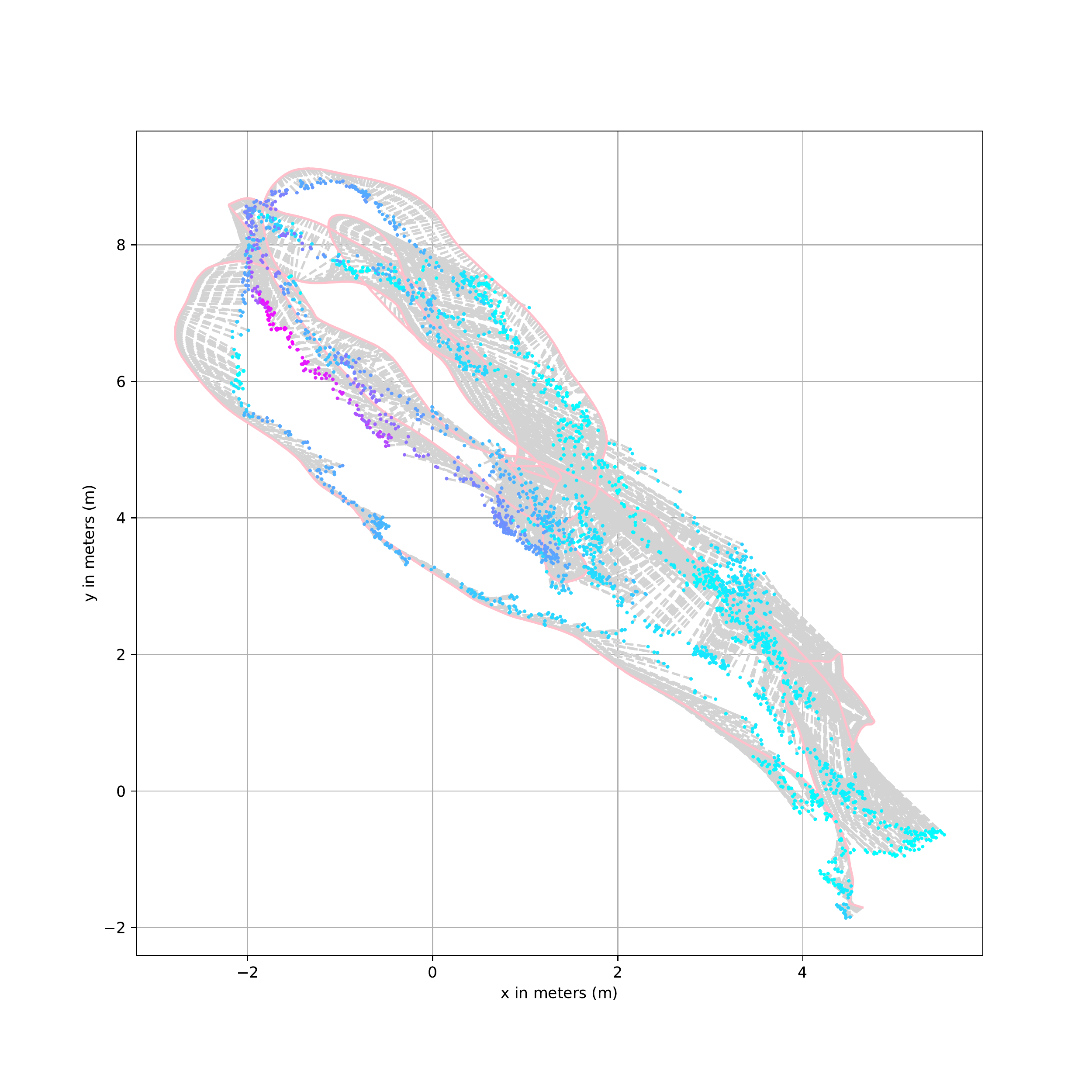}
    \end{minipage}
    \hfill
	\begin{minipage}[b]{0.495\linewidth}
        \centering
    	\includegraphics[trim=40 40 40 40, clip, width=1.0\linewidth]{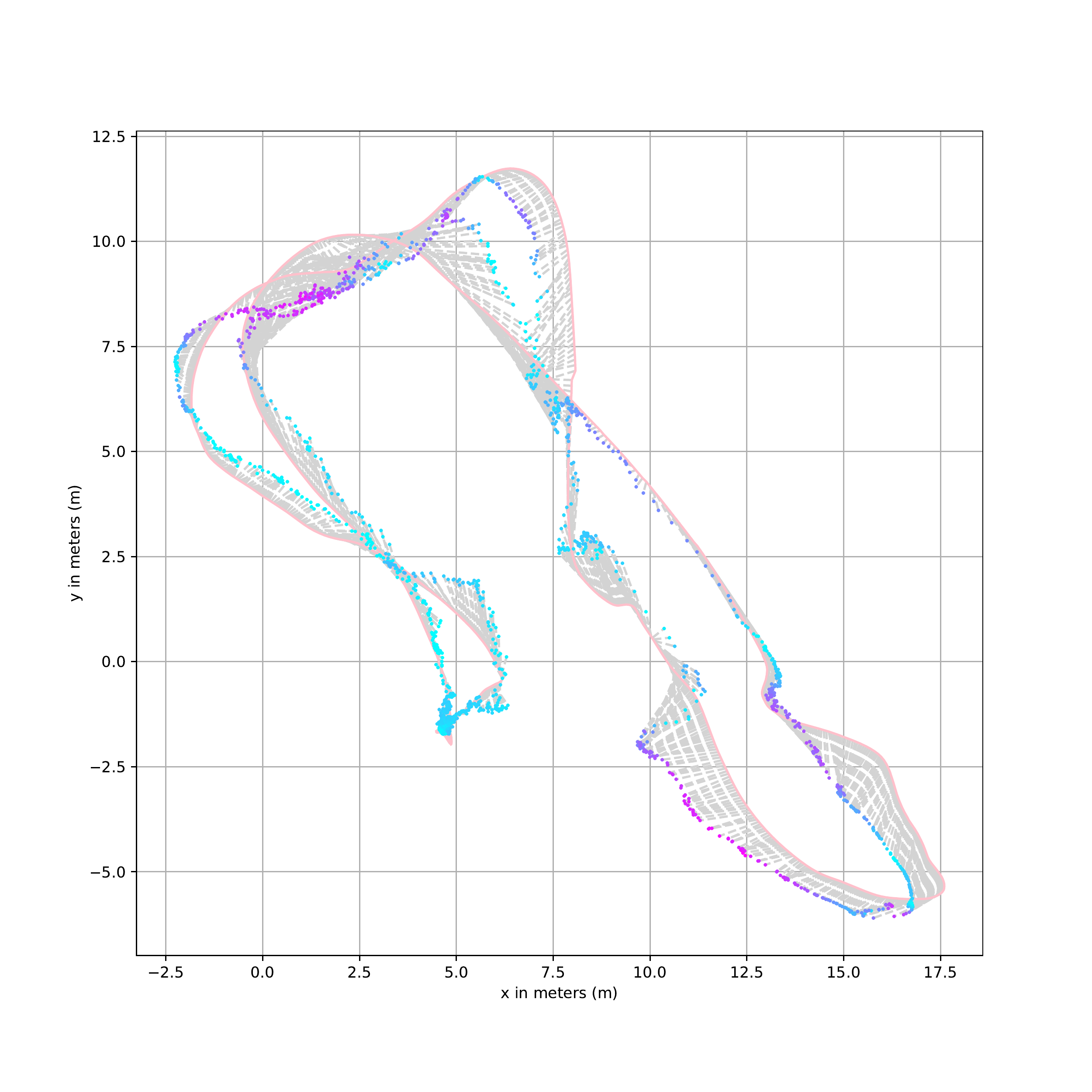}
    \end{minipage}
	\begin{minipage}[b]{0.495\linewidth}
        \centering
    	\includegraphics[width=1.0\linewidth]{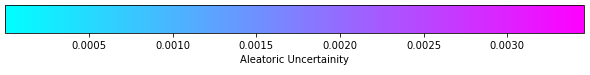}
    	\subcaption{EuRoC MAV~\cite{burri}: MH-02-easy. \cite{nisha_master}}
    	\label{image_euroc_uncertainty1}
    \end{minipage}
    \hfill
	\begin{minipage}[b]{0.495\linewidth}
        \centering
    	\includegraphics[width=1.0\linewidth]{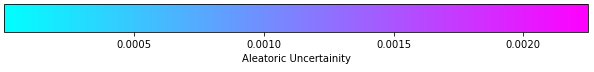}
    	\subcaption{EuRoC MAV~\cite{burri}: MH-04-difficult. \cite{nisha_master}}
    	\label{image_euroc_uncertainty2}
    \end{minipage}
    \caption{Plot of the aleatoric uncertainty for the absolute pose prediction. We plot a dashed line from the predictions to the ground truth trajectory. Note the increased uncertainty for the middle left part of Figure a) and top left part of Figure b).}
    \label{image_euroc_uncertainty}
\end{figure*}

\clearpage

\begin{table*}
\begin{center}
    \caption{Evaluation results on the EuRoC MAV~\cite{burri} dataset and comparison to state-of-the-art techniques. Absolute trajectory error (ATE) $e_{\text{ATE,p}}$ \cite{sturm} calculated with \cite{horn}. \textbf{Bold} are best results. \\ At present, there is a plethora of visual odometry (VO) and inertial odometry (IO) techniques available, which primarily utilize the EuRoC MAV dataset for performance evaluation. Given the absence of a comprehensive survey comparing the results of all techniques, we present a summary of their results in Table~\ref{table_results6} and compare it with the results obtained from our $\text{RPR}_{\text{I}}$ and $\text{RPR}_{\text{V}}$ techniques using the absolute trajectory error (ATE) $e_{\text{ATE,p}}$ metric (see Section~\ref{chap_evaluation}). The EuRoC MAV dataset is cross-validated and evaluated on all testing sequences. The current state-of-the-art techniques employ alternative optimization strategies and exhibit improved performance on the EuRoC MAV dataset. ORB-SLAM~\cite{mur} outperforms other methods on the MH-01, MH-02, MH-03, MH-04, V1-01, and V2-01 datasets, while ORB-SLAM2~\cite{mur2} yields low ATE results on teh MH-05, V1-03, and V2-02 datasets. The variance of results of various methods is very high. The utilization of a stereo camera has a significant impact, as evidenced by the significant improvement in the results of LSD-SLAM~\cite{engel}. In addition, Qin et al.~\cite{qin_pan} improve the method that combines monocular images with IMU data by enhancing the method with stereo images. Our deep learning-based techniques are optimized based on the relative positional error, resulting in favorable performance as evaluated through the median relative metrics. However, it should be noted that the error increases when evaluated using the ATE metric. Comparing the results of IMUNet~\cite{silva} for $\text{RPR}_{\text{I}}$ and FlowNet~\cite{dosovitskiy} for $\text{RPR}_{\text{V}}$ with the fusion techniques, wither FlowNet (MH-01 and V1-01), late fusion utilizing SSF~\cite{chen} and BiLSTM layers (V1-03 and V2-02), and particularly MMTM~\cite{joze} (on the remaining sequences) yield the lowest ATE error.}
    \label{table_results6}
    \footnotesize \begin{tabular}{ p{4.2cm} | p{0.5cm} | p{0.5cm} | p{0.5cm} | p{0.5cm} | p{0.5cm} | p{0.5cm} | p{0.5cm} | p{0.5cm} | p{0.5cm} | p{0.5cm} | p{0.5cm}}
    \multicolumn{1}{c|}{\textbf{Method}} & \multicolumn{1}{c}{\textbf{MH-01}} & \multicolumn{1}{c}{\textbf{MH-02}} & \multicolumn{1}{c}{\textbf{MH-03}} & \multicolumn{1}{c}{\textbf{MH-04}} & \multicolumn{1}{c|}{\textbf{MH-05}} & \multicolumn{1}{c}{\textbf{V1-01}} & \multicolumn{1}{c}{\textbf{V1-02}} & \multicolumn{1}{c|}{\textbf{V1-03}} & \multicolumn{1}{c}{\textbf{V2-01}} & \multicolumn{1}{c}{\textbf{V2-02}} & \multicolumn{1}{c}{\textbf{V2-03}} \\ \hline \hline
    Hong et al.~\cite{hong} & \multicolumn{1}{r}{0.14} & \multicolumn{1}{r}{0.13} & \multicolumn{1}{r}{0.20} & \multicolumn{1}{r}{0.22} & \multicolumn{1}{r|}{0.20} & \multicolumn{1}{r}{0.05} & \multicolumn{1}{r}{0.07} & \multicolumn{1}{r|}{0.16} & \multicolumn{1}{r}{0.04} & \multicolumn{1}{r}{0.11} & \multicolumn{1}{r}{0.17} \\
    SVO + MSF~\cite{faessler} & \multicolumn{1}{r}{0.14} & \multicolumn{1}{r}{0.20} & \multicolumn{1}{r}{0.48} & \multicolumn{1}{r}{1.38} & \multicolumn{1}{r|}{0.51} & \multicolumn{1}{r}{0.40} & \multicolumn{1}{r}{0.63} & \multicolumn{1}{c|}{-} & \multicolumn{1}{r}{0.20} & \multicolumn{1}{r}{0.37} & \multicolumn{1}{c}{-} \\
    OKVIS~\cite{leutenegger_furgale} & \multicolumn{1}{r}{0.16} & \multicolumn{1}{r}{0.22} & \multicolumn{1}{r}{0.24} & \multicolumn{1}{r}{0.34} & \multicolumn{1}{r|}{0.47} & \multicolumn{1}{r}{0.09} & \multicolumn{1}{r}{0.20} & \multicolumn{1}{r|}{0.24} & \multicolumn{1}{r}{0.13} & \multicolumn{1}{r}{0.16} & \multicolumn{1}{r}{0.29} \\
    ROVIO~\cite{bloesch_omari} & \multicolumn{1}{r}{0.21} & \multicolumn{1}{r}{0.25} & \multicolumn{1}{r}{0.25} & \multicolumn{1}{r}{0.49} & \multicolumn{1}{r|}{0.52} & \multicolumn{1}{r}{0.10} & \multicolumn{1}{r}{0.10} & \multicolumn{1}{r|}{0.14} & \multicolumn{1}{r}{0.12} & \multicolumn{1}{r}{0.14} & \multicolumn{1}{r}{0.14} \\
    VINS-monocular~\cite{qin} & \multicolumn{1}{r}{0.27} & \multicolumn{1}{r}{0.12} & \multicolumn{1}{r}{0.13} & \multicolumn{1}{r}{0.23} & \multicolumn{1}{r|}{0.35} & \multicolumn{1}{r}{0.07} & \multicolumn{1}{r}{0.10} & \multicolumn{1}{r|}{0.13} & \multicolumn{1}{r}{0.08} & \multicolumn{1}{r}{0.08} & \multicolumn{1}{r}{0.21} \\
    SVO~\cite{forster}: monocular & \multicolumn{1}{r}{0.17} & \multicolumn{1}{r}{0.27} & \multicolumn{1}{r}{0.43} & \multicolumn{1}{r}{1.36} & \multicolumn{1}{r|}{0.51} & \multicolumn{1}{r}{0.20} & \multicolumn{1}{r}{0.47} & \multicolumn{1}{c|}{-} & \multicolumn{1}{r}{0.30} & \multicolumn{1}{r}{0.47} & \multicolumn{1}{c}{-} \\
    SVO~\cite{forster}: monocular, edge & \multicolumn{1}{r}{0.17} & \multicolumn{1}{r}{0.27} & \multicolumn{1}{r}{0.42} & \multicolumn{1}{r}{1.00} & \multicolumn{1}{r|}{0.60} & \multicolumn{1}{r}{0.22} & \multicolumn{1}{r}{0.35} & \multicolumn{1}{c|}{-} & \multicolumn{1}{r}{0.26} & \multicolumn{1}{r}{0.40} & \multicolumn{1}{c}{-} \\
    SVO~\cite{forster}: monocular, edge + prior & \multicolumn{1}{r}{0.10} & \multicolumn{1}{r}{0.12} & \multicolumn{1}{r}{0.41} & \multicolumn{1}{r}{0.43} & \multicolumn{1}{r|}{0.30} & \multicolumn{1}{r}{0.07} & \multicolumn{1}{r}{0.21} & \multicolumn{1}{c|}{-} & \multicolumn{1}{r}{0.11} & \multicolumn{1}{r}{0.11} & \multicolumn{1}{r}{1.08} \\
    SVO~\cite{forster}: monocular, BA & \multicolumn{1}{r}{0.06} & \multicolumn{1}{r}{0.07} & \multicolumn{1}{c}{-} & \multicolumn{1}{r}{0.40} & \multicolumn{1}{c|}{-} & \multicolumn{1}{r}{0.05} & \multicolumn{1}{c}{-} & \multicolumn{1}{c|}{-} & \multicolumn{1}{c}{-} & \multicolumn{1}{c}{-} & \multicolumn{1}{c}{-} \\
    DSO~\cite{engel_dso} & \multicolumn{1}{r}{0.05} & \multicolumn{1}{r}{0.05} & \multicolumn{1}{r}{0.18} & \multicolumn{1}{r}{2.50} & \multicolumn{1}{r|}{0.11} & \multicolumn{1}{r}{0.12} & \multicolumn{1}{r}{0.11} & \multicolumn{1}{r|}{0.93} & \multicolumn{1}{r}{0.04} & \multicolumn{1}{r}{0.13} & \multicolumn{1}{r}{0.16} \\
    VI-DSO~\cite{stumberg} & \multicolumn{1}{r}{0.041} & \multicolumn{1}{r}{0.041} & \multicolumn{1}{r}{0.116} & \multicolumn{1}{r}{0.129} & \multicolumn{1}{r|}{0.106} & \multicolumn{1}{r}{0.057} & \multicolumn{1}{r}{0.066} & \multicolumn{1}{r|}{0.095} & \multicolumn{1}{r}{0.031} & \multicolumn{1}{r}{0.060} & \multicolumn{1}{r}{0.173} \\
    SVO~\cite{forster}: stereo & \multicolumn{1}{r}{0.08} & \multicolumn{1}{r}{0.08} & \multicolumn{1}{r}{0.29} & \multicolumn{1}{r}{2.67} & \multicolumn{1}{r|}{0.43} & \multicolumn{1}{r}{0.05} & \multicolumn{1}{r}{0.09} & \multicolumn{1}{r|}{0.36} & \multicolumn{1}{r}{0.09} & \multicolumn{1}{r}{0.52} & \multicolumn{1}{c}{-} \\
    ORB-SLAM~\cite{mur} (no LC)& \multicolumn{1}{r}{\textbf{0.03}} & \multicolumn{1}{r}{\textbf{0.02}} & \multicolumn{1}{r}{\textbf{0.02}} & \multicolumn{1}{r}{\textbf{0.20}} & \multicolumn{1}{r|}{0.19} & \multicolumn{1}{r}{\textbf{0.04}} & \multicolumn{1}{c}{-} & \multicolumn{1}{c|}{-} & \multicolumn{1}{r}{\textbf{0.02}} & \multicolumn{1}{r}{0.07} & \multicolumn{1}{c}{-} \\
    ORB-SLAM2~\cite{mur2} & \multicolumn{1}{r}{0.035} & \multicolumn{1}{r}{0.018} & \multicolumn{1}{r}{0.028} & \multicolumn{1}{r}{0.119} & \multicolumn{1}{r|}{\textbf{0.060}} & \multicolumn{1}{r}{0.035} & \multicolumn{1}{r}{0.020} & \multicolumn{1}{r|}{\textbf{0.048}} & \multicolumn{1}{r}{0.037} & \multicolumn{1}{r}{\textbf{0.035}} & \multicolumn{1}{c}{-} \\
    LSD-SLAM~\cite{engel}: monocular, no LC & \multicolumn{1}{r}{0.18} & \multicolumn{1}{r}{0.56} & \multicolumn{1}{r}{2.69} & \multicolumn{1}{r}{2.13} & \multicolumn{1}{r|}{0.85} & \multicolumn{1}{r}{1.24} & \multicolumn{1}{r}{1.11} & \multicolumn{1}{c|}{-} & \multicolumn{1}{c}{-} & \multicolumn{1}{c}{-} & \multicolumn{1}{c}{-} \\
    LSD-SLAM~\cite{engel}: stereo & \multicolumn{1}{c}{-} & \multicolumn{1}{c}{-} & \multicolumn{1}{c}{-} & \multicolumn{1}{r}{} & \multicolumn{1}{r|}{0.066} & \multicolumn{1}{r}{0.074} & \multicolumn{1}{r}{0.089} & \multicolumn{1}{c|}{-} & \multicolumn{1}{c}{-} & \multicolumn{1}{c}{-} & \multicolumn{1}{c}{-} \\
    Qin et al.~\cite{qin_pan}: stereo & \multicolumn{1}{r}{0.54} & \multicolumn{1}{r}{0.46} & \multicolumn{1}{r}{0.33} & \multicolumn{1}{r}{0.78} & \multicolumn{1}{r|}{0.50} & \multicolumn{1}{r}{0.55} & \multicolumn{1}{r}{0.23} & \multicolumn{1}{c|}{-} & \multicolumn{1}{r}{0.23} & \multicolumn{1}{r}{0.20} & \multicolumn{1}{c}{-} \\
    Qin et al.~\cite{qin_pan}: monocular+IMU & \multicolumn{1}{r}{0.24} & \multicolumn{1}{r}{0.18} & \multicolumn{1}{r}{0.23} & \multicolumn{1}{r}{0.39} & \multicolumn{1}{r|}{0.19} & \multicolumn{1}{r}{0.10} & \multicolumn{1}{r}{0.10} & \multicolumn{1}{r|}{0.11} & \multicolumn{1}{r}{0.12} & \multicolumn{1}{r}{0.10} & \multicolumn{1}{r}{0.27} \\
    Qin et al.~\cite{qin_pan}: stereo+IMU & \multicolumn{1}{r}{0.18} & \multicolumn{1}{r}{0.09} & \multicolumn{1}{r}{0.17} & \multicolumn{1}{r}{0.21} & \multicolumn{1}{r|}{0.25} & \multicolumn{1}{r}{0.06} & \multicolumn{1}{r}{0.09} & \multicolumn{1}{r|}{0.18} & \multicolumn{1}{r}{0.06} & \multicolumn{1}{r}{0.11} & \multicolumn{1}{r}{0.26} \\
    Qin et al.~\cite{qin_li} & \multicolumn{1}{r}{0.120} & \multicolumn{1}{r}{0.120} & \multicolumn{1}{r}{0.130} & \multicolumn{1}{r}{0.180} & \multicolumn{1}{r|}{0.210} & \multicolumn{1}{r}{0.068} & \multicolumn{1}{r}{\textbf{0.084}} & \multicolumn{1}{r|}{0.190} & \multicolumn{1}{r}{0.081} & \multicolumn{1}{r}{0.160} & \multicolumn{1}{c}{-} \\
    Mu et a.~\cite{mu_chen} & \multicolumn{1}{r}{0.551} & \multicolumn{1}{r}{0.402} & \multicolumn{1}{r}{0.997} & \multicolumn{1}{r}{0.704} & \multicolumn{1}{r|}{0.672} & \multicolumn{1}{r}{0.521} & \multicolumn{1}{r}{0.652} & \multicolumn{1}{c|}{-} & \multicolumn{1}{r}{0.232} & \multicolumn{1}{r}{0.617} & \multicolumn{1}{c}{-} \\
    Feng et al.~\cite{feng} & \multicolumn{1}{r}{0.111} & \multicolumn{1}{r}{0.074} & \multicolumn{1}{r}{0.173} & \multicolumn{1}{r}{0.143} & \multicolumn{1}{r|}{0.205} & \multicolumn{1}{r}{0.077} & \multicolumn{1}{r}{0.143} & \multicolumn{1}{r|}{0.093} & \multicolumn{1}{r}{0.082} & \multicolumn{1}{r}{0.100} & \multicolumn{1}{r}{0.233} \\ \hline \hline
    IMUNet~\cite{silva} & \multicolumn{1}{r}{\textbf{1.7606}} & \multicolumn{1}{r}{2.8277} & \multicolumn{1}{r}{2.9113} & \multicolumn{1}{r}{3.7231} & \multicolumn{1}{r|}{3.2198}&\multicolumn{1}{r}{\textbf{1.4714}} & \multicolumn{1}{r}{1.7456} & \multicolumn{1}{r|}{1.4709} & \multicolumn{1}{r}{1.7779} & \multicolumn{1}{r}{1.7878} & \multicolumn{1}{r}{1.8935}  \\
    FlowNet~\cite{dosovitskiy} & \multicolumn{1}{r}{2.0312} & \multicolumn{1}{r}{2.1546} & \multicolumn{1}{r}{3.5234} & \multicolumn{1}{r}{3.3650} & \multicolumn{1}{r|}{2.9856} & \multicolumn{1}{r}{1.7920} & \multicolumn{1}{r}{1.8673} & \multicolumn{1}{r|}{1.7761} & \multicolumn{1}{r}{2.0221} & \multicolumn{1}{r}{2.1203} & \multicolumn{1}{r}{2.1952} \\
    Late Fusion (concat) & \multicolumn{1}{r}{2.2651} & \multicolumn{1}{r}{1.9991} & \multicolumn{1}{r}{3.3241} & \multicolumn{1}{r}{3.5760} & \multicolumn{1}{r|}{2.8914} & \multicolumn{1}{r}{1.7451} & \multicolumn{1}{r}{1.7818} & \multicolumn{1}{r|}{1.3156} & \multicolumn{1}{r}{1.8011} & \multicolumn{1}{r}{1.7992} & \multicolumn{1}{r}{1.9185} \\
    Late Fusion (concat) + BiLSTM & \multicolumn{1}{r}{2.1820} & \multicolumn{1}{r}{1.7792} & \multicolumn{1}{r}{2.2406} & \multicolumn{1}{r}{2.9761} & \multicolumn{1}{r|}{2.3476} & \multicolumn{1}{r}{1.6782} & \multicolumn{1}{r}{1.7009} & \multicolumn{1}{r|}{1.2810} & \multicolumn{1}{r}{1.7429} & \multicolumn{1}{r}{1.7224} & \multicolumn{1}{r}{1.8208} \\
    Late Fusion (SSF) \cite{chen} & \multicolumn{1}{r}{2.2827} & \multicolumn{1}{r}{1.9621} & \multicolumn{1}{r}{3.5606} & \multicolumn{1}{r}{3.5281} & \multicolumn{1}{r|}{3.0364} & \multicolumn{1}{r}{1.6774} & \multicolumn{1}{r}{1.7758} & \multicolumn{1}{r|}{1.3459} & \multicolumn{1}{r}{1.7758} & \multicolumn{1}{r}{1.8828} & \multicolumn{1}{r}{1.9256} \\
    Late Fusion (SSF) \cite{chen} + BiLSTM & \multicolumn{1}{r}{2.0110} & \multicolumn{1}{r}{1.8171} & \multicolumn{1}{r}{2.5411} & \multicolumn{1}{r}{3.1201} & \multicolumn{1}{r|}{2.5431} & \multicolumn{1}{r}{1.5901} & \multicolumn{1}{r}{1.6906} & \multicolumn{1}{r|}{\textbf{1.2134}} & \multicolumn{1}{r}{1.7123} & \multicolumn{1}{r}{\textbf{1.7601}} & \multicolumn{1}{r}{1.8567} \\
    MMTM~\cite{joze} (3 modules) & \multicolumn{1}{r}{1.8740} & \multicolumn{1}{r}{\textbf{1.7541}} & \multicolumn{1}{r}{\textbf{1.9451}} & \multicolumn{1}{r}{\textbf{2.6712}} & \multicolumn{1}{r|}{\textbf{2.1182}} & \multicolumn{1}{r}{1.6002} & \multicolumn{1}{r}{\textbf{1.6321}} & \multicolumn{1}{r|}{1.2998} & \multicolumn{1}{r}{\textbf{1.7001}} & \multicolumn{1}{r}{1.7866} & \multicolumn{1}{r}{\textbf{1.8012}} \\
    \end{tabular}
\end{center}
\end{table*}

\end{document}